\documentclass[journal]{IEEEtran}

\usepackage{cite}
\usepackage[cmex10]{amsmath}
\usepackage{algorithmic}
\usepackage{amssymb}
\usepackage{array}
\usepackage{mdwmath}
\usepackage{mdwtab}
\usepackage{xcolor}
\usepackage[caption=false,font=footnotesize]{subfig}
\usepackage{fixltx2e}
\usepackage{multirow}
\usepackage[pdftex]{graphicx}

\newcommand{\PreserveBackSlash}[1]{\let\temp=\\#1\let\\=\temp}
\newcolumntype{C}[1]{>{\PreserveBackSlash\centering}p{#1}}

\newcommand{\GEPPWIDTH}[1]{width=1.8in}

\begin{document}
\title{A Gaussian Mixture MRF\\ for Model-Based Iterative Reconstruction \\
with Applications to Low-Dose X-ray CT}

\author{Ruoqiao Zhang,
	Dong Hye Ye,~\IEEEmembership{Member,~IEEE,}
	Debashish Pal,~\IEEEmembership{Member,~IEEE,} \\
	Jean-Baptiste Thibault,~\IEEEmembership{Member,~IEEE,} 
	Ken D. Sauer,~\IEEEmembership{Member,~IEEE,}	
	and Charles A. Bouman,~\IEEEmembership{Fellow,~IEEE}%
\thanks{This work was supported by GE Healthcare.}
\thanks{R. Zhang is with the Department of Radiology, University of Washington, Seattle, WA 98195, USA
(email: zhangrq@uw.edu).}
\thanks{D. H. Ye, and C. A. Bouman are with the School of Electrical and Computer Engineering,
Purdue University, West Lafayette, IN 47907, USA
(email: yed@purdue.edu; bouman@ecn.purdue.edu).}
\thanks{D. Pal and J.-B. Thibault are with GE Healthcare Technologies,
Waukesha, WI 53188, USA (email: debashish.pal@med.ge.com; jean-baptiste.thibault@med.ge.com).}
\thanks{K. D. Sauer is with the Department of Electrical Engineering, University of Notre Dame,
Notre Dame, IN 46556, USA (email: sauer@nd.edu).}
}

\maketitle

\begin{abstract}
Markov random fields (MRFs) have been widely used as prior models
in various inverse problems such as tomographic reconstruction.
While MRFs provide a simple and often effective way to model
the spatial dependencies in images,
they suffer from the fact that parameter estimation is difficult.
In practice, this means that MRFs typically have very simple structure
that cannot completely capture the subtle characteristics of complex images.

In this paper, we present a novel Gaussian mixture Markov random field model (GM-MRF)
that can be used as a very expressive prior model
for inverse problems such as denoising and reconstruction.
The GM-MRF forms a global image model by merging together
individual Gaussian-mixture models (GMMs) for image patches.
In addition, we present a novel analytical framework for computing MAP estimates
using the GM-MRF prior model through the construction of surrogate functions
that result in a sequence of quadratic optimizations.
We also introduce a simple but effective method to adjust the GM-MRF
so as to control the sharpness in low- and high-contrast regions of the reconstruction separately.
We demonstrate the value of the model with experiments including image denoising and low-dose CT reconstruction.
\end{abstract}


\section{Introduction}
\IEEEPARstart{I}{n} recent years, model-based iterative reconstruction (MBIR) has emerged as a very powerful approach to reconstructing images from sparse or noisy data 
in applications ranging from medical, to scientific, to non-destructive imaging\cite{Boas01, Qi06, Thibault07, Fessler10, Zhang14tmi, Venkat15, Jin15, Mohan15}. 
The power of these methods is due to the synergy that results from modeling both the sensor (i.e., forward model) and the image being reconstructed (i.e., prior model). 
In medical applications, for example, MBIR has been demonstrated to substantially improve image quality by both reducing noise and improving resolution\cite{Nelson11, Yamada12}.

While the MBIR forward model is typically based on the physics of the sensor, accurate prior modeling of real images remains a very challenging problem. 
Perhaps the most commonly used prior model is a very simple Markov random field (MRF) with only very local dependencies and a small number of parameters\cite{Besag74, Sauer93, deman00tns, elbakri02tmi,Thibault07}. 
Alternatively, total-variation (TV) regularization approaches can also be viewed as simple MRF priors\cite{Rudin92, Sidky08, tang09pmb, ritschl11pmb, liu12pmb}. 
Besides direct reconstruction, simple MRF priors have also been employed in applications such as sinogram smoothing and restoration\cite{li04tns, la05mp, wang06tmi}.
While these models have been very useful, their simple form does not allow for accurate or expressive modeling of real images. 

Alternatively, a number of approaches have been proposed for modeling the non-Gaussian distribution of pixels in image reconstruction. 
Hsiao et. al.\cite{hsiao02tip, hsiao03jei} used a gamma mixture model in emission/transmission tomography;
Wang et. al.\cite{wang09tns} used a Gaussian scale mixture to model the distribution of scattering data in muon tomography;
and Mehranian et. al.\cite{mehranian15tmi} used a Gaussian mixture model for PET/MRI reconstruction.
In \cite{wang09tns} and \cite{mehranian15tmi}, the authors also derived a surrogate function for the log of a mixture of Gaussians, which they used in optimization. 
However, none of these mixture models accounted for the spatial correlation of the neighboring pixels in the reconstructed image,
and therefore additional spatial models might be required\cite{mehranian15tmi}.

More recently, methods such as K-SVD\cite{Elad06} and non-local means\cite{Buades05} have been proposed which can be adapted as prior models in MBIR reconstruction.
Though effective in denoising applications, other patch-based methods such as BM3D\cite{Dabov07} 
are not directly suited for application in model-based reconstruction problems,
and integration of such methods into iterative reconstruction is still an active topic of research\cite{venkat13globalsip, chan16arxiv, teodoro16arxiv}. 
While K-SVD can be adapted as a prior model\cite{Ravishankar11, Xu12}, it does not explicitly model the multivariate distribution of the image. 
This can lead to drawbacks in applications. 
For example, the K-SVD algorithm is designed to be invariant to scaling or average gray level of image patches. 
In applications such as CT reconstruction, this is a severe limitation since regions of different densities generally correspond to different tissues (e.g., bone and soft tissue) with distinctly different characteristics. 
While direct incorporation of the non-local mean filtering into iterative reconstruction framework is still under study\cite{huang11cbm, venkat13globalsip, sreehari15arxiv},
there have been efforts to adapt this filter as the regularization in tomographic reconstruction\cite{chen08jmiv, Wang12, ma12pmb, zhang14cmig}.
Though producing promising results, these methods do not explicitly model the statistics of the image,
and therefore are not consistent image models.  
A variety of research also adapted the ideas of dictionary learning to the problem of prior modeling in CT reconstruction\cite{Liao08, Lu12, Pfister14}.

Another approach to prior modeling is to allow different patches of the images to have different distributions. 
This approach has been used by both Zoran et. al.\cite{Zoran11} and Yu et. al.\cite{Yu12} 
to construct non-homogeneous models of images as the composition of patches, each with a distinct Gaussian distribution. 
The Gaussian distribution of each patch is selected from a discrete set of possible distributions (i.e., distinct mean and covariance). 
The reconstruction is then computed by jointly estimating both the image and a discrete class for each patch in the image. 
This approach can be very powerful for modeling the different spatially varying characteristics in real images. 
However, the approaches suffer from the need to make hard classifications of each patch. 
These hard classifications can lead to artifacts when patch distributions overlap, as is typically the case when a large number of classes are used. 
Recently, patch models using Gaussian mixtures have also been applied in various applications\cite{nguyen13, wang13sure, Zhang13fully3d, yang14tip, Nadir15globalsip}.

In this paper, we introduce the Gaussian-mixture MRF (GM-MRF) image prior along with an associated method for computing the MAP estimate using exact surrogate functions. 
(See\cite{Zhang13, Zhang15fully3d} for early conference versions of our method). 
The GM-MRF model is constructed by seaming together patches that are modeled with a single Gaussian mixture (GM) distribution. The advantages of this approach are that:
\begin{itemize}
\item The GM-MRF prior provides a theoretically consistent and very expressive model of the multivariate distribution of the image;
\item The GM-MRF parameters can be easily and accurately estimated by fitting a GM distribution to patch training data using standard methods such as the EM algorithm\cite{Bouman97}; 
\item MAP optimization can be efficiently computed by alternating soft classification of image patches with MAP reconstruction using quadratic regularization. 
\end{itemize}

To create a consistent global image model, we seam together the GM patch models by using the geometric mean of their probability densities. 
This approach, similar to the product-of-experts technique\cite{Salakhutdinov12} employed in deep-learning, produces a single consistent probability density for the entire image. 
Moreover, we also show that the resulting GM-MRF model is a Markov random field (MRF) as its name implies. 

Of course, an accurate image model is of little value if computation of the MAP estimate is difficult. 
Fortunately, we show that  the GM-MRF prior has an exact quadratic surrogate function for its log likelihood. 
This surrogate function allows for tractable minimization of the MAP function using a majorization-minimization approach\cite{Hunter04}. 
The resulting MAP optimization algorithm has the form of alternating minimization. 
The two alternating steps are soft classification for patches followed by MAP optimization using quadratic regularization (i.e., a non-homogeneous Gaussian prior). 
Moreover, our approach to MAP optimization with the GM-MRF prior avoids the need for hard classification of individual patches. 
In practice, this means that patch-based GM-MRF models with a large numbers of overlapping mixture components can be used without adverse modeling effects. 
This allows for the use of very expressive models that capture fine details of image behavior. 
Note that Wang et. al.\cite{wang09tns} and Mehranian et. al.\cite{mehranian15tmi} used similar technique
to construct quadratic surrogate functions for the log of mixture of Gaussians.
However, in this paper we formulate and prove a lemma that is more general and can be applied to a wider range of distributions as compared to previous methods.

It has been reported that MAP estimation with non-Gaussian priors can lead to contrast-dependent noise and resolution properties in reconstructed images\cite{fessler96tip, evans11mp, li14mp}.
Typically, the estimated image has higher noise and resolution at higher-contrast edges, 
and lower noise and resolution at lower-contrast edges. 
In practice, this non-homogeneity may result in undesirable image quality in specific applications.

In order to address this issue of non-homogeneity for the GM-MRF prior, 
we introduce a simple method for adjusting the GM components of the GMMRF prior, 
so as to control the sharpness in low and high contrast regions of the reconstruction separately.

Our experimental results indicate that GM-MRF method results in improved image quality and reduced RMS error in simple denoising problems as compared to simple MRF and K-SVD priors. 
We also show multi-slice helical scan tomographic reconstructions from both phantom and clinical data to demonstrate that the GM-MRF prior produces visually superior images 
as compared to filtered back-projection (FBP) and MBIR using the traditional $q$-GGMRF prior\cite{Thibault07},
especially under the condition of low-dose acquisition.

\section{Gaussian mixture Markov random field}
\label{sec:model}

MBIR algorithms work by computing the maximum \textit{a posteriori} (MAP) estimate of the unknown image, $x$,
given the measured data, $y$, by
\begin{equation}
\label{eq:map1}
\hat{x} \leftarrow \arg \min_{x\in \Omega} \left\{ -\log p(y|x) - \log p(x) \right\} \ .
\end{equation}
In this framework, $p(y|x)$ is the conditional probability of $y$ given $x$,
which comprises the forward model of the measurement process.
The density $p(x)$ is the prior model for $x$, which will be discussed in detail in this section.

Let $x\in \Re^N$ be an image with pixels $s\in S$, where $S$ is the set of all pixels in $x$ with $|S| = N$.
Let $P_s\in \mathcal{Z}^{L\times N}$ be a patch operator that extracts a patch from the image, where the patch is centered at pixel $s$ and contains $L$ pixels.
More precisely, $P_s$ is a rank $L$ matrix that has a value of 1 at locations belonging to the patch and 0 otherwise.
Furthermore, we assume that each patch, $P_s x$, can be modeled as 
having a multivariate Gaussian mixture distribution with $K$ components,
\begin{equation}
\label{eq:GMM}
g(P_s x) = \sum_{k=1}^K  \frac{\pi_k |R_k|^{-\frac12}}{(2\pi)^{\frac{L}2}}  \exp \left\{-\frac12 \|P_s x - \mu_k\|^2_{R_k^{-1}} \right\} \ ,
\end{equation}
where parameters $\pi_k, \mu_k, R_k$ represent the mixture probability, mean, and covariance, respectively, of the $k^{\rm th}$ mixture component.

\begin{figure}[!t]
\centerline{
\includegraphics[width=2.6in]{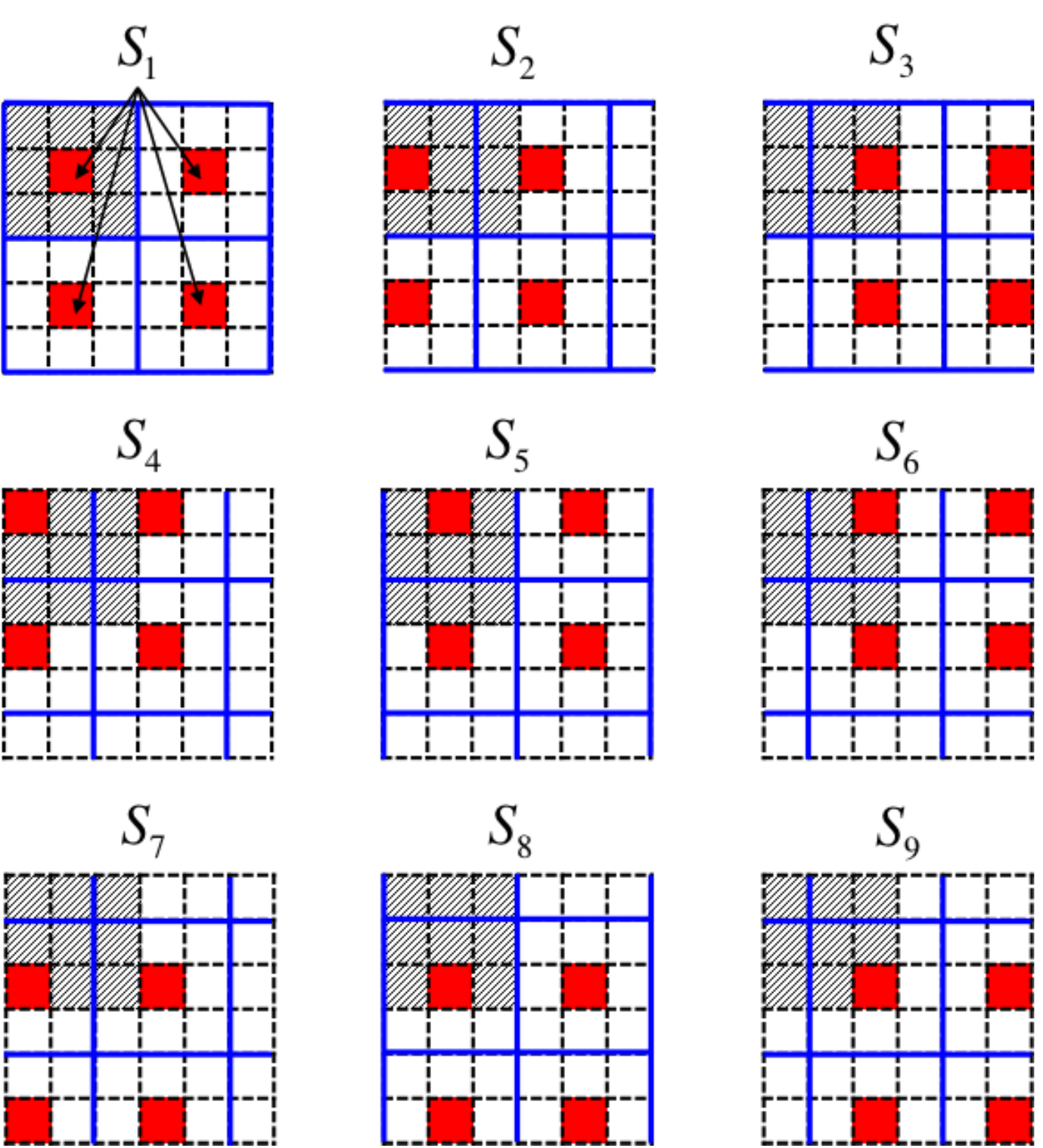}}
\caption{
2-D illustration of the tiling method. 
Each blue grid represents one of nine distinct tilings with $3\times3$ patches on a $6\times6$ grid, i.e., $L=9$,
with the center pixel of each patch marked in red.
Toroidal boundary condition is considered in this illustration.
Note that there are exactly 9 distinct phase shifts of the tiling,
each of which is determined by the center pixel of the first patch in the upper-left corner,
which corresponds to a distinct pixel location in the shadowed patch.}
\label{fig:tiling}
\end{figure}

Then let $\left\{S_m\right\}, m\in\left\{1,\cdots,L \right\},$ be a partition of the set of all pixels into $L$ sets, each of which tiles the image space.
In other words, $\{P_s x\}_{s\in S_m}$ forms a set of non-overlapping patches, which contain all pixels in $x$.
A simple \mbox{2-D} example of this is when each $P_s x$ is a square $r \times r$ patch,
and $S_m$ is the set of pixels at each $r^{\rm th}$ row and column.
Then the set of patches, $\{P_s x\}_{s\in S_m}$, tiles the plane. 

Importantly, there are exactly $L$ distinct tilings of the image space where $L$ is the number of pixels in a patch. 
In order to see why this is true, consider the \mbox{2-D} example in Fig.~\ref{fig:tiling}. 
(Note that this tiling method can be easily extended to \mbox{$n$-D} space with $n\geq3$ by using \mbox{$n$-D} patches.)
Notice that each distinct tiling of the space is determined by the position of the center pixel for the first (e.g., upper left hand) patch 
since the positioning of the first patch determines the phase shift of the tiling. 
With this in mind, there are exactly $L$ distinct phase shifts corresponding to the $L$ pixels in a single patch.
Using this notation, we model the distribution of each tiling as the product of distributions of all its patches, as
\begin{equation}
p_m(x) = \prod_{s\in S_m} g(P_s x) \ .
\end{equation}
In this case, $p_m (x)$ has the desired distribution for each patch.
However, the discrete tiling of the space introduces artificial boundaries between patches.
To remove the boundary artifacts, we use an approach similar to the product-of-experts approach in \cite{Salakhutdinov12}
and take the geometric average of the probability densities for all $L$ tilings of the image space to obtain the resulting distribution
\begin{equation}
\label{eq:POE}
p(x) = \frac1z \left( \prod_{m=1}^{L} p_m(x) \right)^{\frac1{L}} = \frac1z \left( \prod_{s\in S} g(P_sx) \right)^{\frac1{L}} \ ,
\end{equation}
where $z$ is a normalizing factor introduced to assure that
$p(x)$ is a proper distribution after the geometric average is computed.

Let $V(P_s x) = -\log\{ g(P_s x)\}$. Then we formulate a Gaussian mixture MRF (GM-MRF) model directly from (\ref{eq:POE}) as
\begin{equation}
\label{eq:GMMRF}
p(x) = \frac1z  \exp\left\{ -u(x) \right\} \ ,
\end{equation}
with the energy function
\begin{equation}
\label{eq:energy}
u(x) = \frac1{L} \sum_{s\in S} V(P_s x) \ ,
\end{equation}
and the potential function
\begin{equation}
\label{eq:logGMM}
V(P_s x) = -\log \left\{ \sum_{k=1}^K \frac{\pi_k |R_k|^{-\frac12}}{(2\pi)^{\frac{L}2}}  \exp \left\{-\frac{\|P_s x - \mu_k\|^2_{R_k^{-1}}}2\right\} \right\} \ .
\end{equation}

Notice that $p(x)$ is a Gibbs distribution by (\ref{eq:GMMRF}).
Therefore, by the renowned Hammersley-Clifford theorem\cite{Besag74}, $p(x)$ is also an MRF.

\section{MAP estimation with GM-MRF prior}
\label{sec:optimization}
For typical model-based inversion problems, the log-likelihood function may be modeled under the Gaussian assumption as
\begin{equation}
\label{eq:loglikelihood}
-\log p(y|x) = \frac12  \|y -Ax\|^2_D, 
\end{equation}
where $A\in \Re^{M\times N}$ is the projection matrix with $M$ measurements and $N$ unknowns.
The weighting $D$ is a diagonal matrix with each diagonal element inversely proportional to the variance of the corresponding measurement.

\subsection{Surrogate prior}
By substituting (\ref{eq:GMMRF}) and (\ref{eq:loglikelihood}) into (\ref{eq:map1}), 
we can calculate the MAP estimate with the GM-MRF prior as
\begin{equation}
\label{eq:map}
\hat{x} \leftarrow \arg \min_{x\in \Omega} \left\{ \frac12  \|y -Ax\|^2_D + u(x) \right\} \ .
\end{equation}
However, the function $u(x)$ is not well suited for direct optimization
due to the mixture of logarithmic and exponential functions.
Therefore, we will use a majorization-minimization approach,
in which we replace the function $u(x)$ with a quadratic upper-bounding surrogate function.

More precisely, the objective of the majorization-minimization method is to find a surrogate function 
$u(x; x^\prime)$ that satisfies the following two conditions.
\begin{eqnarray}
u(x';x') &=& u(x') \\
u(x;x') &\geq& u(x)
\end{eqnarray}
Intuitively, these conditions state that the surrogate function upper bounds $u(x)$ and that the two functions are equal when $x = x^\prime$. 
Importantly, these conditions also imply that any reduction of $u(x;x^\prime)$ also must reduce $u(x)$.

\begin{figure}[!t]
\centerline{\subfloat[Gaussian mixture distribution]{\includegraphics[height=1.7in]{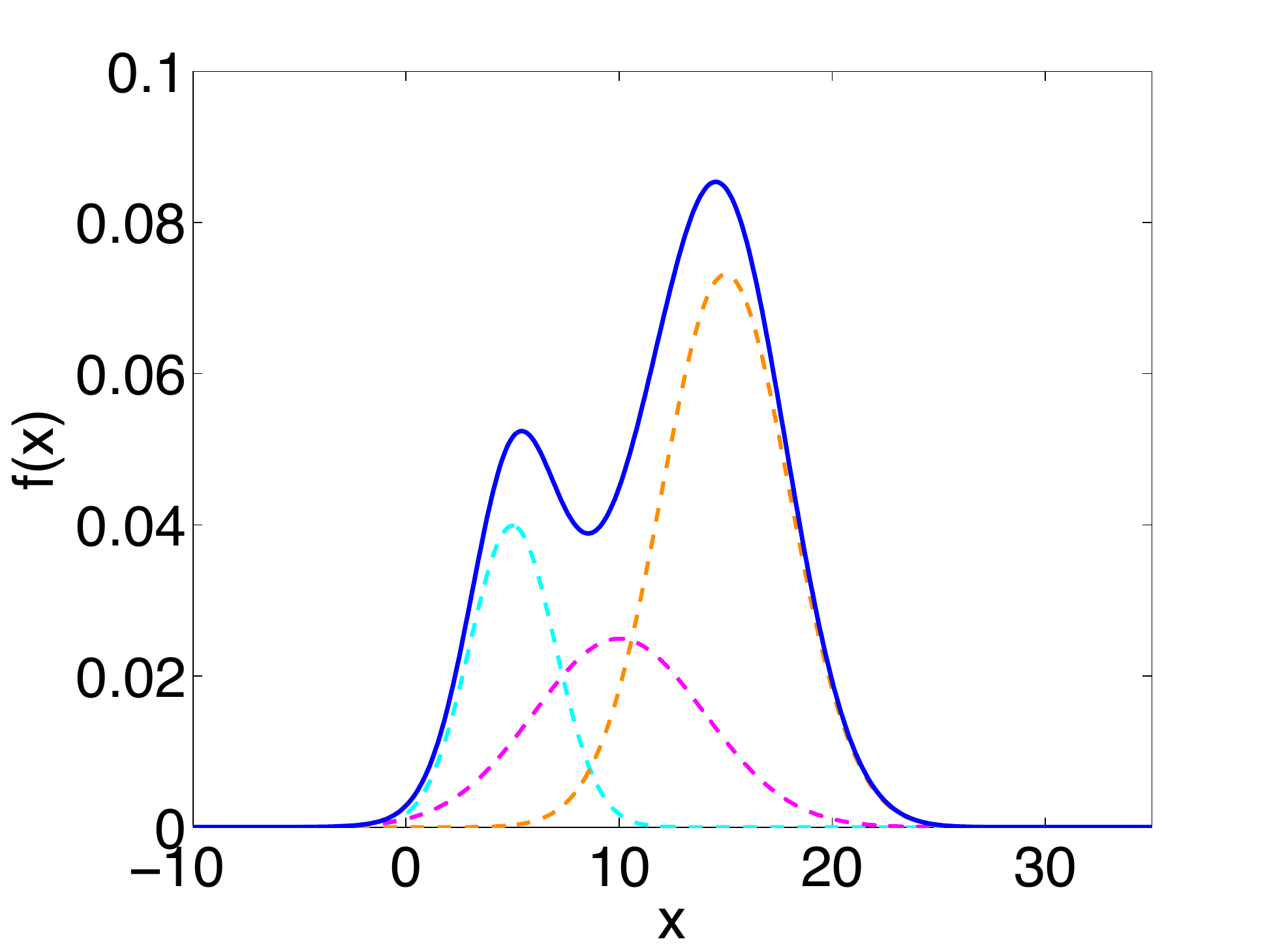}}}
\centerline{\subfloat[surrogate function]{\includegraphics[height=2.1in]{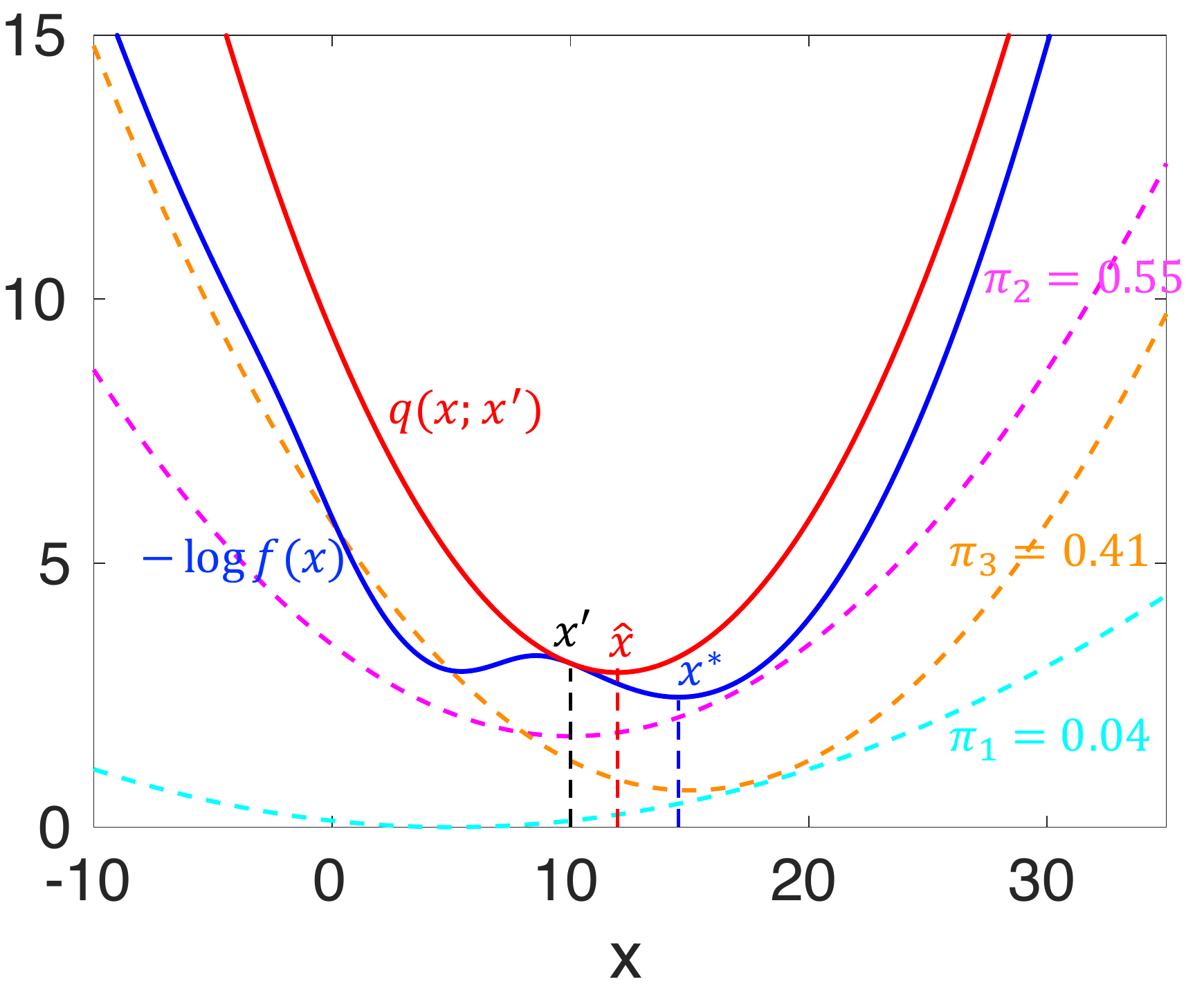}}}
\caption{Figure illustrates the lemma with a 1-D GM distribution.
The quadratic function $q(x;x')$ is a surrogate function for the negative log of the GM distribution $f(x)$ at point $x'$.
The surrogate function is a weighted sum of the quadratic exponents of the exponential functions in the GM distribution.
The weights $\pi_1, \pi_2, \pi_3$ give the posterior probabilities of the point $x'$ belonging to different GM components.}
\label{fig:surrogate}
\end{figure}

In order to construct a surrogate function for our problem, we introduce the following lemma that is proved in Appendix~A.
The lemma provides a surrogate function for a general class of functions formed by the log of a sum of exponential functions.
Since the potential function of (\ref{eq:logGMM}) has this form, we can use this lemma to construct a surrogate function for our MAP estimation problem.
Fig.~\ref{fig:surrogate} illustrates the usage of the lemma for a particular case of Gaussian mixture distribution.
We note that the lemma generalizes the approaches of\cite{wang09tns, mehranian15tmi}, 
since it works for any exponential mixture rather than just mixtures of Gaussians.
\\

\noindent \textit{Lemma (surrogate functions for logs of exponential mixtures):}
Let $f:\Re^N \rightarrow \Re$ be a function of the form
\begin{equation}
f(x)=\sum_k w_k \exp\{ -v_k(x) \} \ ,
\end{equation}
where $w_k \in \Re^+$, $\sum_k w_k >0$, and $v_k:\Re^N \rightarrow \Re$.
Furthermore $\forall (x, x') \in \Re^N\times \Re^N$ define the function
\begin{equation}
q(x;x') \triangleq -  \log f(x') + \sum_k \tilde{\pi}_k (v_k(x)-v_k(x')) \ ,
\end{equation}
where $\tilde{\pi}_k = \frac{w_k \exp \{-v_k(x') \}}{\sum_l w_l \exp\{-v_l(x') \}}$.
Then $q(x;x')$ is a surrogate function for $-\log f(x)$, and $\forall (x, x') \in \Re^N\times \Re^N$,
\begin{eqnarray}
q(x';x') &=& - \log f(x') \\
q(x;x') &\geq& - \log f(x) 
\end{eqnarray}
Proof: see Appendix A. \\

Since the function $u(x)$ specified by (\ref{eq:energy}) and (\ref{eq:logGMM}) has the same form as assumed by the lemma, 
we can use this lemma to find a surrogate function with the following form
\begin{equation}
\label{eq:surrogate_reg}
u(x;x') = \frac1{2L} \sum_{s\in S} \sum_{k=1}^K \tilde{w}_{s,k} \|P_s x - \mu_k\|^2_{R_k^{-1}} + c(x') \ ,
\end{equation}
where $x'$ is the current state of the image, 
$c(x')$ only depends on the current state, and the weights $\tilde{w}_{s,k}$ are given by
\begin{equation}
\label{eq:weight_reg}
\tilde{w}_{s,k} = \frac{\displaystyle \pi_k |R_k|^{-\frac12} \exp \left\{ - \frac12 \|P_s x' - \mu_k\|^2_{R_k^{-1}}  \right\}}
{\displaystyle \sum_{l=1}^K \pi_l |R_l|^{-\frac12} \exp \left\{ - \frac12 \|P_s x' - \mu_l\|^2_{R_l^{-1}}  \right\}} \ .
\end{equation}
Note that the weights $\tilde{w}_{s,k}$ are only functions of the current image $x'$.
Therefore, the optimization in (\ref{eq:map}) can be implemented as a sequence of optimizations as
\begin{align}
\label{eq:cost_surrogate}
\mbox{repeat} \{ \quad  \hat{x} &\leftarrow \arg \min_x \left\{  \frac12  \|y -Ax\|^2_D + u(x;x')\right\} \\
x' &\leftarrow \hat{x} \quad \} \ , \nonumber
\end{align}
with $u(x;x')$ being a quadratic prior that adapts to the current image at each iteration.

Importantly, the weights in (\ref{eq:weight_reg}) represent a soft classification of the current patch into GM components. 
This differs from existing approaches in which each patch is classified to be from a single component of the mixture\cite{Zoran11, Yu12}. 
The previous methods\cite{Zoran11, Yu12} performed discrete optimizations over the Gaussian mixture components to select one single component for each image patch. 
Thus, there is no fixed prior used to form a single consistent MAP estimate of the image. 
Instead, the prior model is iteratively adapted through the choice of a discrete class for each patch. 
	
We believe that this discrete classification approach may cause inaccurate results or artifacts since it requires hard classification of patches into classes even when the class membership of a patch is ambiguous. 
In contrast, our proposed method uses a single consistent and spatially homogeneous non-Gaussian prior model of the entire image. 
A local minimum of the true posterior distribution is then computed by minimizing a series of convex surrogate functions.

Fig.~\ref{fig:surrogate}(b) illustrates a simple 1-D example that demonstrates 
the benefit of the proposed method over previous methods when the distributions of mixture components heavily overlap. 
More precisely, the negative log of the underlying distribution, $Ð\log f(x)$, has a global minimum at $x^* =14.54$. 
For a current point $x^\prime=10$, the previous method will select model 2 (magenta) 
since it produces the greatest posterior probability ($\pi_2 = 0.55$) among the three models. 
However, optimizing the resultant quadratic function (magenta) will give a stationary point at $x=10$, 
which deviates from the global minimum $x^*$ and cannot be improved even with multiple iterations. 
In contrast, the proposed method constructs a quadratic surrogate function (red), $q(x;x^\prime)$, 
of which the optimization will give $\hat{x}=11.97$, 
which moves closer to the global minimum $x^*$ as compared to the current point $x^\prime$. 
Thus, by repeatedly constructing surrogate functions with a new $\hat{x}$ estimated from the previous iteration, 
it will asymptotically reach the global minimum $x^*$ for this particular example.

\subsection{Optimization}

We use the iterative coordinate descent (ICD) algorithm\cite{Sauer93} to solve this quadratic minimization problem in (\ref{eq:cost_surrogate}).
The ICD algorithm sequentially updates each of the pixels by solving a \mbox{1-D} optimization problem, as
\begin{equation}
\label{eq:icdmap}
\hat{x}_j \leftarrow \arg \min_{x_j} \left\{   \frac12  \|y -Ax' + A_{*j} (x'_j - x_j)\|^2_D + u(x_j;x') \right\} \ ,
\end{equation}
with the surrogate prior for $x_j$, as
\begin{equation}
u(x_j;x') = \frac1{2L} \sum_{r\in \mathcal{S}_j} \sum_{k=1}^K  \tilde{w}_{r,k} \|P_rx - \mu_k\|^2_{R_k^{-1}} + c(x') \ ,
\end{equation}
where the weights $\tilde{w}_{r,k}$ are given by (\ref{eq:weight_reg}) and $S_j$ represents a set of center pixels whose patches contain pixel $j$.

By rearranging the terms, we can explicitly write (\ref{eq:icdmap}) as a quadratic function of $x_j$, as
\begin{equation}
\label{eq:quadmap}
\hat{x}_j \leftarrow \arg \min_{x_j} \left\{ (\theta_1 + \varphi_1) x_j + \frac{\theta_2 + \varphi_2}2 (x_j - x'_j)^2 + c(x') \right\} \ ,
\end{equation}
where $c(x')$ is constant to $x_j$ and $\theta_1, \theta_2, \varphi_1, \varphi_2$ are given by
\begin{align}
\theta_1 &= A_{*j}^t D (Ax'-y) \ ,  \\
\theta_2 &= A_{*j}^t D A_{*j} \ , \\
\varphi_1 &=\frac1{L} \sum_{r\in S_j} \sum_k  \tilde{w}_{r,k} (P_r \delta_j)^t R_k^{-1} (P_r x' - \mu_k) \ , \\
\varphi_2 &= \frac1{L} \sum_{r\in S_j} \sum_k  \tilde{w}_{r,k} (P_r \delta_j)^t R_k^{-1} (P_r \delta_j) \ , 
\end{align}
where the calculation of the projection matrix $A$ follows the same procedure in\cite{Thibault07}. 
The function $\delta_j \in \Re^{|S|}$ is a Kronecker delta function, which is a vector with a value of 1 at entry $j$ and with 0 elsewhere.
Therefore, $P_r \delta_j$ is simply an operator that extracts a particular column from a matrix corresponding to the location of the pixel $j$ within the patch operator $P_r$.

Solving (\ref{eq:quadmap}) by rooting the gradient, we then have
\begin{equation}
\hat{x}_j \leftarrow x'_j - \frac{ \theta_1 + \varphi_1}{\theta_2 + \varphi_2} \ .
\end{equation}

\section{Covariance Control for GM-MRF}
\label{sec:regularization}

We will see that the GM-MRF distribution can be used to form a very accurate model of images. 
However, in applications such as CT reconstruction, the MAP estimate may not be visually appealing even with an accurate forward and prior model. 
This is because the MAP estimate with non-Gaussian priors tends to produce a reconstruction that is under-regularized (i.e., too sharp) in high-contrast regions 
and over-regularized (i.e., too smooth) in low-contrast regions\cite{fessler96tip, evans11mp, li14mp}. 
While this variation in spatial resolution may produce a lower mean squared error (MSE), in particular applications it may not be visually appealing. 

In order to address this problem of spatial variation in sharpness, 
in this section we introduce a simple parameterization for systematically controlling the covariance of each GM component of the GM-MRF model. 
In the experimental results section, we will then demonstrate that this simple parameterization can be used to effectively tune the visual quality of the MAP reconstruction. 
In real applications such as medical CT reconstruction, this covariance adjustment can be used to effectively fine-tune the rendering of specific tissue types,
such as soft tissue, lung, and bone, which may have different desired characteristics.

We start by introducing regularization parameters, $\sigma_x$ and $\{ \sigma_k \}_{k=1}^K$, into the distribution given by
\begin{equation}
\label{eq:energy_reg}
u_{\sigma}(x) = -\frac1{L \sigma_x^2} \sum_{s\in S} \log\{ g_{\sigma}(P_s x)\} \ ,
\end{equation}
with the patch Gaussian mixture distribution
\begin{equation}
\label{eq:GMM_reg}
g_{\sigma}(P_s x) =  \sum_{k=1}^K \frac{\pi_k |R_k / \sigma_k^2 |^{-\frac12}}{(2\pi)^{\frac{L}2}}  \exp \left\{ - \frac{\sigma_k^2 \|P_s x - \mu_k\|^2_{ R_k^{-1}}}2\right\} \ .
\end{equation}
Notice that $\sigma_x$ controls the overall level of regularization and that the $K$ values of $\sigma_k$ control the regularization of each individual component of the GMM.
When the value of $\sigma_x$ is increased, the overall reconstruction is made less regularized (i.e., sharper)
and when the value of $\sigma_k$ is increased, the individual GM component is made more regularized.

\begin{figure}[!t]
\centerline{
\includegraphics[height=2.1in]{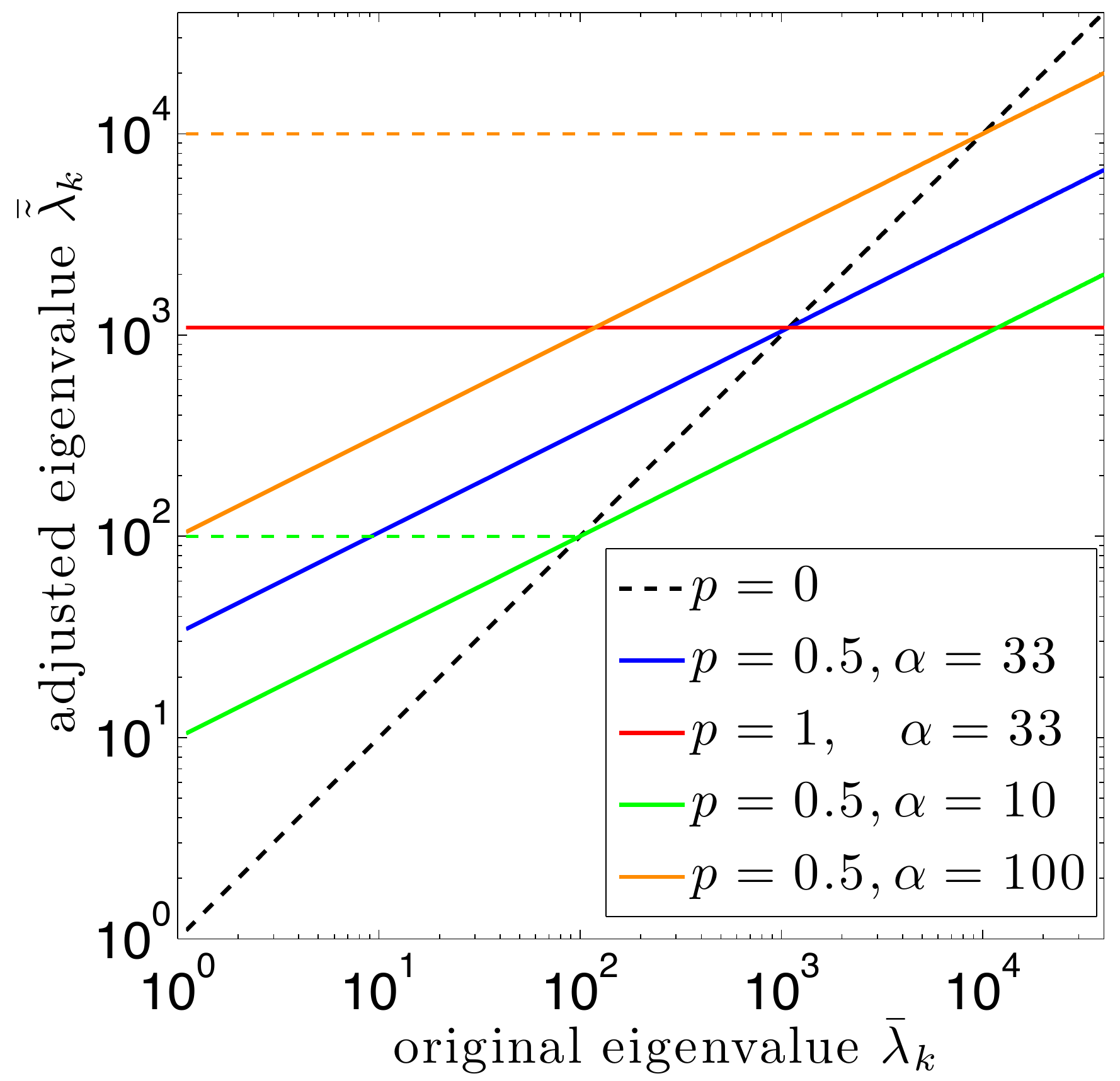}}
\caption{The covariance scaling defined in (\ref{eq:sigmak}) and (\ref{eq:lambda_scaled}) with various values of $p$ and $\alpha$ on a log scale.
The black dotted line shows the case when no scaling is present, i.e., $p=0$.
When $0 < p \leq 1$, the average eigenvalues $\bar{\tilde{\lambda}}_k$ are ``compressed" toward $\alpha^2$,
where eigenvalues further away from $\alpha^2$ lead to greater change and a larger $p$ results in greater compression. 
For a fixed value of $p$, increasing $\alpha$ increases the covariance of each GM component.
}
\label{fig:covariance_scaling}
\end{figure}

\begin{figure*}[!t]
\centerline{
\subfloat[]{\includegraphics[height=1.5in]{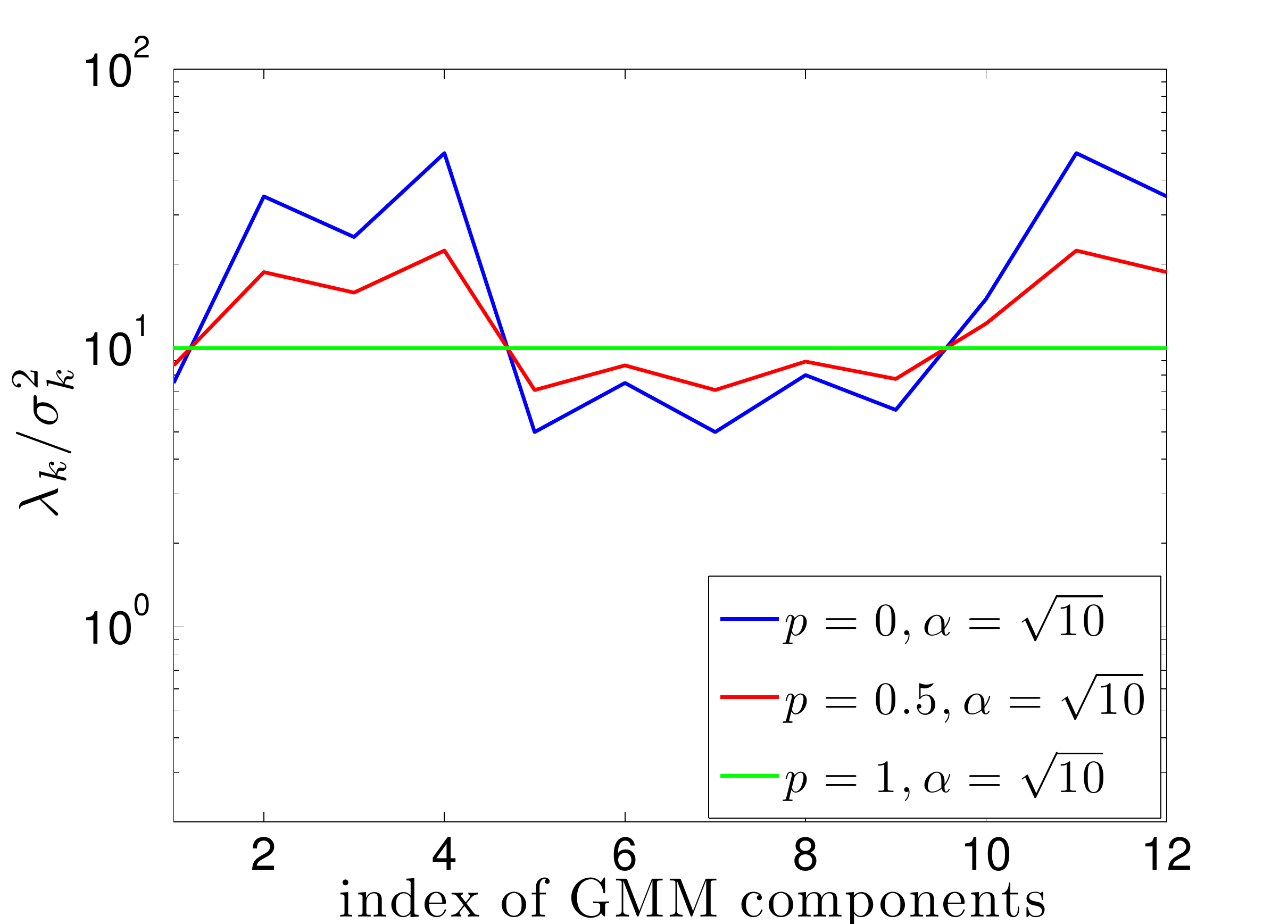}}\ \
\subfloat[]{\includegraphics[height=1.5in]{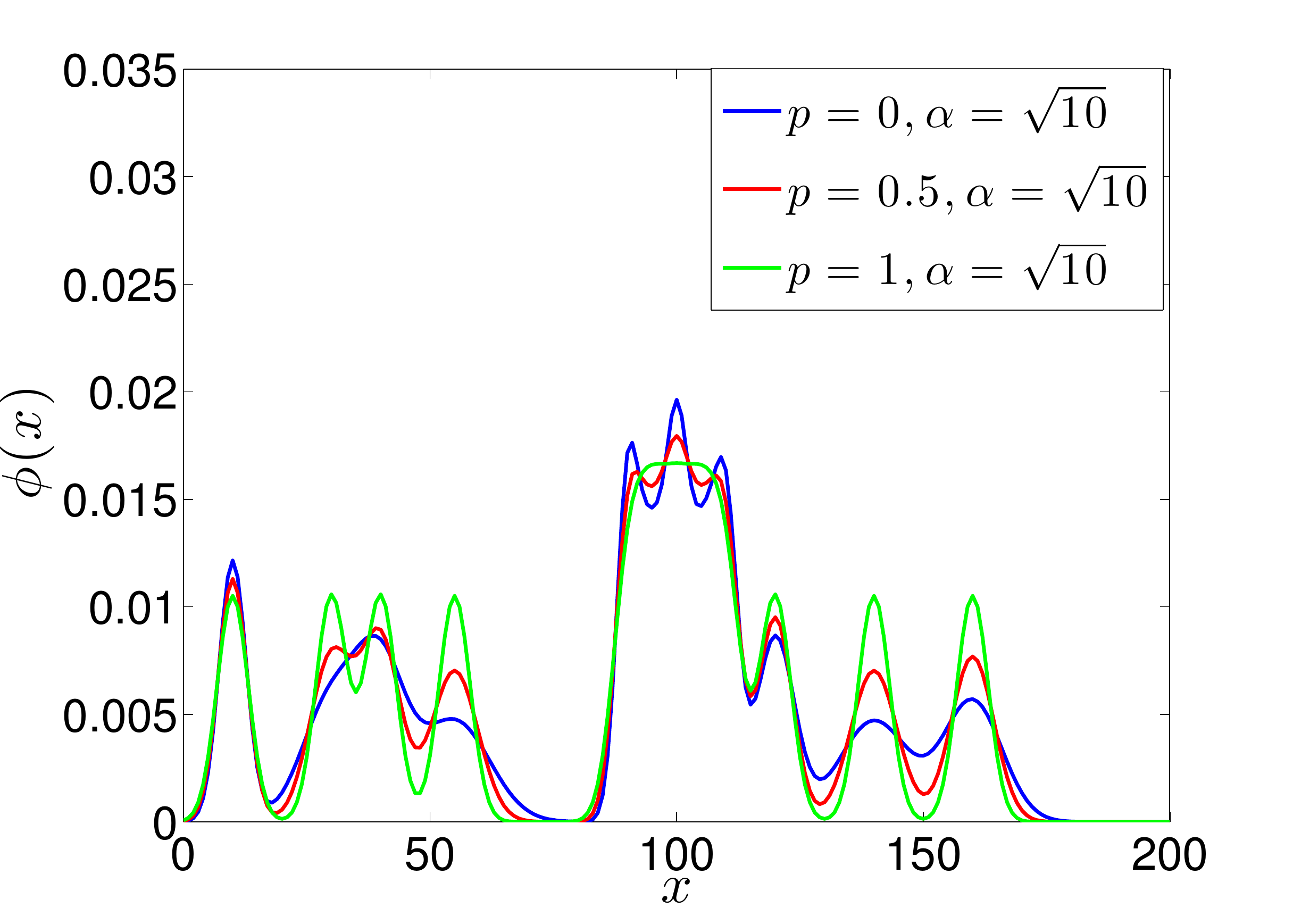}}\
\subfloat[]{\includegraphics[height=1.5in]{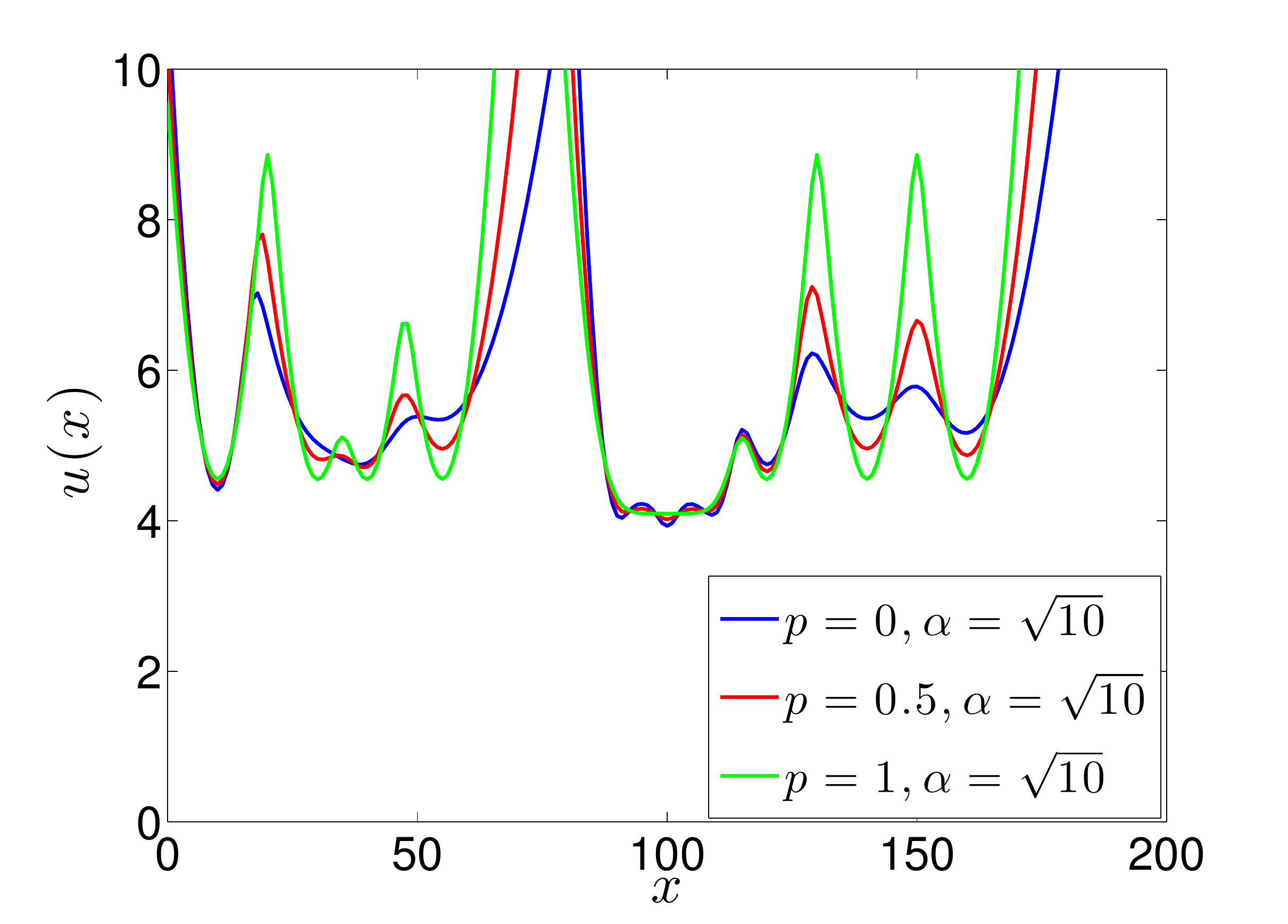}}}
\centerline{
\subfloat[]{\includegraphics[height=1.5in]{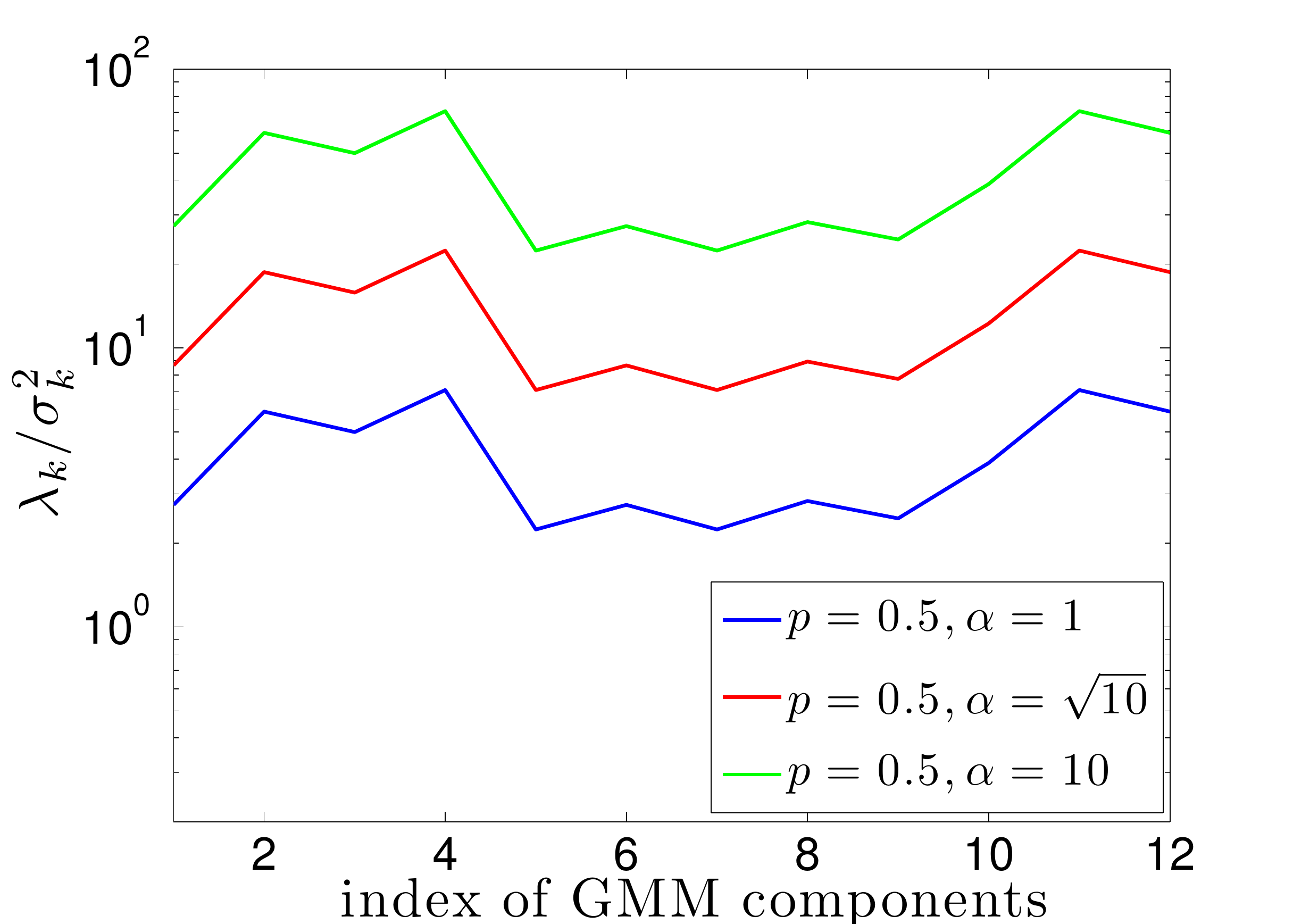}}\ \
\subfloat[]{\includegraphics[height=1.5in]{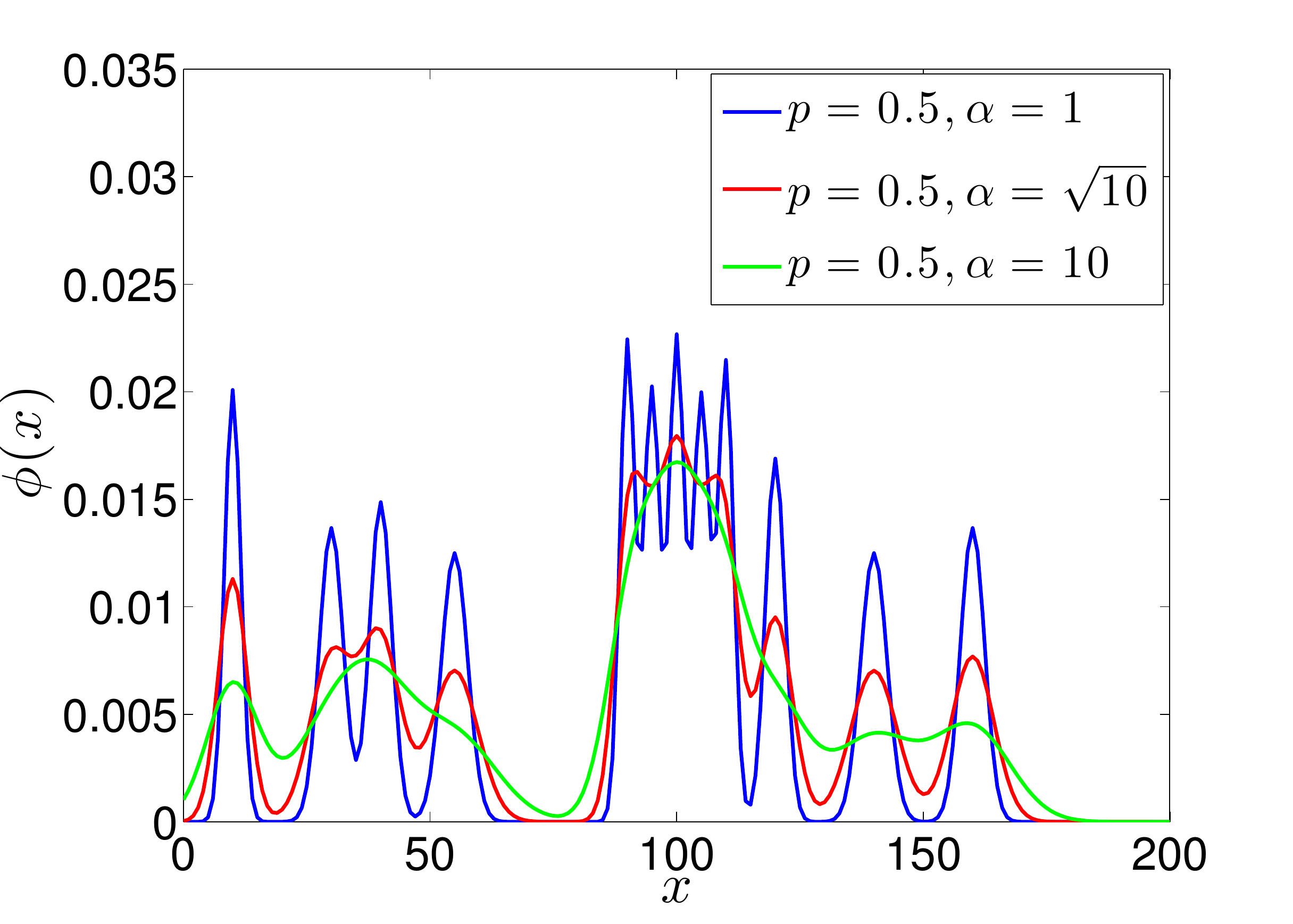}}\
\subfloat[]{\includegraphics[height=1.5in]{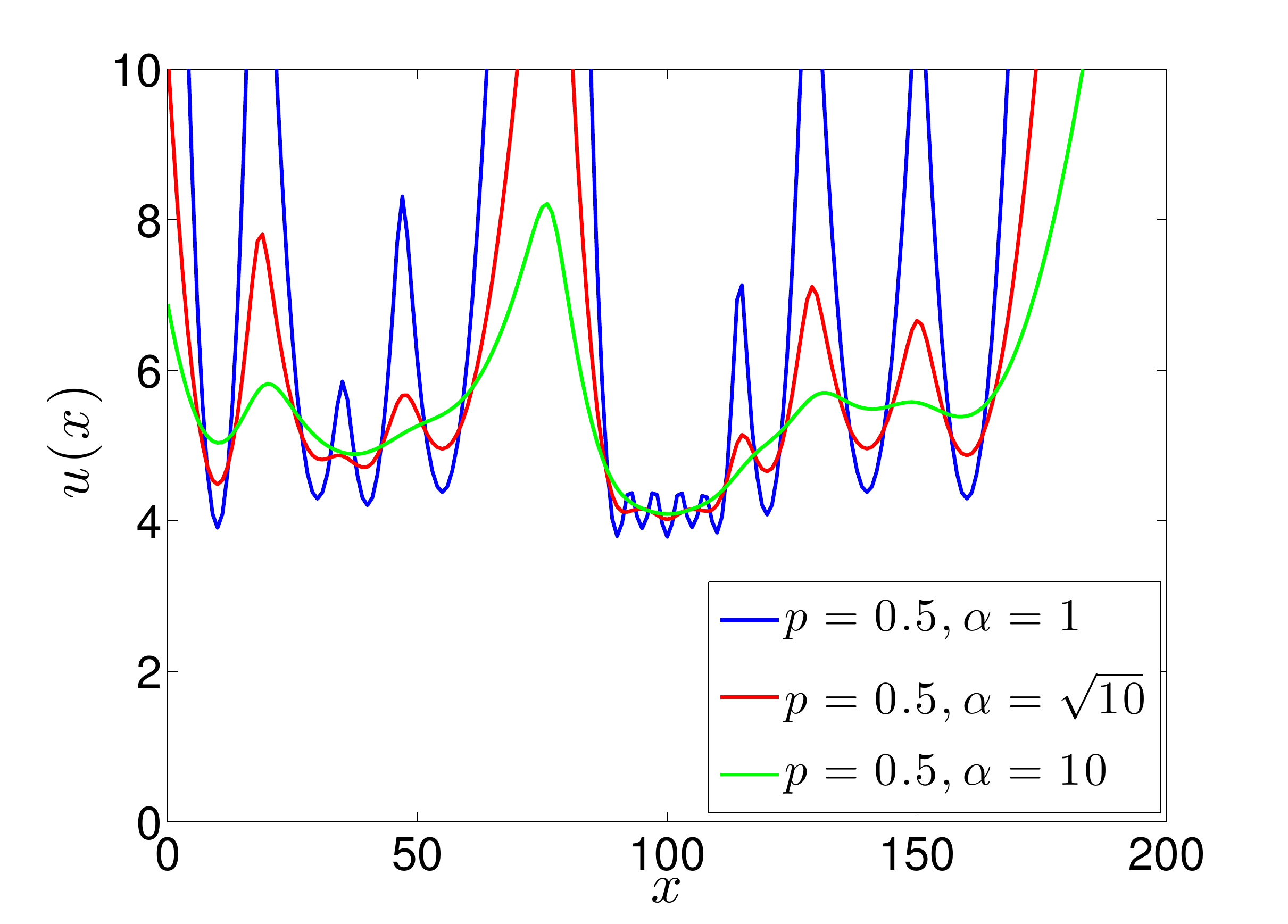}}}
\caption{1-D illustration of the covariance scaling in (\ref{eq:sigmak}).
The \mbox{1-D} energy function is given by $u(x) = -\log(\phi(x))$ with the Gaussian mixture distribution 
$\phi(x)~=~\sum_k \frac1{\sqrt{2\pi \lambda_k/ \sigma_k^2}} \exp \left\{ -\frac1{2 \lambda_k / \sigma_k^2} (x-\mu_k)^2\right\}$,
with $\lambda_k$ the original variance and the scaling $\sigma_k = (\lambda_k / \alpha^2)^{p/2}$.
(a) varying $p$ with $\alpha$ fixed, (b) the resulting distribution, and (c) the resulting energy function; 
(d) varying $\alpha$ with $p$ fixed, (e) the resulting distribution, and (f) the resulting energy function.}
\label{fig:sigmak}
\end{figure*}

Now for a typical GM-MRF model there may be many components, 
so this would require the choice of many values of $\sigma_k$.
Therefore, we introduce a simple method to specify these $K$ parameters using the following equation,
\begin{equation}
\label{eq:sigmak}
\sigma_k =  \left( \bar{\lambda}_k / \alpha^2\right)^{p/2} \ ,
\end{equation}
where $p$ and $\alpha$ are two user-selectable parameters such that $0\leq p \leq 1$, $\alpha>0$,
and $\bar{\lambda}_k = |R_k|^{\frac1L}$ is the geometric average of the eigenvalues of $R_k$.
Define $\tilde{R}_k = R_k / \sigma_k^2$ as the covariance matrix after scaling. 
Then its corresponding average eigenvalue is given by
\begin{equation}
\label{eq:lambda_scaled}
\bar{\tilde{\lambda}}_k =\alpha^{2p} \bar{\lambda}_k^{1-p} \ .
\end{equation}
Fig.~\ref{fig:covariance_scaling} illustrates this scaling with various values of $p$ and $\alpha$.

In this model, the parameters $p$ and $\alpha$ collectively compress the dynamic range of the average eigenvalues $\bar{\lambda}_k$ of all GM covariance matrices.
That is, for those GM components with large average eigenvalues of covariance, which typically correspond to high-contrast or structural regions,
applying the scaling in (\ref{eq:sigmak}) decreases the eigenvalues, which leads to increased regularization.
Conversely, for those GM components with small average eigenvalues of covariance, which are typically associated with low-contrast or homogeneous regions,
applying the scaling increases the eigenvalues and subsequently results in reduced regularization. 

More specifically, $p$ is the compression rate with a larger value resulting in greater compression of the dynamic range,
and $\alpha$ defines a stationary point during the compression, i.e.,
\begin{equation}
\bar{\tilde{\lambda}}_k = \bar{\lambda}_k \ , \quad {\rm if }\ \ \bar{\lambda}_k = \alpha^2  \ .
\end{equation}
When $0 < p < 1$, the average eigenvalues $\bar{\tilde{\lambda}}_k$ are ``compressed" toward $\alpha^2$,
with eigenvalues further away from $\alpha^2$ leading to greater change. 
When $p=1$, all GM components have the same average eigenvalue $\alpha^2$,
while they maintain the original eigenvalues when $p=0$.  
Fig.~\ref{fig:sigmak}(a)-(c) illustrate the change in the distribution and the energy function as $p$ varies.

In addition, the parameter $\alpha$ controls the ``smoothness" of the GM distribution.
With $p$ fixed, increasing the value of $\alpha$ leads to a smoother distribution of (\ref{eq:GMM_reg}), 
which potentially reduces the degree of non-convexity of the energy function in (\ref{eq:energy_reg}).
Moreover, an increased $\alpha$ also reduces the overall regularization.
Fig.~\ref{fig:sigmak}(d)-(f) illustrate the change in the distribution and the energy function as $\alpha$ varies.

Table~\ref{tab:param} presents the selection of the regularization parameters and the corresponding effect.
Note that the parameter $\alpha$ is related to the reconstruction noise and therefore has the same unit as the reconstruction.
For instance, in X-ray CT reconstruction, the parameter $\alpha$ is in Hounsfield Unit (HU).

\begin{table}[!t]
\renewcommand{\arraystretch}{1.3}
\caption{Parameter selection for GM-MRF model.}
\label{tab:param}
\centering
\begin{tabular}{|C{.45in}|C{.45in}|C{2.2in}|} \hline
Parameter & Selection & Effect \\ \hline
\multirow{3}{*}{$\sigma_x$} & $\approx 1$ & MAP estimate \\   \cline{2-3}
 & $>> 1$ &  large value $\rightarrow$ weak overall regularization \\ \cline{2-3}
 & $<< 1$ & small value $\rightarrow$ heavy overall regularization \\ \hline
 \multirow{5}{*}{$p$} & $0$ &  unmodified GM-MRF \\   \cline{2-3}
 & \multirow{3}{*}{0.5} & regularization strength increases for GM \\ 
 & & components with large average eigenvalues;  \\
 & & reduces for those with small average eigenvalues \\ \cline{2-3}
 & \multirow{2}{*}{1} & same regularization strength \\ 
 & & for all GM components \\ \hline
 \multirow{4}{*}{$\alpha$} & \multirow{4}{*}{33 (HU)} & large value $\rightarrow$ smooth prior distribution \\   
 & &   $\rightarrow$ weak regularization \\ \cline{3}
 & & small value $\rightarrow$ peaky prior distribution \\ 
 & &  $\rightarrow$ heavy regularization \\ \hline
  \end{tabular}
\end{table} 

\section{Methods}
\label{sec:method}
In this section, we present the datasets used for training and testing in our experiments.
We also provide description for the training procedure in detail.
In addition, we describe the different methods that used for comparison.

\subsection{Training}
\label{sec:training}
We trained the GMM patch distribution, $g(P_s x)$ in (\ref{eq:GMM}), on clinical CT images using the standard EM algorithm with the software in \cite{Bouman97}.
Training data consisted of \mbox{2-D} or \mbox{3-D} overlapping patches extracted from the \mbox{3-D} reconstruction of a normal-dose scan,
acquired with a GE Discovery CT750 HD scanner  
in \mbox{$64\times0.625$ mm} helical mode with \mbox{100 kVp}, \mbox{500 mA}, \mbox{0.8 s/rotation}, \mbox{pitch 0.984:1}, and reconstructed in \mbox{360 mm} field-of-view (FOV).
We have supplied typical images of the training dataset in the supplementary material.

Instead of training one GMM using all the patches, we partitioned the patches into different groups and then trained one GMM from each of the groups.
In this way, we were able to collect sufficiently many samples from underrepresented groups to obtain accurate parameter estimates,
while simultaneously limiting the data size for other groups to retain training efficiency.
For the $i^{\rm th}$ group, we trained the parameters, $\{ \pi_{i,k}, \mu_{i,k}, R_{i,k} \}_k^{K_i}$, for one GMM, $g_i(P_s x)$, with $K_i$ components.
Then we merged all GMMs trained from different groups into a single GMM by weighted summation, 
\begin{equation}
g(P_s x) = \sum_{i=1}^I \pi_i g_i(P_s x),
\end{equation}
where the mixture weights $\pi_i$ were determined by the natural proportions of corresponding groups in the whole training data.

\begin{figure*}[!t]
\centerline{
\subfloat[original]{\includegraphics[width=.9in]{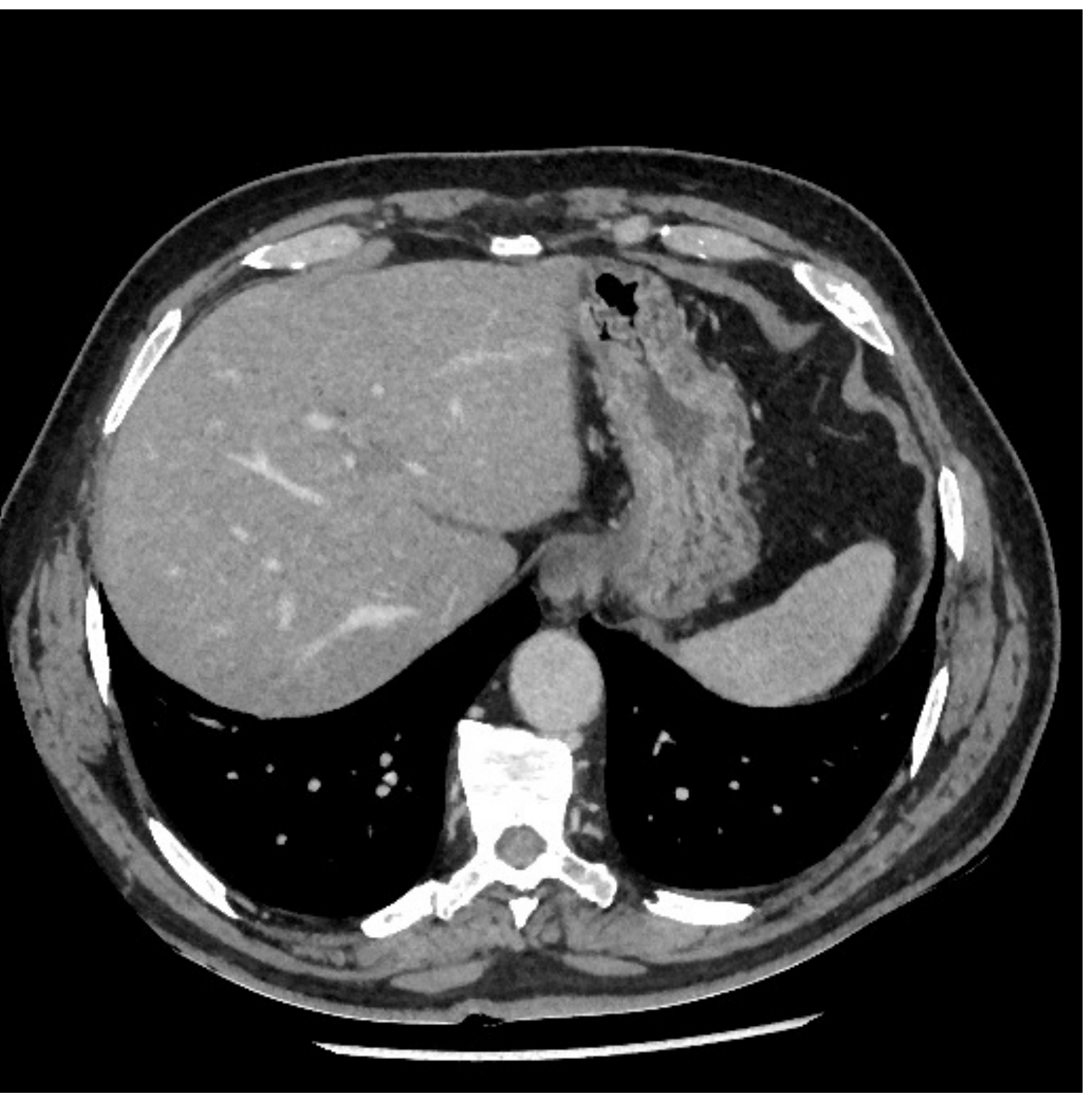}} \qquad
\subfloat[group 1]{\includegraphics[width=.9in]{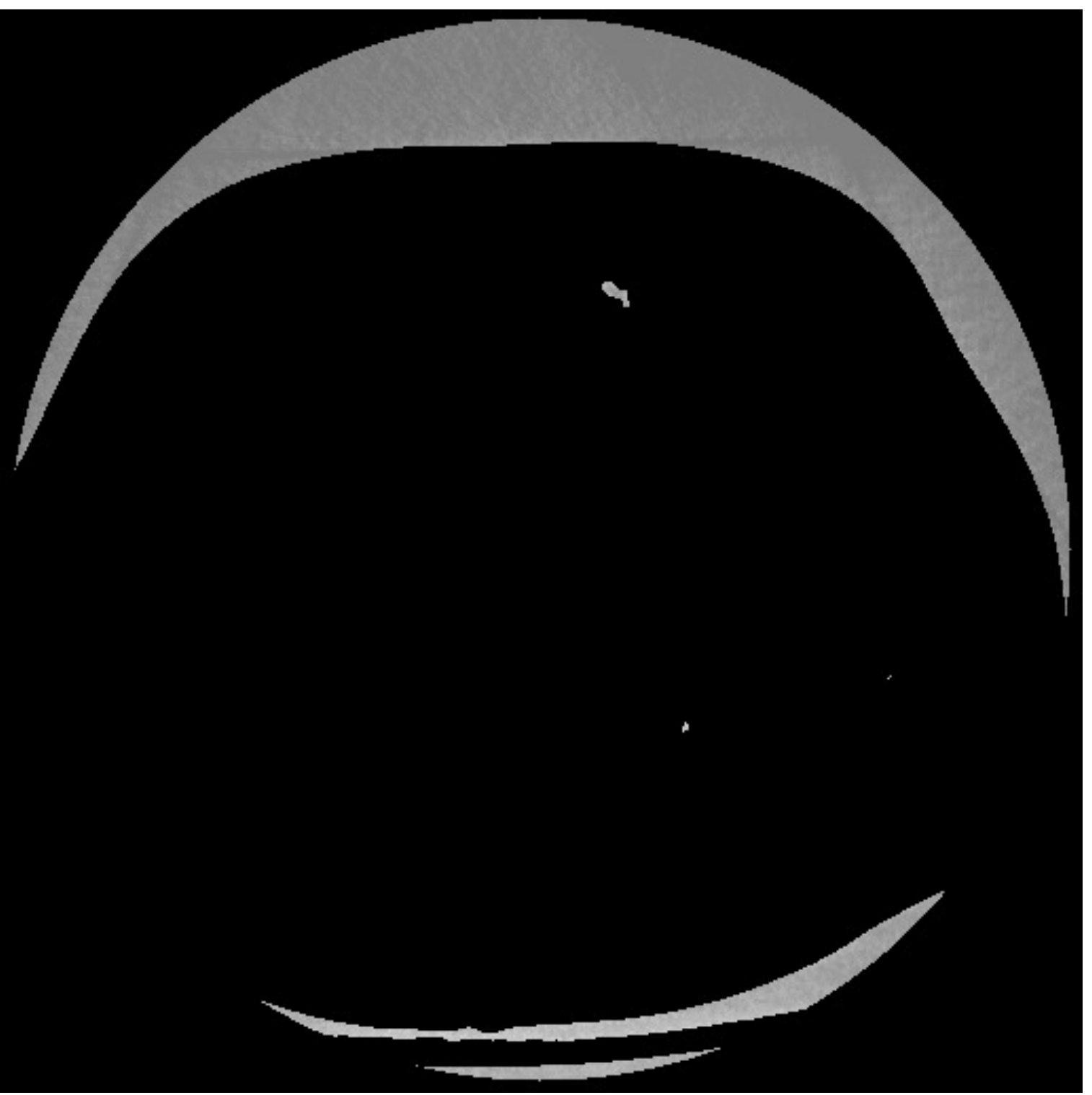}}
\subfloat[group 2]{\includegraphics[width=.9in]{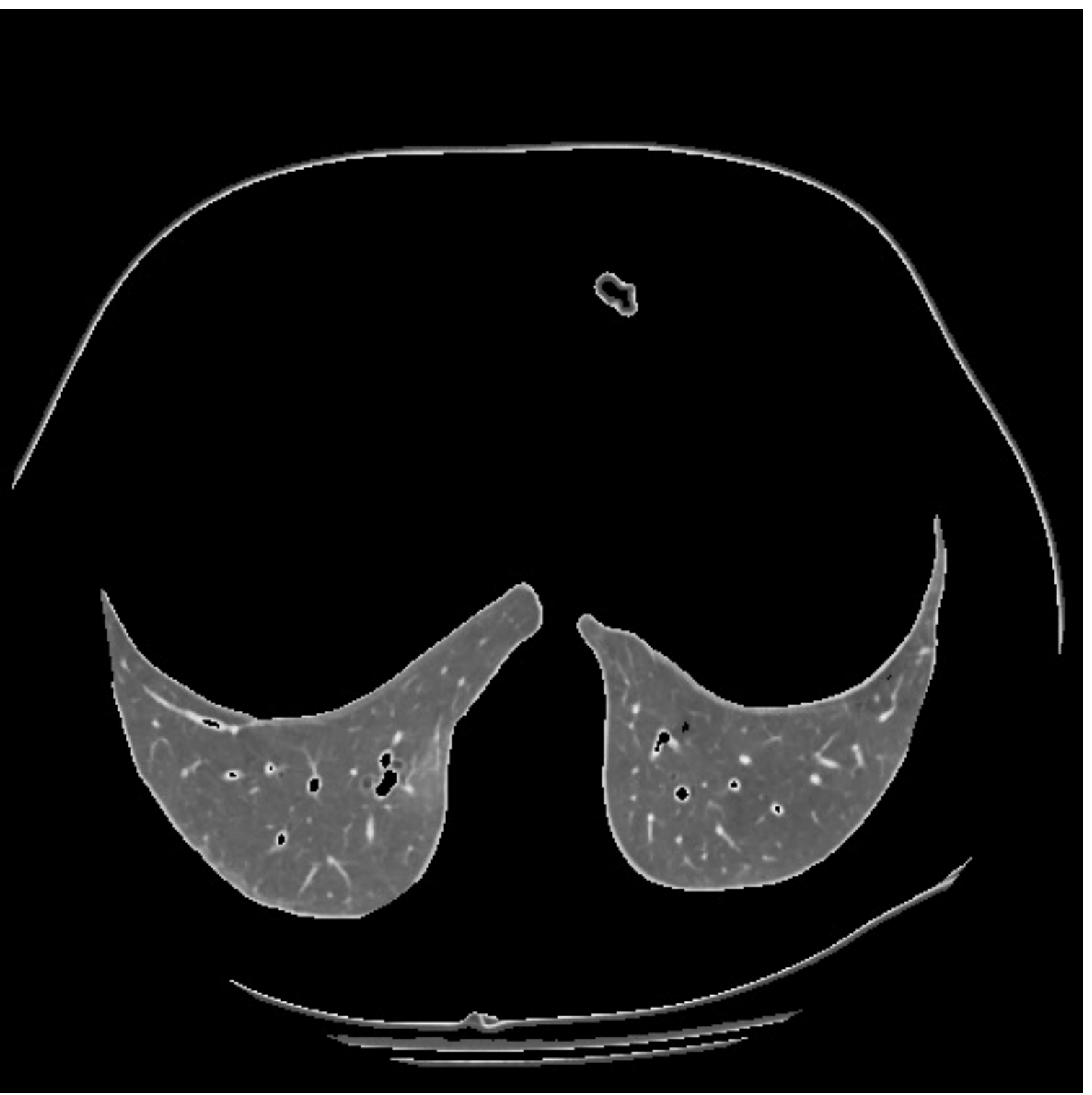}}
\subfloat[group 3]{\includegraphics[width=.9in]{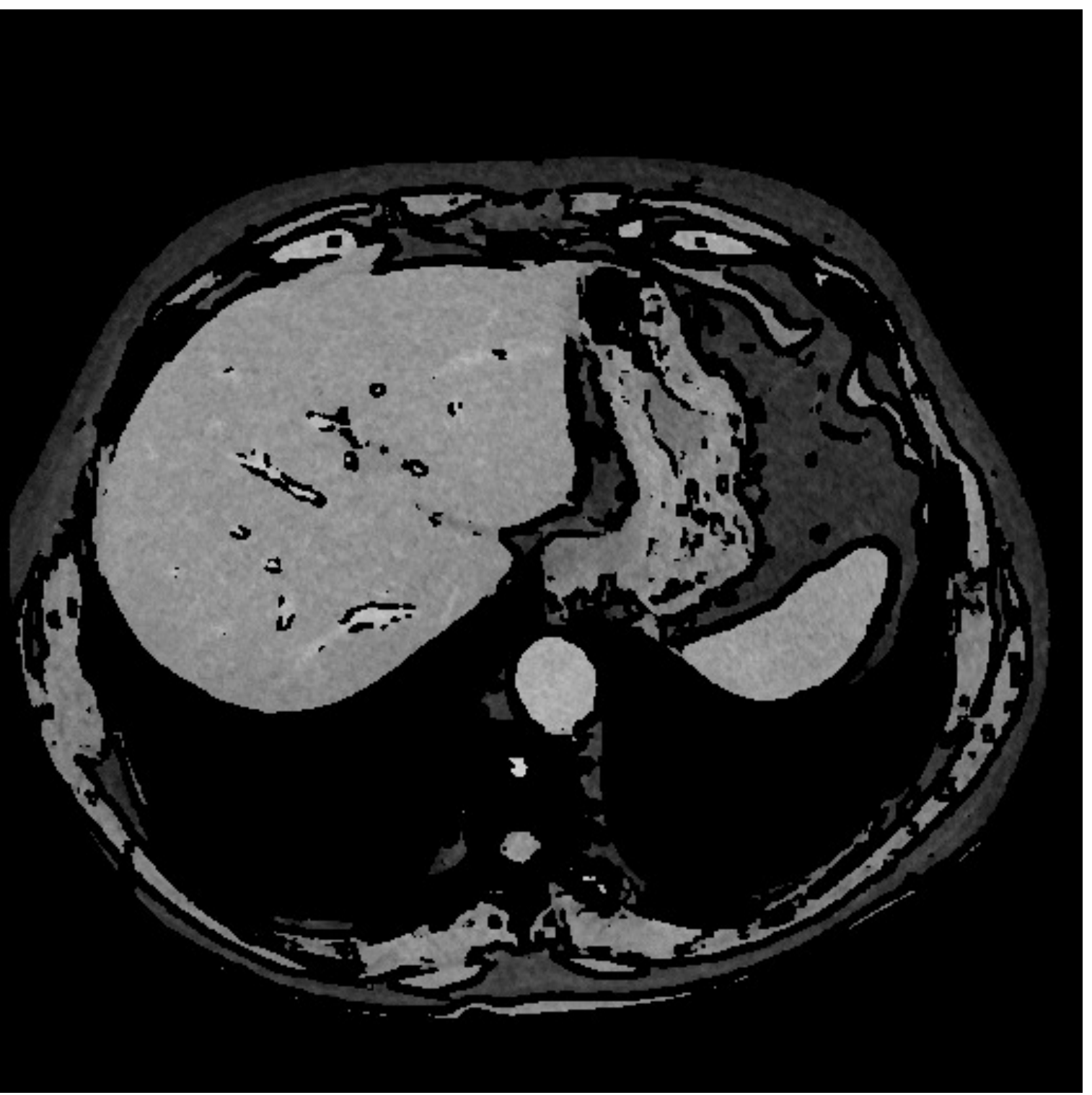}}
\subfloat[group 4]{\includegraphics[width=.9in]{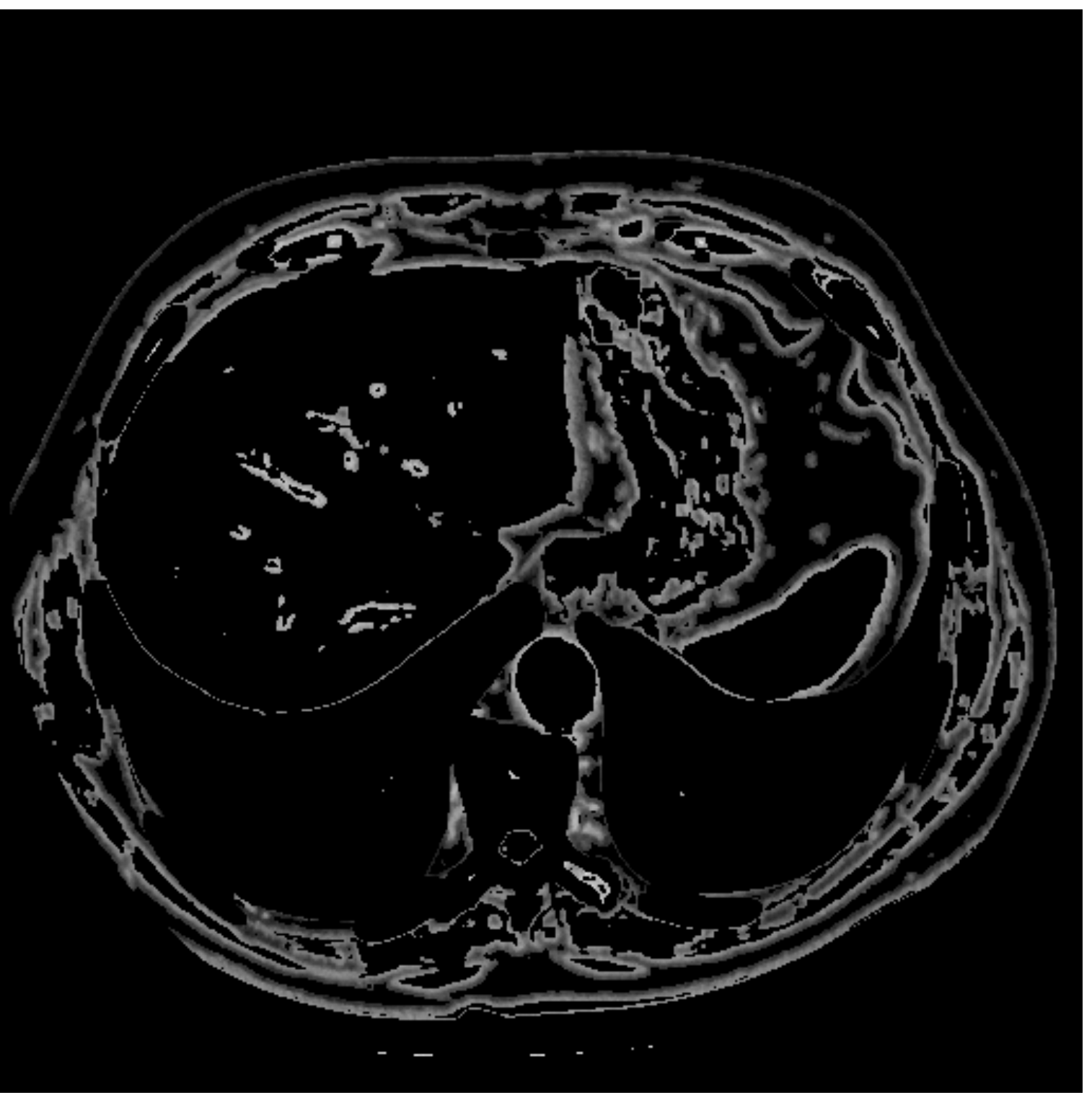}}
\subfloat[group 5]{\includegraphics[width=.9in]{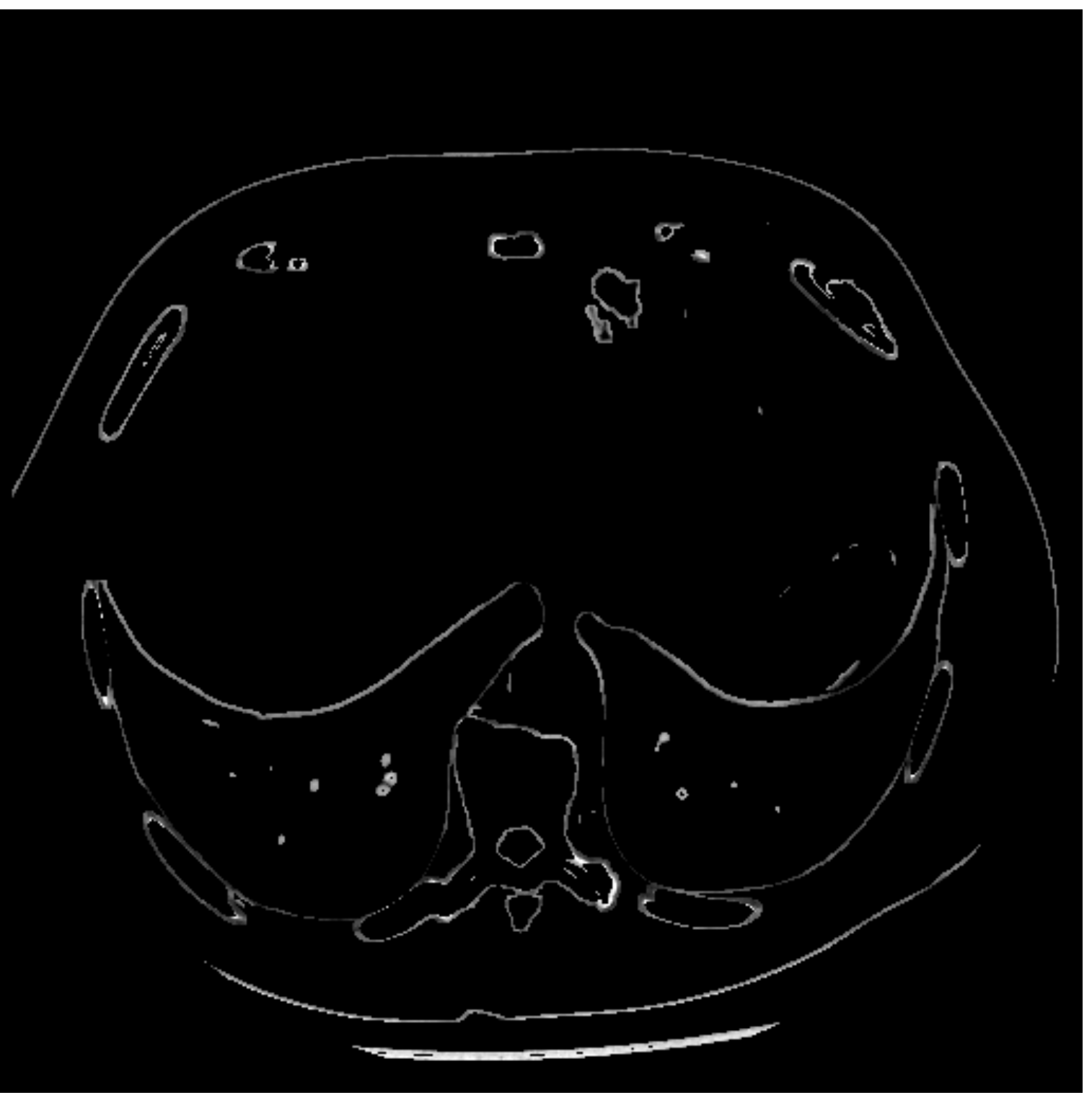}}
\subfloat[group 6]{\includegraphics[width=.9in]{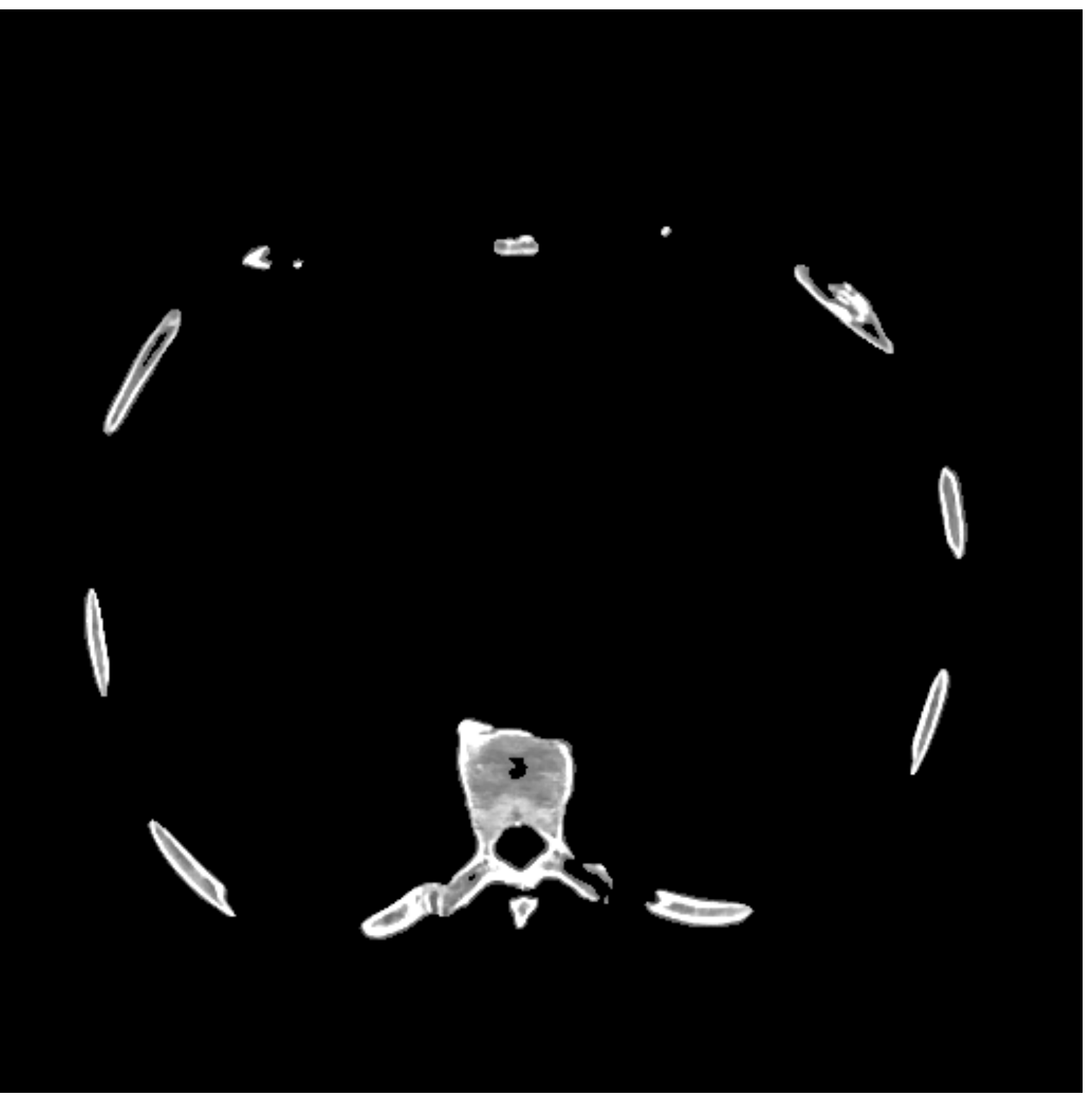}}}
\caption{Partition of the training data. 
(a): a \mbox{2-D} axial slice from the \mbox{3-D} image volume where training patches were extracted;
(b)-(g): partition of the data based on the criteria in Table~\ref{tab:training}.
Display window: (a) \mbox{[-160, 240] HU}, (b) \mbox{[-1250, -750] HU}, (c) \mbox{[-1400, 200] HU}, (d)-(f) \mbox{[-210, 290] HU}, (g) \mbox{[-300, 700] HU}.
Notice that different groups roughly capture different materials or tissue types in the image, 
as group 1 for air, group 2 for lung tissue, group 3 for smooth soft tissue, group 4 for low-contrast soft-tissue edge,
group 5 for high-contrast edge, and group 6 for bone. }
\label{fig:train_group}
\end{figure*}

More specifically, we partitioned the patches into six groups based on the mean and standard deviation as listed in Table~\ref{tab:training},
where the partition thresholds were empirically determined to roughly reflect typical tissue types in a medical CT image.
Fig.~\ref{fig:train_group} illustrates different groups on a \mbox{2-D} image slice.
As shown in Fig.~\ref{fig:train_group}, different groups roughly capture different materials or tissue types in the image,
as group 1 for air, group 2 for lung tissue, group 3 for smooth soft tissue, group 4 for low-contrast soft-tissue edge,
group 5 for high-contrast edge, and group 6 for bone. 
With this partition, we were able to collect adequate patches for individual groups separately,
especially for the underrepresented ones as group 4, 5, and 6.  
During the separate training process, we empirically fixed the number of GMM components, $K_i$, in the EM algorithm for each group. 
Table~\ref{tab:training} also presents the mixture weights $\pi_i$ for different GMMs,
which were determined by the natural proportions of corresponding groups in the whole training data.

We trained three 2-D GM-MRF models consisting of 66 GM components with different patch sizes,
i.e., $3\times3$, $5\times5$, and $7\times7$ patches, for an image denoising experiment.
We also trained a number of 3-D GM-MRF models with different parameter settings to study the influence of model parameters in phantom studies.
More precisely, we fixed the number of GM components to 66 and then trained different GM-MRFs with 
$3\times3\times3$ patches, $5\times5\times3$ patches, and $7\times7\times3$ patches, respectively. 
Similarly, we fixed the patch size to $5\times5\times3$ and then 
trained several GM-MRF models with various number of GM components, namely 6, 15, 31, 66, and 131 components.
Details of the training procedures are provided in the supplementary material. 
Note that the covariance scaling introduced in Sec.~\ref{sec:regularization} is applied to the trained model and hence does not require additional training.

\begin{figure}[!t]
\centerline{\includegraphics[height=2in]{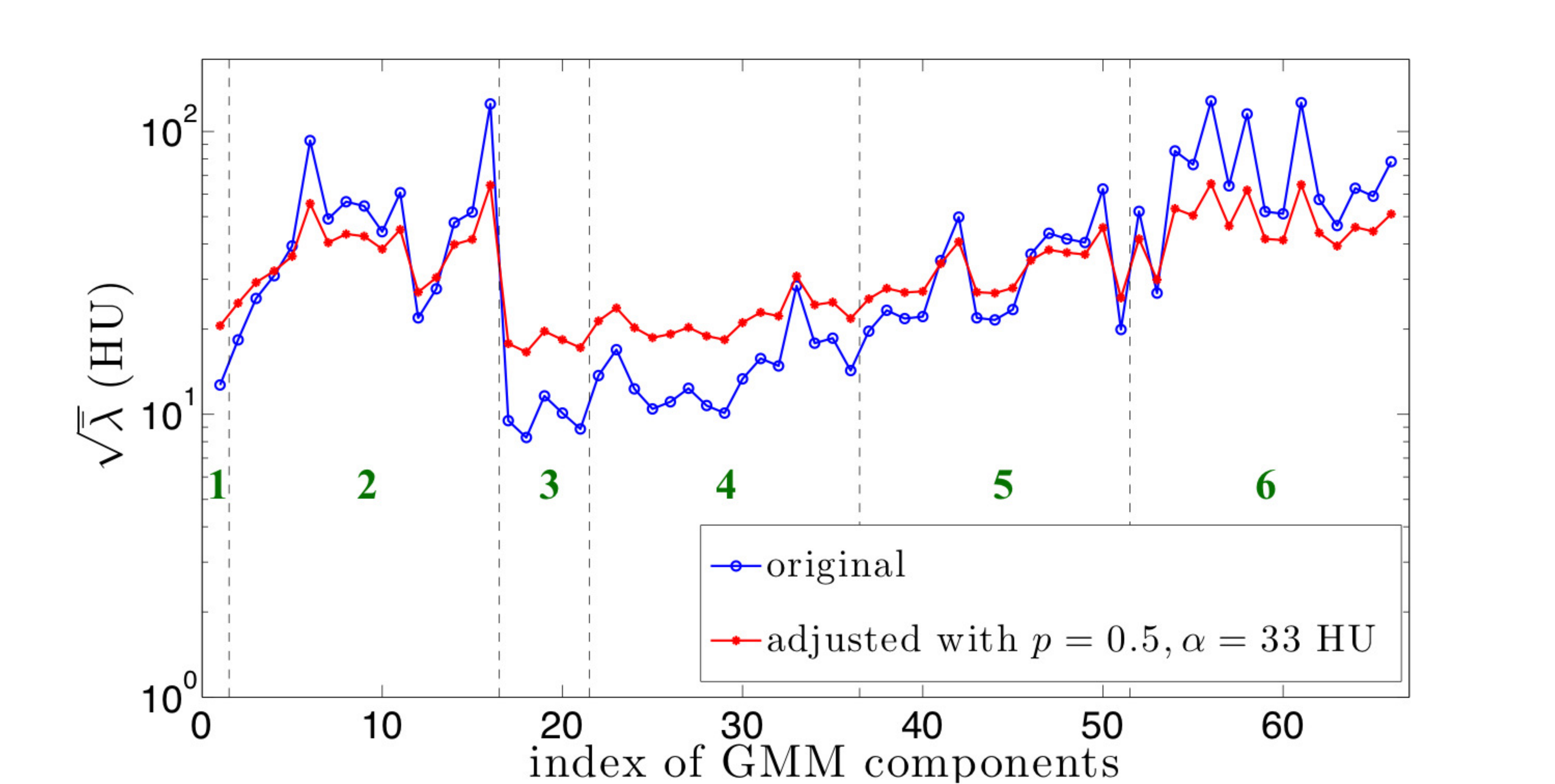}}
\caption{The covariances of the originally trained GMM and the adjusted GMM with $p=0.5$ and $\alpha = 33\ {\rm HU}$ in (\ref{eq:sigmak}), for a $5\times5\times3$ patch case.
More precisely, the figure plots the square root of geometrically-averaged eigenvalues $\bar{\lambda}_k$ of the GMM covariances $R_k$, 
as $\bar{\lambda}_k = |R_k|^{\frac1L}$ with $L=75$.
Numbers within the figure correspond to the indices of the training groups.
Within each group, the GMM components are sorted from the most probable to the least probable.
The adjusted model increases the eigenvalues for group 1, 3, 4, and decreases the eigenvalues for most of group 2, 5, and 6.}
\label{fig:covariance_adjusted}
\end{figure}

The blue plot in Fig. \ref{fig:covariance_adjusted} illustrates the 
square-rooted geometrically-averaged eigenvalues, $\bar{\lambda}_k$, of trained GM covariance matrices for the $5\times5\times3$ GM-MRF with 66 components.
The numbers within the figure correspond to the indices of training groups in Table \ref{tab:training}.
Within each group, the GM components are sorted from the most probable to the least probable based on the trained mixture probabilities, $\pi_{i,k}$.
Fig.~\ref{fig:covariance_adjusted} shows that different groups present different amounts of regularization strength.
Note that there is a large variation in average eigenvalues for different groups, which leads to highly varying regularization strength for different image contents.
For example, group 3 has much smaller average eigenvalues than groups 2 and 6,
which indicates that during the reconstruction, patches dominated by group 3, typically the smooth soft-tissue patches, 
will be regularized more heavily than patches dominated by group 2 and 6, typically lung and bone patches respectively, 
and therefore will contain less noise in the reconstructed image.

As introduced in Sec. \ref{sec:regularization}, we will apply the simple parameterization of (\ref{eq:sigmak}) to the trained GM covariances to tune the visual quality.
The red plot in Fig. \ref{fig:covariance_adjusted} illustrates this adjusted model with $p=0.5$ and $\alpha=33\ {\rm HU}$.
By adjusting the model parameters,
we increase the eigenvalues of group 3 and 4, which will consequently reduce the regularization for smooth and low-contrast soft-tissue contents,
while we decrease the eigenvalues of group 2, 5, and 6, which will lead to stronger regularization for lung, high-contrast edge, and bone.

\begin{table*}[!t]
\renewcommand{\arraystretch}{1.3}
\caption{Partition of the training data. 
Each image patch was classified into one of the six groups based on its mean and standard deviation.
The number of GM components for each group was empirically chosen.
The mixture weights are determined by the proportions of corresponding groups in the whole training data
and will be used when combining different GMMs to form a single model.}
\label{tab:training}
\centering
\begin{tabular}{|C{4cm}|C{1.5cm}|C{1.5cm}|C{1.8cm}|C{2.1cm}|C{2.1cm}|C{1.5cm}|} \hline
Group index, $i$ & 1 & 2 & 3 & 4 & 5 & 6 \\ \hline
Mean (HU) &  [-1000 -850) & [-850 -200) & [-200 200) & [-200 200) & [-200 200) & $\geq$ 200 \\ \hline
Standard deviation (HU) &   &   & [0 25) & [25 80) & $\geq$ 80 &   \\ \hline
Number of patch samples & $5\times10^3$ & $1\times10^5$ & $5\times10^4$ & $1\times10^5$ & $1\times10^5$ & $1\times10^5$ \\ \hline
Number of GM components, $K_i$ &  1 & 15 & 5 & 15 & 15 & 15 \\ \hline
Mixture weight, $\pi_i$ &  0.05 &  0.17 & 0.40 & 0.25 & 0.04 & 0.09  \\ \hline
\end{tabular}
\end{table*}

\subsection{Testing}

We tested the proposed GM-MRF as a prior for model-based inversion problems.
The testing data included different scans on the same and different patients as compared to the training data. 
All the data were acquired using GE Discovery CT750 HD scanners with \mbox{$64\times0.625$ mm} helical mode. 
For testing purposes, the same patient whose normal-dose scan was used for training, was scanned on the same imaging site with the same CT scanner, but using a different protocol with \mbox{pitch 1.375:1} and a much lower current of \mbox{40 mA}. 
Different patients for testing, including a GE Performance Phantom (GEPP), were scanned on various imaging sites with a variety of acquisition protocols. 
More precisely, the GEPP datasets were acquired with \mbox{120 kVp}, \mbox{1 s/rotation}, \mbox{pitch 0.516:1}, 
with four different magnitudes of tube current as \mbox{290 mA}, \mbox{145 mA}, \mbox{75 mA}, and \mbox{40 mA},
and were reconstructed in \mbox{135 mm} FOV.
A normal-dose scan of a new patient was acquired with \mbox{120 kVp}, \mbox{200 mA}, \mbox{0.5 s/rotation}, \mbox{pitch 0.984:1}, and reconstructed in \mbox{320 mm} FOV.
This particular dataset was used for both denoising and reconstruction experiments.
The supplementary material includes reconstruction results of a low-dose scan acquired with another new patient.
 \\

\subsubsection{2-D image denoising} 
\begin{figure*}[!t]
\centerline{
\subfloat[ground truth]{\includegraphics[width=2.1in]{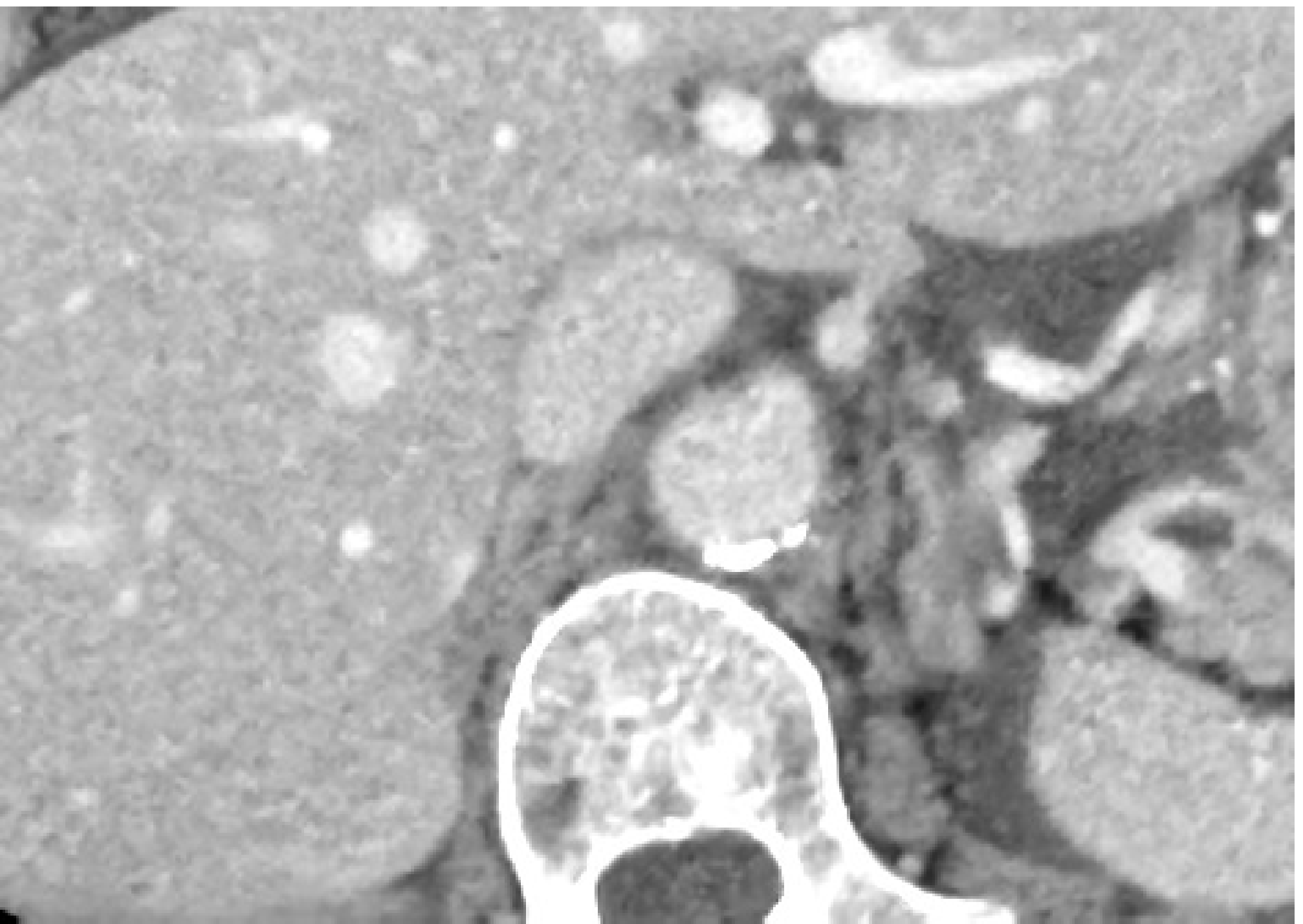}}\
\subfloat[noisy (39.88 HU)]{\includegraphics[width=2.1in]{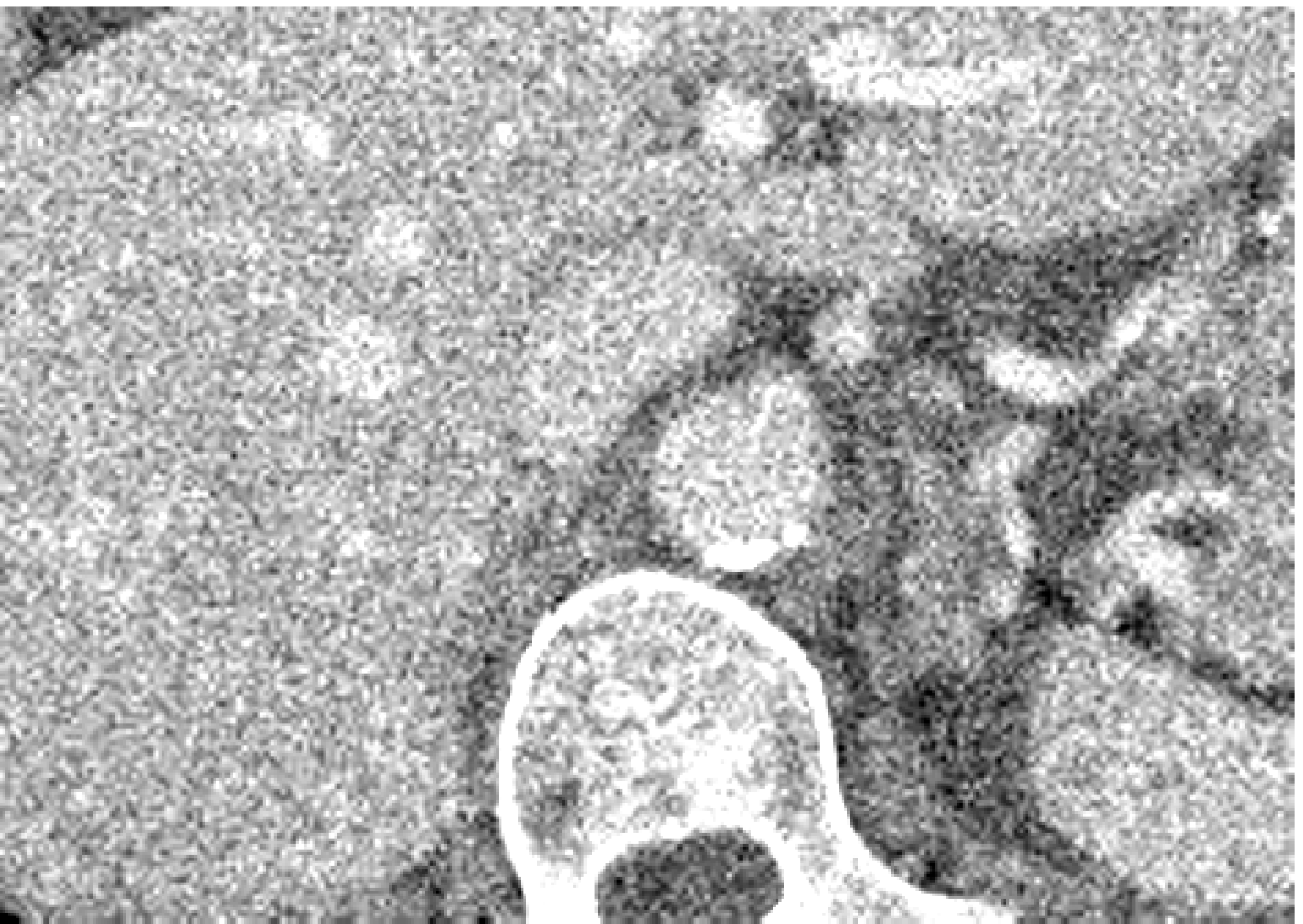}}}
\centerline{
\subfloat[BM3D (13.35 HU)]{\includegraphics[width=2.1in]{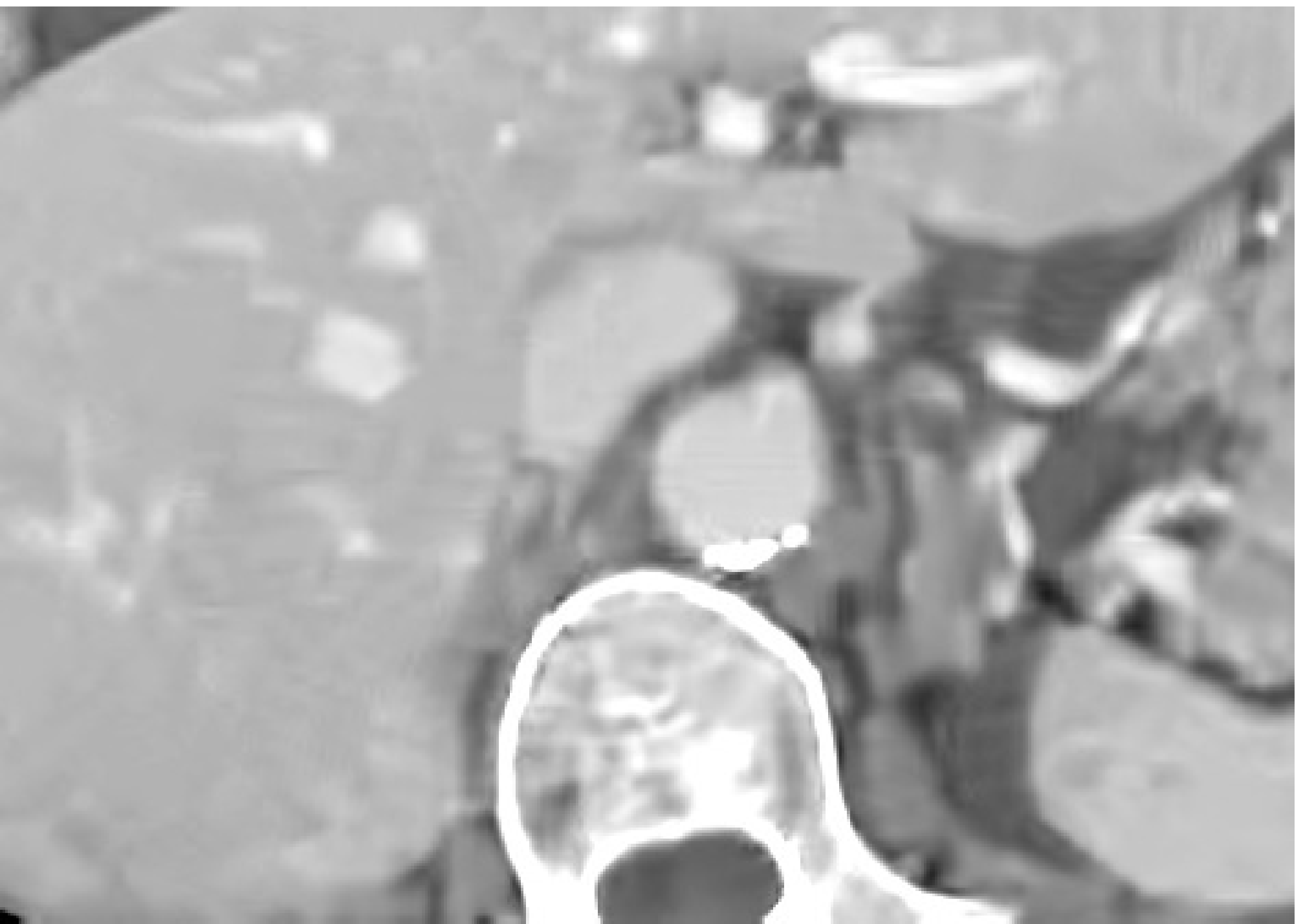}}\
\subfloat[K-SVD (14.57 HU)]{\includegraphics[width=2.1in]{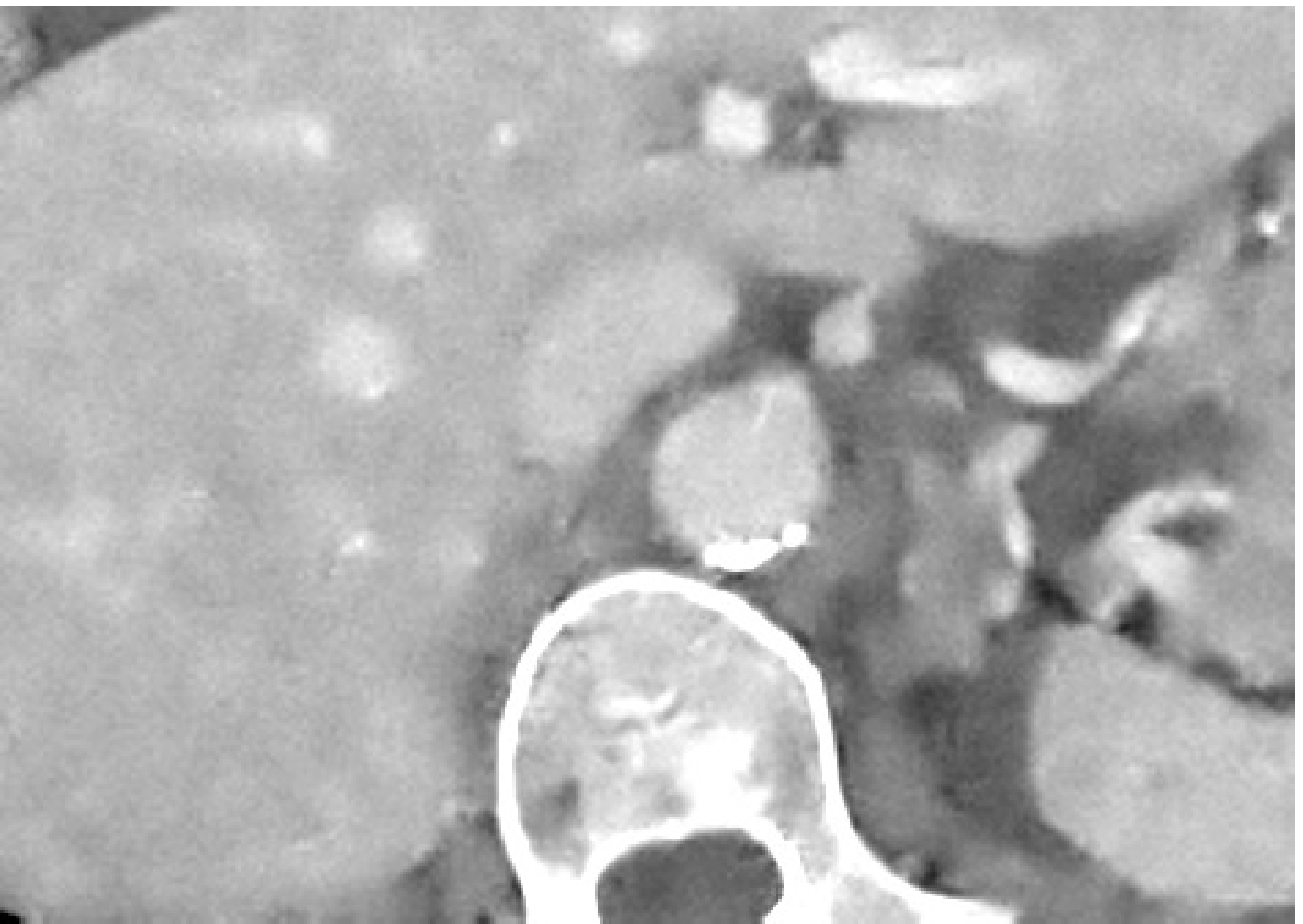}}\
\subfloat[NLM (14.82 HU)]{\includegraphics[width=2.1in]{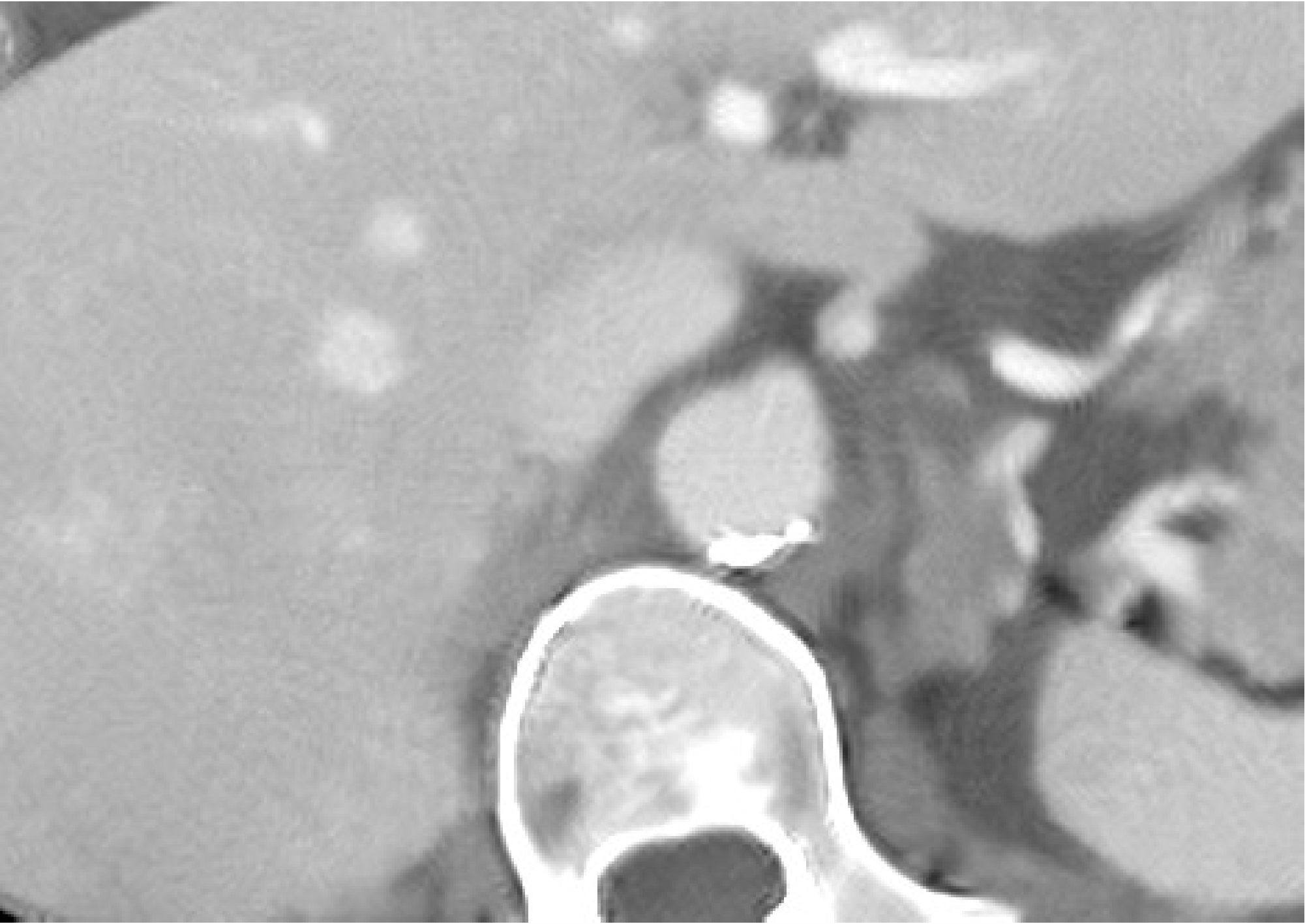}}}
\centerline{
\subfloat[$q$-GGMRF (15.96 HU)]{\includegraphics[width=2.1in]{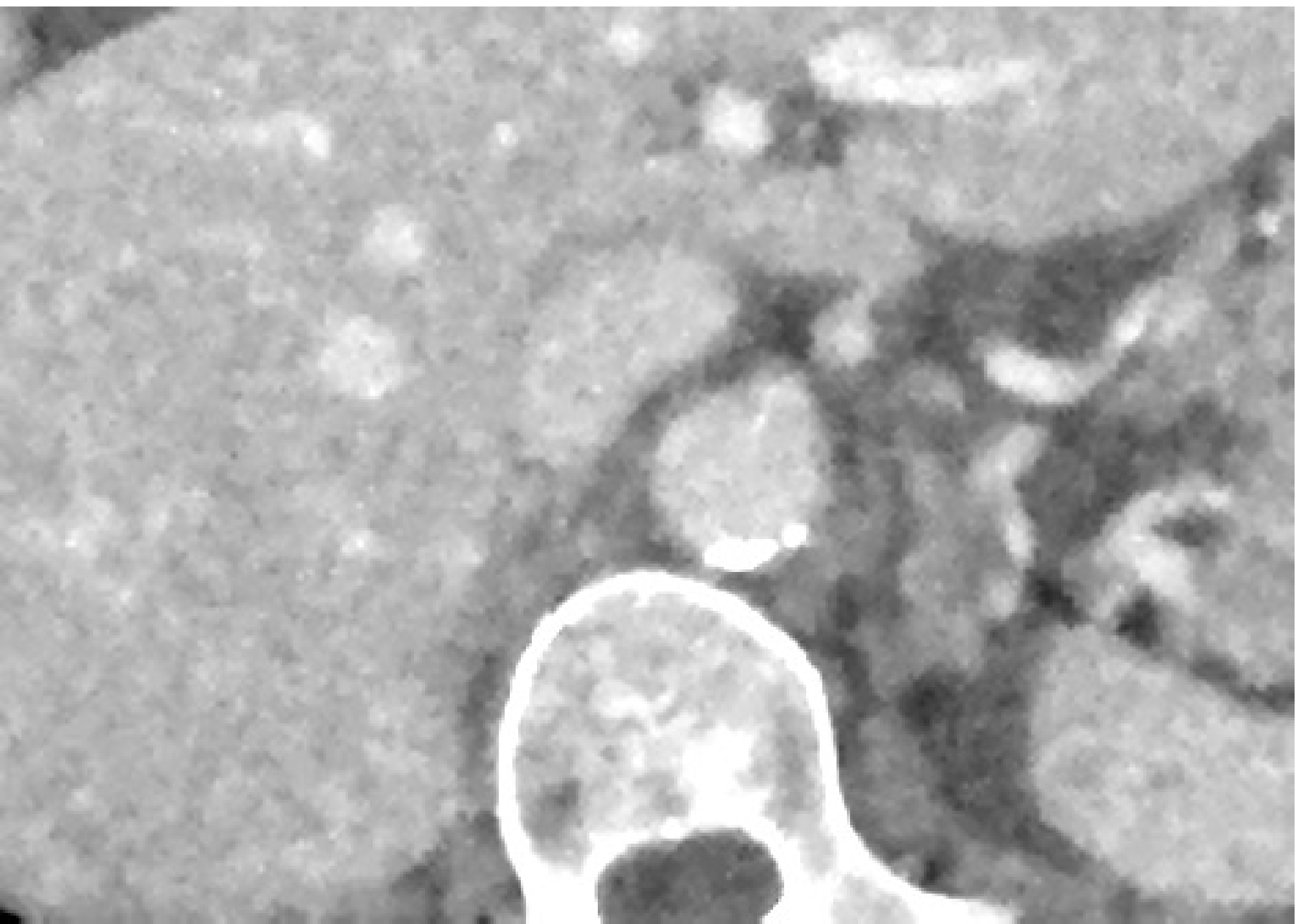}}\
\subfloat[original $5\times5$ GM-MRF (13.78 HU)]{\includegraphics[width=2.1in]{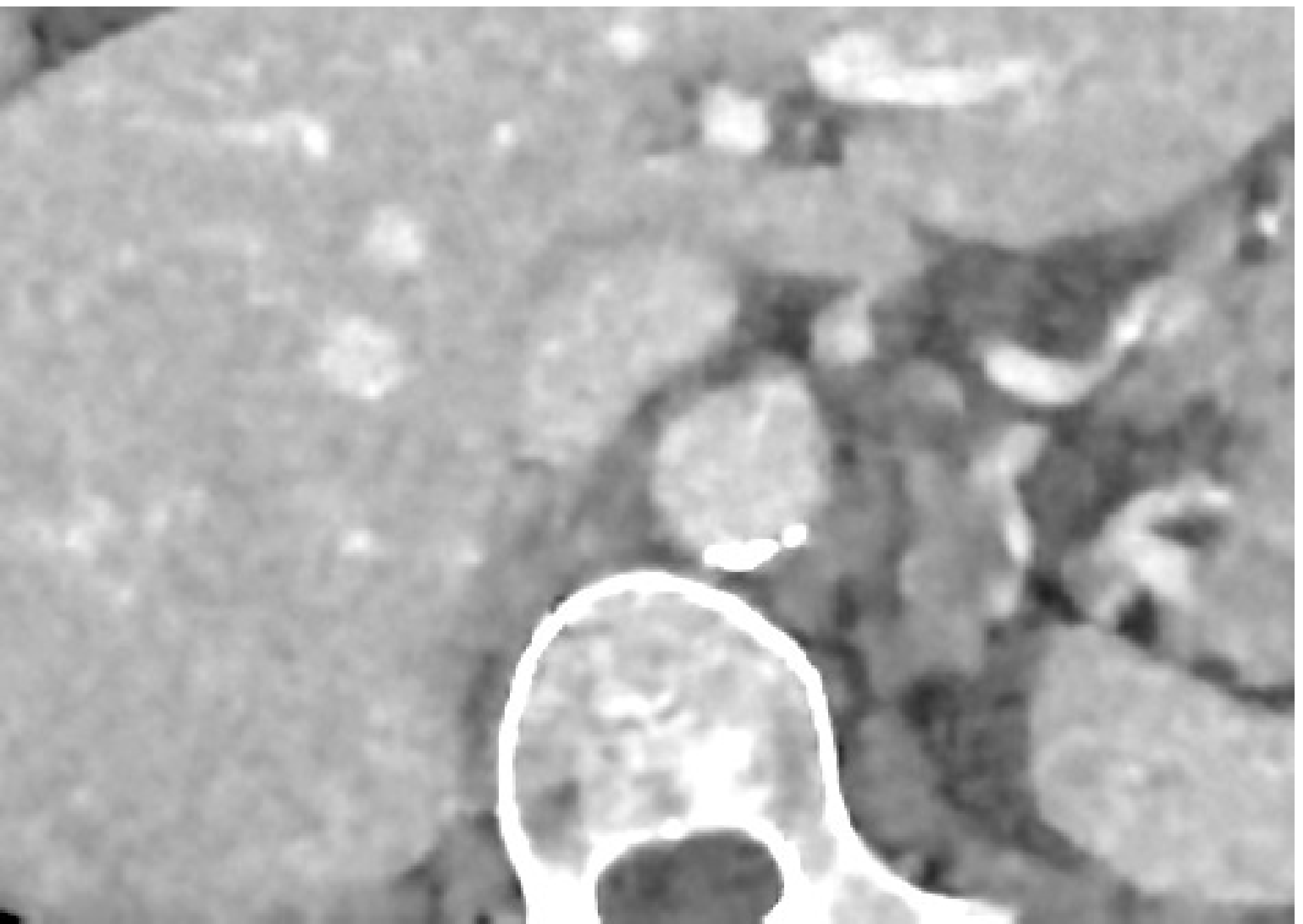}}\
\subfloat[adjusted $5\times5$ GM-MRF (14.33 HU)]{\includegraphics[width=2.1in]{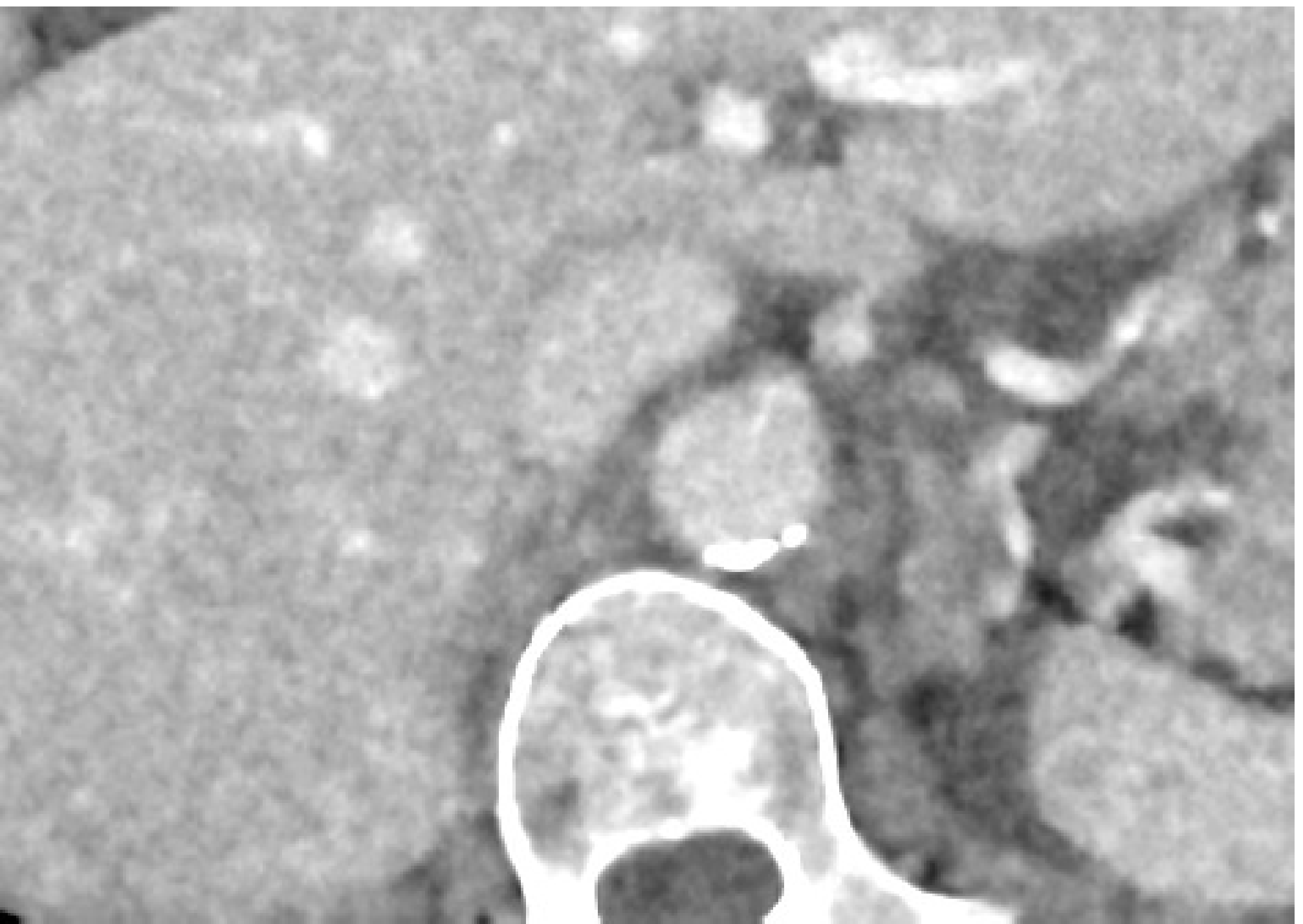}}}
\caption{Denoising results with different methods (with RMSE value reported). 
Individual image is zoomed to a region containing soft tissue, contrast, and bone for display purposes.
The RMSE value between each reconstructed image and the ground truth is reported. 
Display window: [-100 200] HU.
GM-MRF methods achieve lower RMSEs and better visual quality than $q$-GGMRF, K-SVD, and NLM methods.
Though having a slightly higher RMSE, GM-MRF with original model preserves real textures in soft tissue without creating severe artifacts,
while BM3D tends to over-smooth the soft tissue and introducing artificial, ripple-like structures.
In addition, though compromising the RMSE than the original model, GM-MRF with the adjusted model produces better visual quality, especially for the soft-tissue texture.}
\label{fig:denoising}
\end{figure*}

\quad  \vspace{0.05in} \\
\indent We first tested the GM-MRF model in a 2-D image denoising experiment.
The ground-truth image in Fig.~\ref{fig:denoising}(a) was obtained from the reconstructed images of the aforementioned normal-dose scan of a new patient whose data was not used for training.
Then, we added Gaussian white noise to the ground truth to generate the noisy image in Fig.~\ref{fig:denoising}(b).
Different denoising methods were then applied to the noisy image.

We experimented with a few GM-MRFs with various sizes of patch models,
that is, 2-D GM-MRF with $3\times3$, $5\times5$, and $7\times7$ patch models,
to study the impact of patch sizes in the proposed GM-MRF model.
In addition, for each patch size, we experimented with the original model obtained directly from training 
and the adjusted model with $p=0.5, \alpha=33\ {\rm HU}$ in (\ref{eq:sigmak}),
to study the effect of covariance scaling.

We will compare our GM-MRF methods with a number of widely used methods, including 
the $q$-GGMRF method\cite{Thibault07}, K-SVD method\cite{Elad06}, BM3D method\cite{Dabov07}, 
and non-local mean (NLM) method\cite{Buades05}.
The $q$-GGMRF method was implemented with $3\times3$ neighborhood with parameters $p=2$, $q=1.2$, $c=10\ {\rm HU}$.
The K-SVD method was performed by using the software in\cite{Rubinstein13} with $7\times7$ patch size and 512 dictionary entries.
The BM3D method was performed by using the software in\cite{Dabov14} with $8\times8$ patch.
The NLM method was performed by using the software in\cite{Kroon10} with $9\times9$ patch and $21\times21$ search window.
We adjusted the regularization strength for all methods to achieve the lowest root-mean-square error (RMSE) 
between the recovered image and the ground truth.
\\
\vspace{-0.1in}

\subsubsection{3-D CT reconstruction} \quad  \vspace{0.05in} \\
\indent We tested the GM-MRF method in a number of 3-D CT reconstruction experiments.
We first conducted phantom studies using different GM-MRF models with various parameter settings 
to study the impact of change in model parameters on the final reconstructed images.
Then, we picked only one GM-MRF model and then applied this model to all subsequent experiments with clinical datasets.
Thus, no additional training was conducted for the test data.

We conducted both qualitative and quantitative comparisons by using GEPP reconstructions.
The GEPP contains a plexiglas insert with cyclic water bars and a 50~$\mu$m diameter tungsten wire placed in water.
We measured the mean and standard deviation within fixed ROIs in flat regions to assess the reconstruction accuracy and noise.
In addition, we measured the modulation transfer function (MTF) using the wire to assess the in-plane resolution and contrast.
We reported the 10\% MTF since it reflects the visual resolution of the image,
with higher value indicating finer texture, which is a desirable image quality especially for a low-dose condition.
We compared the visual quality for clinical reconstructions.

We studied the influence of different model parameters in the GM-MRF model using GEPP reconstructions,
including the number of GM components ($K$), the patch size ($L$), and parameters $p$ and $\alpha$ for covariance scaling. 
For fair comparison, we matched the noise level between different models by adjusting the global regularization parameter,
$\sigma_x$, in (\ref{eq:energy_reg}).
That is, for a given dose level, we matched the standard deviations within a selected flat region between different models
such that the absolute difference of the two is within \mbox{1 HU}.

In addition to evaluating the behavior of different parameters within the GM-MRF prior, 
we will also compare the MBIR using GM-MRF prior with two widely used reconstruction methods:
FBP using a standard kernel and MBIR using $q$-GGMRF prior\cite{Thibault07} with reduced regularization.
Note that we intentionally reduced the regularization for MBIR with the $q$-GGMRF prior
so as to obtain higher resolution, which lead to much higher noise as well.
We matched the noise level between the $q$-GGMRF and GM-MRF methods.
However, for low-dose datasets, it is challenging to match the noise between those two methods
due to the excessive speckle noise produced by using an under-regularized $q$-GGMRF prior.
Thus, we will instead demonstrate that the GM-MRF prior achieves higher resolution with even less noise than the $q$-GGMRF prior.
Besides direct comparison, we also use FBP reconstruction as an illustration of the current dose level.

\setlength\tabcolsep{0in}
\begin{figure*}[!t]
\centering
\begin{tabular}{C{1.75in}C{1.75in}C{1.75in}C{1.75in}} 
\includegraphics[width=1.7in]{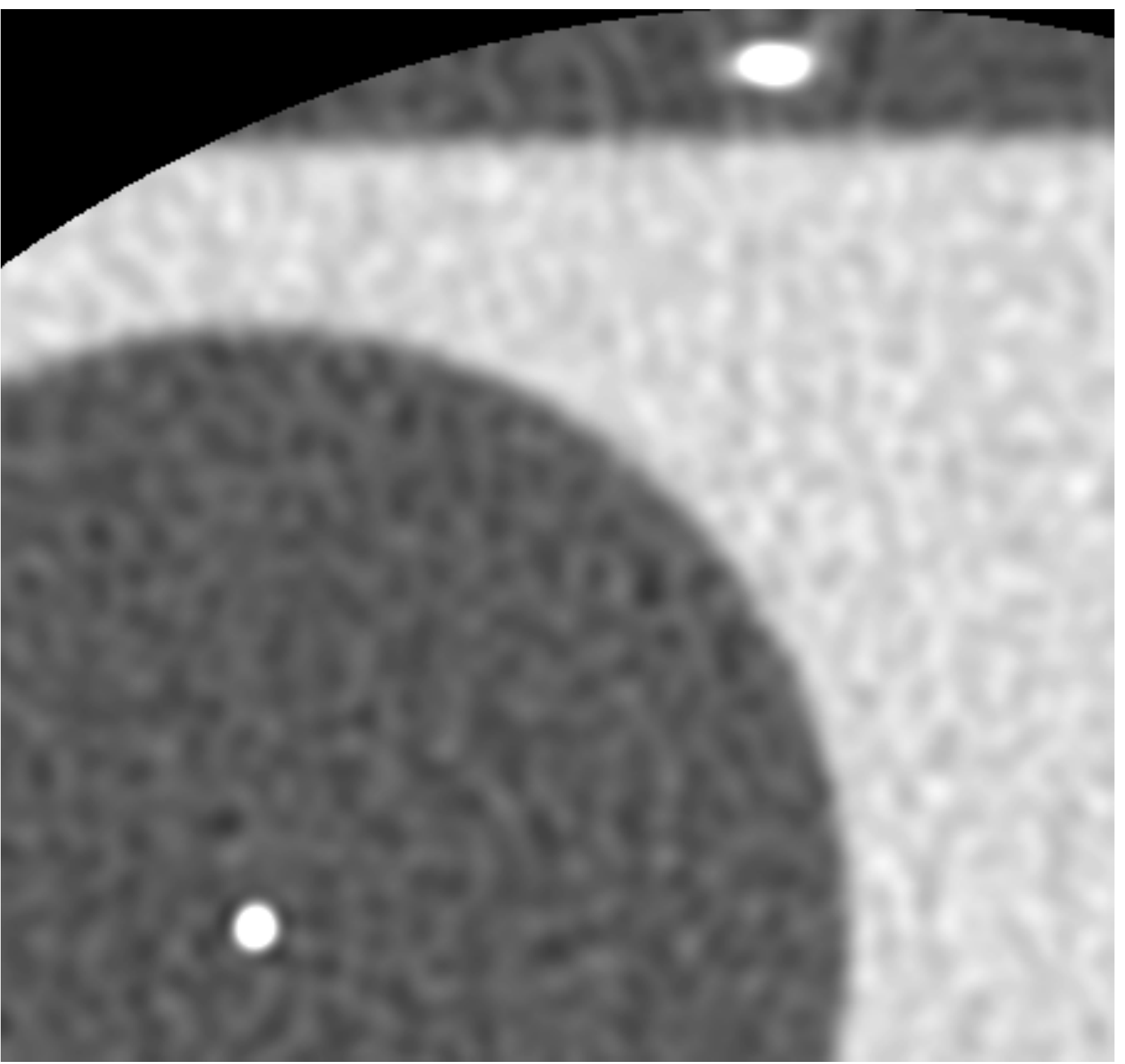} &
\includegraphics[width=1.7in]{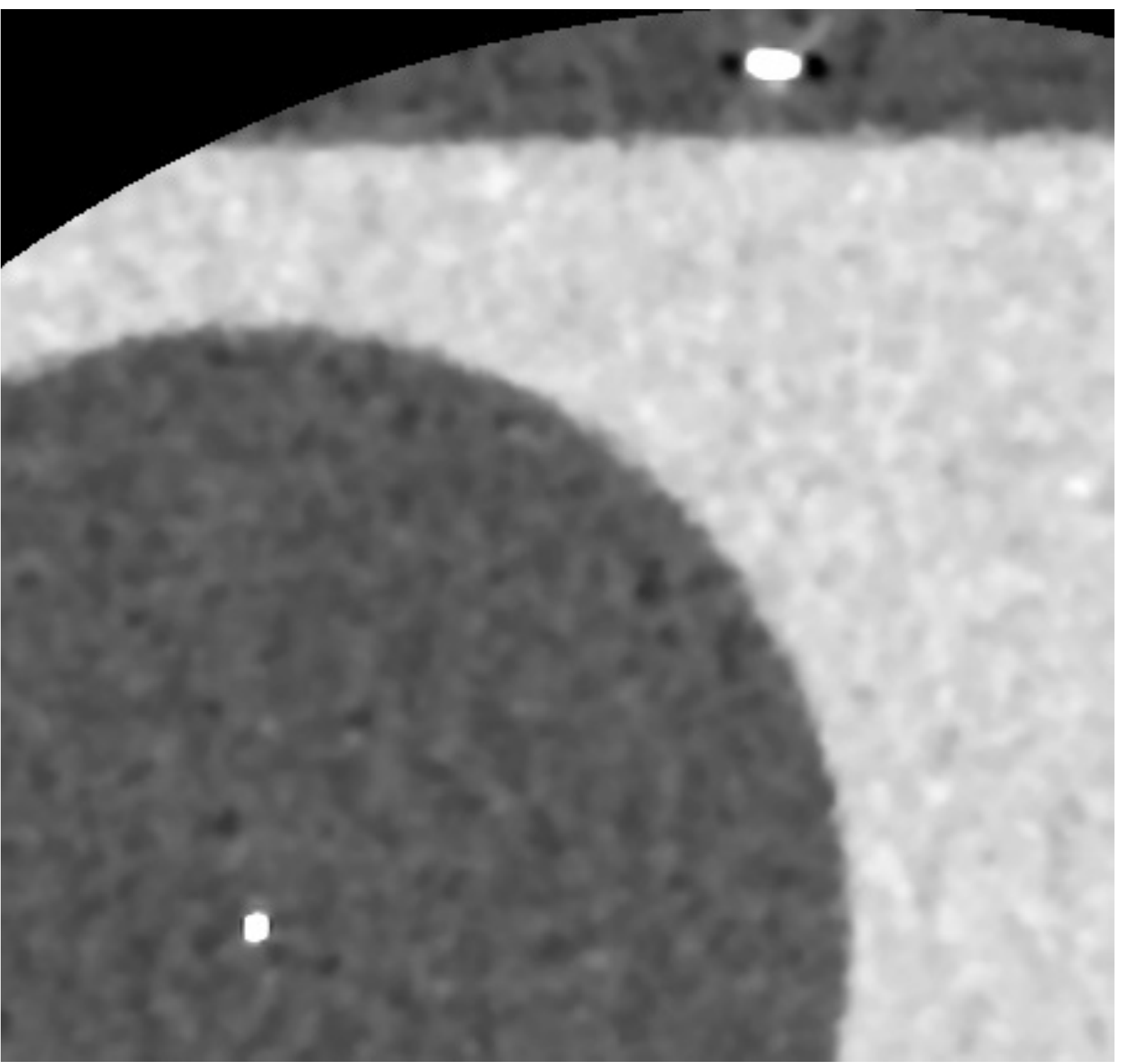} &
\includegraphics[width=1.7in]{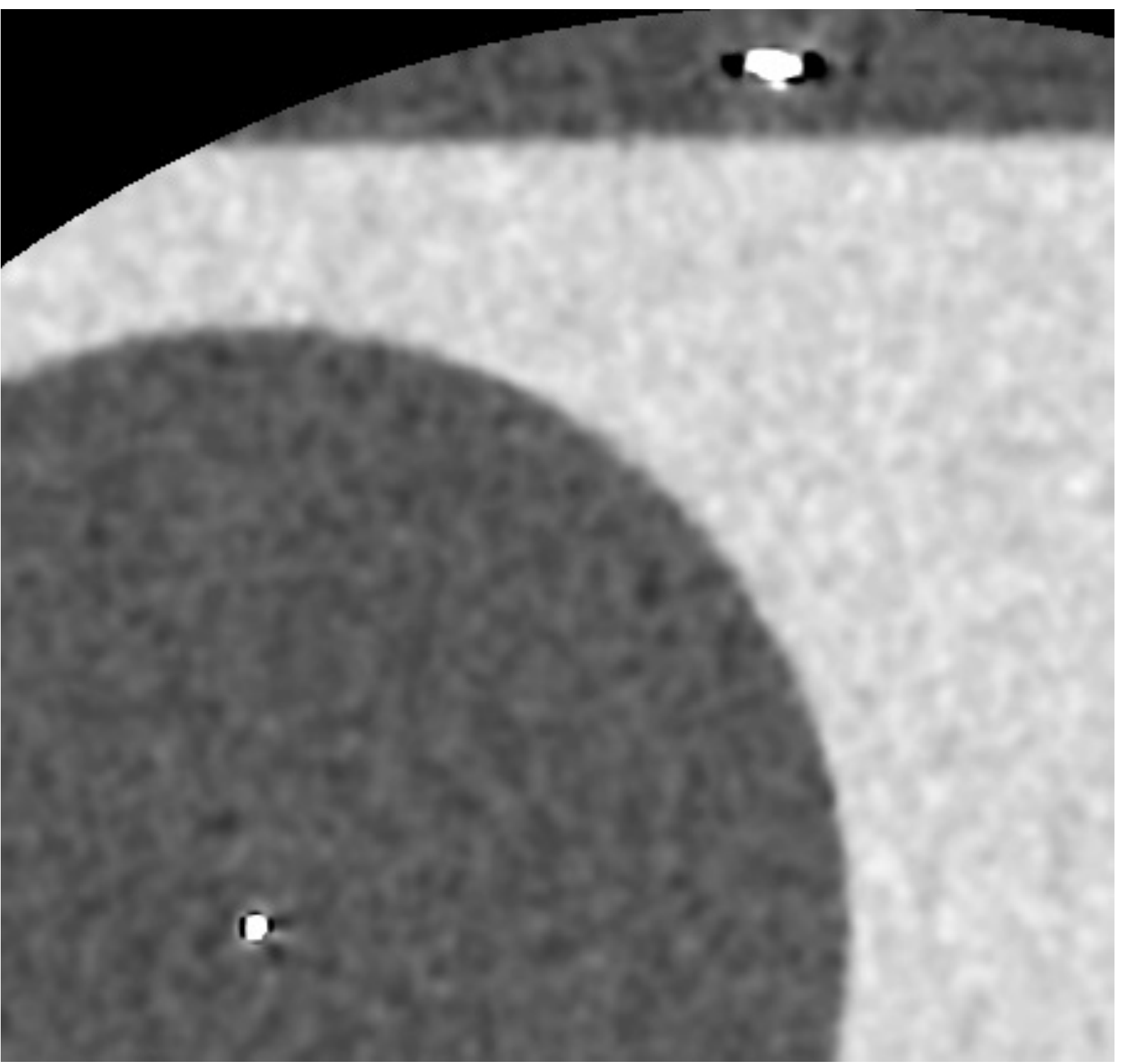} &
\includegraphics[width=1.7in]{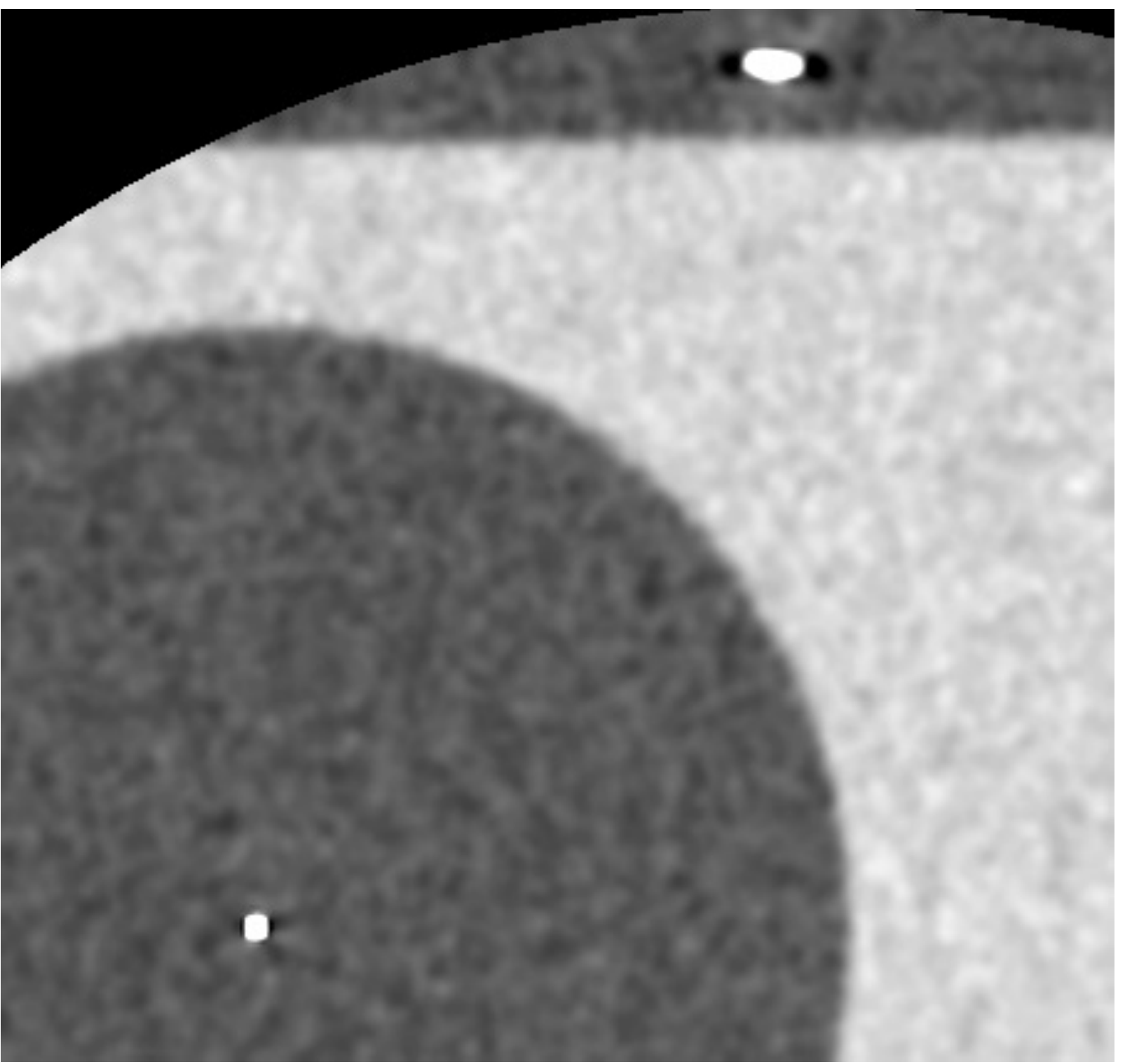} \\
\includegraphics[width=1.7in]{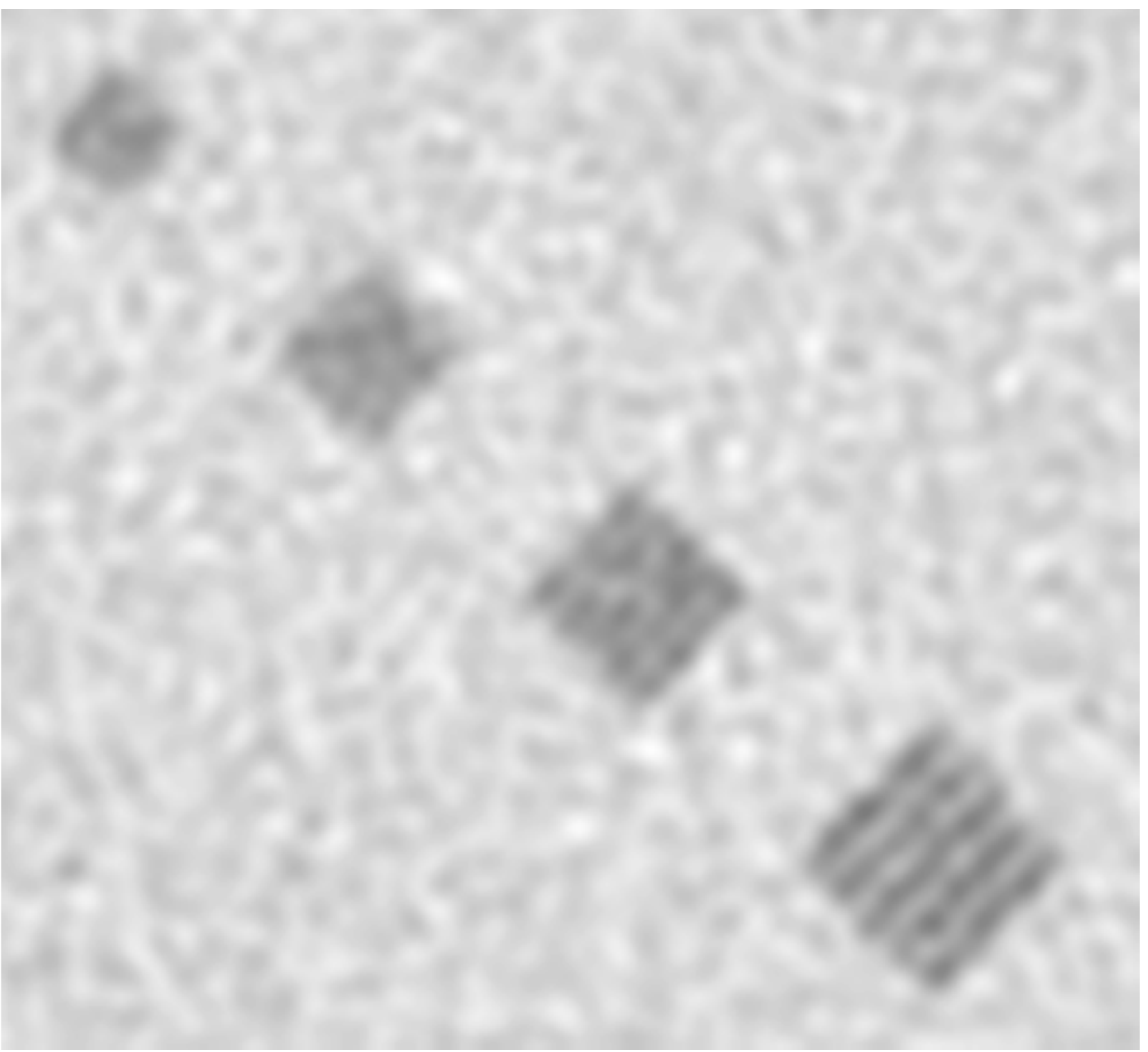} &
\includegraphics[width=1.7in]{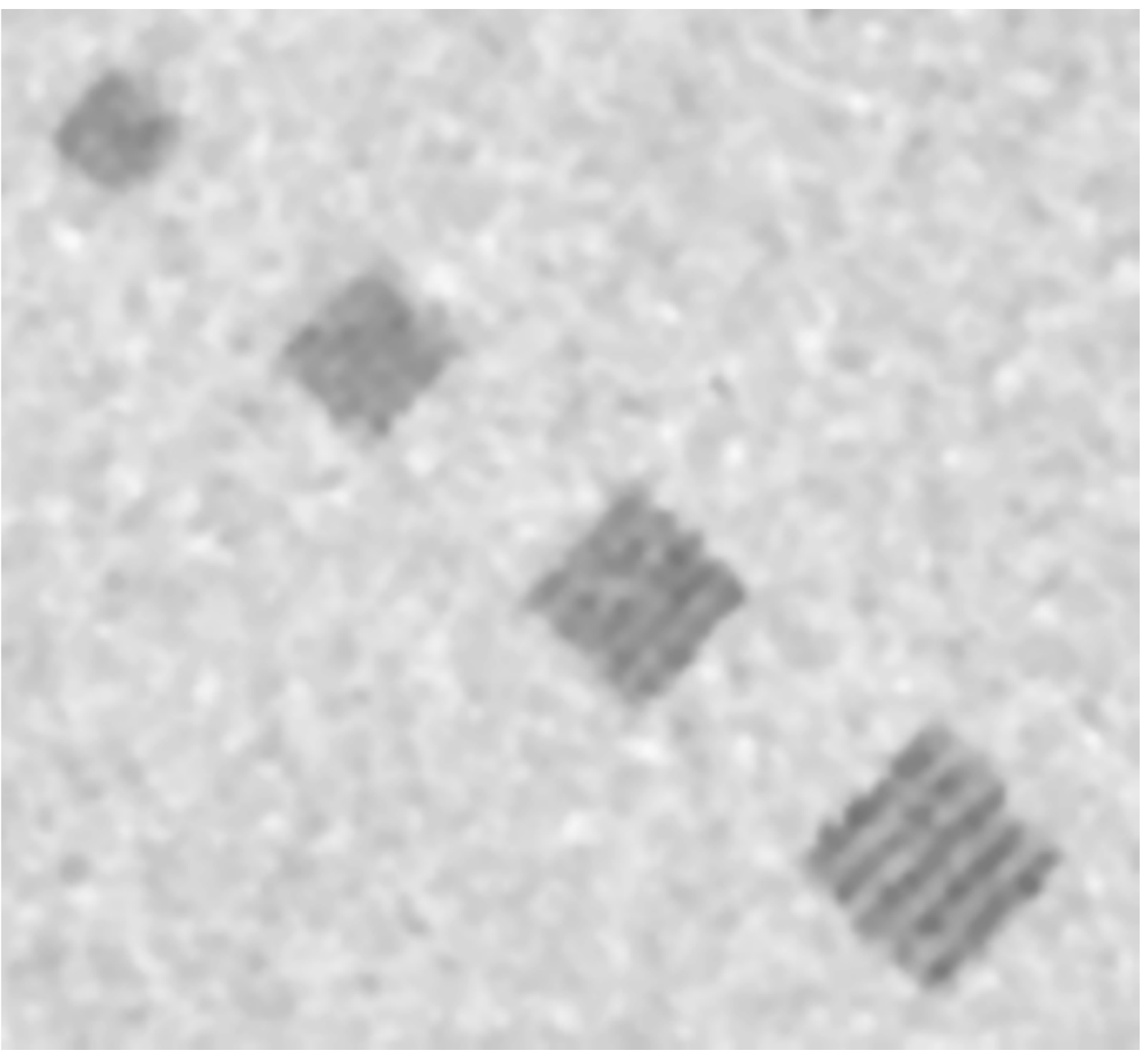} &
\includegraphics[width=1.7in]{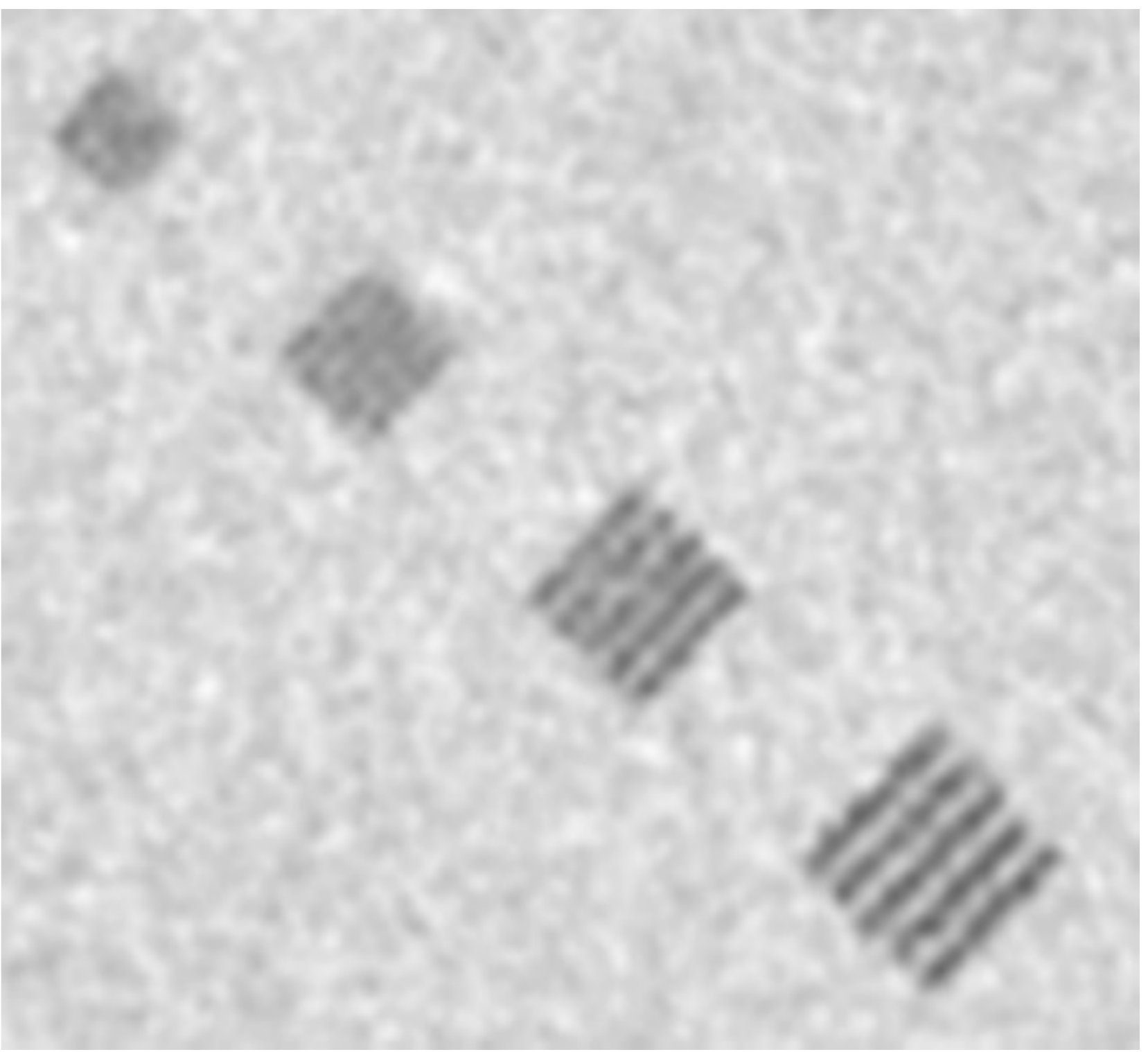} &
\includegraphics[width=1.7in]{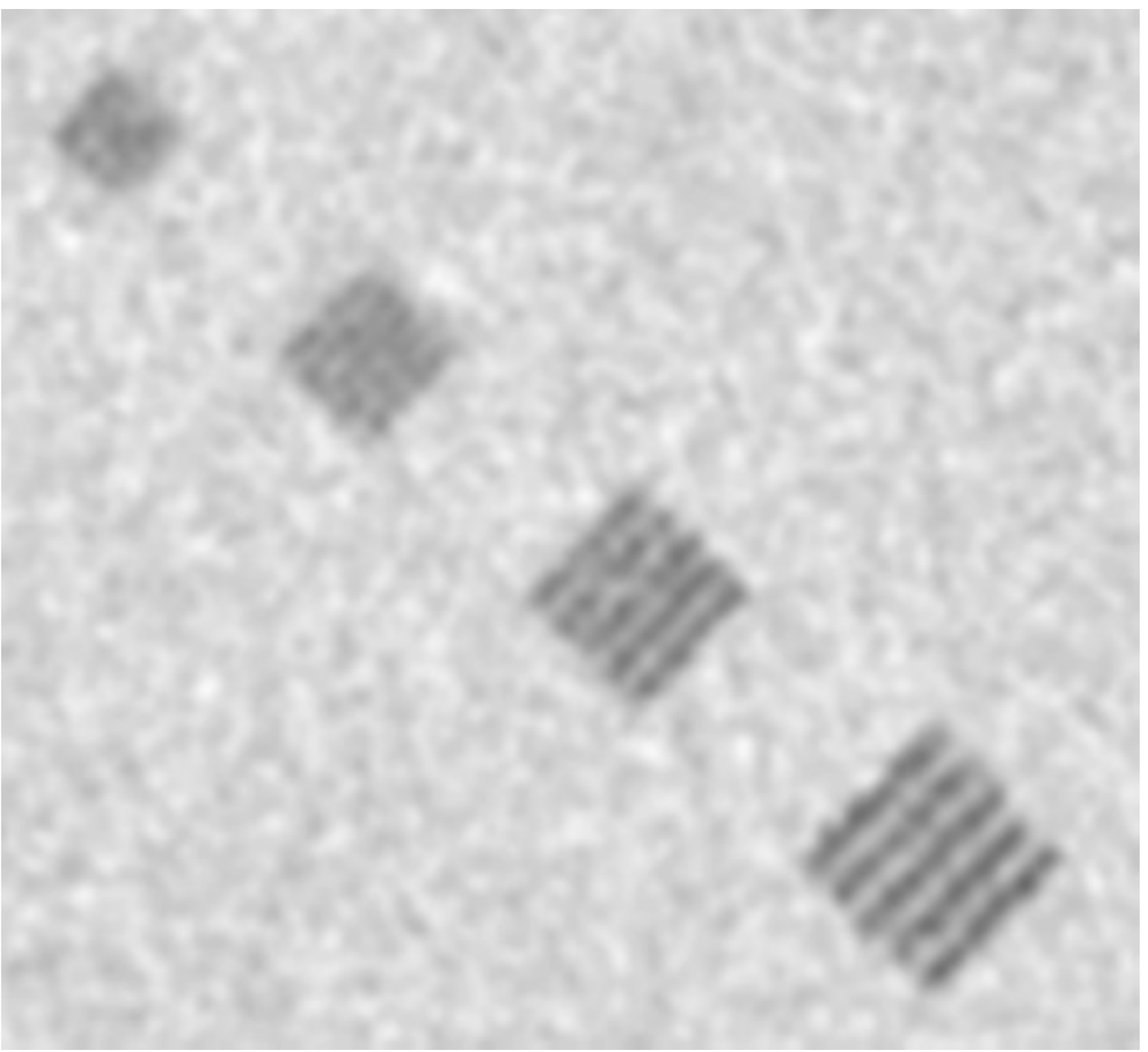} \\
 (a) FBP & (b) MBIR w/ $q$-GGMRF w/ reduced regularization & (c) MBIR w/ original $5\times5\times3$ GM-MRF & (d) MBIR w/ adjusted $5\times5\times3$ GM-MRF
\end{tabular}
\caption{GEPP reconstruction with data collected under \mbox{290 mA}. 
Individual image is zoomed to small FOVs for display purposes.
From left to right, the columns represent (a) FBP, (b) MBIR with $q$-GGMRF with reduced regularization,
(c) MBIR with original $5\times5\times3$ GM-MRF, and (d) MBIR with adjusted $5\times5\times3$ GM-MRF with $p=0.5, \alpha=33\ {\rm HU}$.
The top row shows the wire section and the bottom row shows the resolution bars with the display window as [-85 165] HU.
MBIR with GM-MRF priors generate images with sharper high-contrast objects and better texture than MBIR with $q$-GGMRF prior at a comparable noise level.
Moreover, for GM-MRF priors, the adjusted model produces better rendering of the high-contrast objects than the original model.}
\label{fig:gepp_290ma}
\end{figure*}

\begin{figure}[!t]
\centerline{
\includegraphics[height=2in,trim={0 0.05in 0 0},clip]{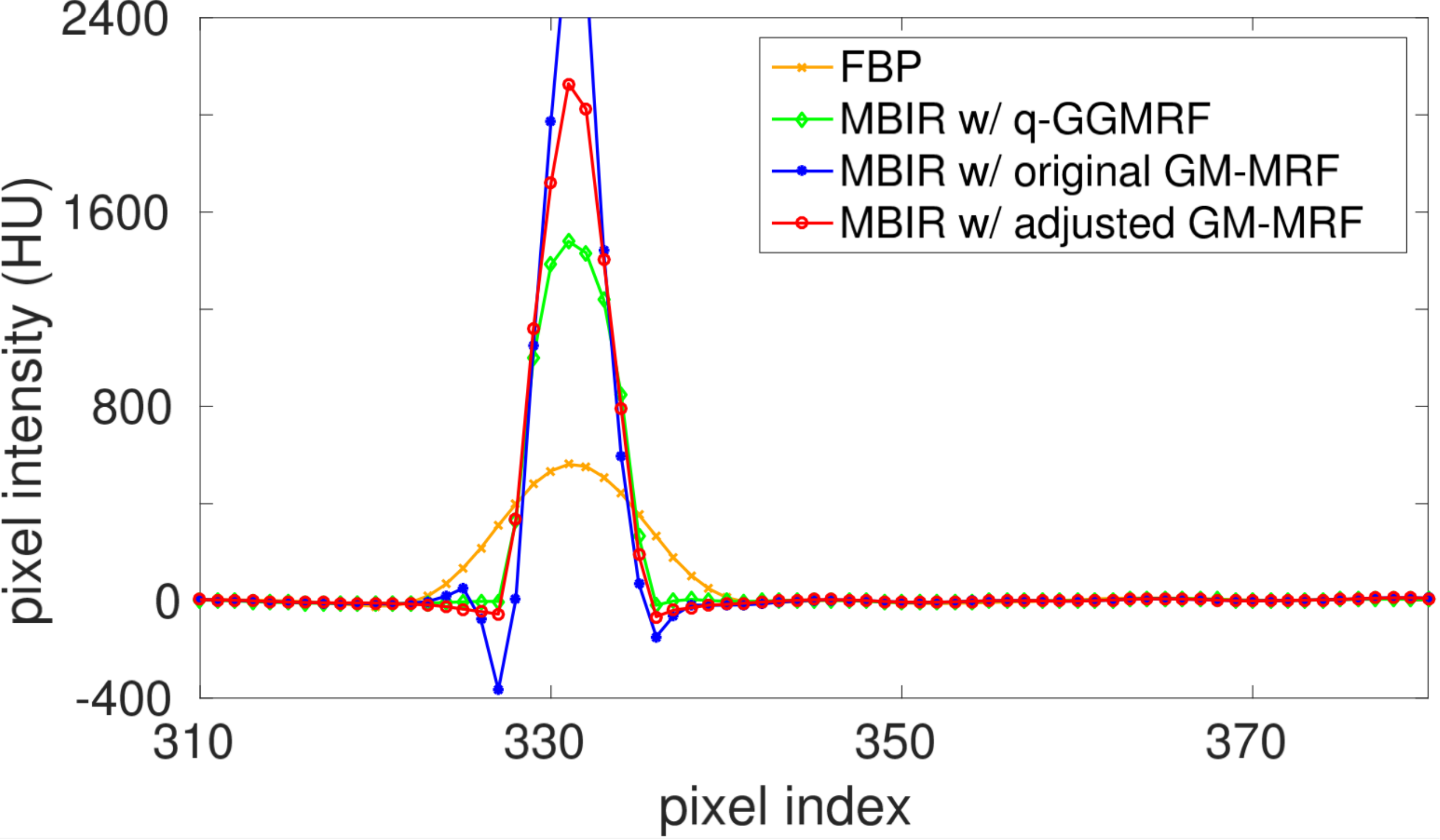}}
\caption{Profile plot through the center of the tungsten wire in Fig.~\ref{fig:gepp_290ma}.
As compared to the $q$-GGMRF result (green), there is an undershoot-like artifact near the wire in the MBIR result with the original GM-MRF (blue), which is substantially reduced by using the adjusted GM-MRF model (red). 
Also notice that increasing the regularization for high-contrast regions tends to reduce the contrast (reduced peak from blue to red); however, the resultant contrast still remains more substantial than the $q$-GGMRF result.
}
\label{fig:profile_wire}
\end{figure}

\section{Results}
\label{sec:experiment}

\subsection{2-D image denoising} 

Fig.~\ref{fig:denoising} presents the denoising result with different methods.
It shows that the mean-square-error (MSE) achieved by the $5\times5$ GM-MRF method 
with original covariances is slightly higher than the BM3D method,
but significantly lower than the $q$-GGMRF, K-SVD, and NLM methods.
Qualitatively, the GM-MRF method with the original model produces sharper edges and less speckle noise than the $q$-GGMRF method,
and preserves more fine structures and detail than K-SVD and NLM methods.
The BM3D method seems to produce more enhanced fine structures than the GM-MRF method due to its strong structure-preserving behavior.
However, it also tends to over-smooth the soft tissue region while creating some artificial, ripple-like structures and texture.
Alternatively, the GM-MRF method is able to preserve some real texture in soft tissue without inducing severe artifacts,
which can be important in some medical applications.

The supplementary material include the 2-D denoising results by using GM-MRF methods with different sizes of patch models.
It is observed that the RMSE value decreases as the patch size increases in the GM-MRF models, 
since the model becomes more expressive as patch size increases.
Moreover, denoising results produced by using the GM-MRF with larger patches appear visually more natural than 
results associated with the GM-MRF with smaller patches.

Interestingly, though compromising the MSE, 
the GM-MRF methods with adjusted models produce images with better visual quality than those with original models.
The better visual quality is achieved with improved soft-tissue texture and better rendering of high-contrast structures.
This is because the MAP estimate tends to over-regularize the low-contrast regions and under-regularize the high-contrast regions in the image. 
Therefore, by adjusting the regularization strength in different contrast regions, 
we may achieve desirable visual quality but with the compromise in the MSE.

\begin{figure*}[!t]
\centerline{
\qquad \subfloat[measurements]{\includegraphics[width=1.6in]{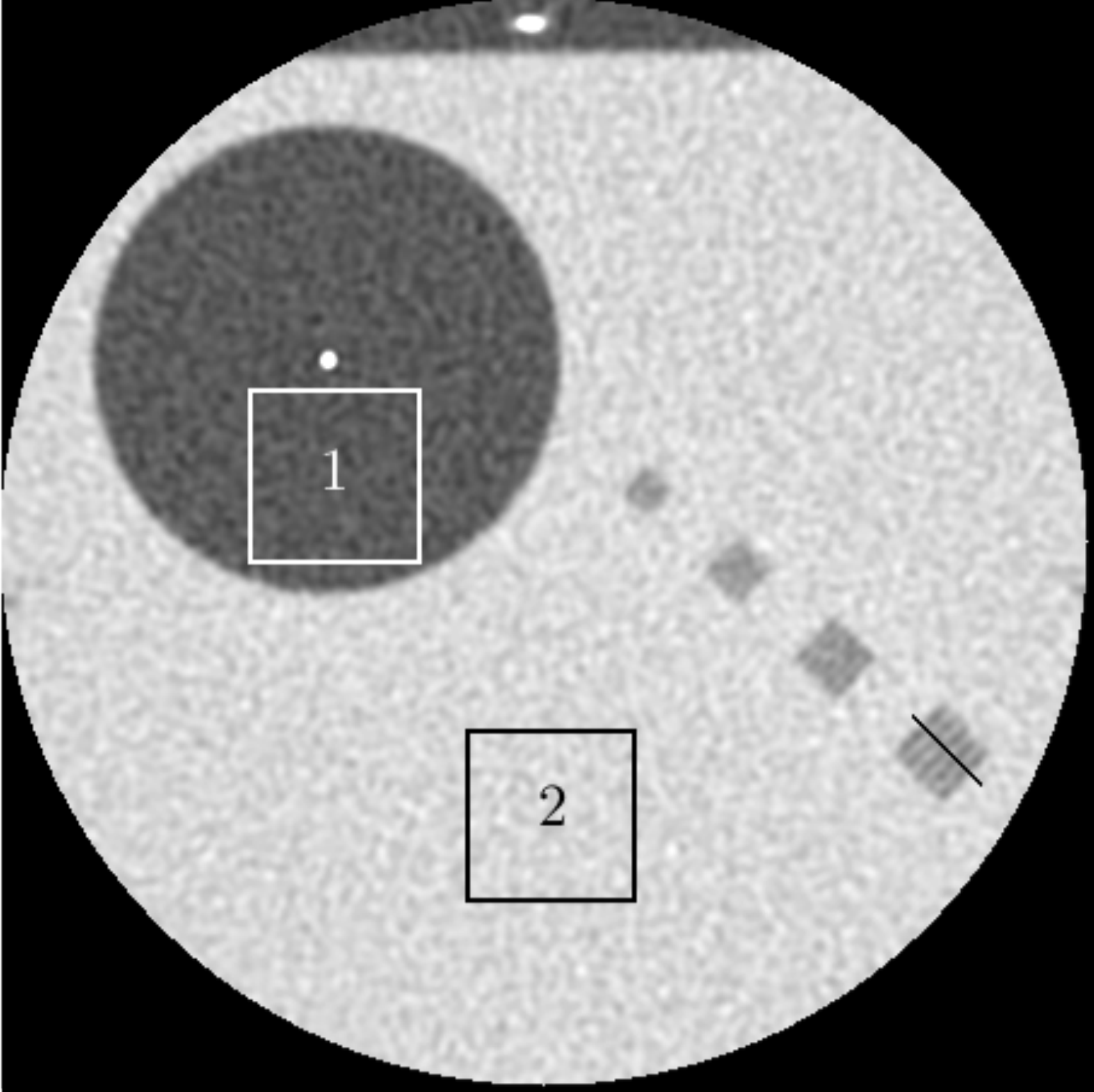}}\quad 
\quad \subfloat[mean within ROI 1]{\includegraphics[width=2.1in]{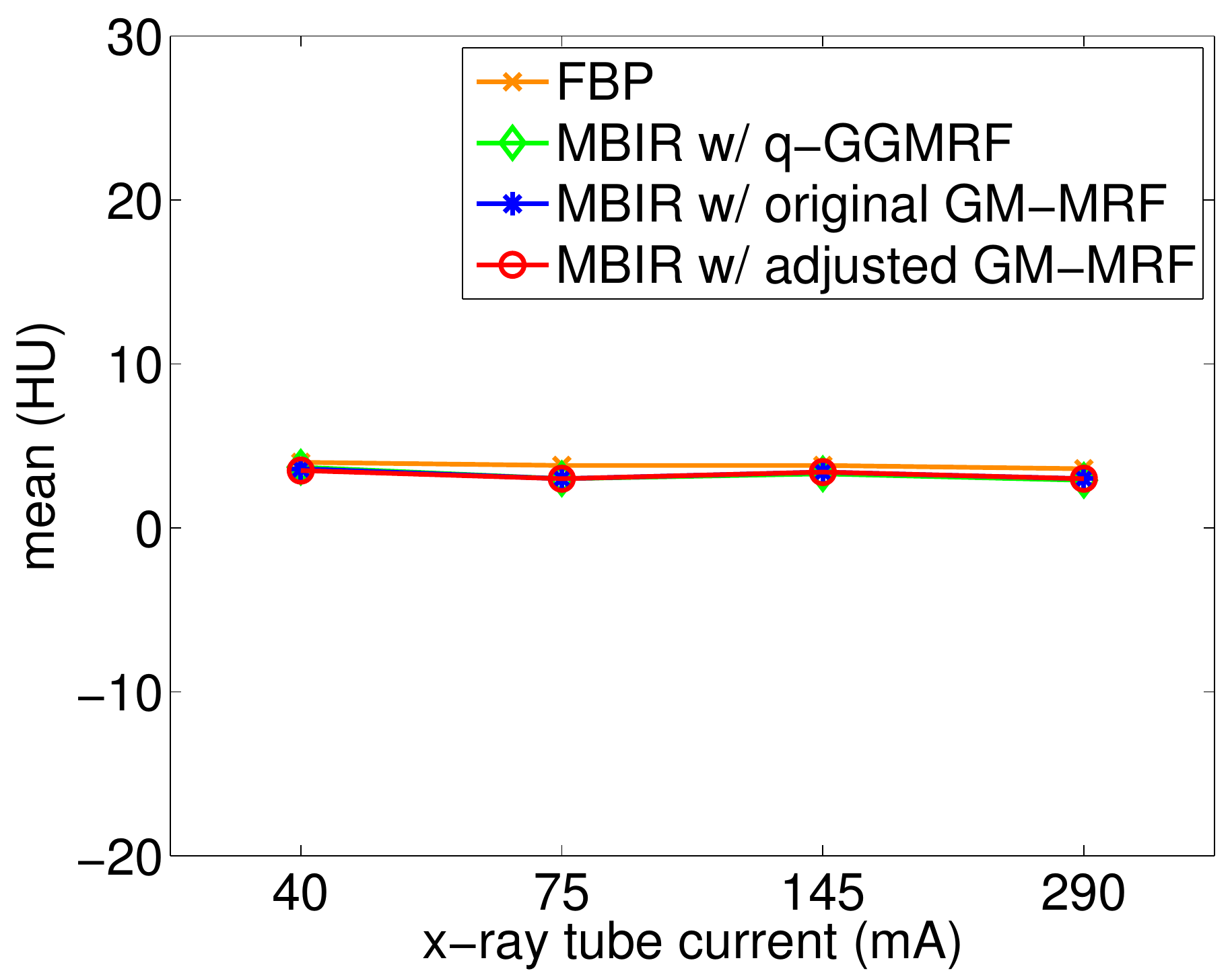}}\quad
\subfloat[mean within ROI 2]{\includegraphics[width=2.1in]{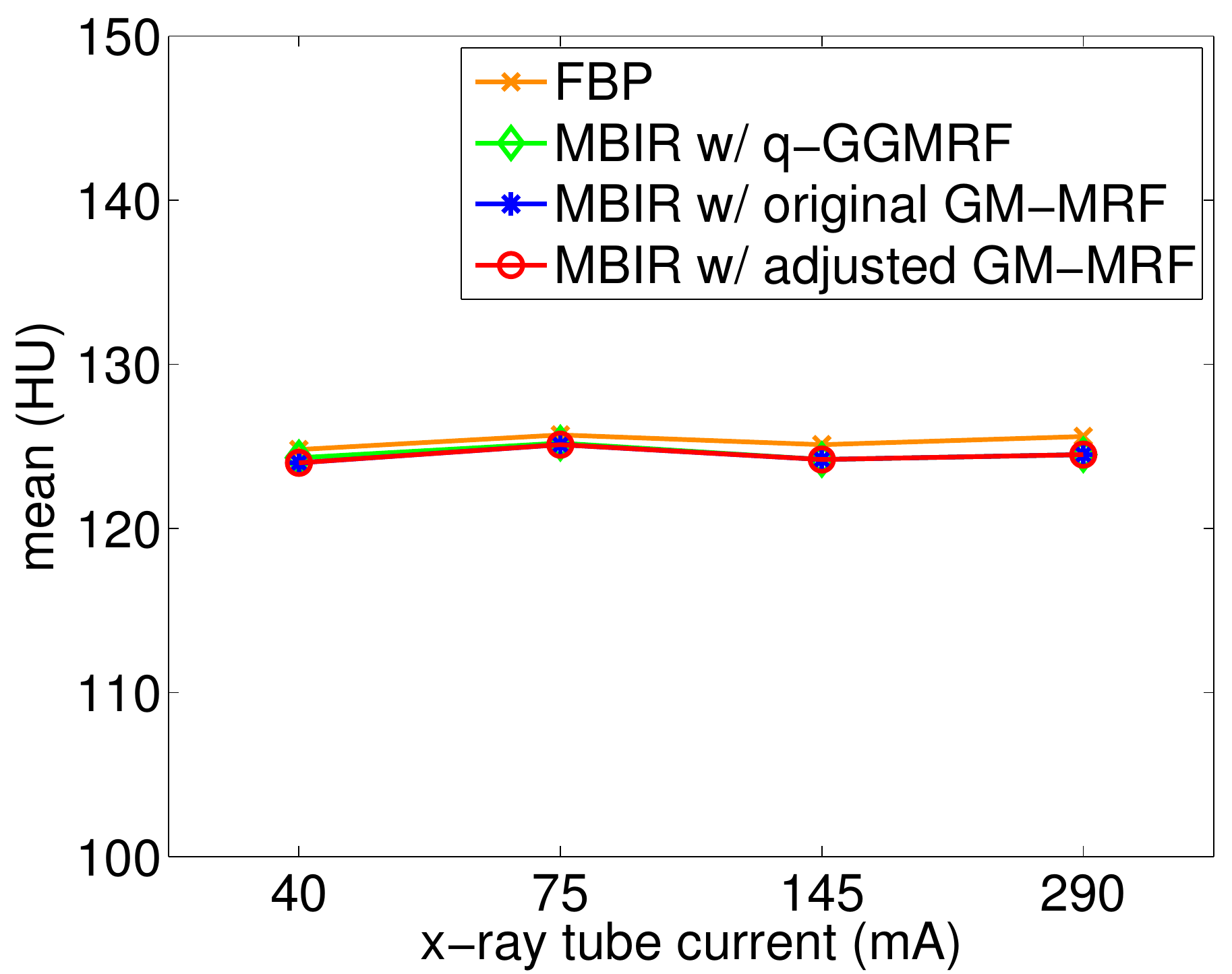}}}
\centerline{ 
\subfloat[noise within ROI 1]{\includegraphics[width=2.1in]{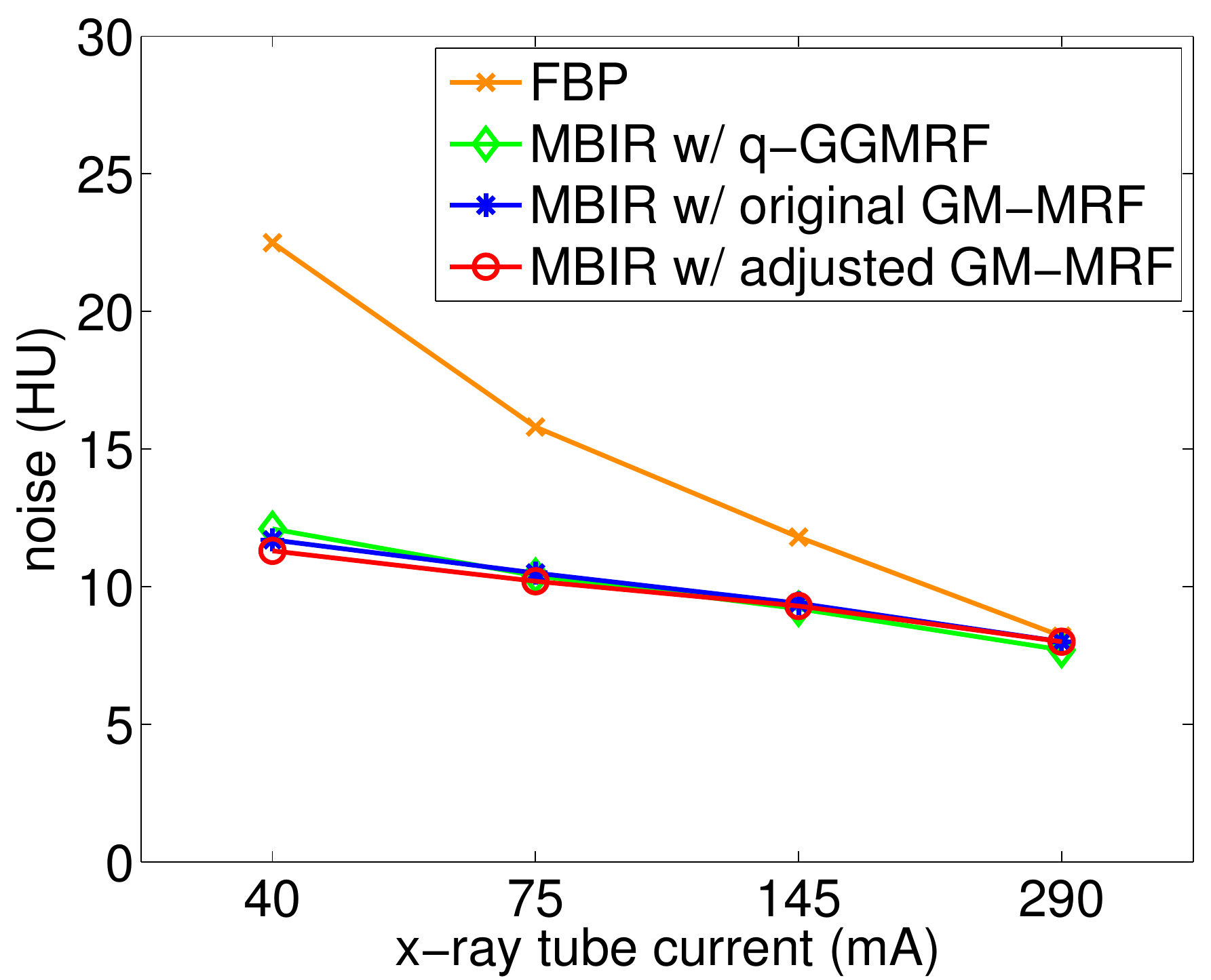}}\quad
\subfloat[noise within ROI 2]{\includegraphics[width=2.1in]{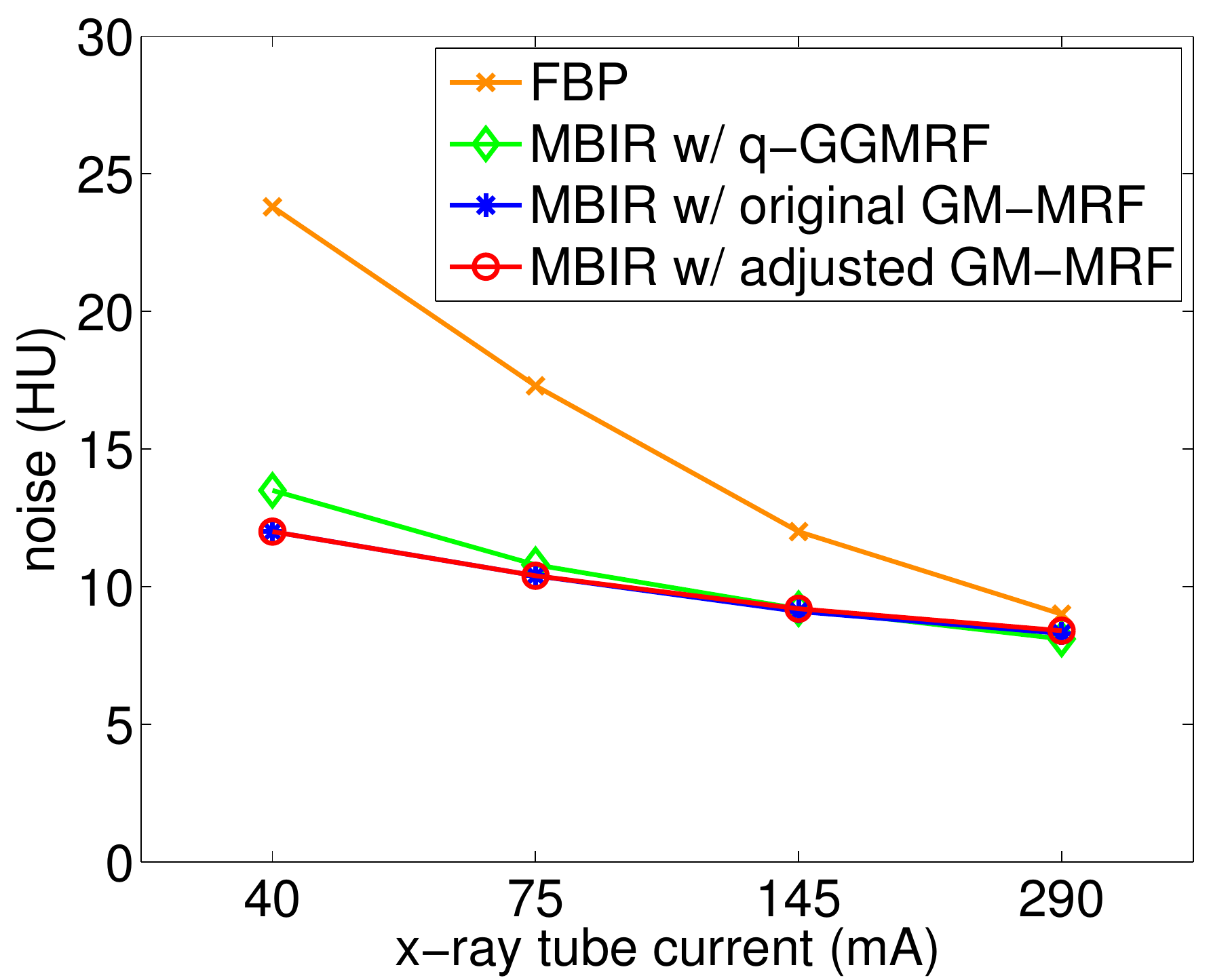}}\quad
\subfloat[10\% MTF]{\includegraphics[width=2.1in]{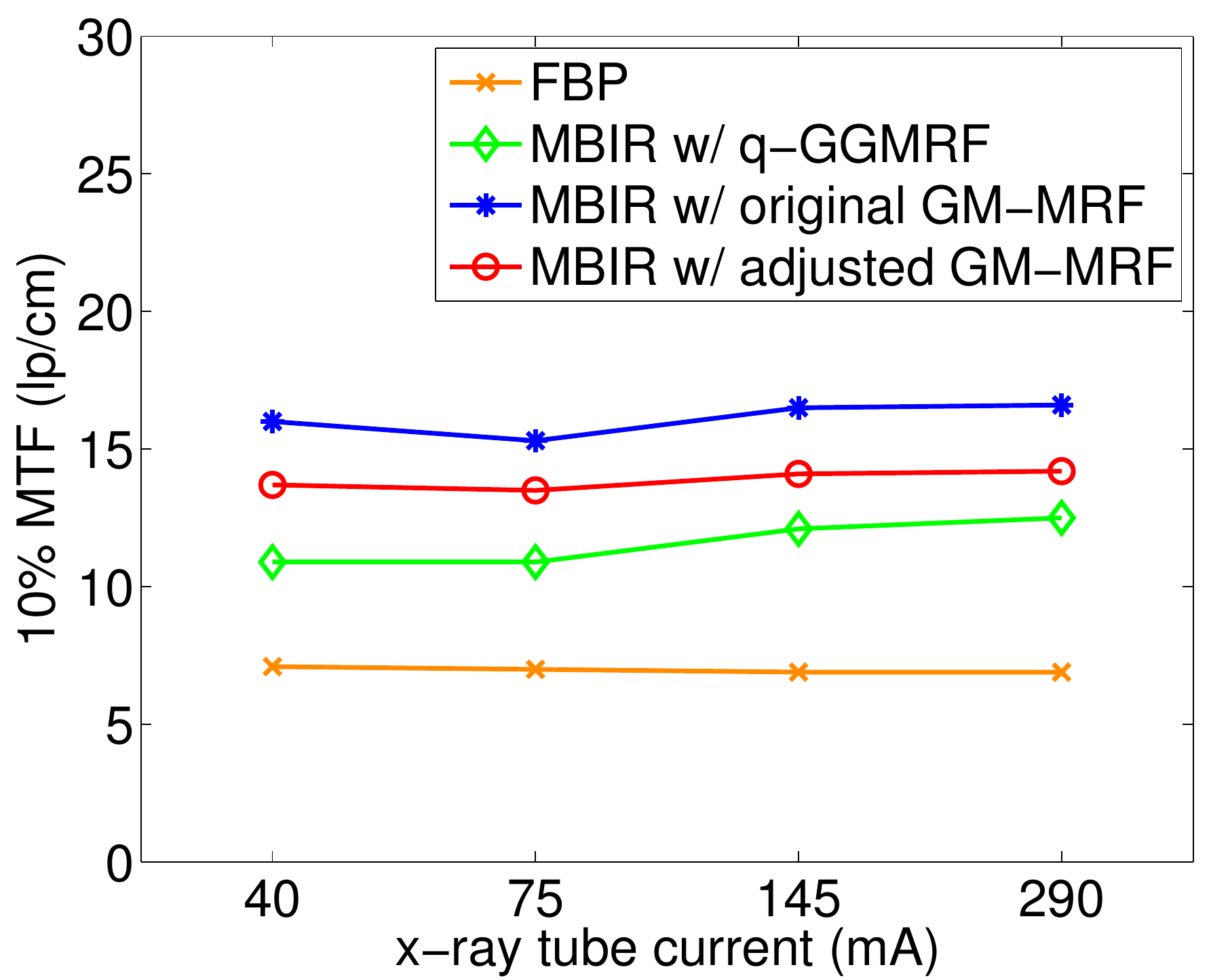}}}
\caption{Quantitative measurements for GEPP reconstructions. 
Four different magnitudes of X-ray tube current were used in data acquisition to achieve different X-ray dose levels.
The mean values along with noise were measured within two different ROIs in (a).
The MTF values were measured at the tungsten wire.
Figure demonstrates that MBIR with $5\times5\times3$ GM-MRF priors improve the in-plane resolution in (f) 
while producing comparable or even less noise than FBP and MBIR with the $q$-GGMRF prior in (d) and (e),
without affecting the reconstruction accuracy in (b) and (c).
}
\label{fig:gepp_measurement}
\end{figure*}

\subsection{3-D phantom reconstruction} 

Fig.~\ref{fig:gepp_290ma} shows the GEPP reconstruction under normal X-ray dosage,
with zoomed-in images for the tungsten wire and cyclic bars.
It shows that MBIR with the traditional $q$-GGMRF prior produces sharper images with less noise than FBP,
as indicated by smoother homogeneous regions, a smaller reconstructed wire, and more enhanced cycling bars.
As a further improvement, MBIR with the $5\times5\times3$ GM-MRF priors produce even sharper image than MBIR with the $q$-GGMRF prior at a comparable noise level.
The GM-MRF priors also improve the texture in smooth regions over the $q$-GGMRF method 
by reducing the speckle noise and grainy texture.
For the GM-MRF priors, the original model shows a sharper tungsten wire as compared to the adjusted model, 
since the adjusted model increases regularization for high-contrast edge (group 5) and bone (group 6), as shown in Fig.~\ref{fig:covariance_adjusted}.
However, the limited regularization for high-contrast edge and bone in the original model also leads to noisy rendering of high-attenuation objects,
such as the non-circular tungsten wire and irregularly shaped small metal insertion.

Fig.~\ref{fig:profile_wire} presents a profile line through the tungsten wire. As compared to the $q$-GGMRF result, 
there is clearly an undershoot near the wire in the original GM-MRF result, 
which, however, is substantially reduced by using the adjusted GM-MRF. 
We believe that this type of artifact is caused by the under-regularization of high-contrast regions in the original GM-MRF model, 
and can be mitigated by increasing the regularization in those regions using the proposed covariance scaling method. 
To further improve the result, one may continue increasing the regularization for high-contrast regions 
by using the proposed systematic approach, 
or simply change the regularization for mixture components associated with high-contrast regions. 
Additionally, notice that increasing the regularization for high-contrast regions tends to reduce the contrast; 
however, the resultant contrast still remains more substantial than the $q$-GGMRF result.

\begin{figure*}[!t]
\centerline{ 
\subfloat[different patch sizes, 290 mA]{\includegraphics[width=2.7in,trim={0 0.05in 0 0},clip]{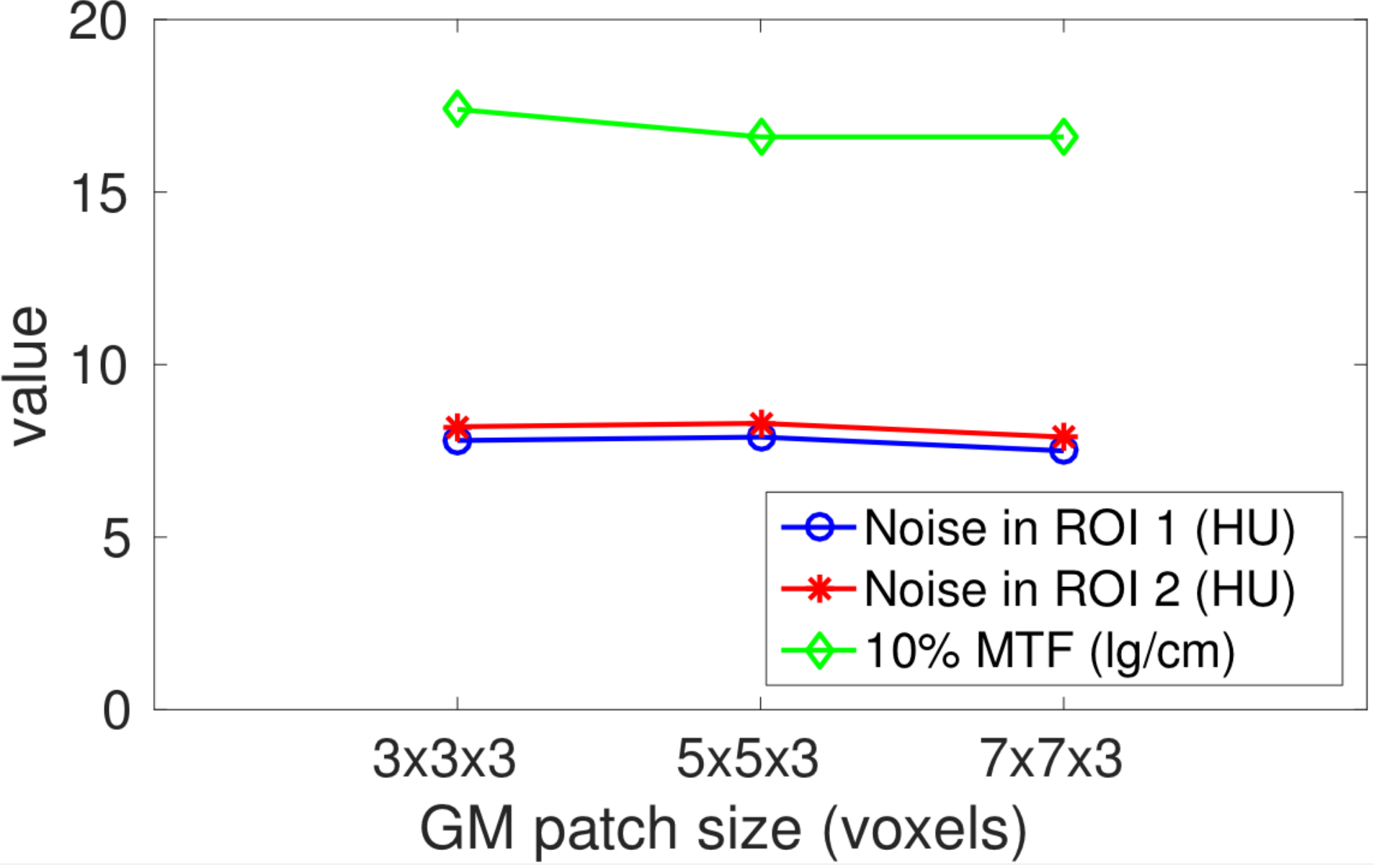}}\qquad
\subfloat[different patch sizes, 40 mA]{\includegraphics[width=2.7in,trim={0 0.05in 0 0},clip]{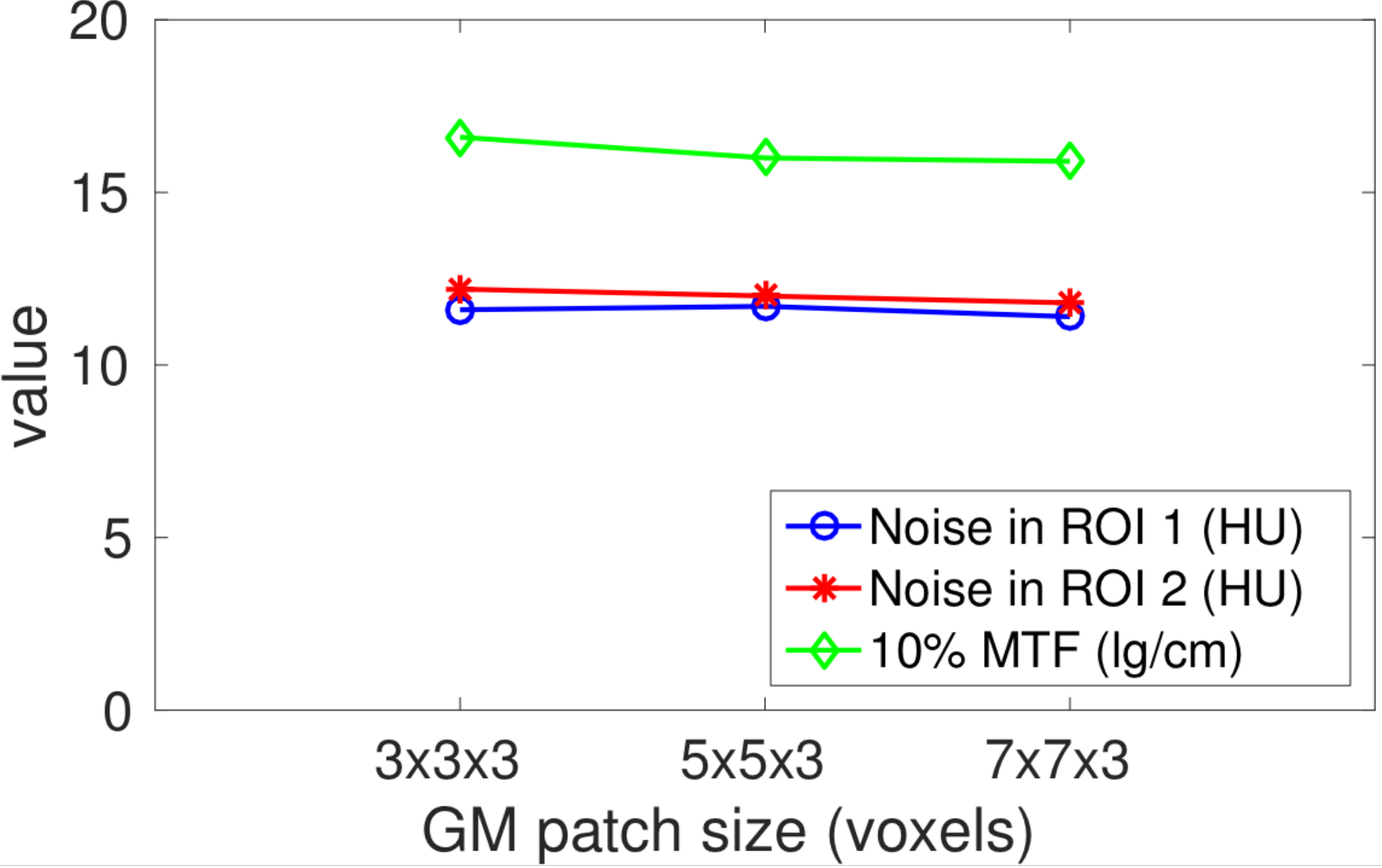}}}
\caption{Quantitative measurements for GEPP reconstructions with different patch sizes with
(a) 290 mA data and (b) 40 mA data.
With matched noise level in homogeneous regions, 
trained GM-MRF models with various patch sizes can achieve similar high-contrast resolution.}
\label{fig:gepp_measurement_diffsize}
\end{figure*}

\setlength\tabcolsep{0in}
\begin{figure*}[!t]
\centering
\begin{tabular}{cC{1.42in}C{1.42in}C{1.42in}C{1.42in}C{1.42in}} 
\rotatebox[origin=l]{90}{\qquad \quad 290 mA} &
\includegraphics[width=1.4in]{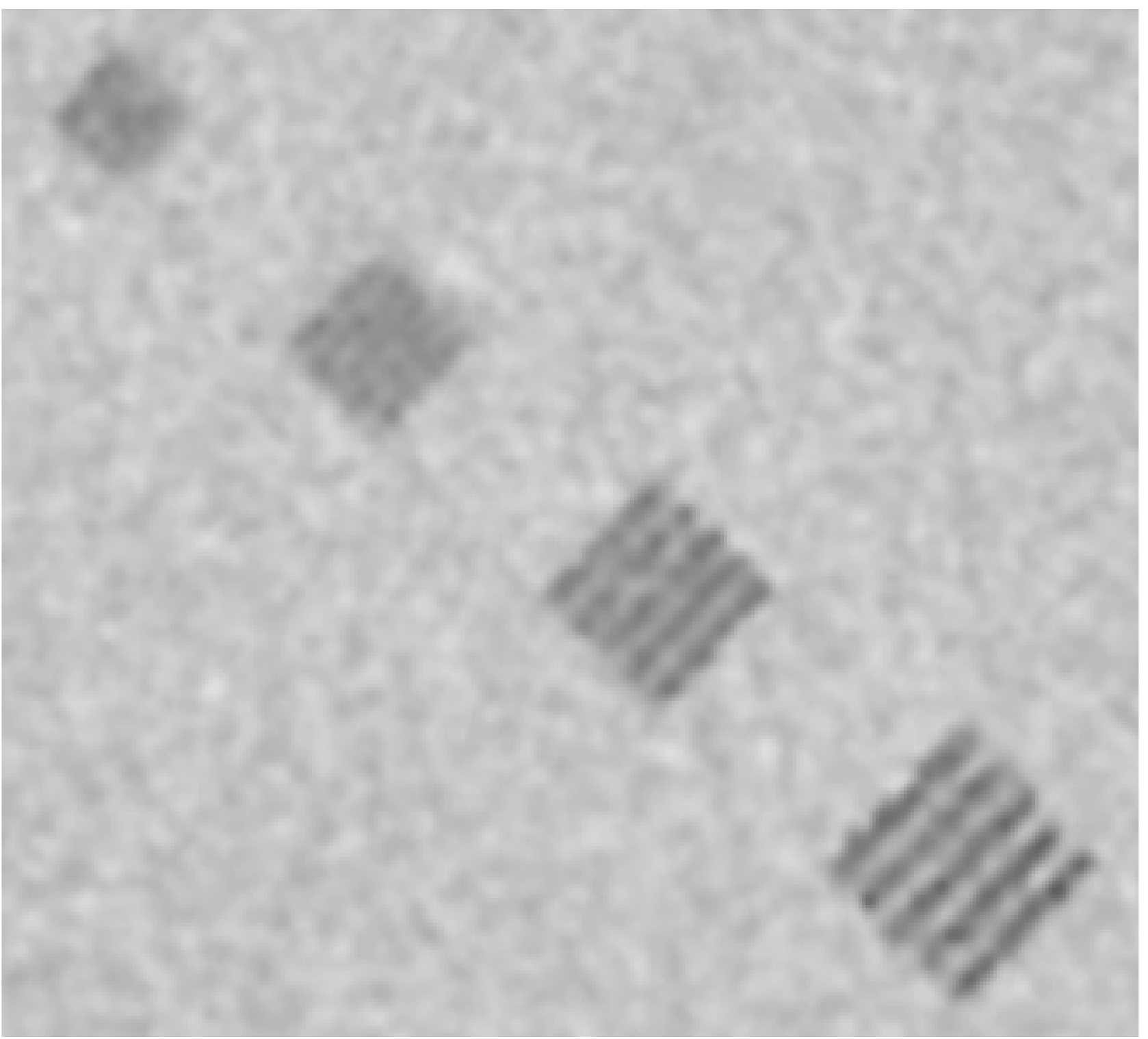} &
\includegraphics[width=1.4in]{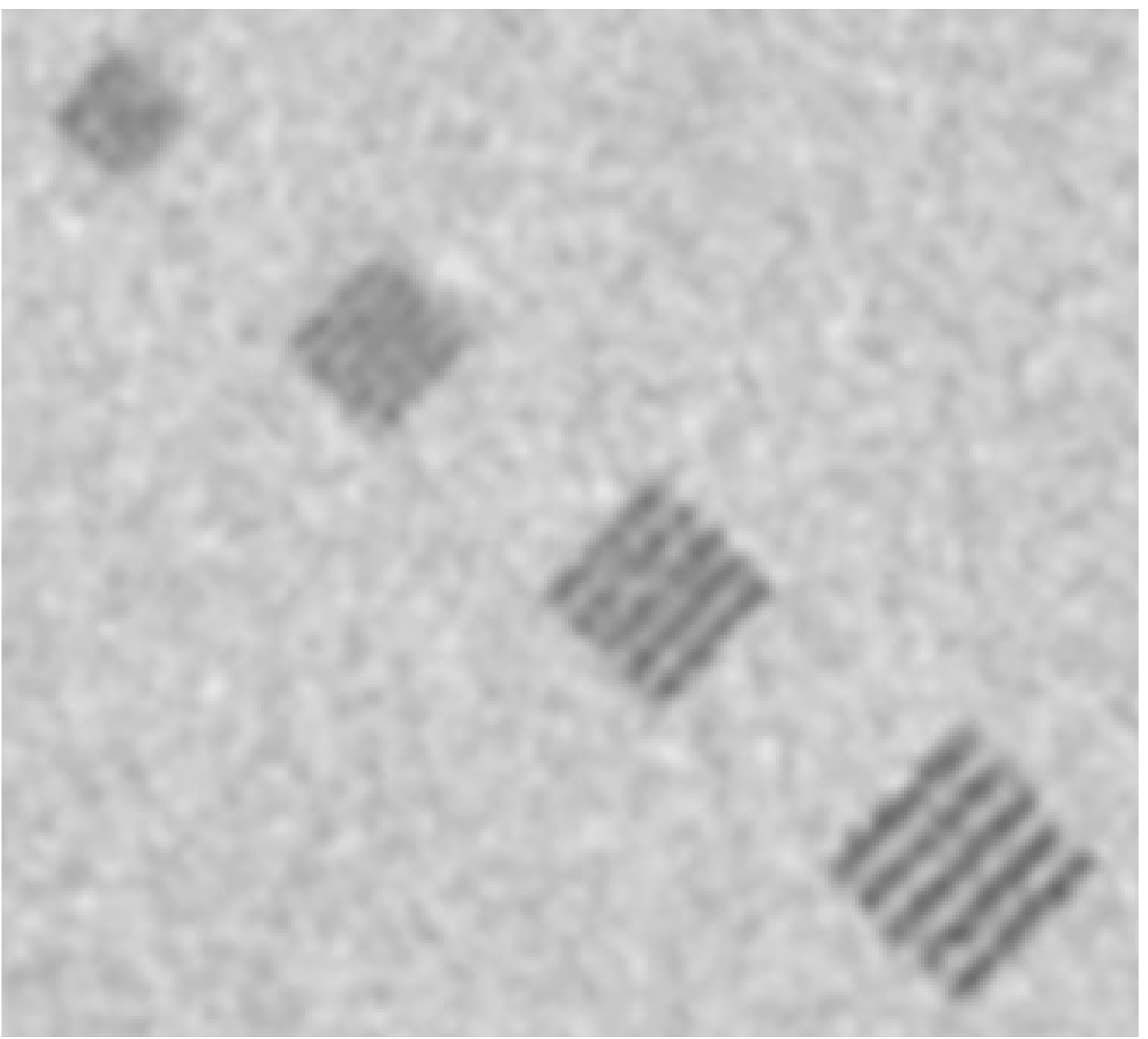} &
\includegraphics[width=1.4in]{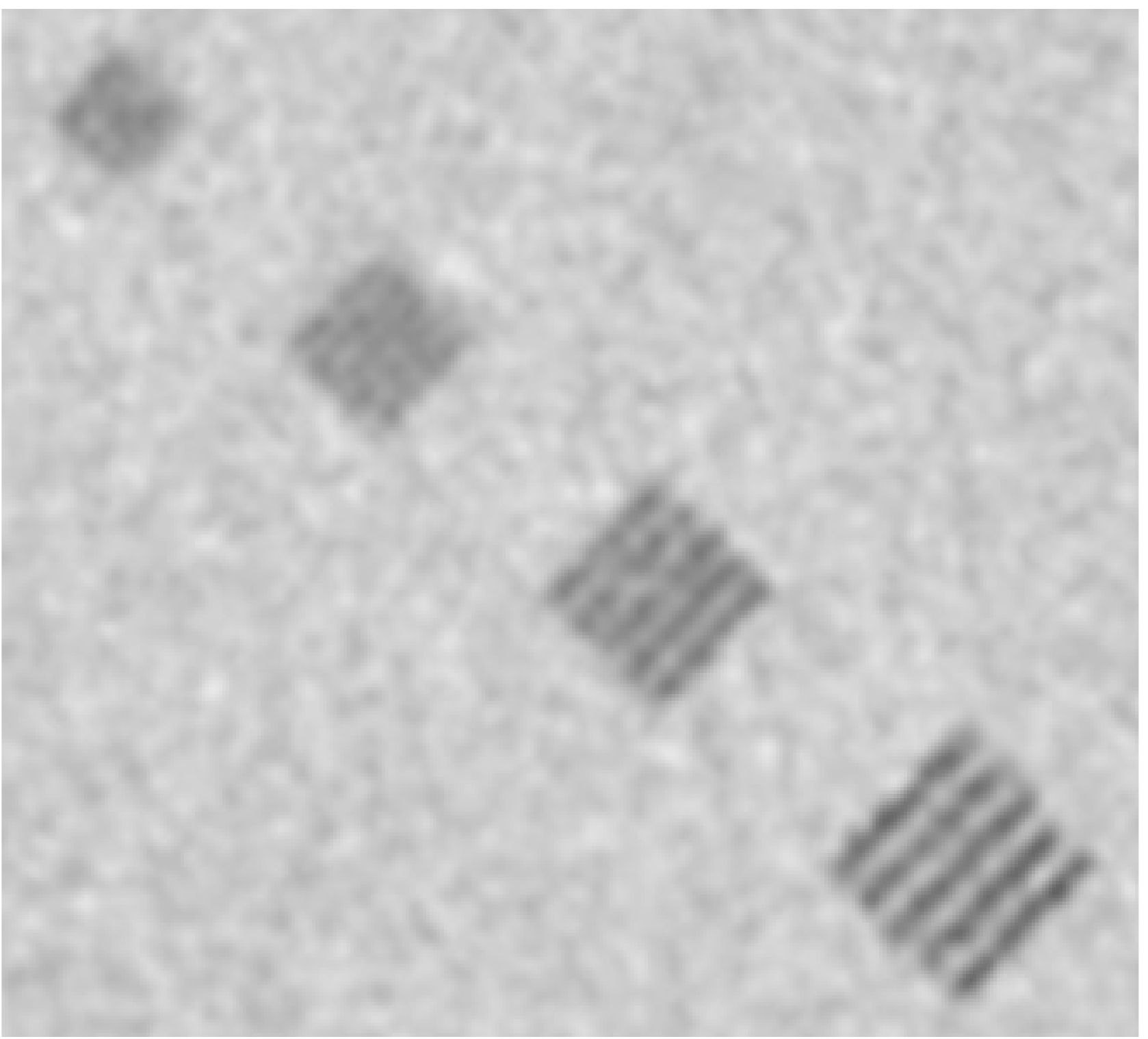} &
\includegraphics[width=1.4in]{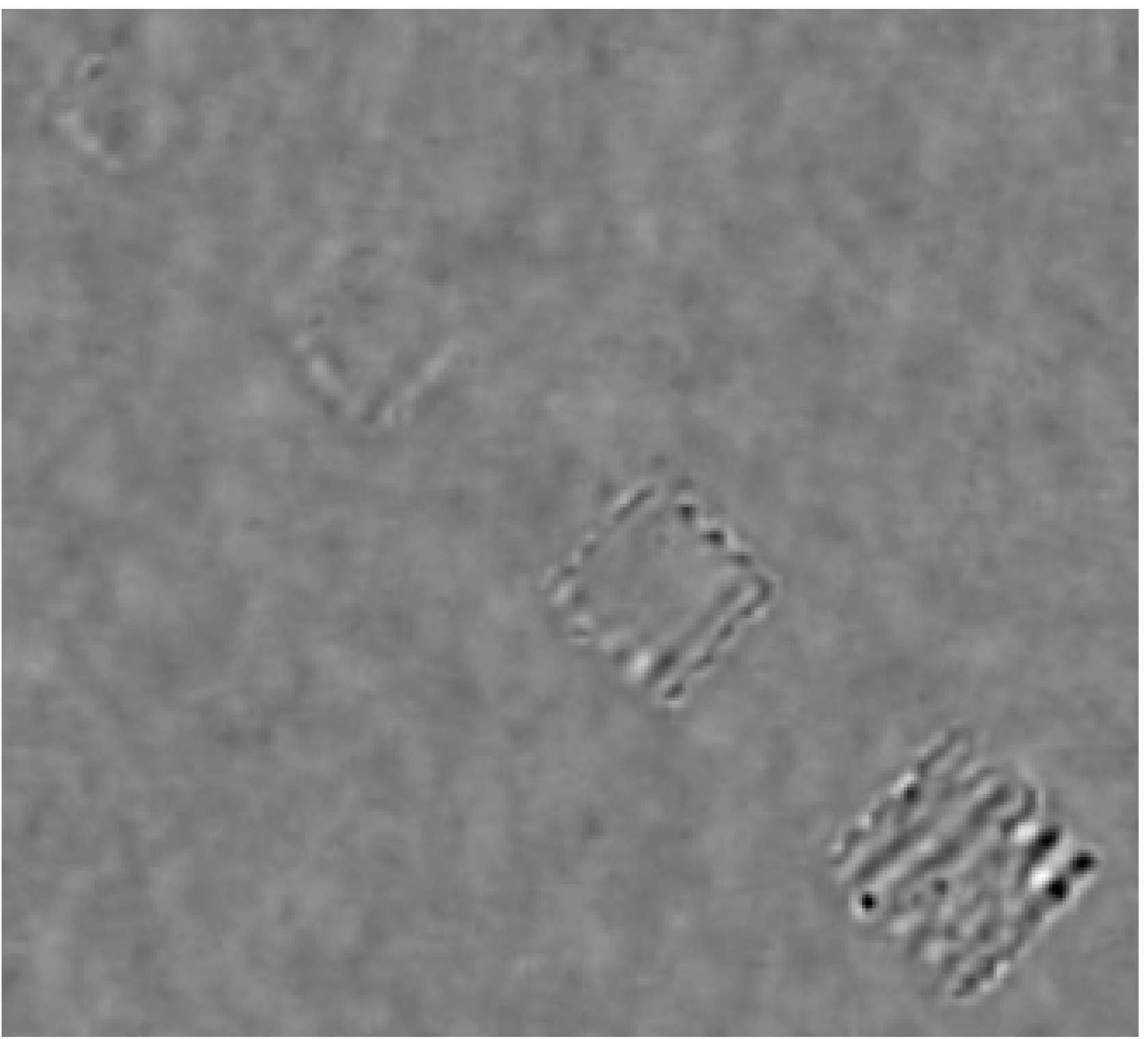} &
\includegraphics[width=1.4in]{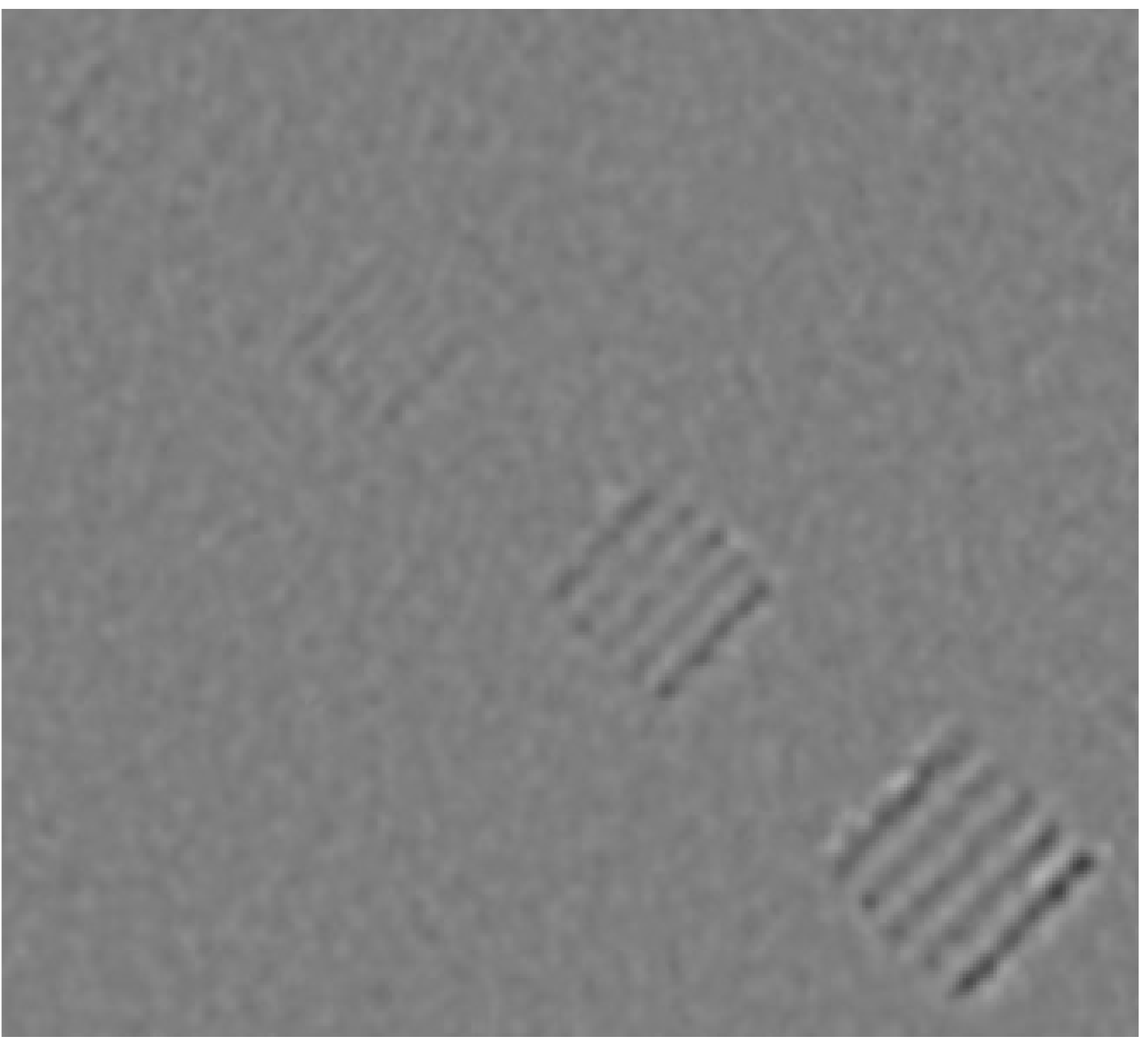}  \\
\rotatebox[origin=l]{90}{\qquad \quad 40 mA} &
\includegraphics[width=1.4in,trim={0 0.15in 0 0},clip]{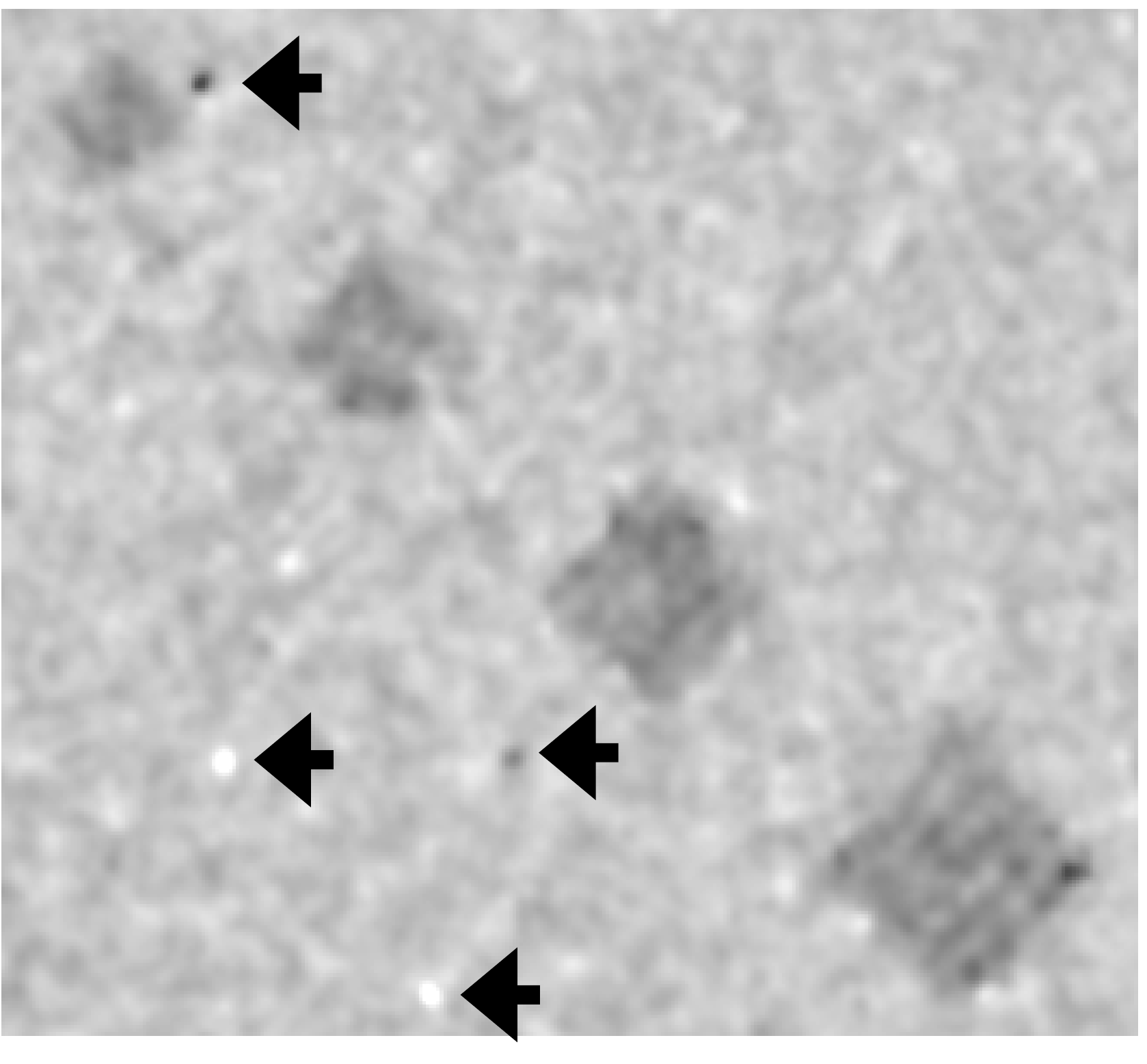} &
\includegraphics[width=1.4in,trim={0 0.15in 0 0},clip]{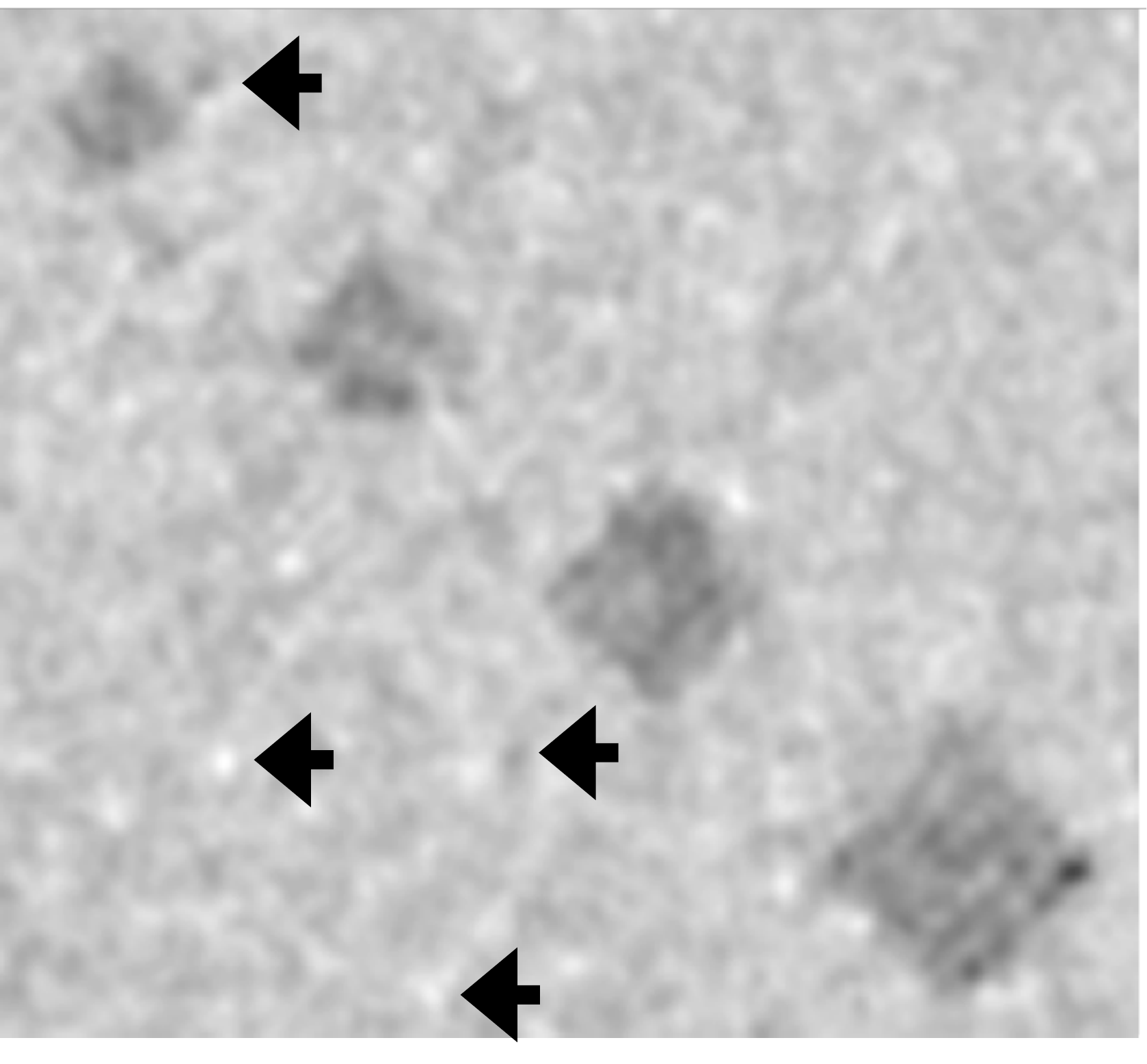} &
\includegraphics[width=1.4in]{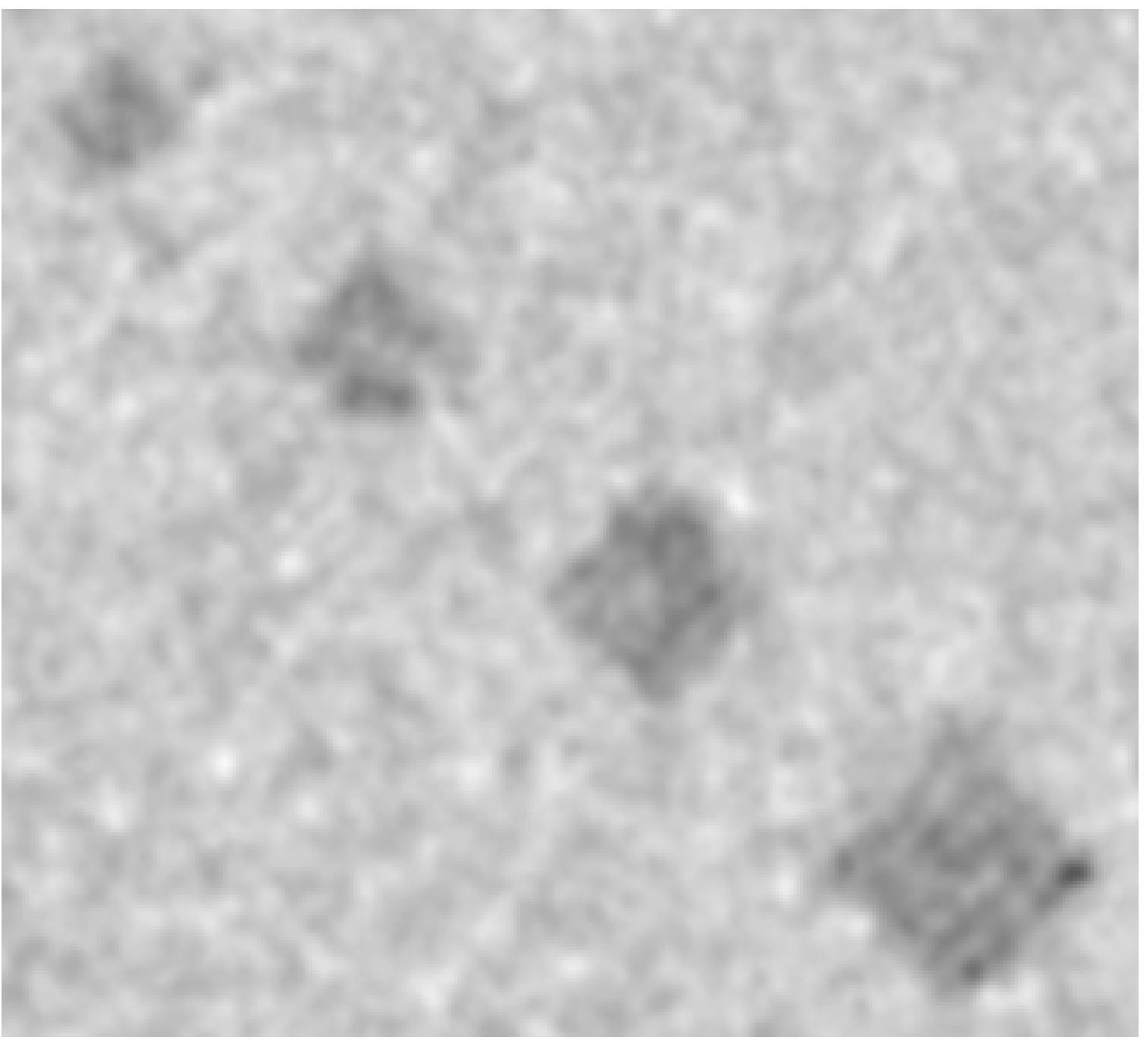} &
\includegraphics[width=1.4in,trim={0 0.15in 0 0},clip]{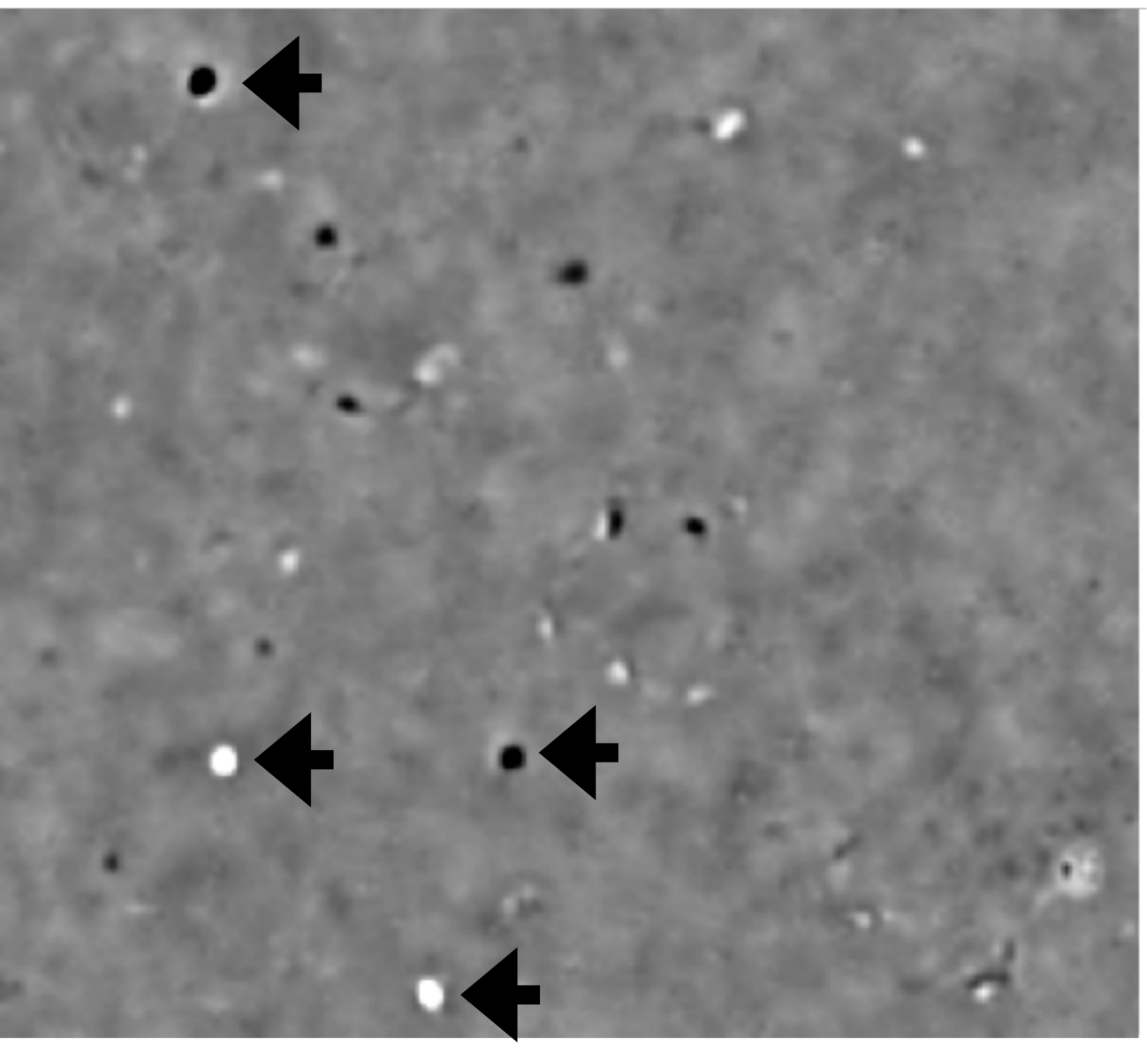} &
\includegraphics[width=1.4in]{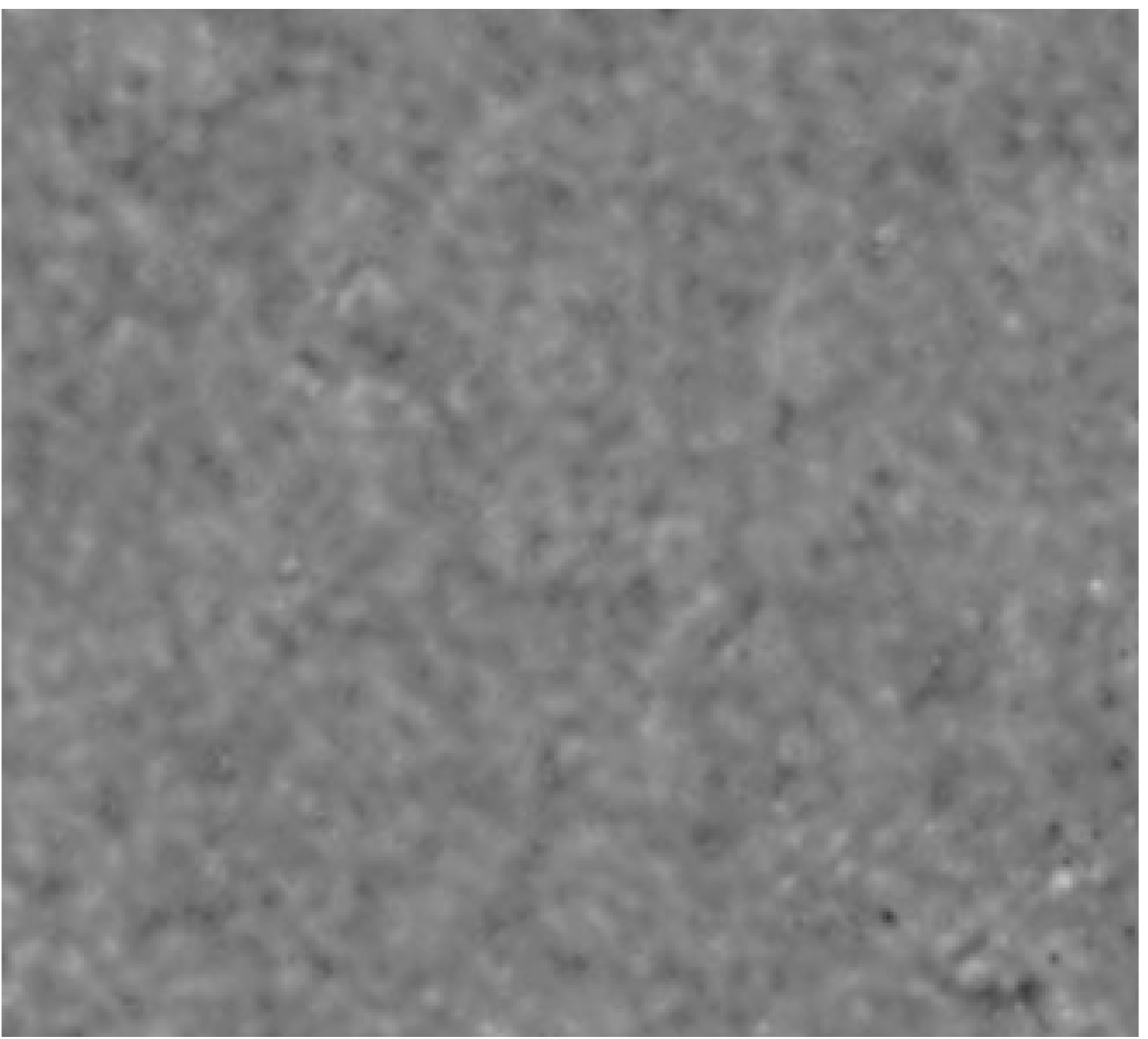}  \\
 & (a) $3\times3\times3$ patch & (b) $5\times5\times3$ patch & (c) $7\times7\times3$ patch & 
 (d) Diff: (a) -- (b) & (e) Diff: (b) -- (c)
 \end{tabular}
\caption{GEPP reconstructions with 66-component GM-MRF model with different sizes of patch models. 
Individual image is zoomed to a small FOV containing cyclic bars for display purposes.
From top to bottom: GEPP reconstructions with 290 mA data and 40 mA data, respectively. 
Display window: for GEPP images: [-110 190] HU; for difference images: [-15 15] HU.
The $3\times3\times3$ result with 40 mA data contains many noisy spots, as indicated by the arrows,
which are suppressed substantially in the result with larger patch models.
It is also observed that the $7\times7\times3$ result with 290 mA data has slightly blurred cyclic bars 
as compared to results with smaller patch models.}
\label{fig:gepp_diffsize}
\end{figure*}

\setlength\tabcolsep{0in}
\begin{figure*}[!t]
\centering
\begin{tabular}{C{2.2in}C{2.2in}C{2.2in}} 
\includegraphics[width=2.2in,trim={0 0.27in 0 0.23in},clip]{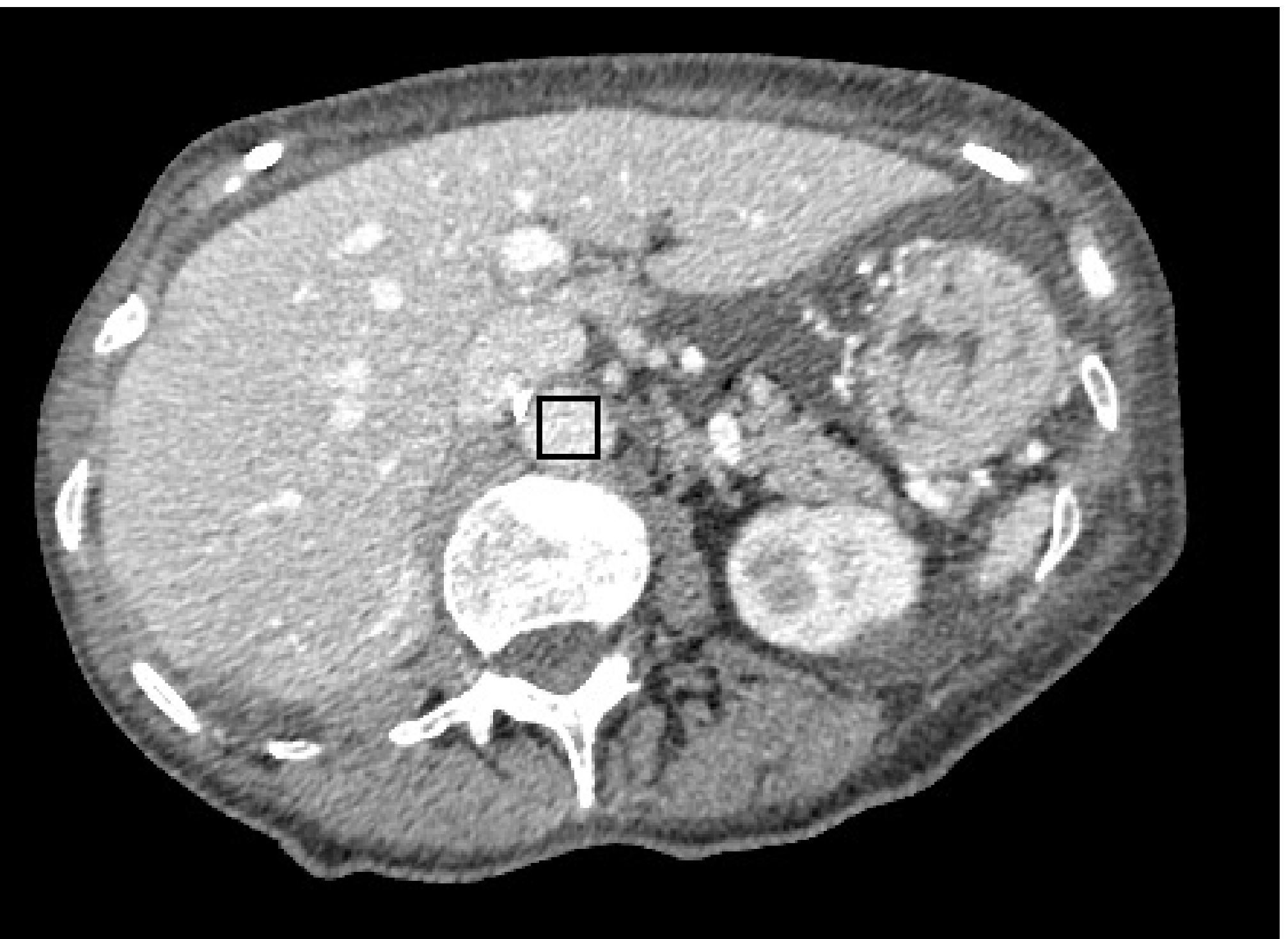} &
\includegraphics[width=2.2in,trim={0 0.27in 0 0.23in},clip]{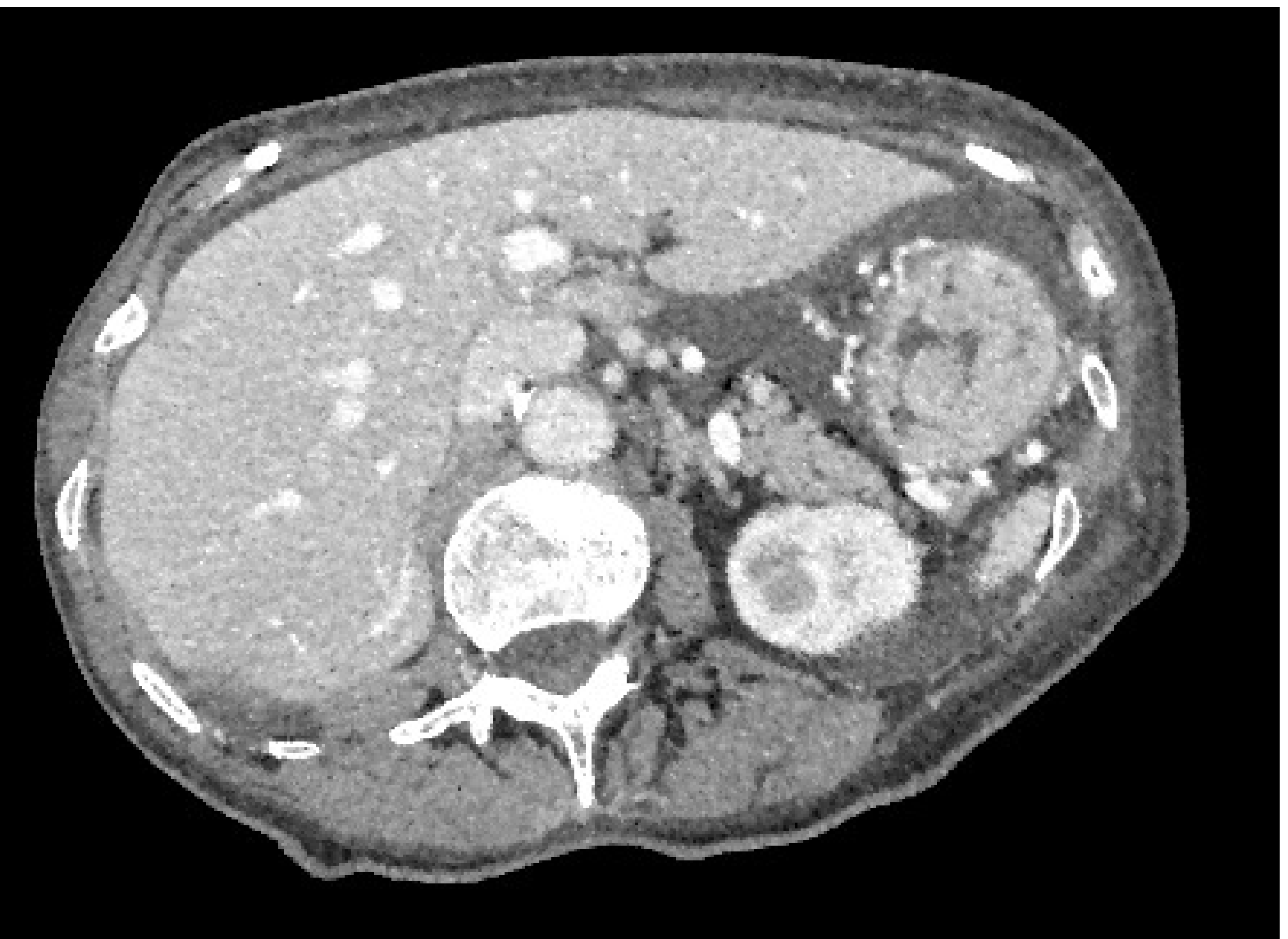} &
\includegraphics[width=2.2in,trim={0 0.27in 0 0.23in},clip]{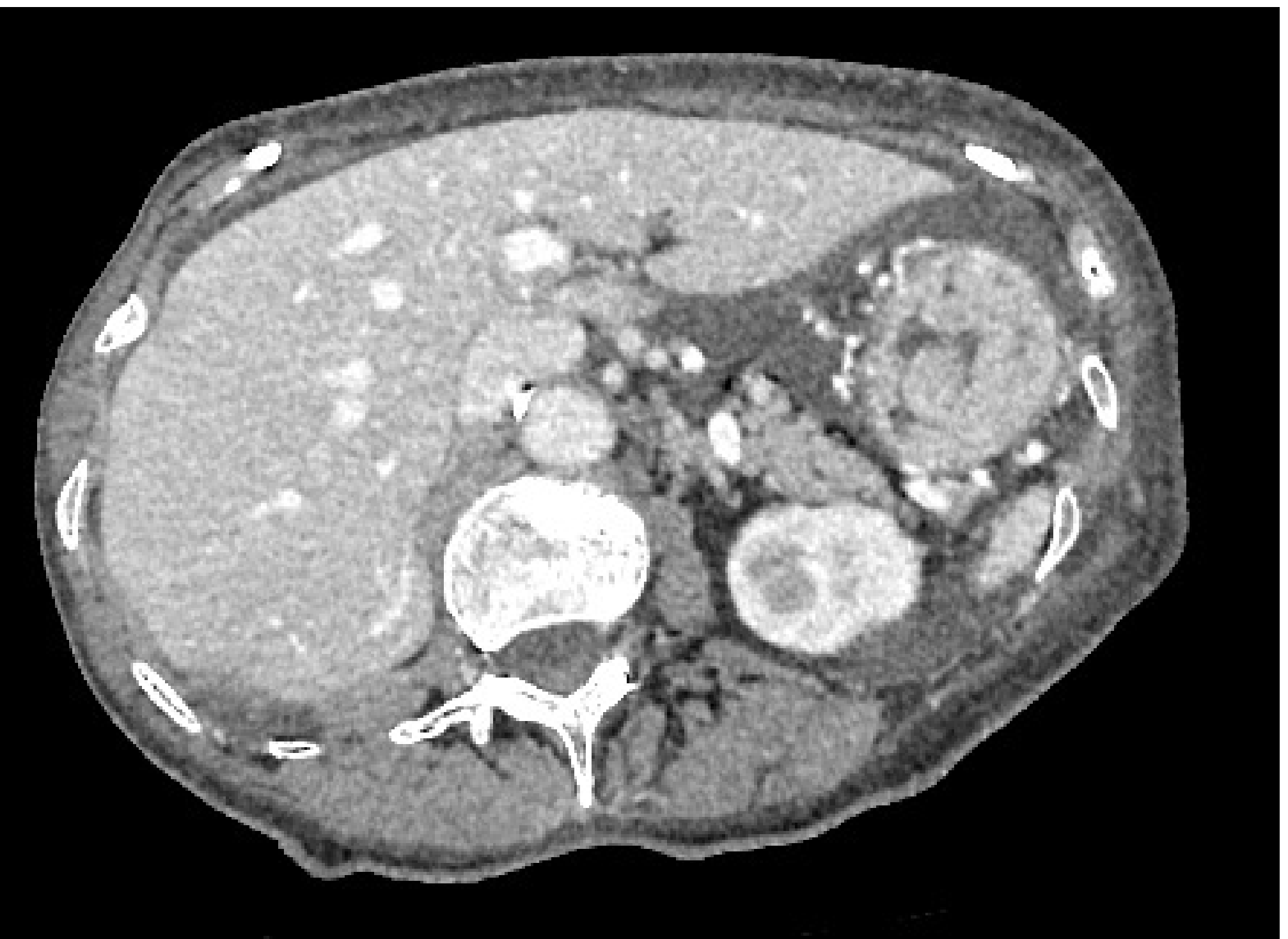} \\
\includegraphics[width=2.2in]{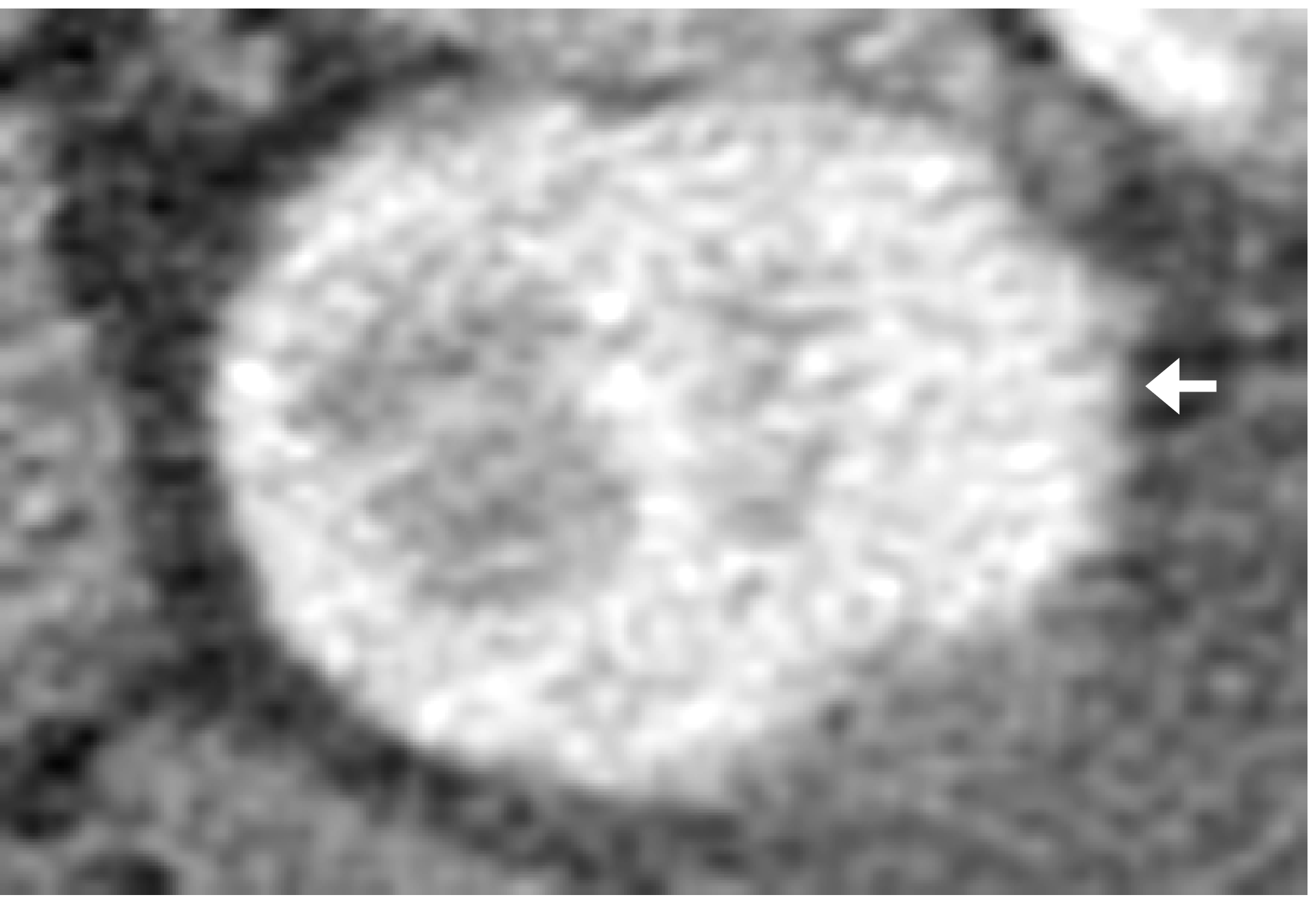} &
\includegraphics[width=2.2in]{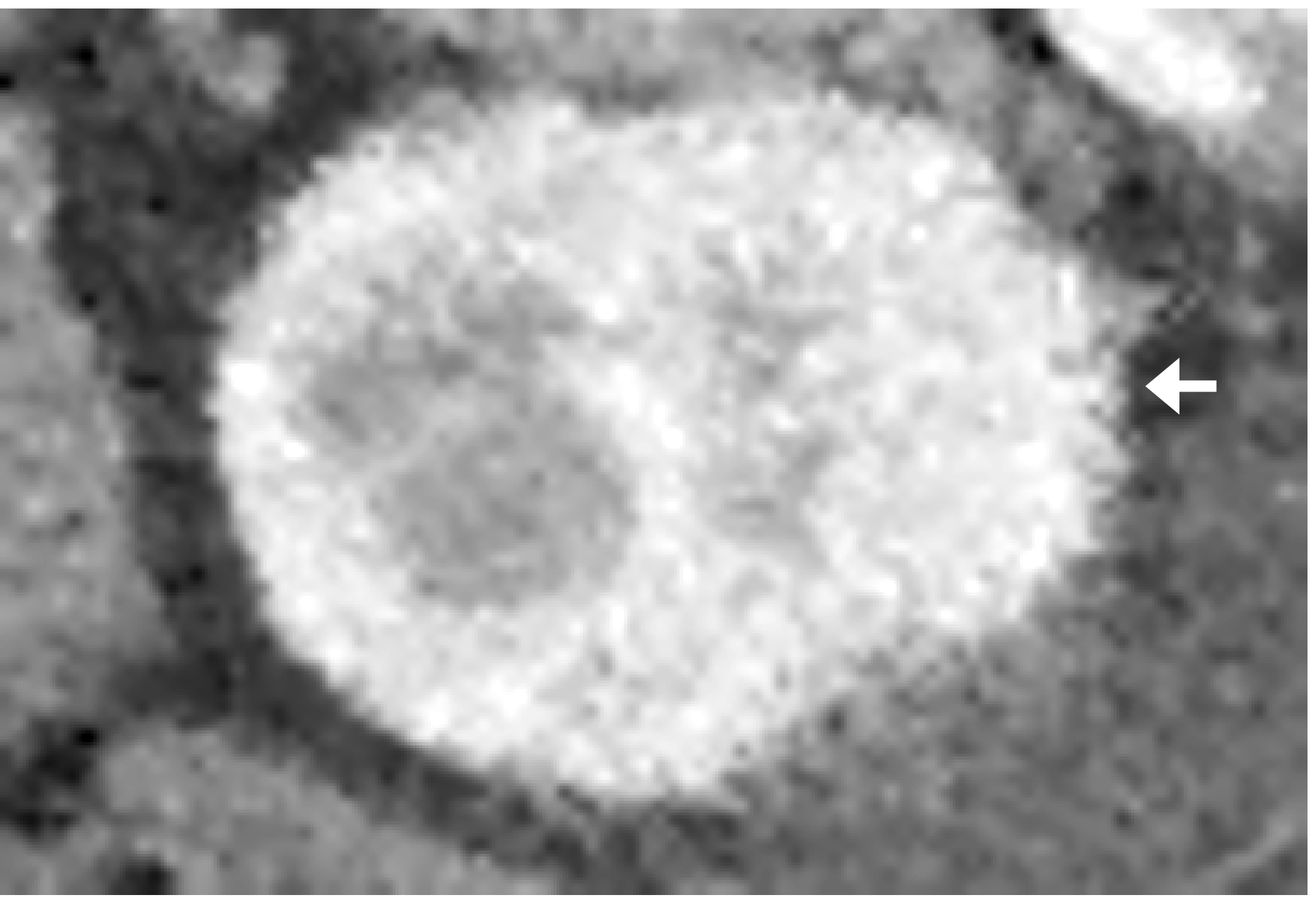} &
\includegraphics[width=2.2in]{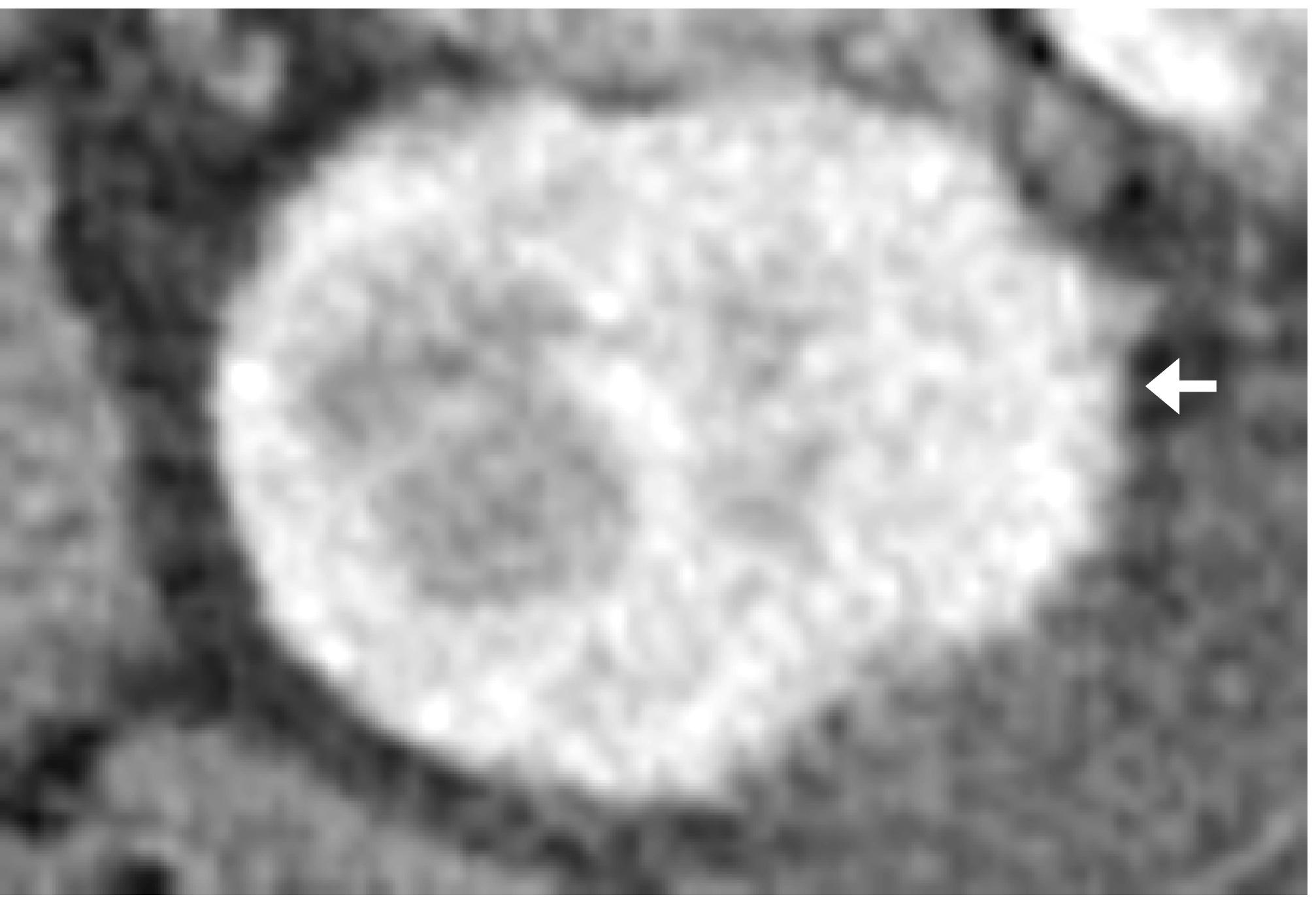} \\
 (a) FBP (19.14 HU) & (b) MBIR w/ $q$-GGMRF w/ reduced regularization (14.43~HU) & (c) MBIR w/ adjusted GM-MRF (14.02~HU)
\end{tabular}
\caption{An abdominal axial slice of the normal-dose clinical reconstruction (with noise standard deviation reported).
From left to right, the columns represent (a) FBP, (b) MBIR with $q$-GGMRF with reduced regularization, and
(c) MBIR with adjusted GM-MRF with $p=0.5, \alpha=33\ {\rm HU}$.
Top row shows the full FOV of the reconstructed images, 
while the bottom row shows a zoomed-in FOV.
Noise standard deviation is measured within an ROI in aorta, as illustrated in the FBP image, and is reported for each method.
Display window is [-110 190] HU.
Note the reduced jagged appearance in the GM-MRF reconstruction.}
\label{fig:clinical_abd}
\end{figure*}

\setlength\tabcolsep{0in}
\begin{figure*}[!t]
\centering
\begin{tabular}{C{2.2in}C{2.2in}C{2.2in}} 
\includegraphics[width=2.2in,trim={0 0 0 0.5in},clip]{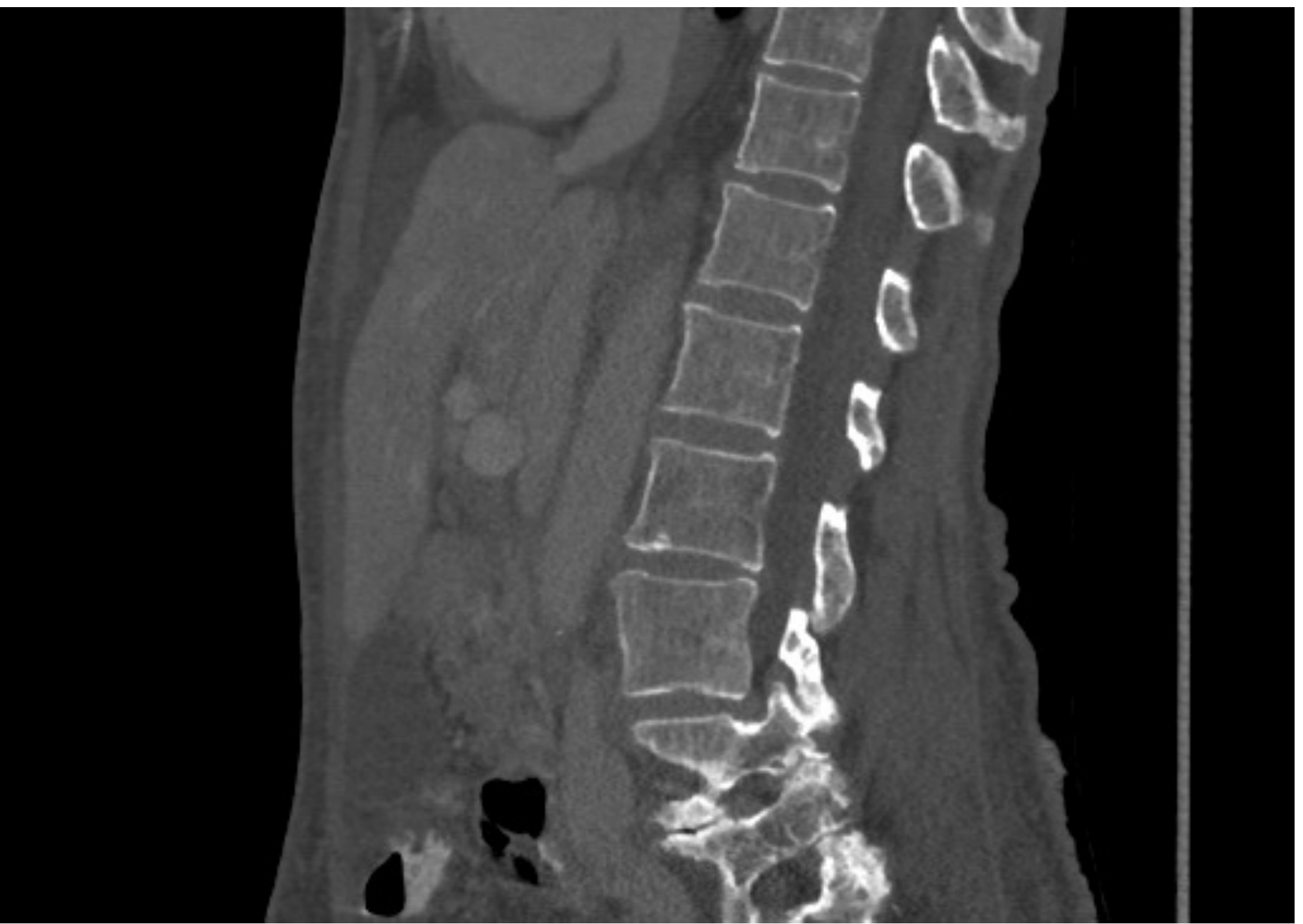} &
\includegraphics[width=2.2in,trim={0 0 0 0.5in},clip]{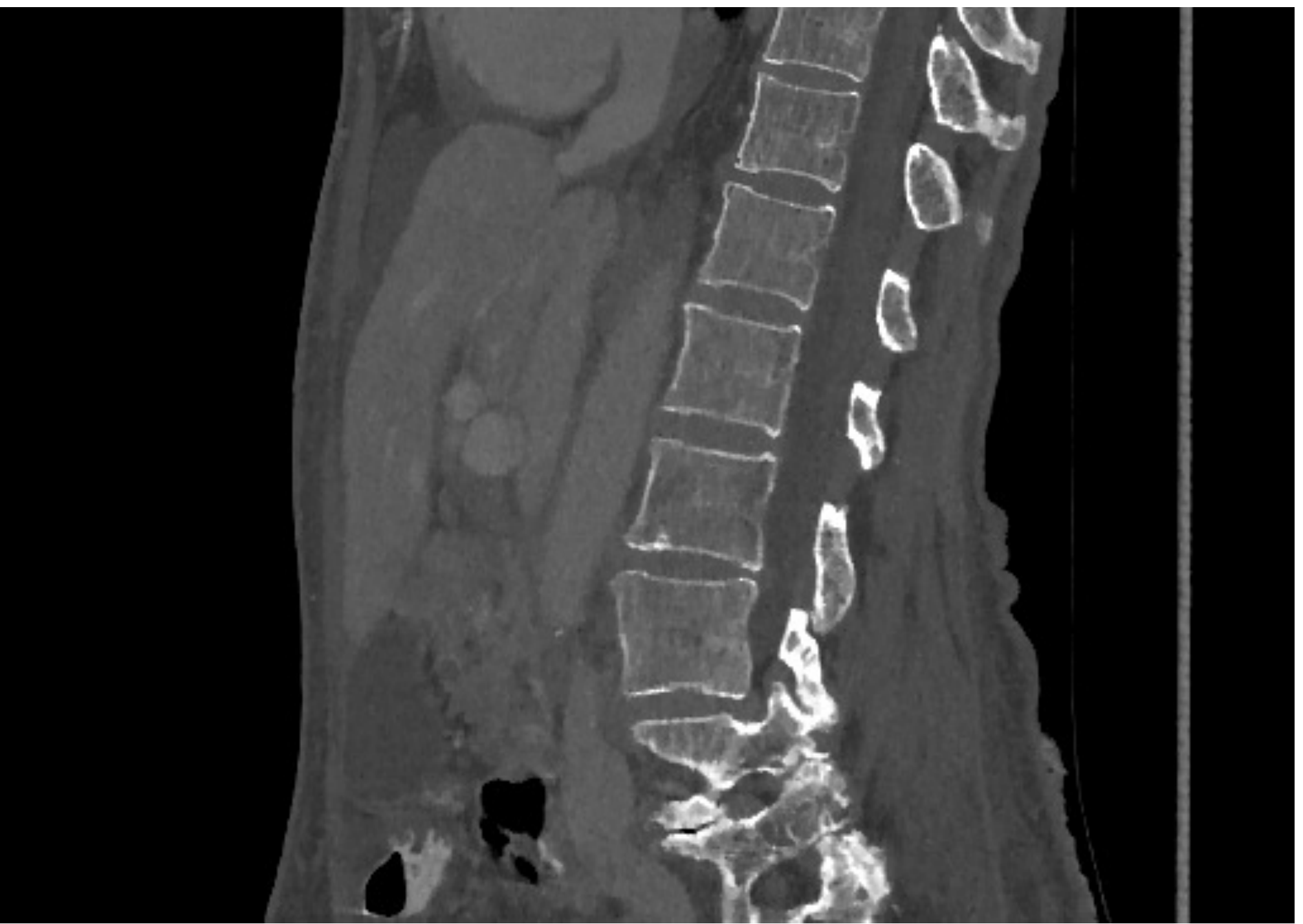} &
\includegraphics[width=2.2in,trim={0 0 0 0.5in},clip]{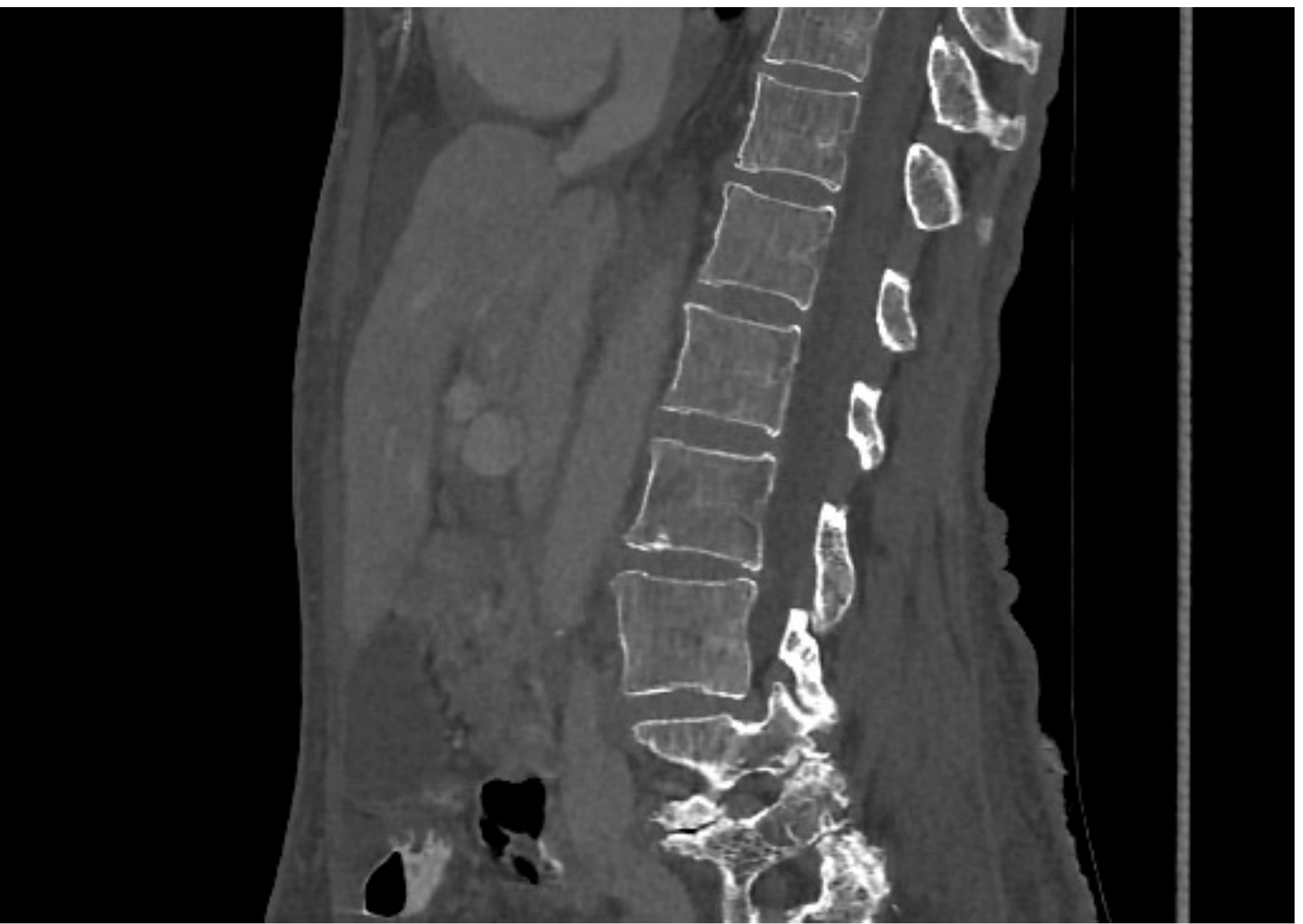} \\
\includegraphics[width=2.2in]{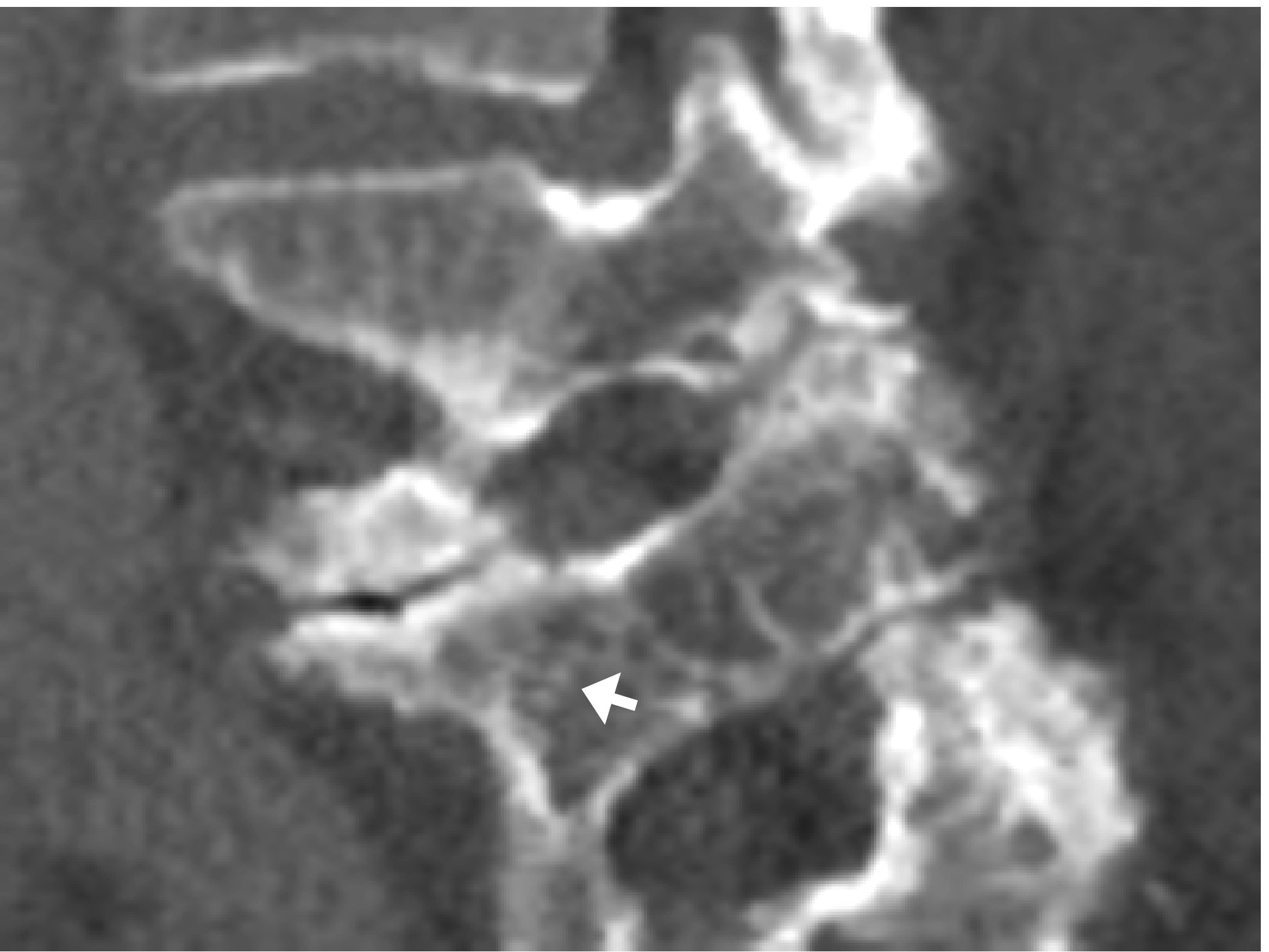} &
\includegraphics[width=2.2in]{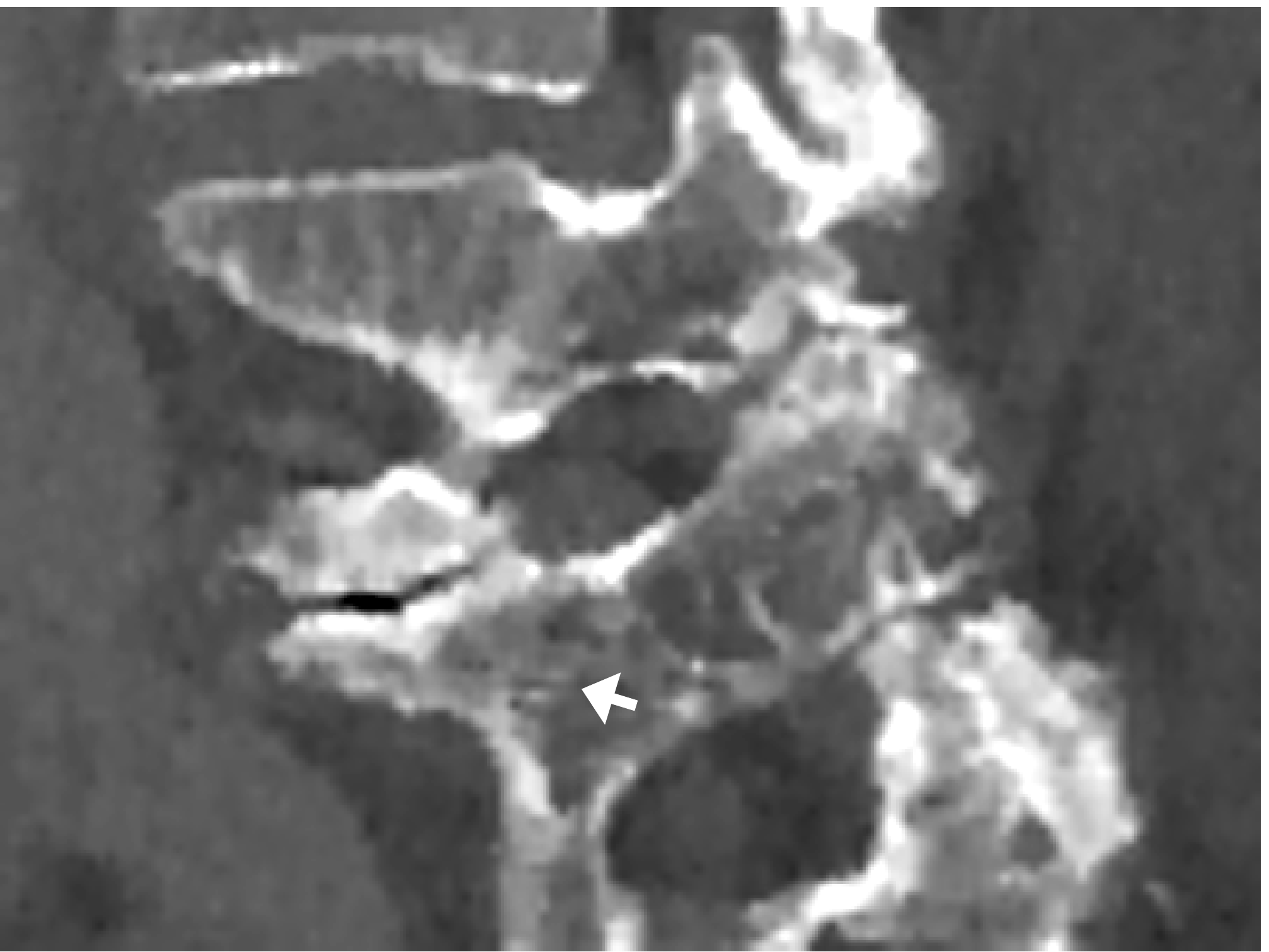} &
\includegraphics[width=2.2in]{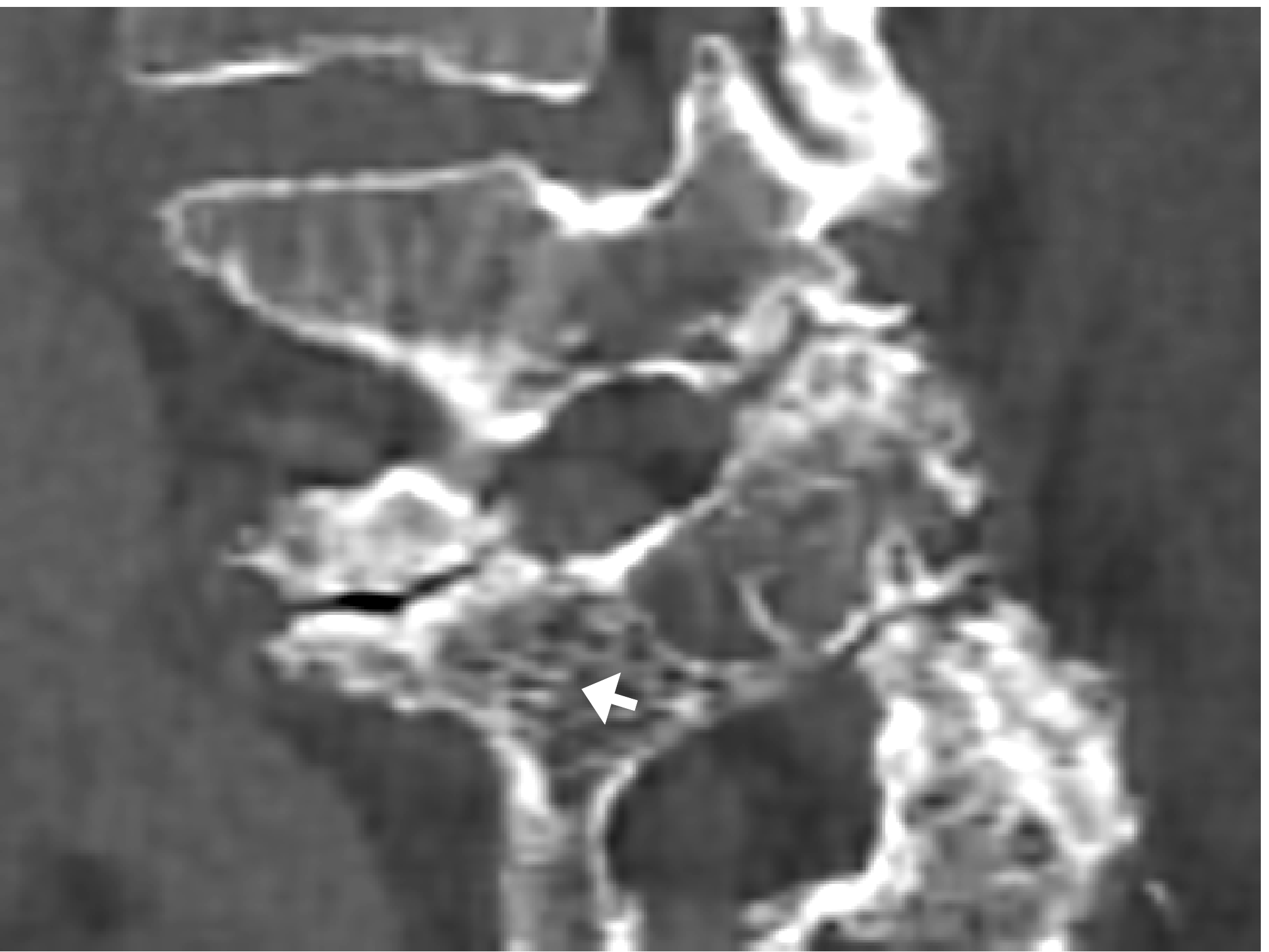} \\
 (a) FBP & (b) MBIR w/ $q$-GGMRF w/ reduced regularization  & (c) MBIR w/ adjusted GM-MRF 
\end{tabular}
\caption{A sagittal view of the normal-dose clinical reconstruction in bone window.
From left to right, the columns represent (a) FBP, (b) MBIR with $q$-GGMRF with reduced regularization, and
(c) MBIR with adjusted GM-MRF with $p=0.5, \alpha=33\ {\rm HU}$.
Top row shows the full FOV of the reconstructed images, 
while the bottom row shows a zoomed-in FOV.
Display window is \mbox{[-300 900] HU}.
Note the honeycomb structure of the trabecular bone reconstructed by the GM-MRF prior,
which is missing in other methods.}
\label{fig:clinical_sag}
\end{figure*}

\setlength\tabcolsep{0in}
\begin{figure*}[!t]
\centering
\begin{tabular}{C{2.2in}C{2.2in}C{2.2in}} 
\includegraphics[width=2.2in,trim={0 0.3in 0 0.8in},clip]{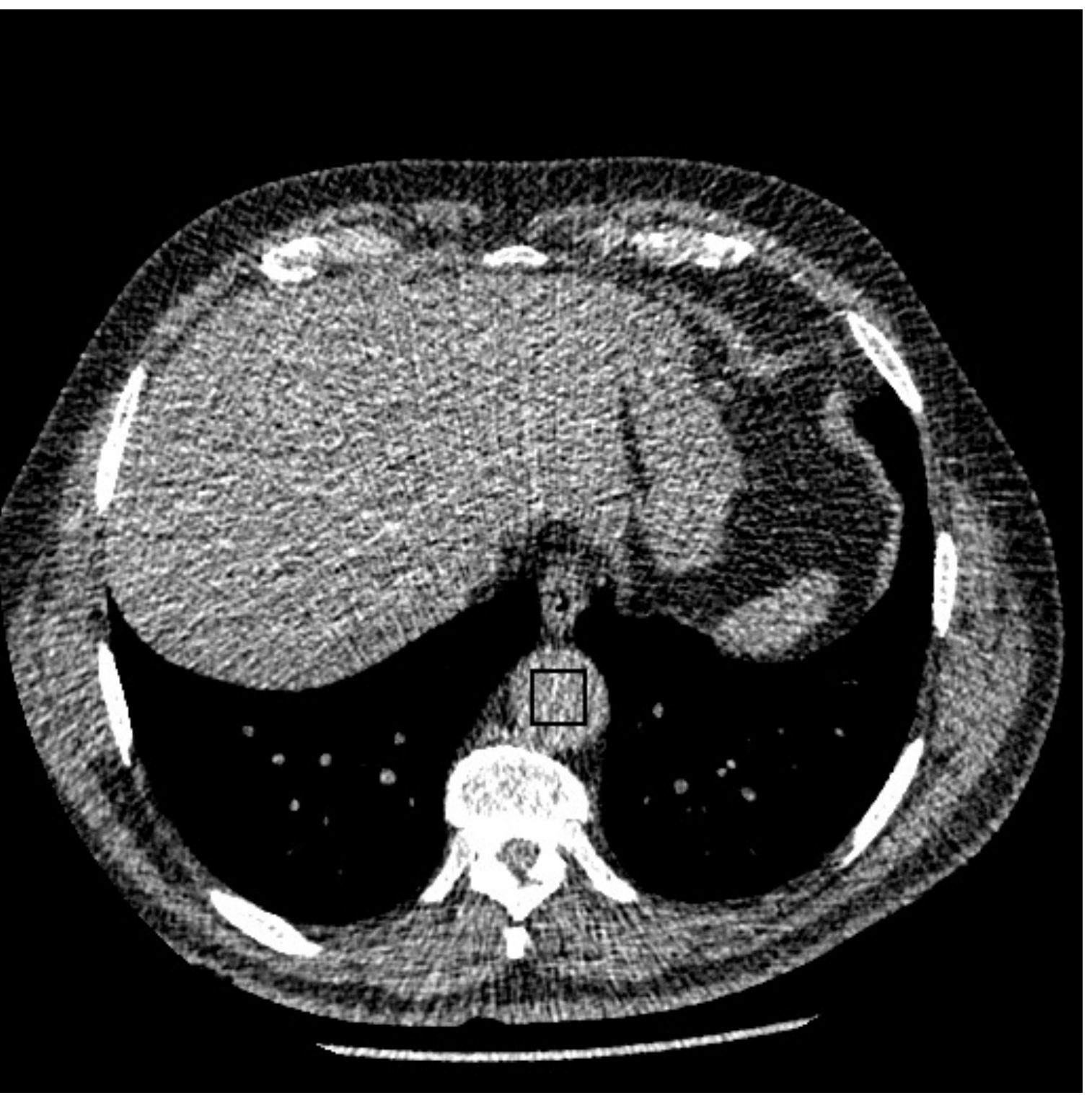} &
\includegraphics[width=2.2in,trim={0 0.3in 0 0.8in},clip]{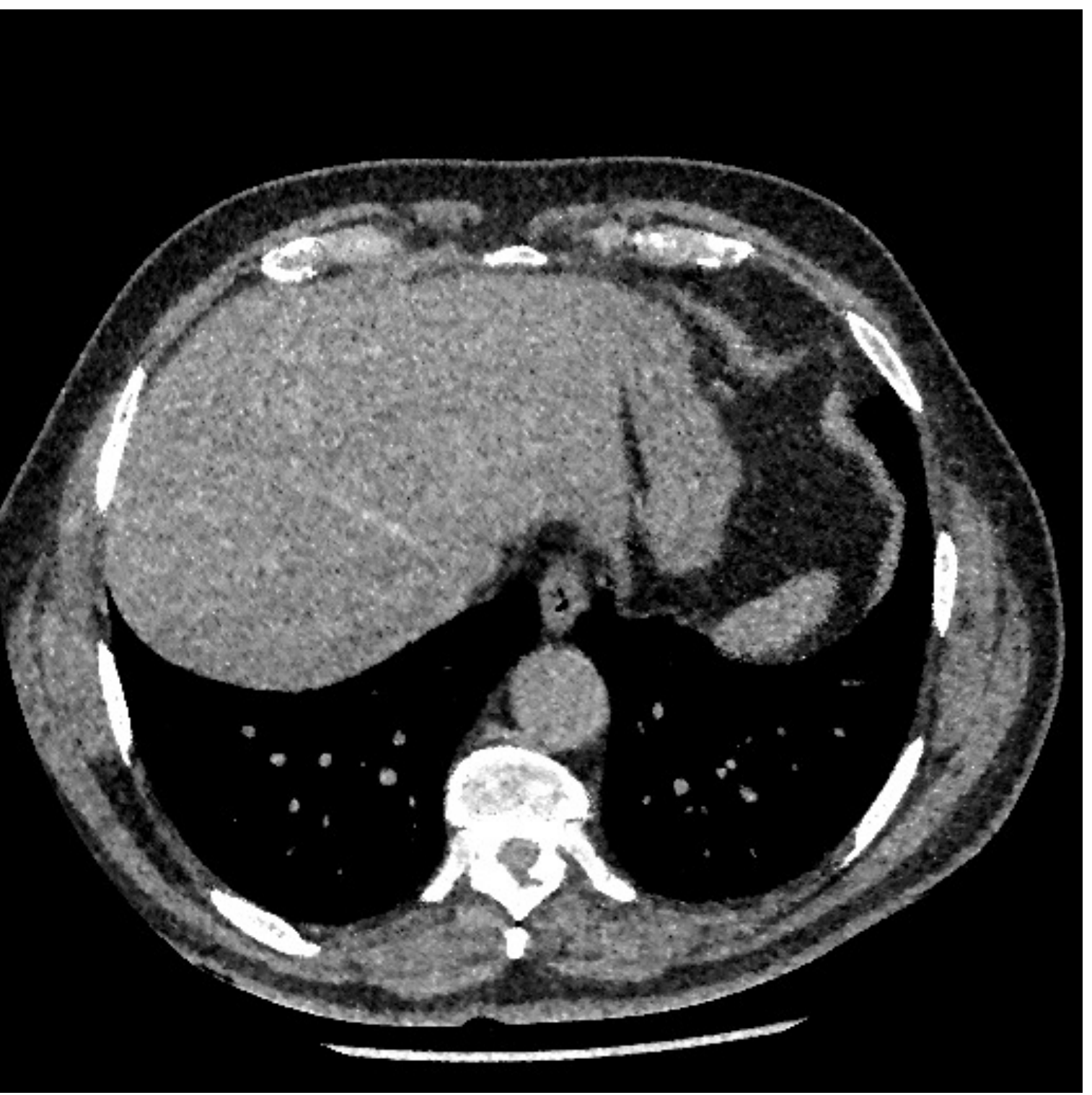} &
\includegraphics[width=2.2in,trim={0 0.3in 0 0.8in},clip]{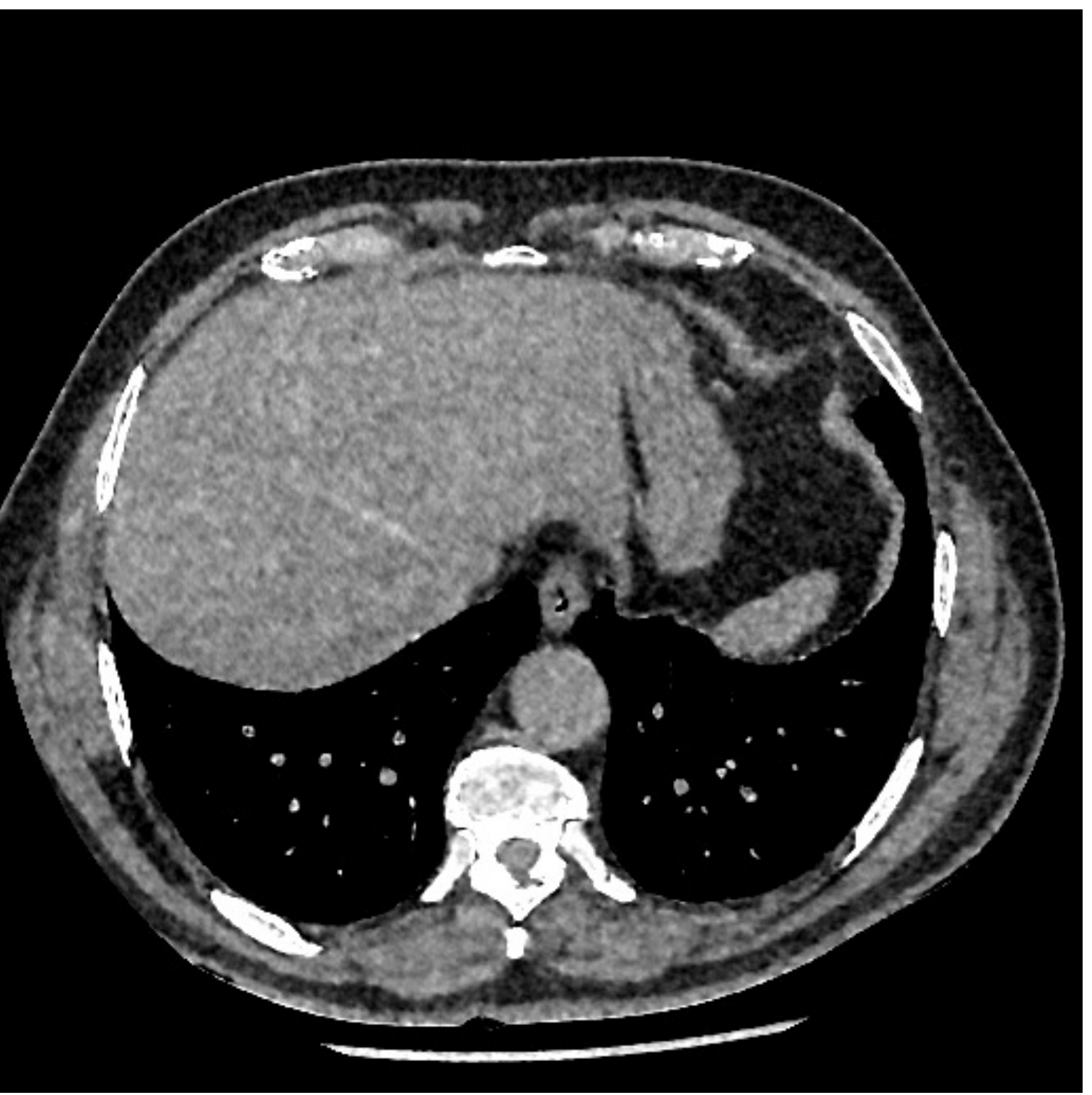} \\
\includegraphics[width=2.2in]{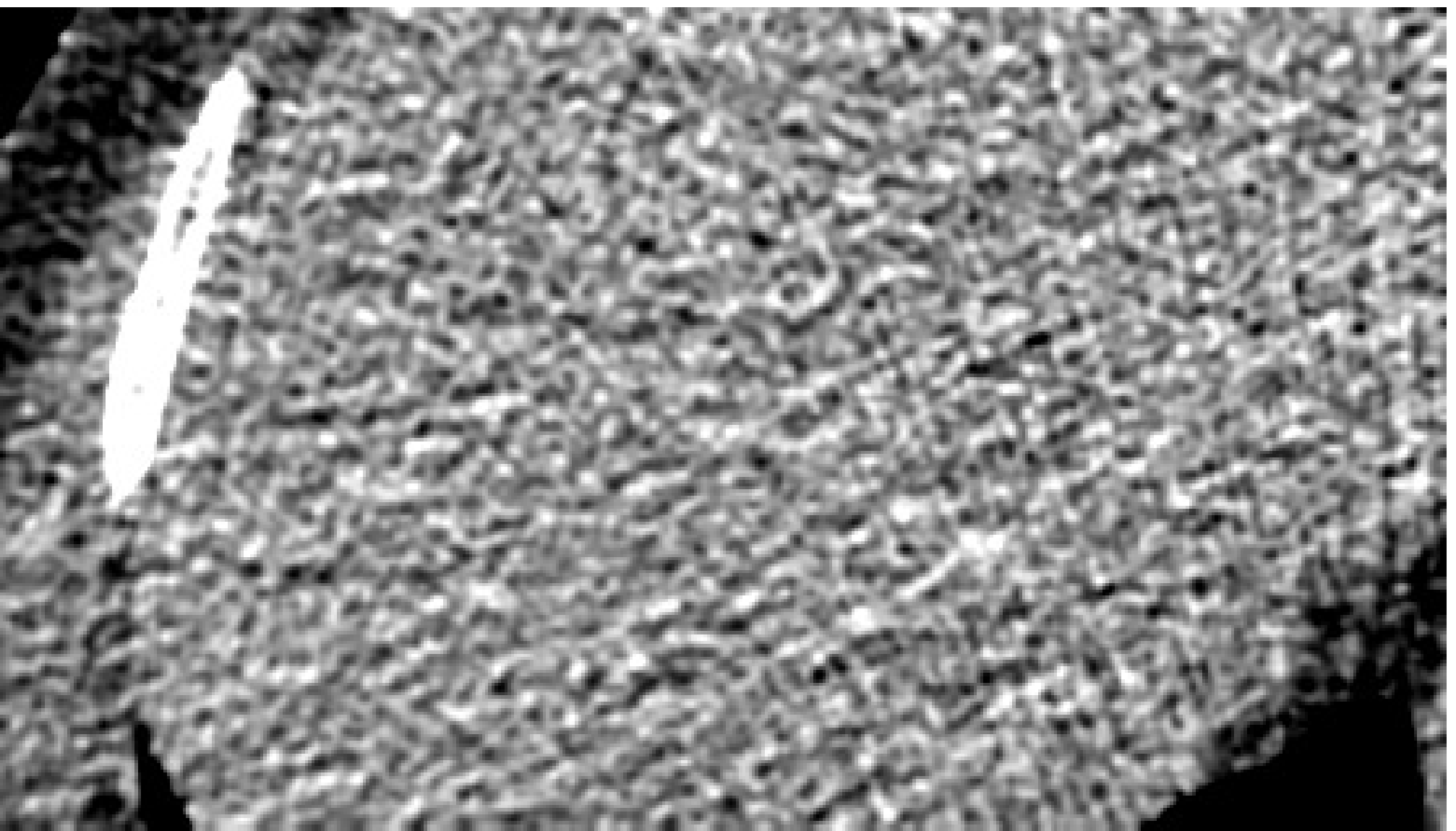} &
\includegraphics[width=2.2in]{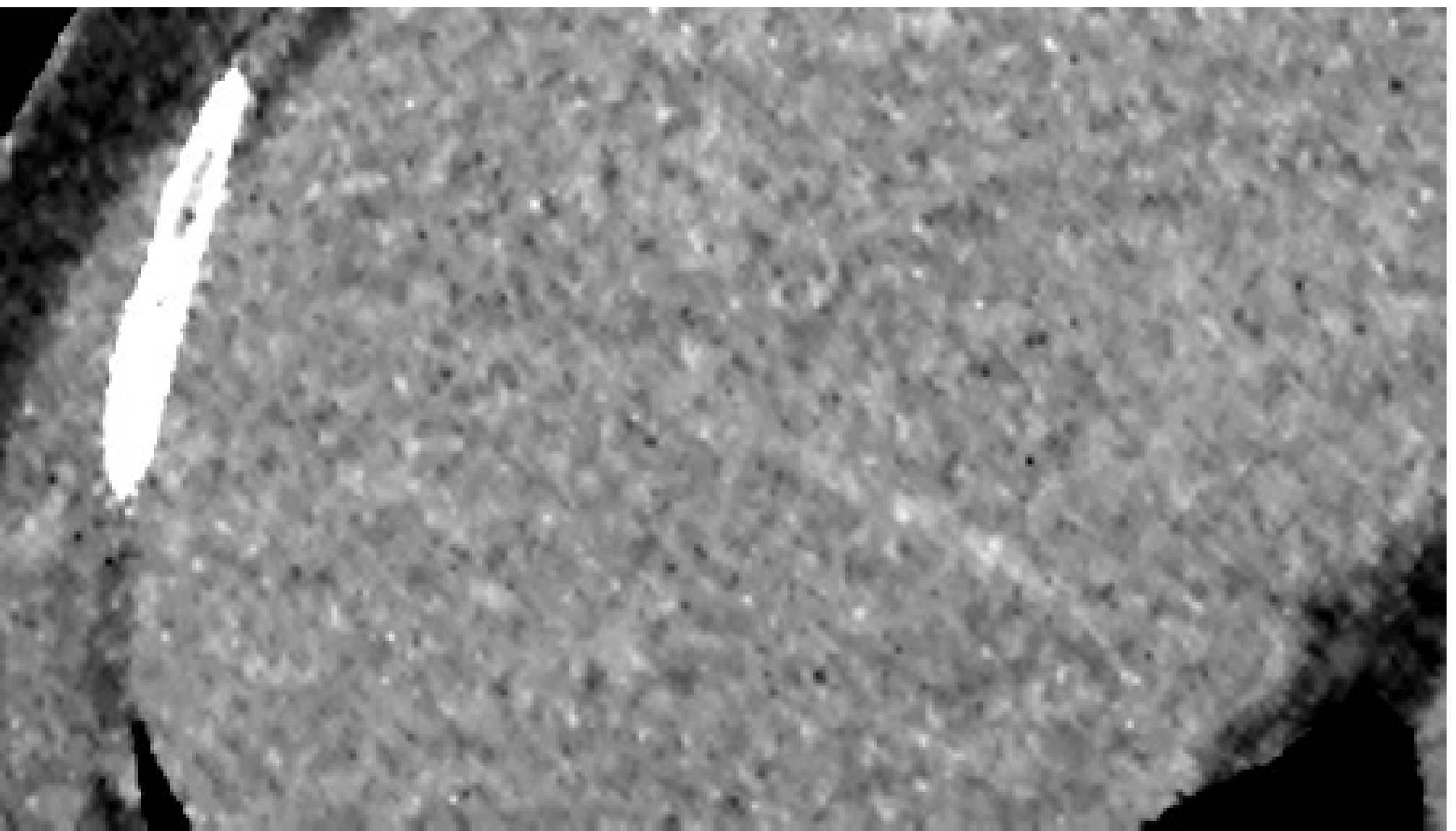} &
\includegraphics[width=2.2in]{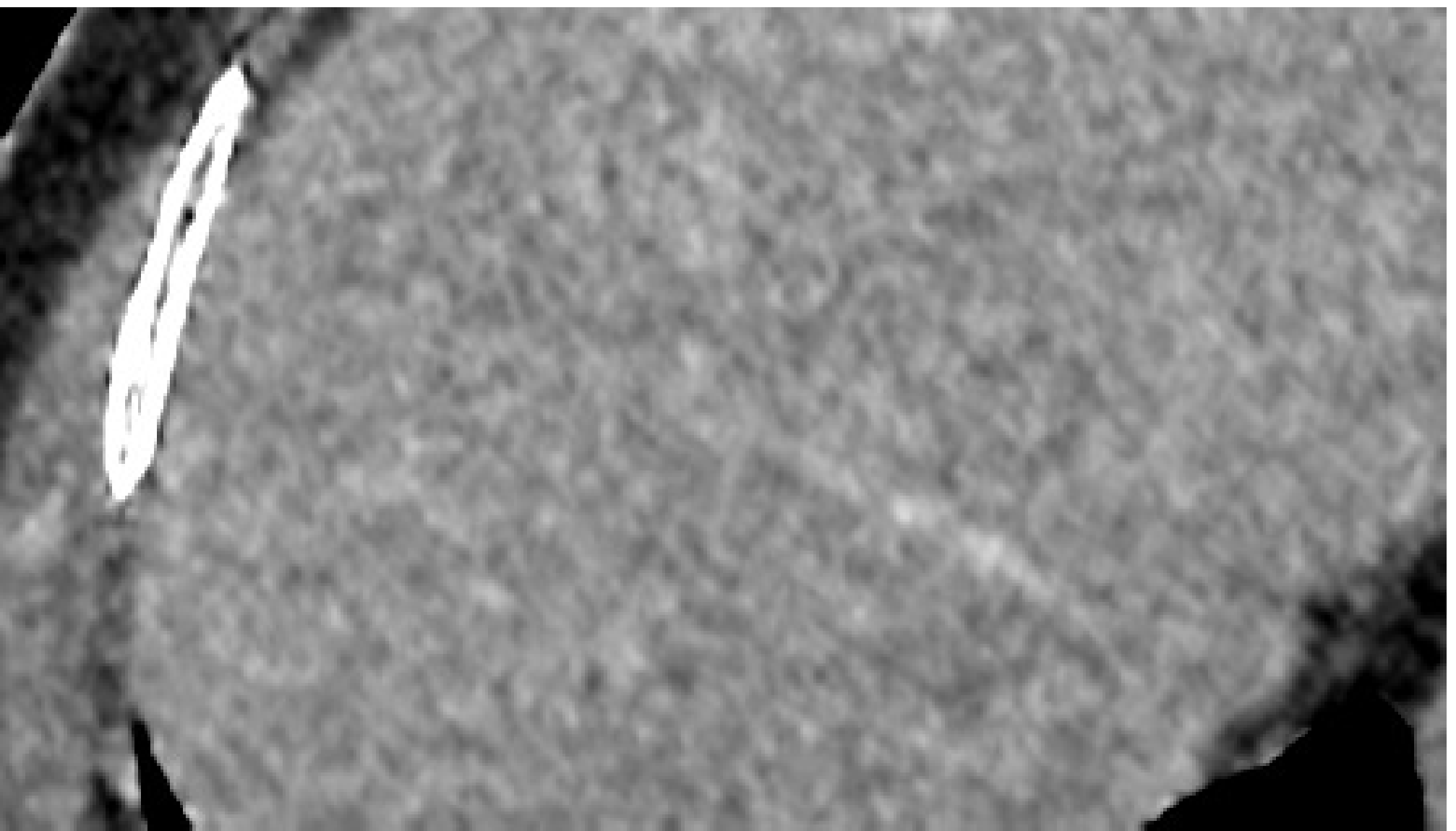} \\
 (a) FBP (62.15 HU) & (b) MBIR w/ $q$-GGMRF w/ reduced regularization (25.46 HU)  & (c) MBIR w/ adjusted GM-MRF (18.41~HU)
\end{tabular}
\caption{An abdominal axial slice of the low-dose clinical reconstruction.
From left to right, the columns represent (a) FBP, (b) MBIR with $q$-GGMRF with reduced regularization, and
(c) MBIR with adjusted GM-MRF with $p=0.5, \alpha=33\ {\rm HU}$.
Top row shows the full FOV of the reconstructed images, 
while the bottom row shows a zoomed-in FOV.
Noise standard deviation is measured within an ROI in aorta, as illustrated in the FBP image, and is reported for each method.
Display window is \mbox{[-160 240] HU}.
Note the suppression of speckle noise in soft tissue and improvement of sharpness in bone provided by MBIR with the GM-MRF prior.}
\label{fig:clinical_low_axl_abd}
\end{figure*}

The visual comparison is further verified by quantitative measurements in Fig.~\ref{fig:gepp_measurement},
which presents the measurements of reconstruction accuracy, noise, and resolution, of GEPP reconstructions at different dose levels.
It shows that MBIR with $5\times5\times3$ GM-MRF priors improve the resolution (in Fig.~\ref{fig:gepp_measurement}(f))
while producing comparable or even less noise than FBP and MBIR with the $q$-GGMRF prior (in Fig.~\ref{fig:gepp_measurement}(d)(e)),
without affecting the reconstruction accuracy (in Fig.~\ref{fig:gepp_measurement}(b)(c)).

Fig.~\ref{fig:gepp_measurement_diffsize} presents the quantitative measurements for GEPP reconstructions produced by various GM-MRF models. 
It shows that, with matched noise level in homogeneous regions, 
trained GM-MRF models with different patch sizes can achieve similar high-contrast resolution.

Fig.~\ref{fig:gepp_diffsize} presents the GEPP reconstructions for 290 mA and 40 mA data, 
produced by using GM-MRF with three different sizes of patch models.
As shown in the figure, larger patch is more robust to individual noise spots and can generate less grainy texture. 
This is more significant in 40 mA images due to the low SNR condition.
It is also observed in the 290 mA images that increased patch size introduces some blurriness to the cyclic bars.
We believe this is because larger patches belong to a higher-dimensional feature space and therefore contains more potential variations. While the GM-MRF with larger patch is able to provide more expressive patch models than that with smaller patch, it also requires more components to capture all possible variations. With limited number of components, GM-MRFs with larger patches may compromise the modeling of fine structures, 
and therefore perform inferiorly for fine structures as compared to GM-MRFs with sufficient structure modeling, i.e., smaller patch.

The supplementary material includes 3-D GEPP reconstruction results produced by using GM-MRF with various number of components. 
The results show that the GM-MRF model tends to produce more enhanced fine structures with increasing number of components. 
This implies that additional GM components will capture the behavior of edges and structures.

In addition, the supplementary material presents the study on the impact of parameters $p$ and $\alpha$ 
of GM covariance scaling on the reconstructed images.
With matched noise level in homogeneous regions,
increasing the value of $p$ with fixed $\alpha$ leads to reduction of the covariances of high-contrast edges (group 5) and bones (group 6),
and therefore results in decreasing high-contrast resolution.
On the other hand, increasing the value of $\alpha$ with fixed $p$ increases the covariances for all GM components,
and therefore encourages distribution overlapping and may consequently reduce the specification of structures and edges, 
which leads to blurriness in cyclic bars. 

\setlength\tabcolsep{0in}
\begin{figure*}[!t]
\centering
\begin{tabular}{C{2.2in}C{2.2in}C{2.2in}} 
\includegraphics[width=2.2in,trim={0 0.4in 0 0},clip]{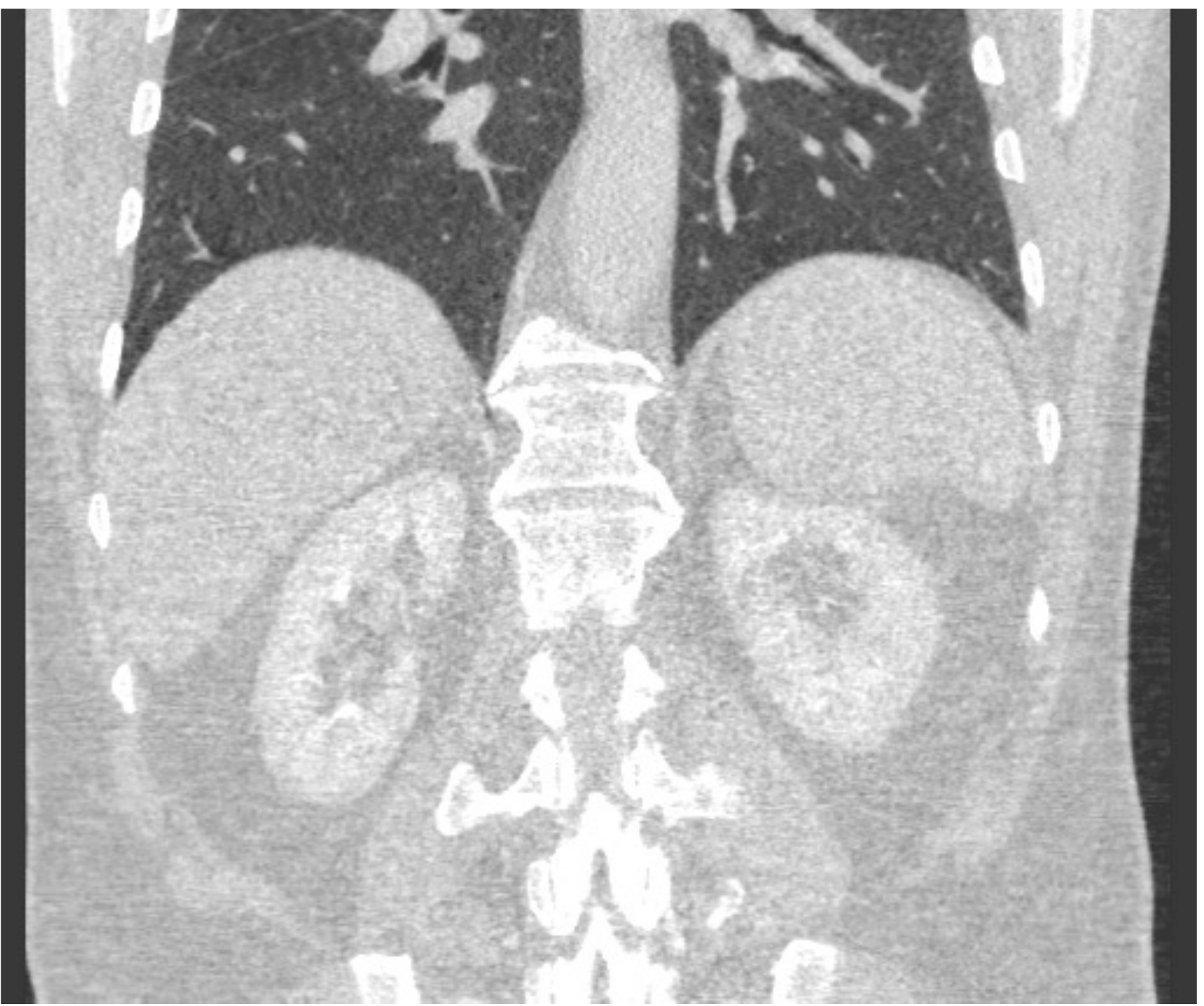} &
\includegraphics[width=2.2in,trim={0 0.4in 0 0},clip]{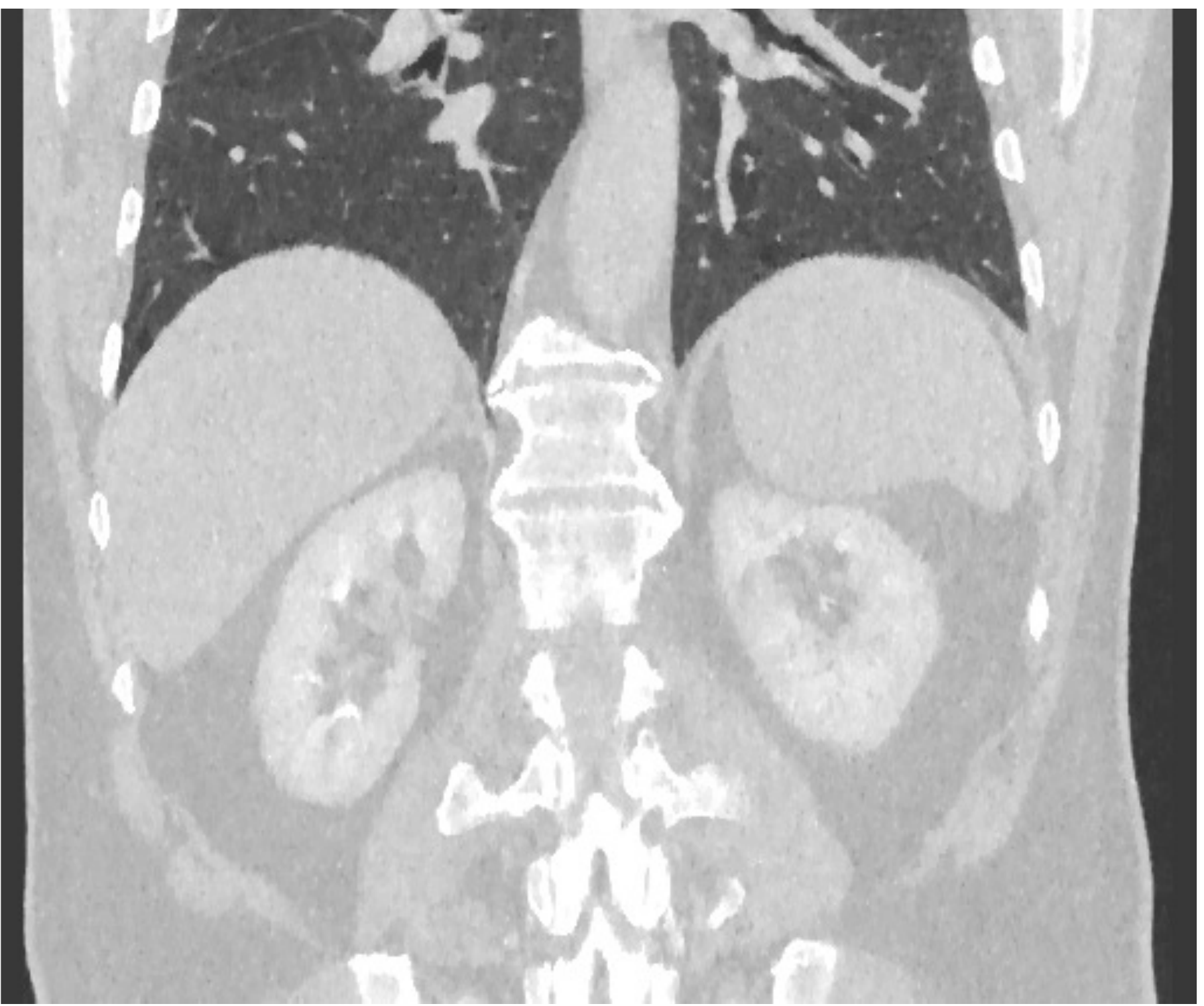} &
\includegraphics[width=2.2in,trim={0 0.4in 0 0},clip]{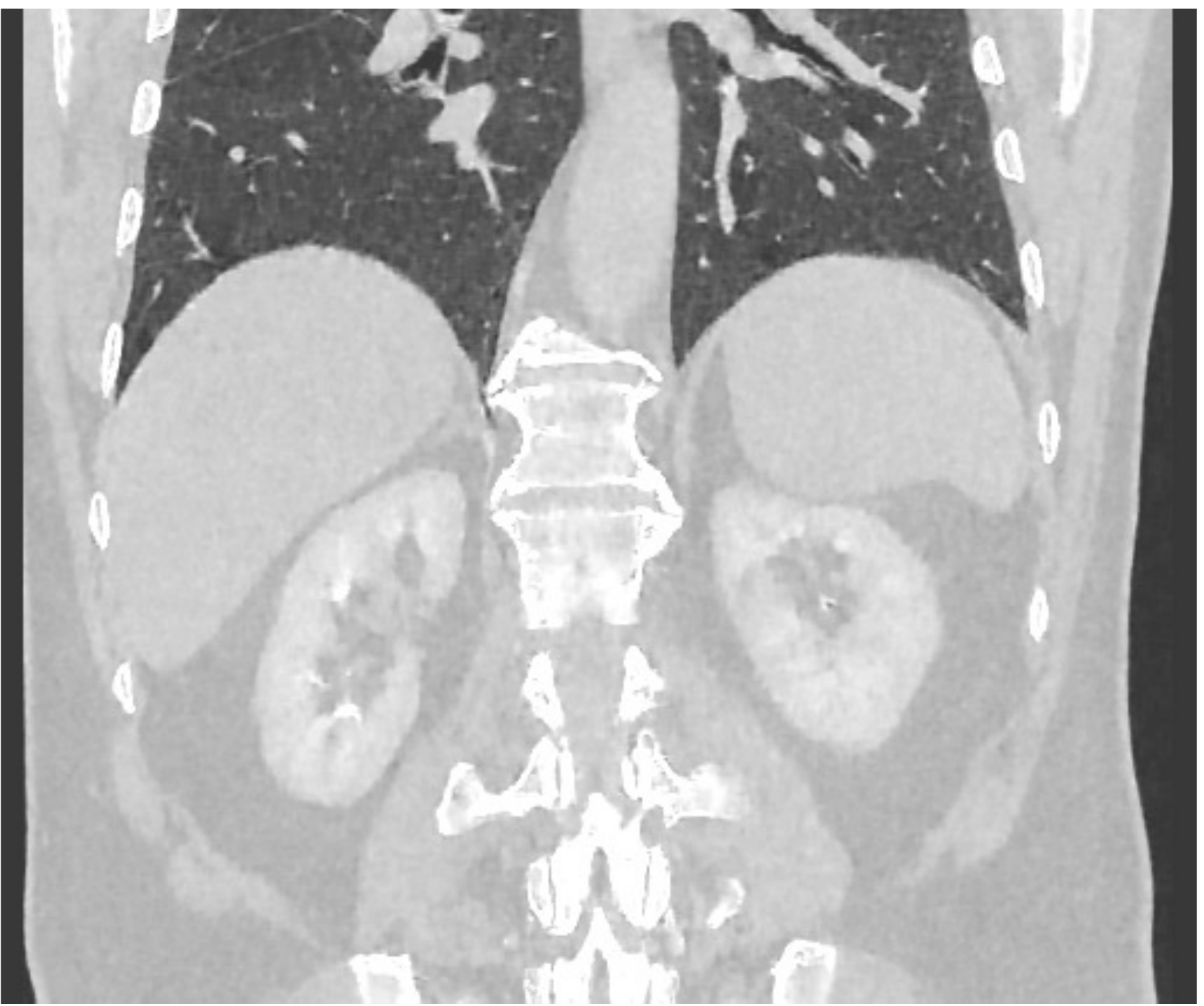} \\
\includegraphics[width=2.2in]{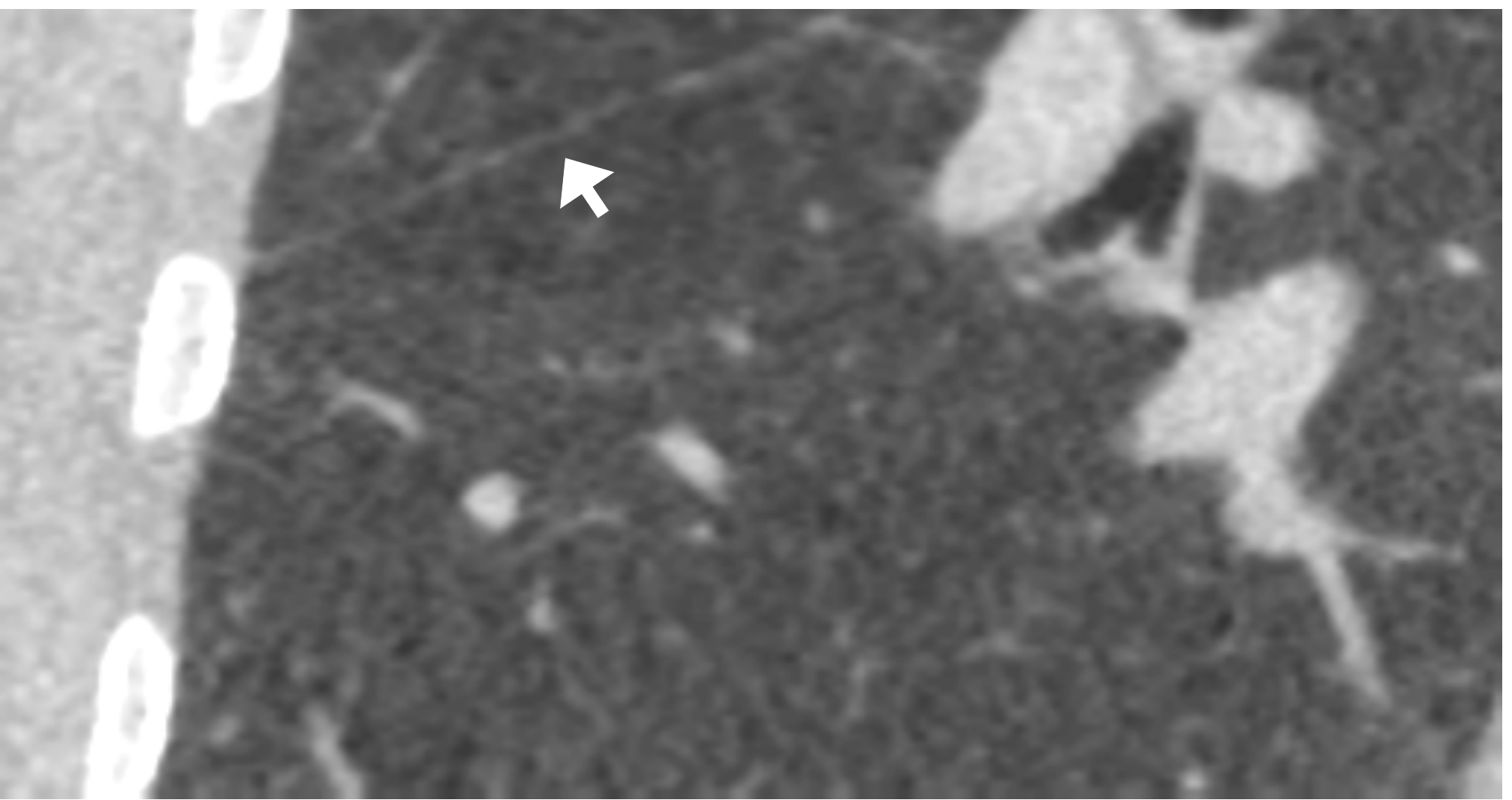} &
\includegraphics[width=2.2in]{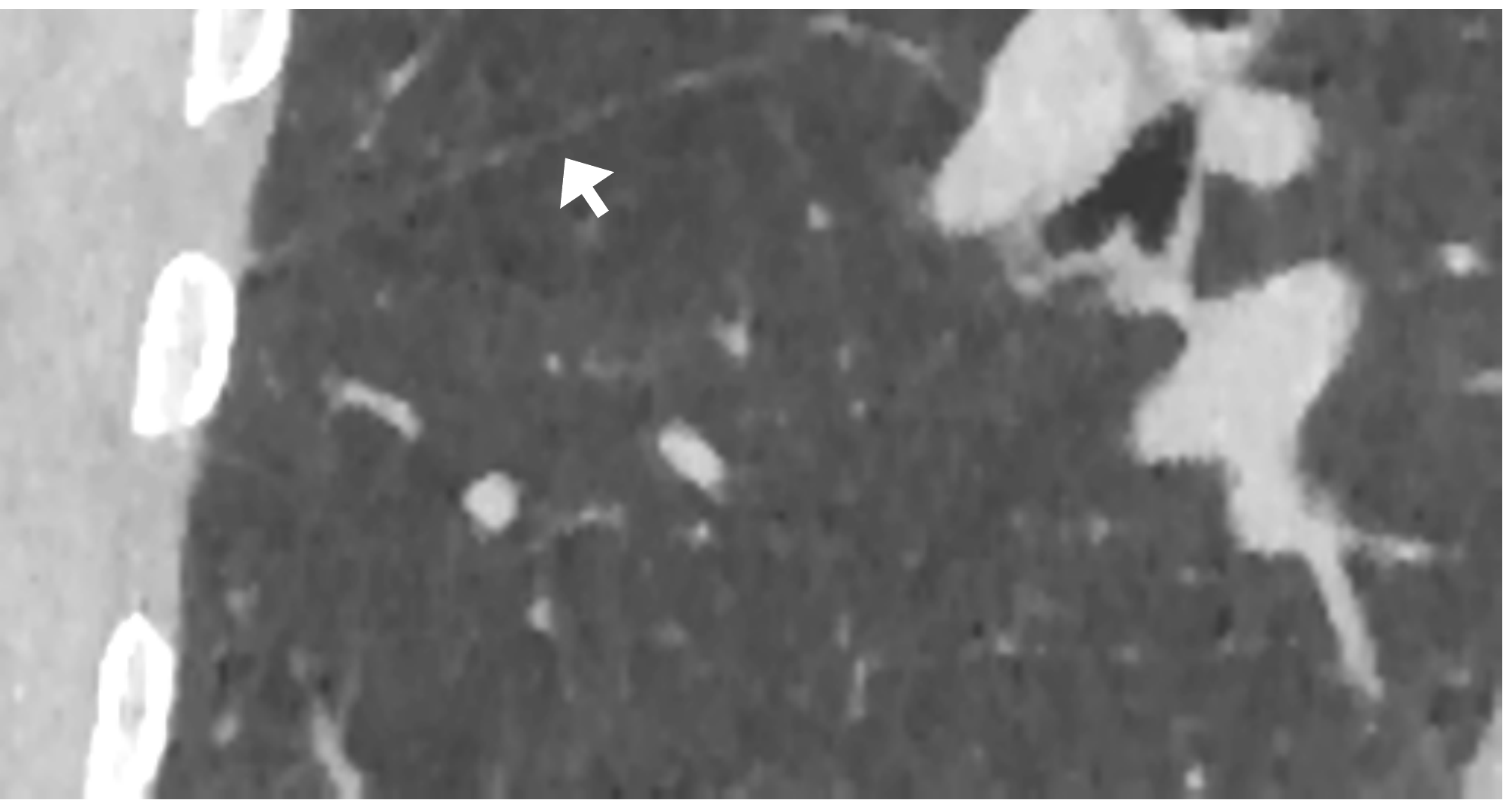} &
\includegraphics[width=2.2in]{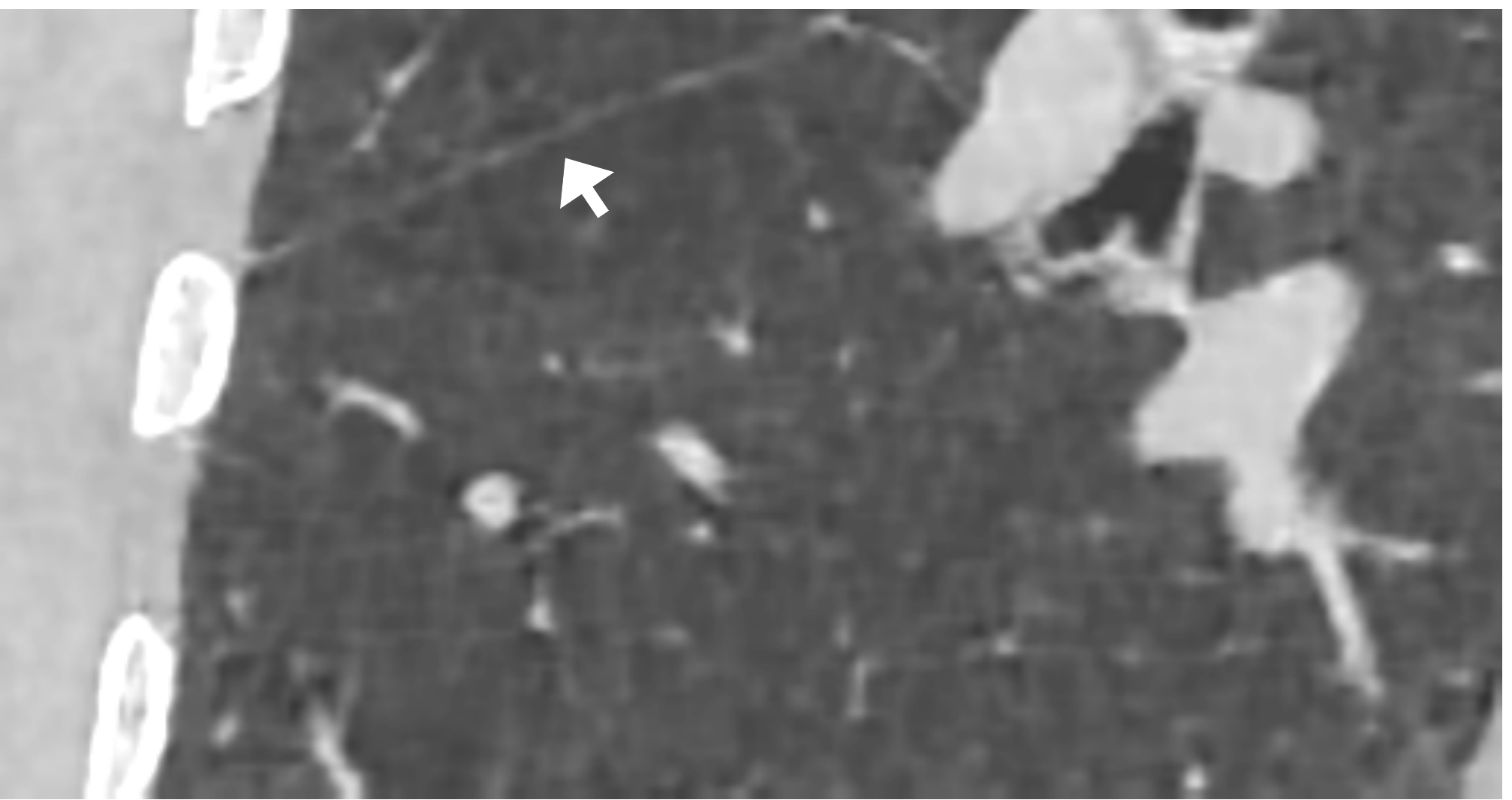} \\
\includegraphics[width=2.2in]{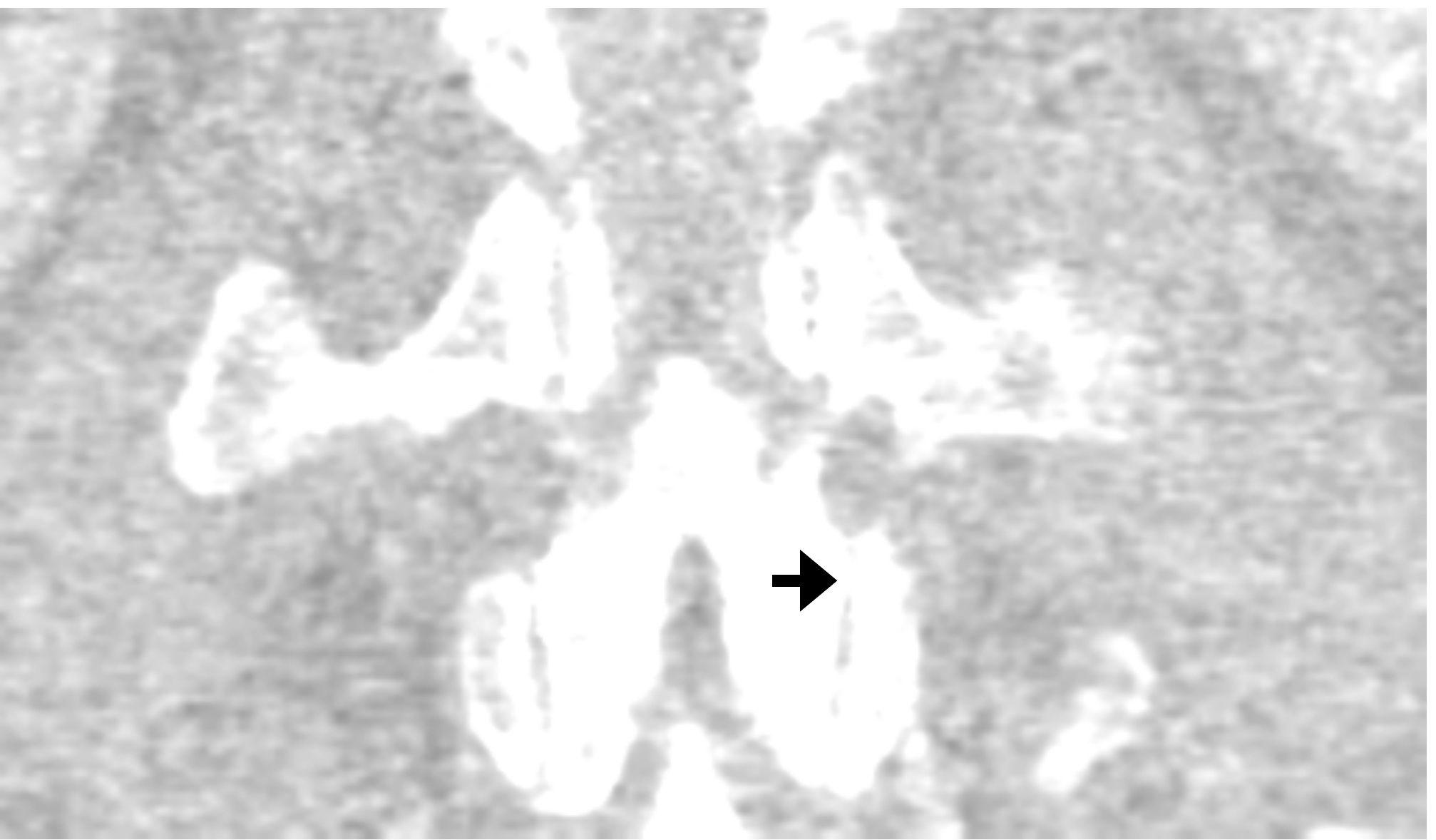} &
\includegraphics[width=2.2in]{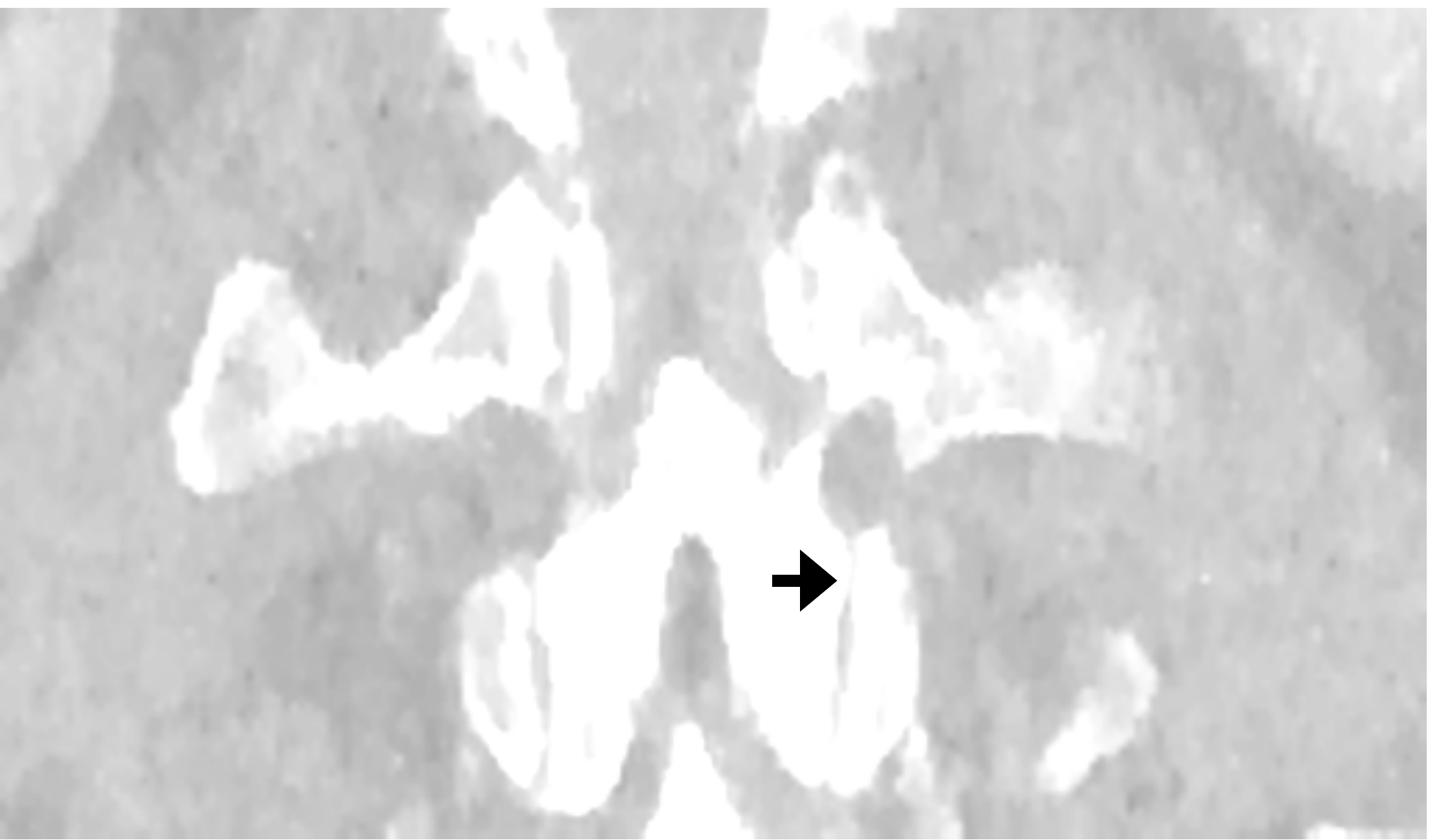} &
\includegraphics[width=2.2in]{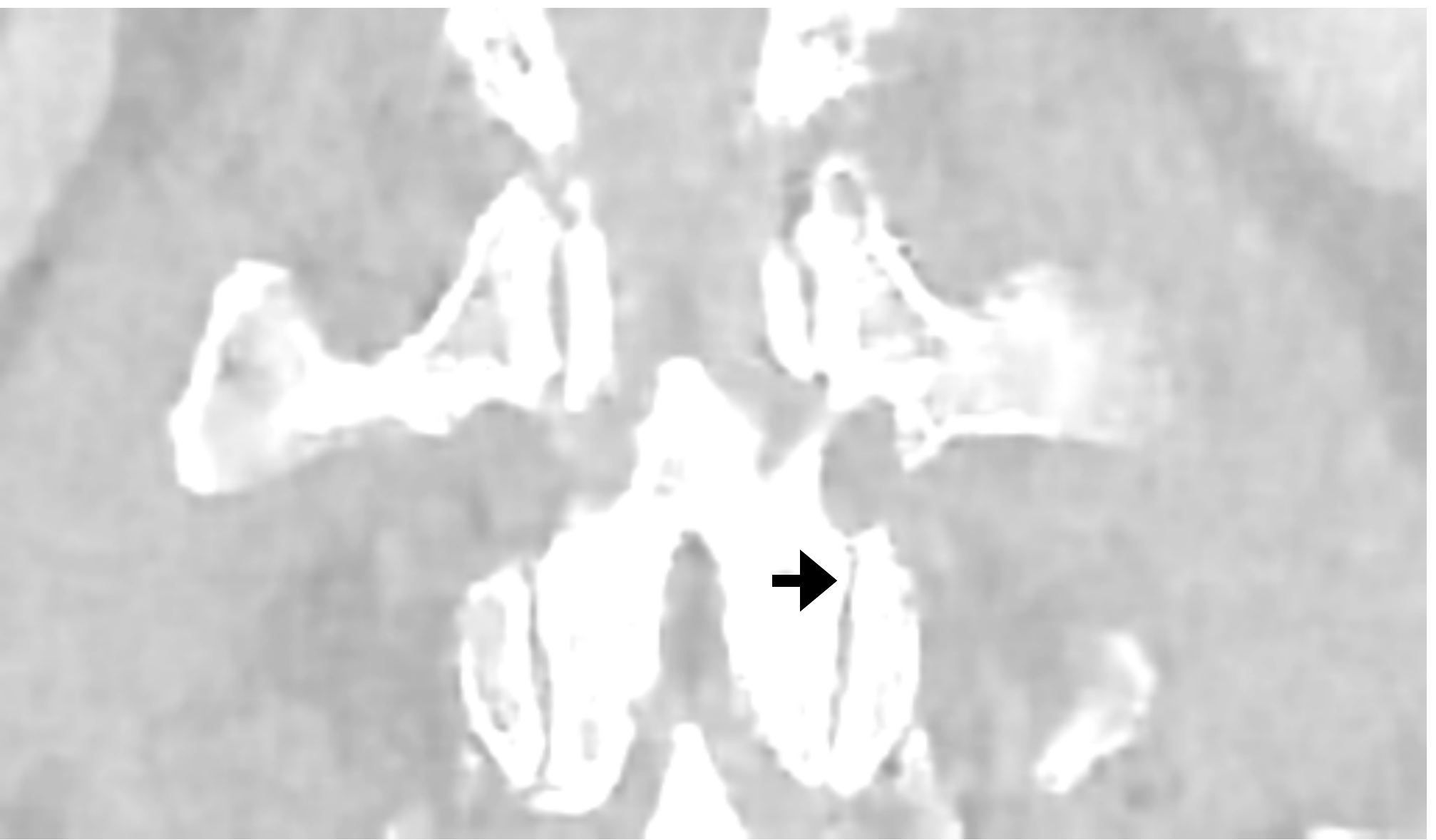} \\
 (a) FBP  & (b) MBIR w/ $q$-GGMRF w/ reduced regularization & (c) MBIR w/ adjusted GM-MRF
\end{tabular}
\caption{A coronal view of the normal-dose clinical reconstruction in lung window.
From left to right, the columns represent (a) FBP, (b) MBIR with $q$-GGMRF with reduced regularization, and
(c) MBIR with adjusted GM-MRF with $p=0.5, \alpha=33\ {\rm HU}$.
Top row shows the full FOV of the reconstructed images, 
while the bottom row shows a zoomed-in FOV. 
Display window is [-1400 400] HU.
Note the lung fissure and bone structure are reconstructed much more clearly by using the GM-MRF prior as compared to the other methods.}
\label{fig:clinical_low_cor_lung}
\end{figure*}

\subsection{3-D reconstruction for clinical datasets}

For clinical datasets, we used the $5\times5\times3$ GM-MRF model with 66 components with covariances adjusted with parameters $p=0.5$ and $\alpha=33\ {\rm HU}$,
to achieve balanced visual quality between low- and high-contrast regions.

\mbox{Figs. \ref{fig:clinical_abd} - \ref{fig:clinical_sag}} present the reconstructions with the normal-dose scan of the new patient whose data was not used for training.
As compared to FBP, MBIR with GM-MRF priors produce images with sharper bones, more lung details, as well as less noise in soft tissues. 
When compared to MBIR with traditional $q$-GGMRF prior with similar noise level, MBIR with GM-MRF priors reduce the jagged appearance in edges in Fig.~\ref{fig:clinical_abd}.
These improvements are due to better edge definition in the patch-based model over traditional pair-wise models. 
Moreover, MBIR with GM-MRF prior reveals more fine structures in bone, such as the honeycomb structure of trabecular bones in Fig.~\ref{fig:clinical_sag}.
This indicates that the GM-MRF model is also a flexible prior and 
inherently allows different regularization strategies for different tissues in CT images. 
This flexibility allows CT reconstructions with great soft-tissue quality
while simultaneously preserving the resolution in regions with larger variation, such as bone and lung.

Figs.~\ref{fig:clinical_low_axl_abd}~-~\ref{fig:clinical_low_cor_lung} present the reconstructions with the low-dose of the same patient whose normal-dose scan was used for training.
It is shown that all the improvements revealed by experiment with normal-dose data can be observed more clearly in the low-dose situation,
where the better image prior model is perhaps more valuable.
Fig.~\ref{fig:clinical_low_axl_abd} shows that the GM-MRF prior improves the texture in soft tissue 
without compromising the fine structures and details, as compared to the other methods.
Particularly, when compared to MBIR with traditional $q$-GGMRF prior, 
MBIR with GM-MRF prior reduces the speckle noise in liver while still maintaining the normal texture and edge definition.  
Fig.~\ref{fig:clinical_low_cor_lung} shows the improved resolution in lung and bone as produced by the GM-MRF prior.
More specifically, the zoomed-in images show that the lung fissure reconstructed by MBIR with GM-MRF  
have comparable resolution as that produced by FBP, which is blurred by MBIR with $q$-GGMRF.
The GM-MRF prior also leads to much clearer bone structure as compared to other methods.
These improvements demonstrate the material-specific regularization capability of the GM-MRF prior.

In addition, the supplementary material includes reconstruction results of a low-dose scan of another new patient. 
It is worth emphasizing that we only trained the GM-MRF model with data from a single patient scan, 
and then applied it to different scans on the same and different patients. 
However, we did not notice any systematic difference between the two cases.
We believe that ideally this GM-MRF model can be trained off-line on a pool of normal-dose images from various patients and the resulting prior model can then be fixed and applied to future patients.

\section{Conclusion}
In this paper, we introduced a novel Gaussian-mixture Markov random field (GM-MRF) image model 
along with the tools to use it as a prior for model-based iterative reconstruction (MBIR).
The proposed method constructs an image model by seaming together Gaussian-mixture (GM) patch models. 
In addition, we presented an analytical framework for computing the MAP estimate with the GM-MRF prior
using an exact surrogate function.
We also proposed a systematic approach to adjust the covariances of the GM components of the GM-MRF model,
in order to control the sharpness in low- and high-contrast regions of the reconstruction separately.
The results in image denoising and multi-slice CT reconstruction experiments demonstrate improved image quality 
and material-specific regularization by the GM-MRF prior.

\appendices
\section{Proof of lemma: surrogate functions for logs of exponential mixtures}
\label{app:lemma}

\noindent \textit{Lemma:}
Let $f:\Re^N \rightarrow \Re$ be a function of the form
\begin{equation}
f(x)=\sum_k w_k \exp\{ -v_k(x) \} \nonumber \ ,
\end{equation}
where $w_k \in \Re^+$, $\sum_k w_k >0$, and $v_k:\Re^N \rightarrow \Re$.
Furthermore $\forall (x, x') \in \Re^N\times \Re^N$ define the function
\begin{equation}
q(x;x') \triangleq -  \log f(x') + \sum_k \tilde{\pi}_k (v_k(x)-v_k(x')) \nonumber \ ,
\end{equation}
where $\tilde{\pi}_k = \frac{w_k \exp \{-v_k(x') \}}{\sum_l w_l \exp\{-v_l(x') \}}$.
Then $q(x;x')$ is a surrogate function for $-\log f(x)$, and $\forall (x, x') \in \Re^N\times \Re^N$,
\begin{eqnarray}
q(x';x') &=& - \log f(x') \nonumber \\
q(x;x') &\geq& - \log f(x) \nonumber 
\end{eqnarray}

\noindent {\it Proof:}
\begin{align*}
\log f(x) &= \log f(x') + \log \left\{\frac{f(x)}{f(x')}\right\}\\
&= \log f(x') + \log \left\{ \sum_k \left( \frac{w_k }{f(x')} \right) \exp\{-v_k(x) \} \right\} \\
&= \log f(x') + \log \left\{ \sum_k \left( \frac{w_k \exp\{-v_k(x')\}}{\sum_l w_l \exp\{-v_l(x')\} } \right) \right.\\
&\left. \times \exp \left\{ -v_k(x)+v_k(x')\right\}\right\}\\
&= \log f(x') + \log \left\{ \sum_k \tilde{\pi}_k \exp\{-v_k(x)+v_k(x')\}\right\}\\
&\geq \log f(x') + \sum_k \tilde{\pi}_k \{-v_k(x)+v_k(x')\}
\end{align*}
where
\begin{equation*}
\tilde{\pi}_k \triangleq \frac{w_k \exp\{-v_k(x')\}}{\sum_l w_l \exp\{-v_l(x')\} } \ .
\end{equation*}
The last inequality results from Jensen's inequality.
Taking the negative of the final expression results in
$$ -\log f(x) \leq - \log f(x') + \sum_k \tilde{\pi}_k \{v_k(x) - v_k(x')\}  \triangleq q(x;x') \ , $$
and evaluating this result at $x=x'$ results in $$-\log f(x') = q(x';x') \ .$$

\bibliographystyle{IEEEtran}
\bibliography{zhang}

\begin{thebibliography}{10}
\providecommand{\url}[1]{#1}
\csname url@samestyle\endcsname
\providecommand{\newblock}{\relax}
\providecommand{\bibinfo}[2]{#2}
\providecommand{\BIBentrySTDinterwordspacing}{\spaceskip=0pt\relax}
\providecommand{\BIBentryALTinterwordstretchfactor}{4}
\providecommand{\BIBentryALTinterwordspacing}{\spaceskip=\fontdimen2\font plus
\BIBentryALTinterwordstretchfactor\fontdimen3\font minus
  \fontdimen4\font\relax}
\providecommand{\BIBforeignlanguage}[2]{{%
\expandafter\ifx\csname l@#1\endcsname\relax
\typeout{** WARNING: IEEEtran.bst: No hyphenation pattern has been}%
\typeout{** loaded for the language `#1'. Using the pattern for}%
\typeout{** the default language instead.}%
\else
\language=\csname l@#1\endcsname
\fi
#2}}
\providecommand{\BIBdecl}{\relax}
\BIBdecl

\bibitem{Boas01}
D.~Boas, D.~Brooks, E.~Miller, C.~DiMarzio, M.~Kilmer, R.~Gaudette, and
  Q.~Zhang, ``Imaging the body with diffuse optical tomography,'' \emph{IEEE
  Signal Process. Mag.}, vol.~18, no.~6, pp. 57--75, 2001.

\bibitem{Qi06}
J.~Qi and R.~M. Leahy, ``Iterative reconstruction techniques in emission
  computed tomography,'' \emph{Phys. Med. Biol.}, vol.~51, no.~15, p. R541,
  2006.

\bibitem{Thibault07}
J.-B. Thibault, K.~D. Sauer, J.~Hsieh, and C.~A. Bouman, ``A three-dimensional
  statistical approach to improve image quality for multislice helical {CT},''
  \emph{Med. Phys.}, vol.~34, no.~11, pp. 4526--4544, Nov. 2007.

\bibitem{Fessler10}
J.~A. Fessler, ``Model-based image reconstruction for {MRI},'' \emph{IEEE
  Signal Process. Mag.}, vol.~27, no.~4, pp. 81--89, 2010.

\bibitem{Zhang14tmi}
R.~Zhang, J.-B. Thibault, C.~A. Bouman, K.~D. Sauer, and J.~Hsieh,
  ``Model-based iterative reconstruction for dual-energy {X}-ray {CT} using a
  joint quadratic likelihood model,'' \emph{IEEE Trans. Med. Imag.}, vol.~33,
  no.~1, pp. 117--134, 2014.

\bibitem{Venkat15}
S.~Venkatakrishnan, L.~Drummy, M.~Jackson, M.~De~Graef, J.~Simmons, and C.~A.
  Bouman, ``Model-based iterative reconstruction for bright-field electron
  tomography,'' \emph{IEEE Trans. Computational Imag.}, vol.~1, no.~1, pp.
  1--15, 2015.

\bibitem{Jin15}
P.~Jin, C.~A. Bouman, and K.~D. Sauer, ``A model-based image reconstruction
  algorithm with simultaneous beam hardening correction for {X-ray CT},''
  \emph{IEEE Trans. Computational Imag.}, vol.~1, no.~3, pp. 200--216, 2015.

\bibitem{Mohan15}
A.~K. Mohan, S.~Venkatakrishnan, J.~Gibbs, E.~Gulsoy, X.~Xiao, M.~De~Graef,
  P.~Voorhees, and C.~A. Bouman, ``{TIMBIR}: A method for time-space
  reconstruction from interlaced views,'' \emph{IEEE Trans. Computational
  Imag.}, vol.~1, no.~2, pp. 96--111, 2015.

\bibitem{Nelson11}
R.~C. Nelson, S.~Feuerlein, and D.~T. Boll, ``New iterative reconstruction
  techniques for cardiovascular computed tomography: how do they work, and what
  are the advantages and disadvantages?'' \emph{J. Cardiovascular Computed
  Tomography}, vol.~5, no.~5, pp. 286--292, 2011.

\bibitem{Yamada12}
Y.~Yamada, M.~Jinzaki, Y.~Tanami, E.~Shiomi, H.~Sugiura, T.~Abe, and
  S.~Kuribayashi, ``Model-based iterative reconstruction technique for
  ultralow-dose computed tomography of the lung: A pilot study,''
  \emph{Investigative Radiology}, vol.~47, no.~8, pp. 482--489, 2012.

\bibitem{Besag74}
J.~Besag, ``Spatial interaction and the statistical analysis of lattice
  systems,'' \emph{J. Roy. Statistical Soc. Series B (Methodological)}, pp.
  192--236, 1974.

\bibitem{Sauer93}
K.~D. Sauer and C.~A. Bouman, ``A local update strategy for iterative
  reconstruction from projections,'' \emph{IEEE Trans. Signal Process.},
  vol.~41, no.~2, pp. 534--548, Feb. 1993.

\bibitem{deman00tns}
B.~De~Man, J.~Nuyts, P.~Dupont, G.~Marchal, and P.~Suetens, ``Reduction of
  metal streak artifacts in x-ray computed tomography using a transmission
  maximum a posteriori algorithm,'' \emph{IEEE Trans. Nucl. Sci.}, vol.~47,
  no.~3, pp. 977--981, 2000.

\bibitem{elbakri02tmi}
I.~A. Elbakri and J.~A. Fessler, ``Statistical image reconstruction for
  polyenergetic {X}-ray computed tomography,'' \emph{IEEE Trans. Med. Imag.},
  vol.~21, no.~2, pp. 89--99, 2002.

\bibitem{Rudin92}
L.~I. Rudin, S.~Osher, and E.~Fatemi, ``Nonlinear total variation based noise
  removal algorithms,'' \emph{Physica D: Nonlinear Phenomena}, vol.~60, no.~1,
  pp. 259--268, 1992.

\bibitem{Sidky08}
E.~Y. Sidky and X.~Pan, ``Image reconstruction in circular cone-beam computed
  tomography by constrained, total-variation minimization,'' \emph{Phys. Med.
  Biol.}, vol.~53, no.~17, pp. 4777--4807, 2008.

\bibitem{tang09pmb}
J.~Tang, B.~E. Nett, and G.-H. Chen, ``Performance comparison between total
  variation ({TV})-based compressed sensing and statistical iterative
  reconstruction algorithms,'' \emph{Phys. Med. Biol.}, vol.~54, no.~19, p.
  5781, 2009.

\bibitem{ritschl11pmb}
L.~Ritschl, F.~Bergner, C.~Fleischmann, and M.~Kachelrie{\ss}, ``Improved total
  variation-based {CT} image reconstruction applied to clinical data,''
  \emph{Phys. Med. Biol.}, vol.~56, no.~6, p. 1545, 2011.

\bibitem{liu12pmb}
Y.~Liu, J.~Ma, Y.~Fan, and Z.~Liang, ``Adaptive-weighted total variation
  minimization for sparse data toward low-dose x-ray computed tomography image
  reconstruction,'' \emph{Phys. Med. Biol.}, vol.~57, no.~23, p. 7923, 2012.

\bibitem{li04tns}
T.~Li, X.~Li, J.~Wang, J.~Wen, H.~Lu, J.~Hsieh, and Z.~Liang, ``Nonlinear
  sinogram smoothing for low-dose x-ray {CT},'' \emph{IEEE Trans. Nucl. Sci.},
  vol.~51, no.~5, pp. 2505--2513, 2004.

\bibitem{la05mp}
P.~J. La~Riviere, ``Penalized-likelihood sinogram smoothing for low-dose
  {CT},'' \emph{Med. Phys.}, vol.~32, no.~6, pp. 1676--1683, 2005.

\bibitem{wang06tmi}
J.~Wang, T.~Li, H.~Lu, and Z.~Liang, ``Penalized weighted least-squares
  approach to sinogram noise reduction and image reconstruction for low-dose
  {X}-ray computed tomography,'' \emph{IEEE Trans. Med. Imag.}, vol.~25,
  no.~10, pp. 1272--1283, 2006.

\bibitem{hsiao02tip}
I.-T. Hsiao, A.~Rangarajan, and G.~Gindi, ``{Joint-MAP} {Bayesian} tomographic
  reconstruction with a gamma-mixture prior,'' \emph{IEEE Trans. Image
  Process.}, vol.~11, no.~12, pp. 1466--1477, 2002.

\bibitem{hsiao03jei}
------, ``Bayesian image reconstruction for transmission tomography using
  deterministic annealing,'' \emph{J. Electron. Imaging}, vol.~12, no.~1, pp.
  7--16, 2003.

\bibitem{wang09tns}
G.~Wang, L.~Schultz, and J.~Qi, ``Statistical image reconstruction for muon
  tomography using a {G}aussian scale mixture model,'' \emph{IEEE Trans. Nucl.
  Sci.}, vol.~56, no.~4, pp. 2480--2486, 2009.

\bibitem{mehranian15tmi}
A.~Mehranian and H.~Zaidi, ``Joint estimation of activity and attenuation in
  whole-body {TOF PET/MRI} using constrained {Gaussian} mixture models,''
  \emph{IEEE Trans. Med. Imag.}, vol.~34, no.~9, pp. 1808--1821, 2015.

\bibitem{Elad06}
M.~Elad and M.~Aharon, ``Image denoising via sparse and redundant
  representations over learned dictionaries,'' \emph{IEEE Trans. Image
  Process.}, vol.~15, no.~12, pp. 3736--3745, 2006.

\bibitem{Buades05}
A.~Buades, B.~Coll, and J.-M. Morel, ``A non-local algorithm for image
  denoising,'' in \emph{Proc. IEEE Conf. Comput. Vision Pattern Recognition
  (CVPR)}, vol.~2, 2005, pp. 60--65.

\bibitem{Dabov07}
K.~Dabov, A.~Foi, V.~Katkovnik, and K.~Egiazarian, ``Image denoising by sparse
  3-d transform-domain collaborative filtering,'' \emph{IEEE Trans. Image
  Process.}, vol.~16, no.~8, pp. 2080--2095, 2007.

\bibitem{venkat13globalsip}
S.~Venkatakrishnan, C.~A. Bouman, and B.~Wohlberg, ``Plug-and-play priors for
  model based reconstruction,'' in \emph{Global Conf. Signal and Inform.
  Process. (GlobalSIP), 2013 {IEEE}}, Dec 2013, pp. 945--948.

\bibitem{chan16arxiv}
S.~H. Chan, ``Algorithm-induced prior for image restoration,''
  \emph{arXiv:1602.00715}, 2016.

\bibitem{teodoro16arxiv}
A.~M. Teodoro, J.~M. Bioucas-Dias, and M.~A. Figueiredo, ``Image restoration
  and reconstruction using variable splitting and class-adapted image priors,''
  \emph{arXiv:1602.04052}, 2016.

\bibitem{Ravishankar11}
S.~Ravishankar and Y.~Bresler, ``{MR} image reconstruction from highly
  undersampled k-space data by dictionary learning,'' \emph{IEEE Trans. Med.
  Imag.}, vol.~30, no.~5, pp. 1028--1041, 2011.

\bibitem{Xu12}
Q.~Xu, H.~Yu, X.~Mou, L.~Zhang, J.~Hsieh, and G.~Wang, ``Low-dose {X}-ray {CT}
  reconstruction via dictionary learning,'' \emph{IEEE Trans. Med. Imag.},
  vol.~31, no.~9, pp. 1682--1697, 2012.

\bibitem{huang11cbm}
J.~Huang, J.~Ma, N.~Liu, H.~Zhang, Z.~Bian, Y.~Feng, Q.~Feng, and W.~Chen,
  ``Sparse angular {CT} reconstruction using non-local means based
  iterative-correction {POCS},'' \emph{Comput. in biology and medicine},
  vol.~41, no.~4, pp. 195--205, 2011.

\bibitem{sreehari15arxiv}
S.~Sreehari, S.~Venkatakrishnan, B.~Wohlberg, L.~F. Drummy, J.~P. Simmons, and
  C.~A. Bouman, ``{Plug-and-Play} priors for bright field electron tomography
  and sparse interpolation,'' \emph{arXiv:1512.07331}, 2015.

\bibitem{chen08jmiv}
Y.~Chen, J.~Ma, Q.~Feng, L.~Luo, P.~Shi, and W.~Chen, ``Nonlocal prior
  {Bayesian} tomographic reconstruction,'' \emph{J. Math. Imag. and Vision},
  vol.~30, no.~2, pp. 133--146, 2008.

\bibitem{Wang12}
G.~Wang and J.~Qi, ``Penalized likelihood {PET} image reconstruction using
  patch-based edge-preserving regularization,'' \emph{IEEE Trans. Med. Imag.},
  vol.~31, no.~12, pp. 2194--2204, 2012.

\bibitem{ma12pmb}
J.~Ma, H.~Zhang, Y.~Gao, J.~Huang, Z.~Liang, Q.~Feng, and W.~Chen, ``Iterative
  image reconstruction for cerebral perfusion {CT} using a pre-contrast scan
  induced edge-preserving prior,'' \emph{Phys. Med. Biol.}, vol.~57, no.~22, p.
  7519, 2012.

\bibitem{zhang14cmig}
H.~Zhang, J.~Ma, J.~Wang, Y.~Liu, H.~Lu, and Z.~Liang, ``Statistical image
  reconstruction for low-dose {CT} using nonlocal means-based regularization,''
  \emph{Computerized Medical Imag. and Graph.}, vol.~38, no.~6, pp. 423--435,
  2014.

\bibitem{Liao08}
H.~Y. Liao and G.~Sapiro, ``Sparse representations for limited data
  tomography,'' in \emph{Proc. IEEE Int. Symp. Biomed. Imag. (ISBI): From Nano
  to Macro}, 2008, pp. 1375--1378.

\bibitem{Lu12}
Y.~Lu, J.~Zhao, and G.~Wang, ``Few-view image reconstruction with dual
  dictionaries,'' \emph{Phys. Med. Biol.}, vol.~57, no.~1, p. 173, 2012.

\bibitem{Pfister14}
L.~Pfister and Y.~Bresler, ``Tomographic reconstruction with adaptive
  sparsifying transforms,'' in \emph{Proc. IEEE Int. Conf. Acoust., Speech and
  Signal Process. (ICASSP)}, 2014, pp. 6914--6918.

\bibitem{Zoran11}
D.~Zoran and Y.~Weiss, ``From learning models of natural image patches to whole
  image restoration,'' in \emph{Proc. Int. Conf. Comput. Vision (ICCV)}, 2011,
  pp. 479--486.

\bibitem{Yu12}
G.~Yu, G.~Sapiro, and S.~Mallat, ``Solving inverse problems with piecewise
  linear estimators: from {G}aussian mixture models to structured sparsity,''
  \emph{IEEE Trans. Image Process.}, vol.~21, no.~5, pp. 2481--2499, 2012.

\bibitem{nguyen13}
T.~M. Nguyen and Q.~Wu, ``Fast and robust spatially constrained gaussian
  mixture model for image segmentation,'' \emph{IEEE Trans. Circuits Syst.
  Video Technol.}, vol.~23, no.~4, pp. 621--635, 2013.

\bibitem{wang13sure}
Y.-Q. Wang and J.-M. Morel, ``{SURE} guided {G}aussian mixture image
  denoising,'' \emph{SIAM J. Imag. Sci.}, vol.~6, no.~2, pp. 999--1034, 2013.

\bibitem{Zhang13fully3d}
R.~Zhang, J.-B. Thibault, C.~A. Bouman, and K.~D. Sauer, ``Soft classification
  with {G}aussian mixture model for clinical dual-energy {CT}
  reconstructions,'' in \emph{12th Int. Mtg. Fully Three-Dimensional Image
  Reconstruction in Radiology and Nuclear Medicine}, June 2013, pp. 408--411.

\bibitem{yang14tip}
J.~Yang, X.~Yuan, X.~Liao, P.~Llull, D.~J. Brady, G.~Sapiro, and L.~Carin,
  ``Video compressive sensing using {Gaussian} mixture models,'' \emph{IEEE
  Trans. Image Process.}, vol.~23, no.~11, pp. 4863--4878, 2014.

\bibitem{Nadir15globalsip}
Z.~Nadir, M.~S. Brown, M.~L. Comer, and C.~A. Bouman, ``Gaussian mixture prior
  models for imaging of flow cross sections from sparse hyperspectral
  measurements,'' in \emph{Global Conf. Signal and Inform. Process.
  (GlobalSIP), 2015 IEEE}, Dec 2015, pp. 527--531.

\bibitem{Zhang13}
R.~Zhang, C.~Bouman, J.-B. Thibault, and K.~Sauer, ``Gaussian mixture {M}arkov
  random field for image denoising and reconstruction,'' in \emph{Global Conf.
  Signal and Inform. Process. (GlobalSIP), 2013 IEEE}, Dec 2013, pp.
  1089--1092.

\bibitem{Zhang15fully3d}
R.~Zhang, D.~Pal, J.-B. Thibault, K.~D. Sauer, and C.~A. Bouman, ``Model-based
  iterative reconstruction with a {G}aussian mixture {MRF} prior for {X}-ray
  {CT},'' in \emph{13th Int. Mtg. Fully Three-Dimensional Image Reconstruction
  in Radiology and Nuclear Medicine}, June 2015, pp. 407--410.

\bibitem{Bouman97}
C.~A. Bouman, ``Cluster: {A}n unsupervised algorithm for modeling {G}aussian
  mixtures,'' Apr. 1997, available from
  http://engineering.purdue.edu/\string~bouman.

\bibitem{Salakhutdinov12}
R.~Salakhutdinov and G.~Hinton, ``An efficient learning procedure for deep
  {B}oltzmann machines,'' \emph{Neural Computation}, vol.~24, no.~8, pp.
  1967--2006, 2012.

\bibitem{Hunter04}
D.~R. Hunter and K.~Lange, ``A tutorial on {MM} algorithms,'' \emph{The Amer.
  Statistician}, vol.~58, no.~1, pp. 30--37, 2004.

\bibitem{fessler96tip}
J.~A. Fessler, ``Mean and variance of implicitly defined biased estimators
  (such as penalized maximum likelihood): {A}pplications to tomography,''
  \emph{IEEE Trans. Image Process.}, vol.~5, no.~3, pp. 493--506, 1996.

\bibitem{evans11mp}
J.~D. Evans, D.~G. Politte, B.~R. Whiting, J.~A. OÕSullivan, and J.~F.
  Williamson, ``Noise-resolution tradeoffs in x-ray {CT} imaging: a comparison
  of penalized alternating minimization and filtered backprojection
  algorithms,'' \emph{Med. Phys.}, vol.~38, no.~3, pp. 1444--1458, 2011.

\bibitem{li14mp}
K.~Li, J.~Garrett, Y.~Ge, and G.-H. Chen, ``Statistical model based iterative
  reconstruction ({MBIR}) in clinical {CT} systems. {Part II. Experimental}
  assessment of spatial resolution performance,'' \emph{Med. Phys.}, vol.~41,
  no.~7, pp. 071\,911--1 -- 071\,911--12, 2014.

\bibitem{Rubinstein13}
R.~Rubinstein, M.~Zibulevsky, and M.~Elad, ``Efficient implementation of the
  {K-SVD} algorithm using batch orthogonal matching pursuit,'' Feb. 2013,
  available from www.cs.technion.ac.il/\string~ronrubin/software.

\bibitem{Dabov14}
K.~Dabov, A.~Danieyan, and A.~Foi, ``{BM3D} demo software for image/video
  restoration and enhancement,'' Jan. 2014, available from
  www.cs.tut.fi/\string~foi/GCF-BM3D/index.html\string#ref\_software.

\bibitem{Kroon10}
D.-J. Kroon, ``Fast non-local means 1{D}, 2{D} color and 3{D},'' Sept. 2010,
  available from
  http://www.mathworks.com/matlabcentral/fileexchange/27395-fast-non-local-means-1d--2d-color-and-3d.

\end{thebibliography}

\begin{table*}[!h]
\renewcommand{\arraystretch}{3}
\centering
{\Huge A Gaussian Mixture MRF\\ for Model-Based Iterative Reconstruction \\
with Applications to Low-Dose X-ray CT:  \\
Supplementary Material \\ } \ \\
\centering
{\large Ruoqiao Zhang,
	Dong Hye Ye,~\IEEEmembership{Member,~IEEE,} \\
		Debashish Pal,~\IEEEmembership{Member,~IEEE,} 
	Jean-Baptiste Thibault,~\IEEEmembership{Member,~IEEE,} \\
	Ken D. Sauer,~\IEEEmembership{Member,~IEEE,}	
	and Charles A. Bouman,~\IEEEmembership{Fellow,~IEEE,} \\ }
\end{table*}

\pagebreak

\setcounter{equation}{0}
\setcounter{figure}{0}
\setcounter{table}{0}
\setcounter{page}{1}
\setcounter{section}{0}
\makeatletter

\section{Typical images from the training dataset}

\begin{figure*}[!t]
\centering
\begin{tabular}{C{2.2in}C{2.2in}C{2.2in}} 
\includegraphics[width=2.2in]{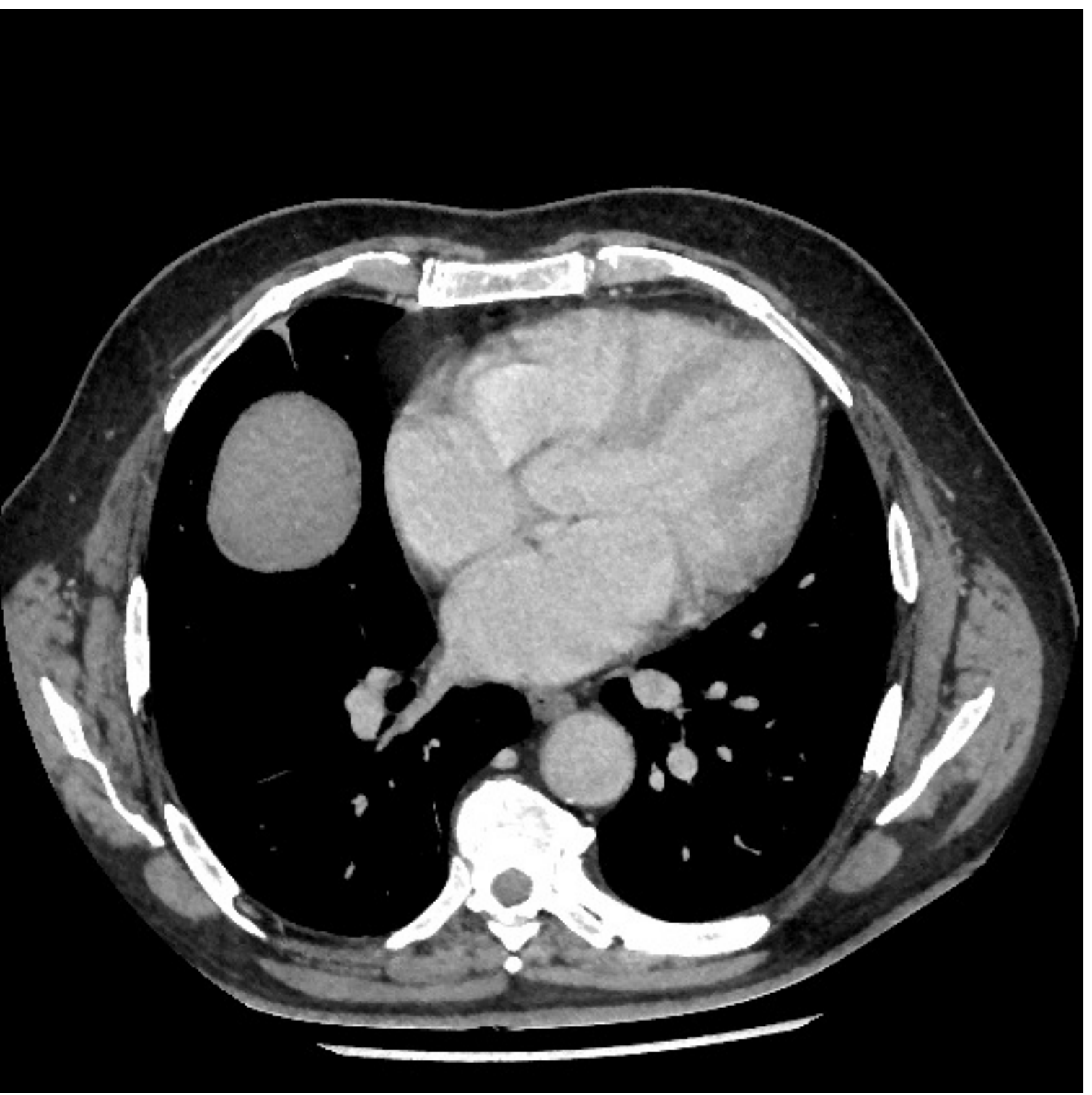} &
\includegraphics[width=2.2in]{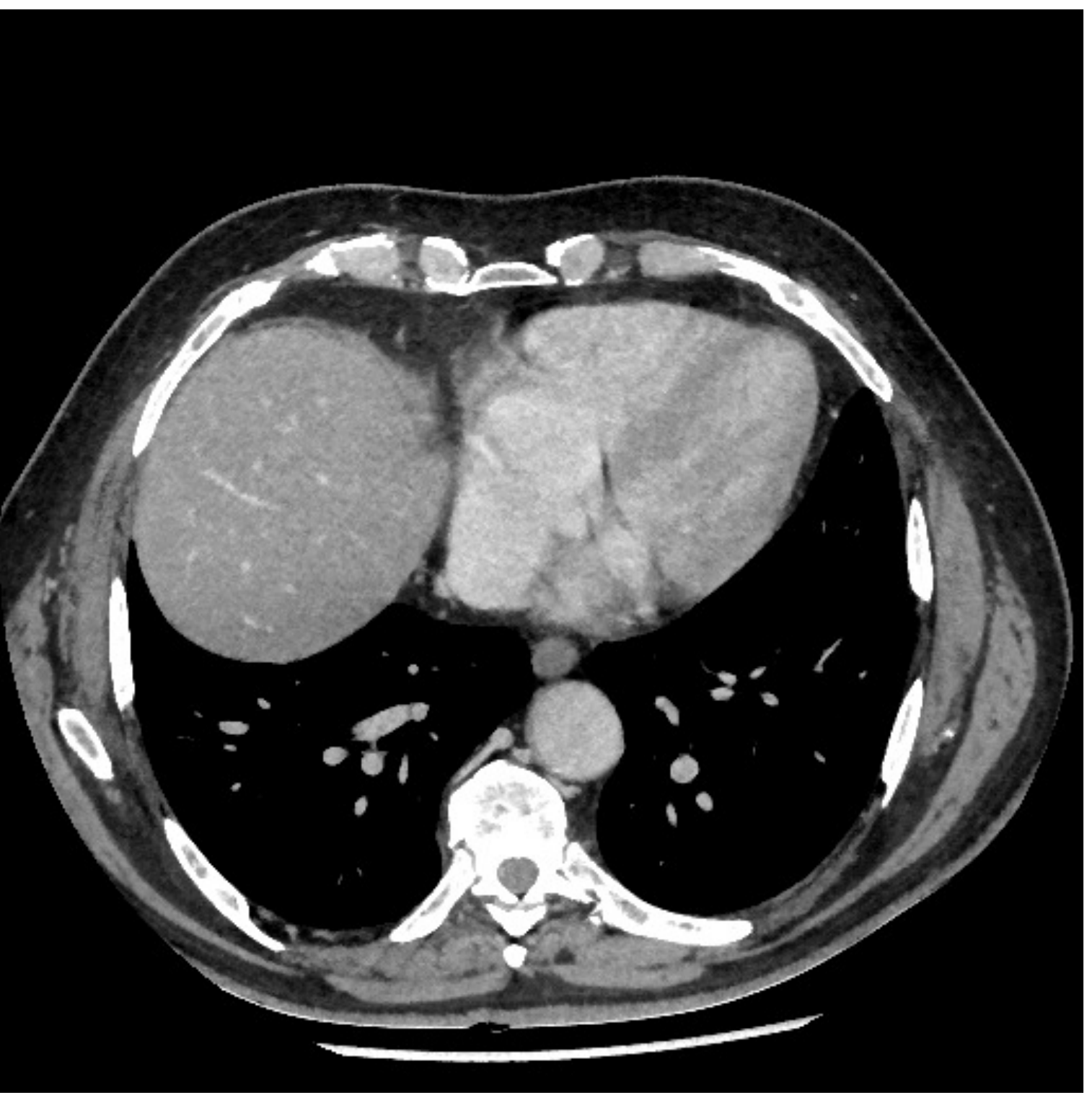} &
\includegraphics[width=2.2in]{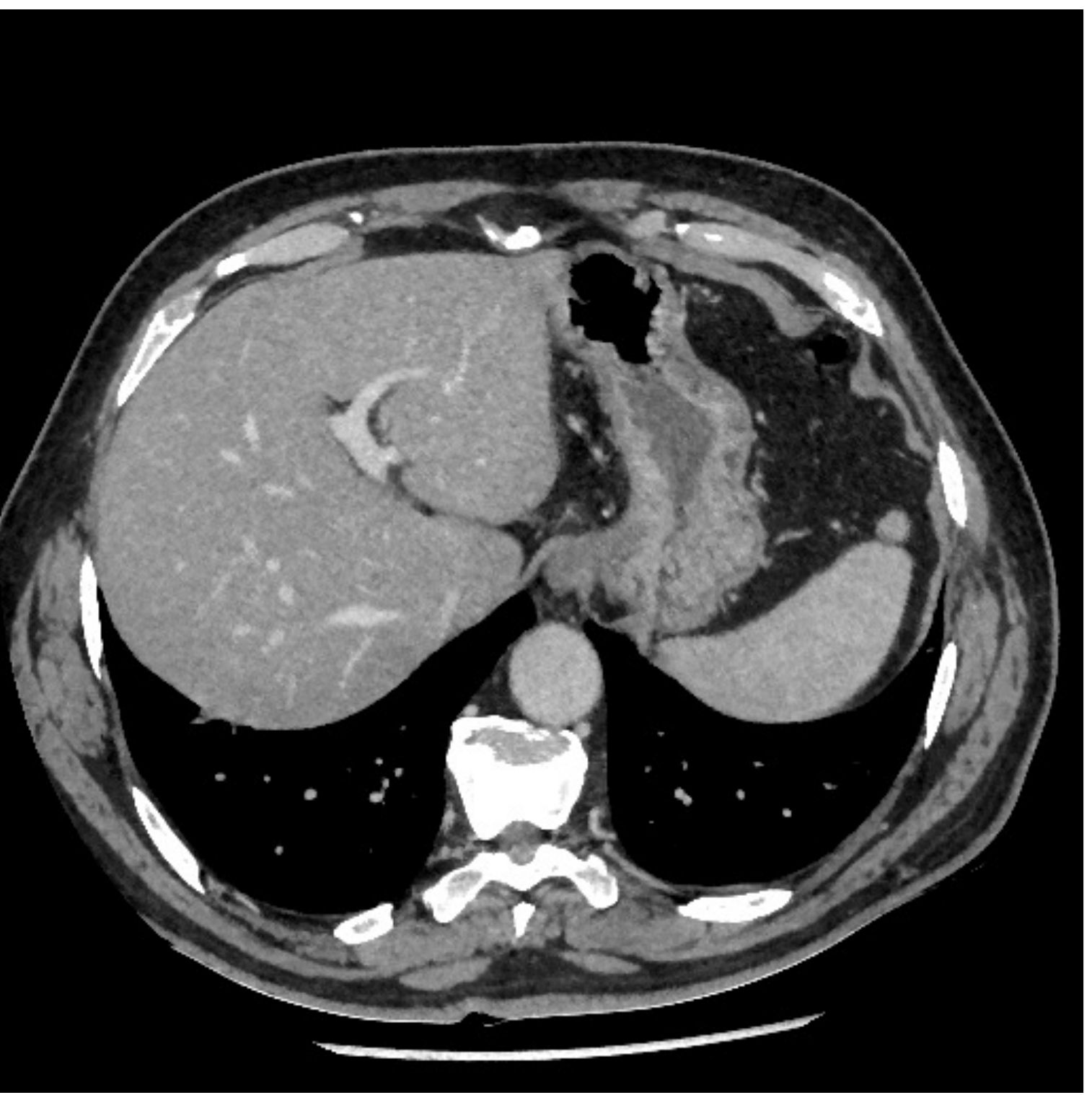} \\
\includegraphics[width=2.2in]{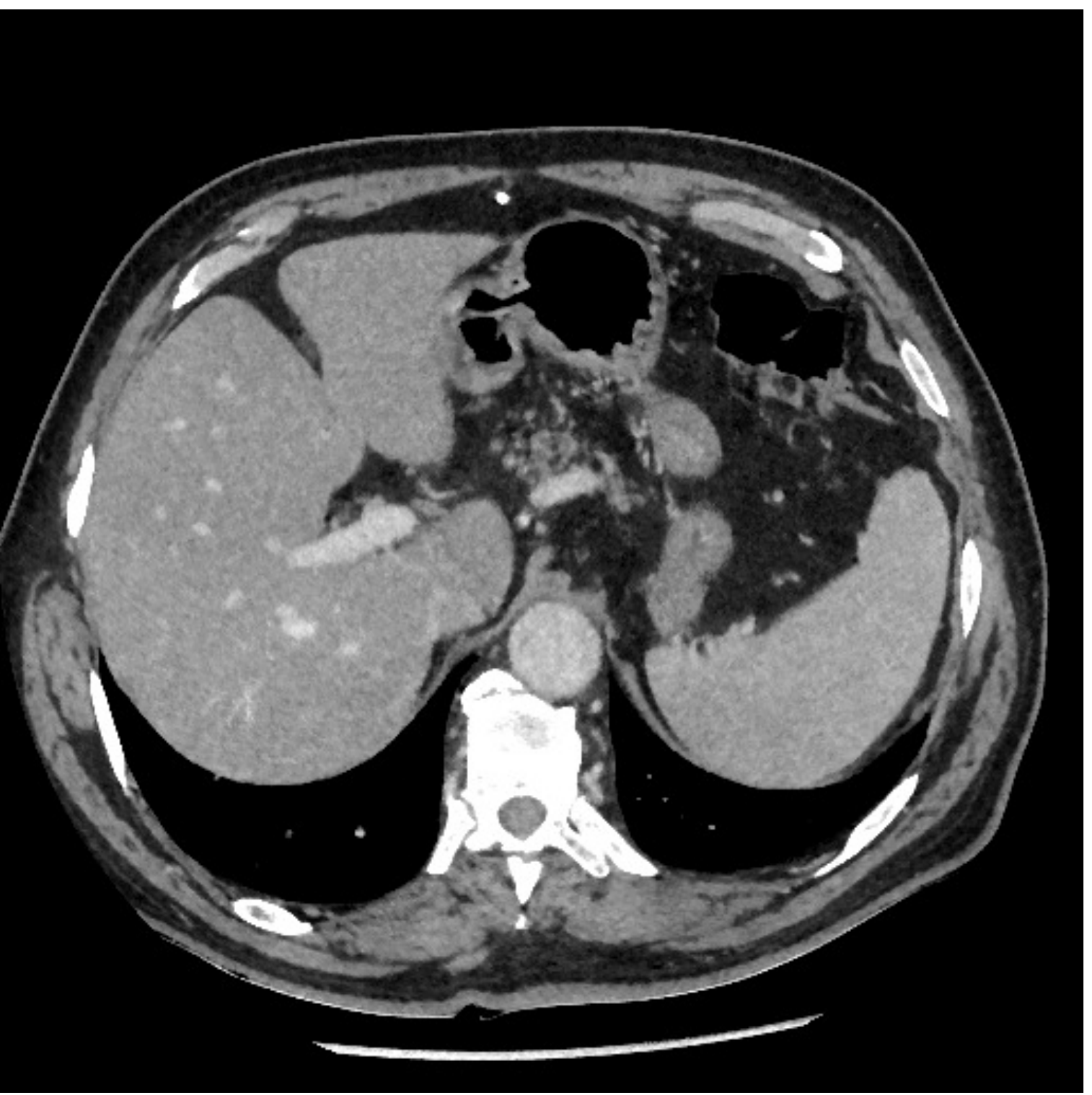} &
\includegraphics[width=2.2in]{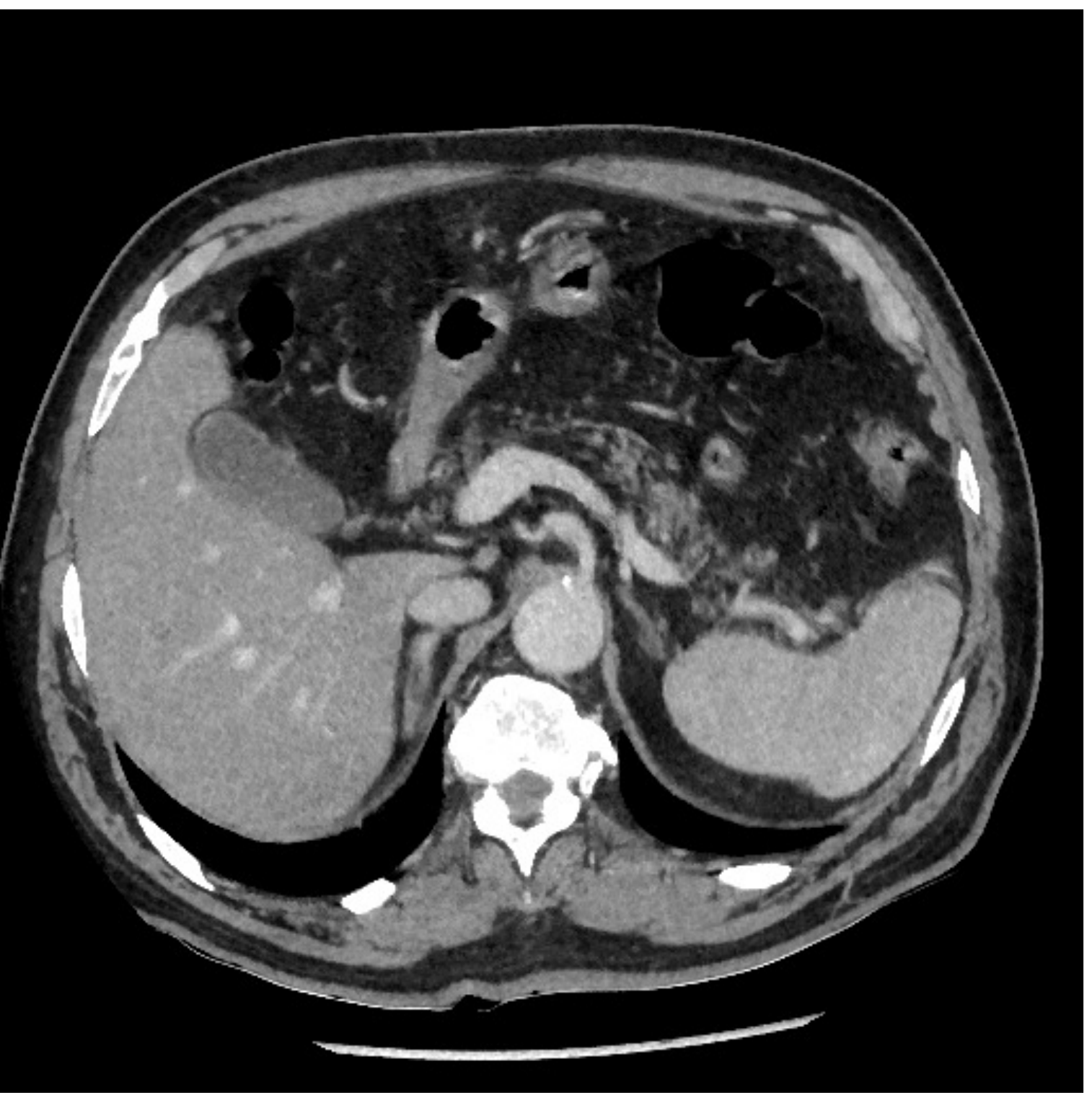} &
\includegraphics[width=2.2in]{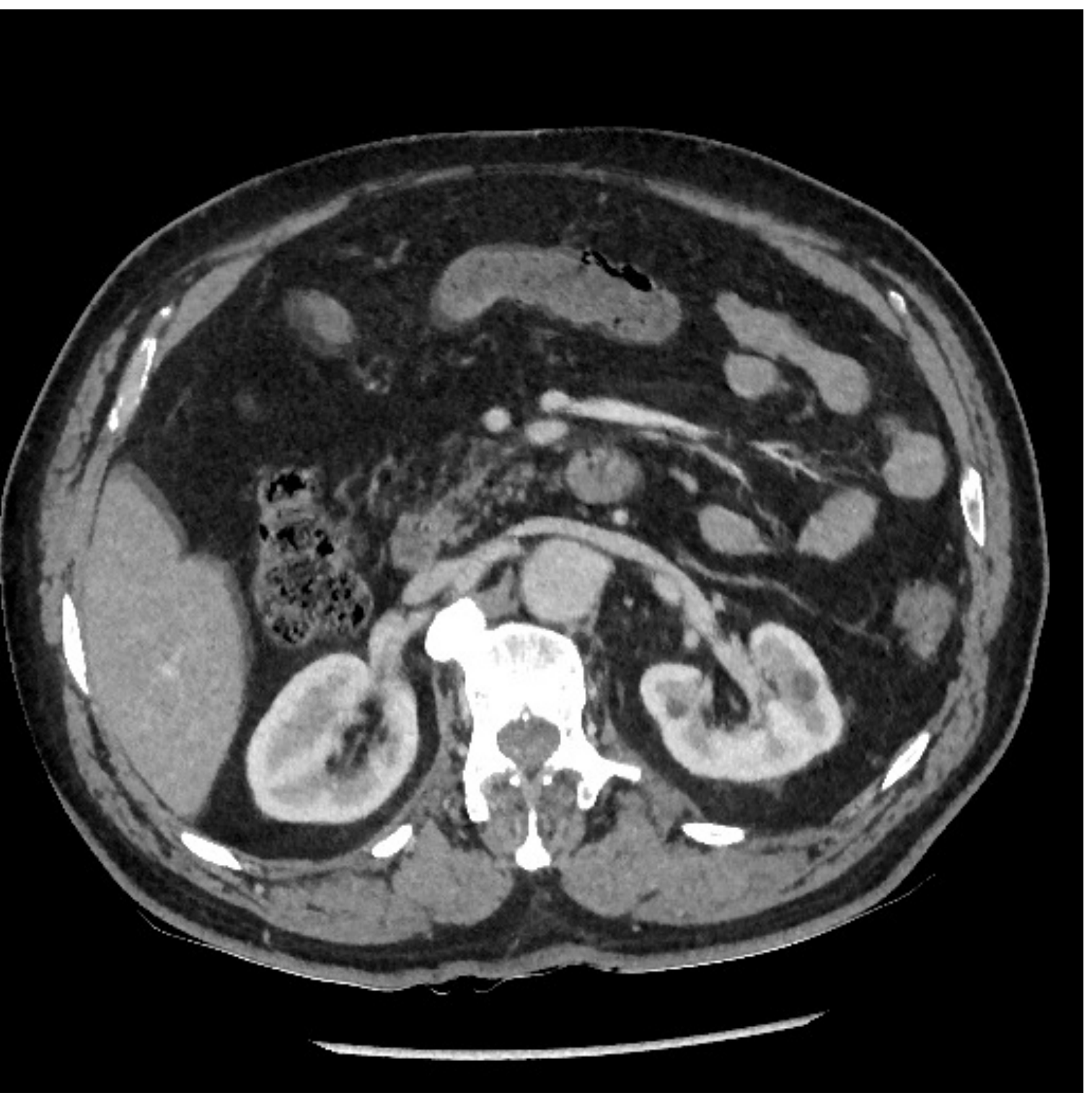} \\
\includegraphics[width=2.2in]{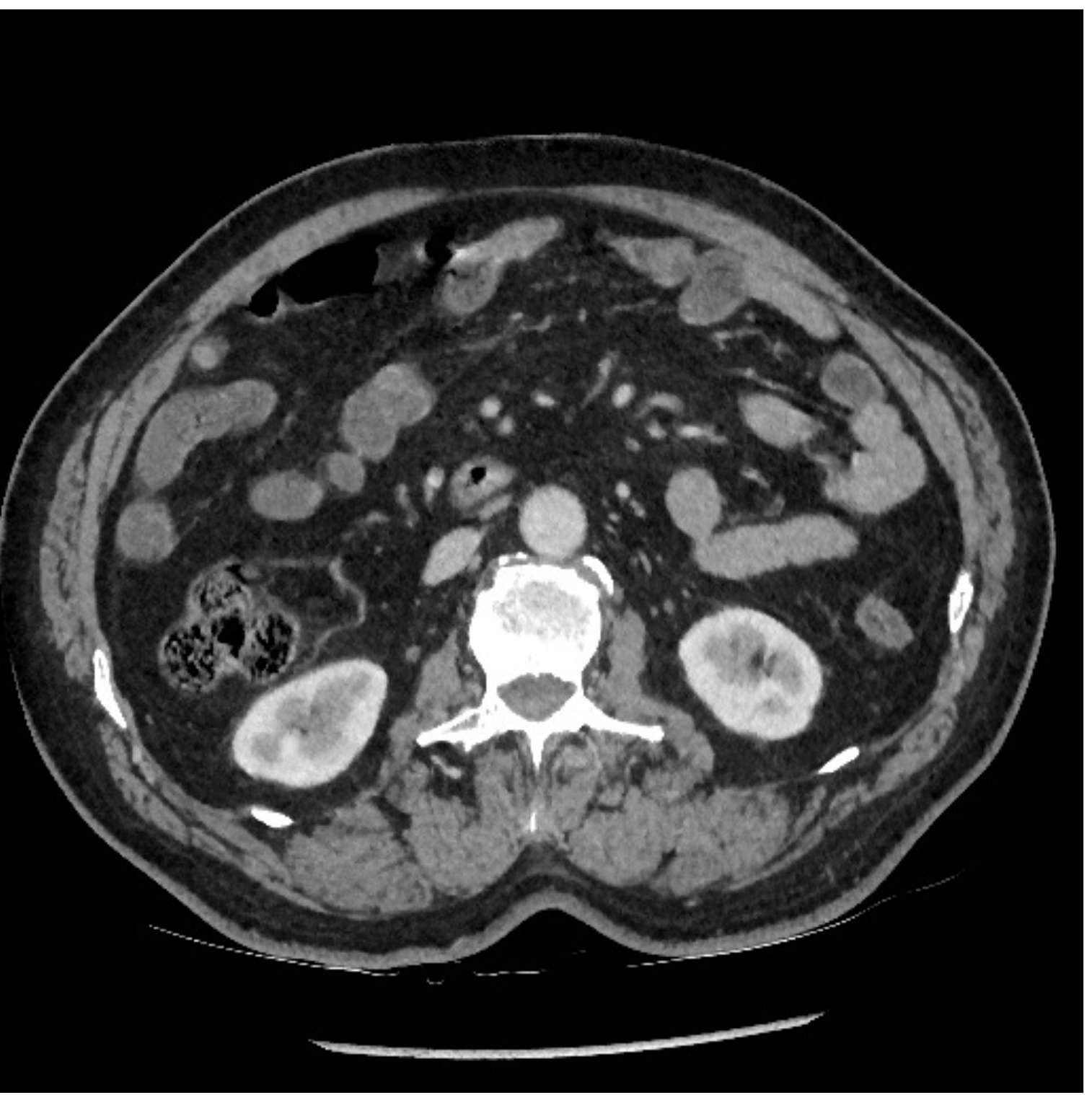} &
\includegraphics[width=2.2in]{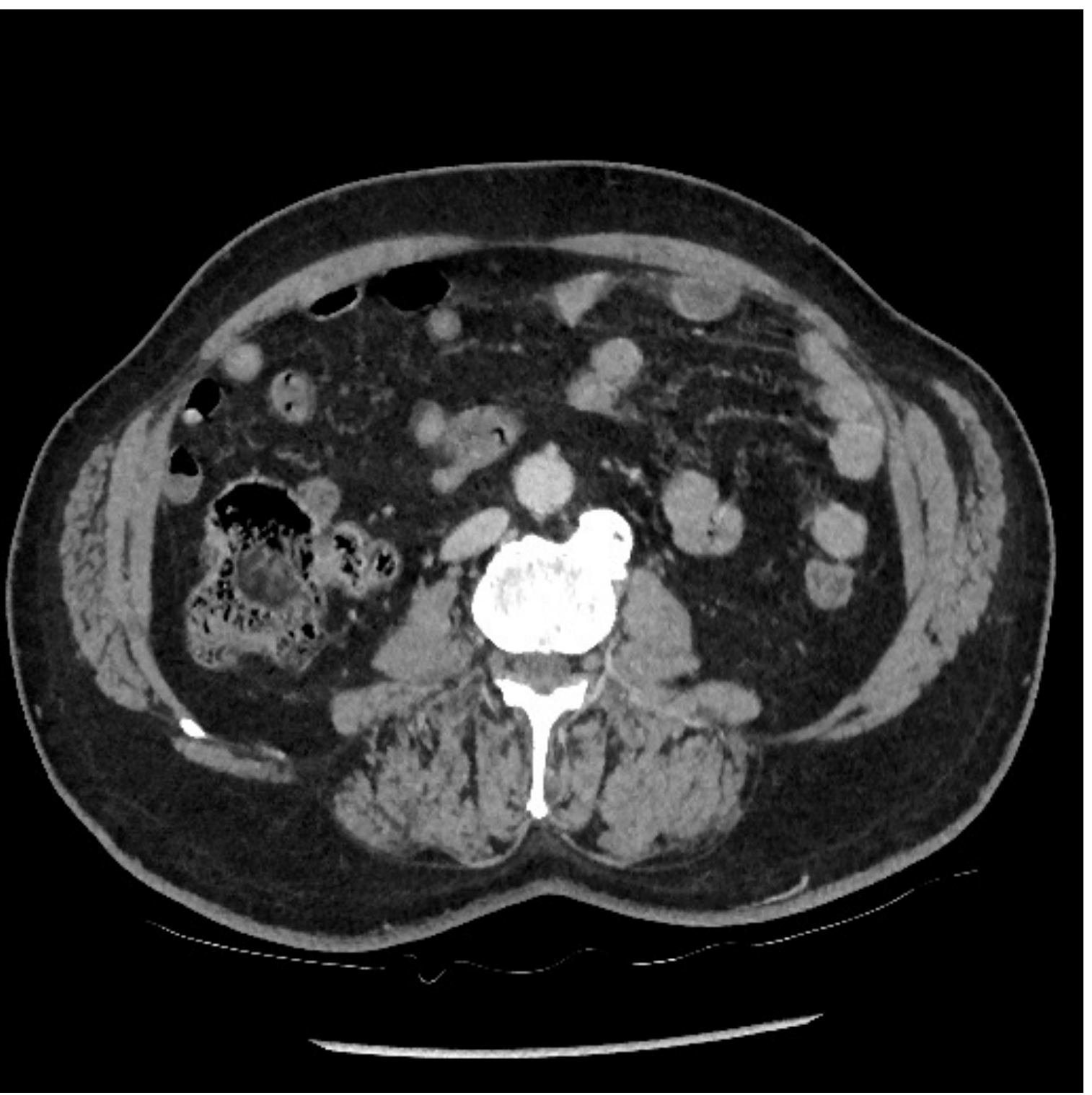} &
\includegraphics[width=2.2in]{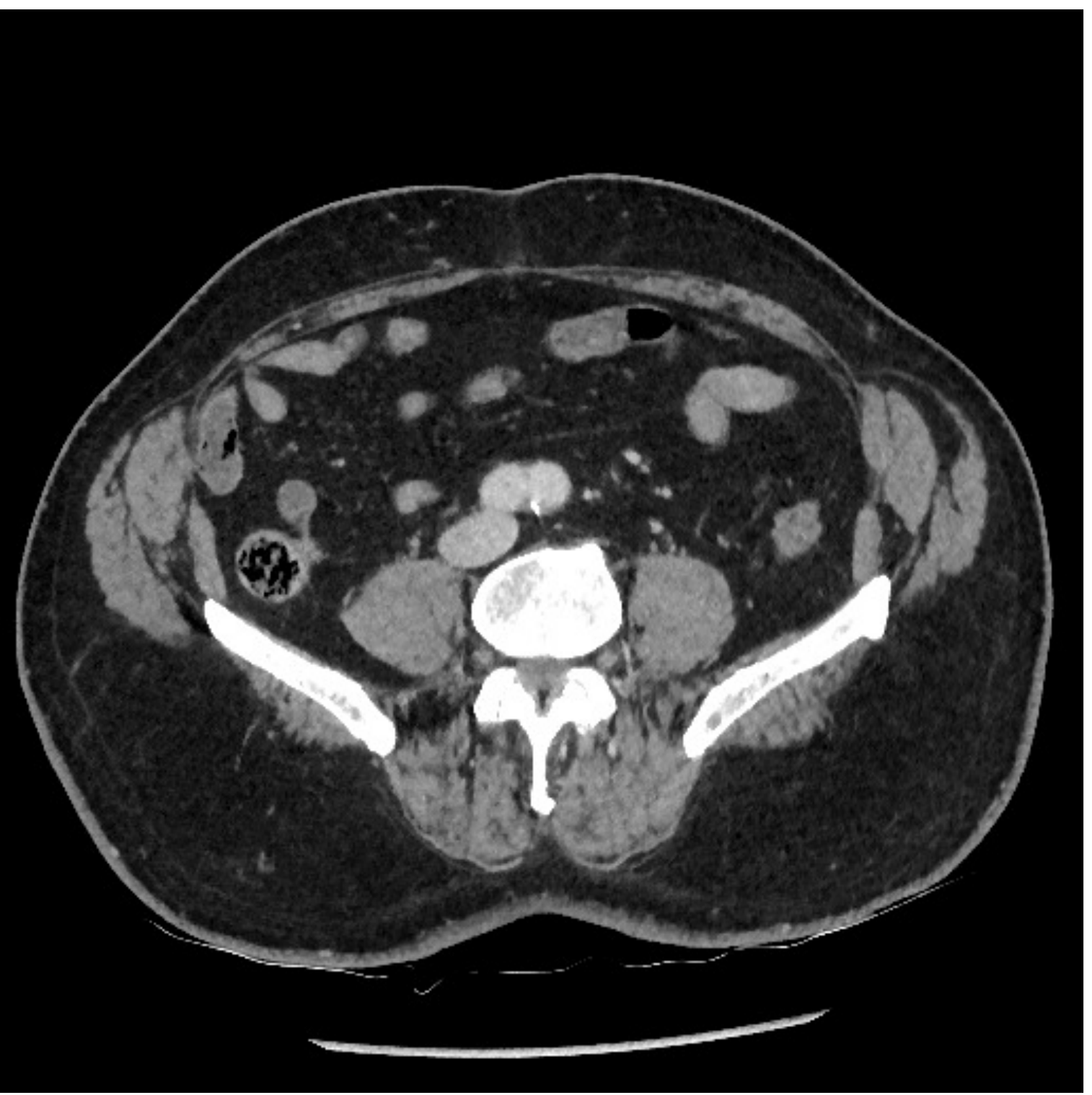} 
\end{tabular}
\caption{Typical images from the training dataset.
2-D axial slices are presented.
Display window is \mbox{[-160 240] HU}.}
\label{fig:train_typical}
\end{figure*}

Fig.~\ref{fig:train_typical} presents a number of typical images from the training dataset used in Sec. V-A.
The training dataset was collected on a GE Discovery CT750 HD scanner  
in \mbox{$64\times0.625$ mm} helical mode with \mbox{100 kVp}, \mbox{500 mA}, \mbox{0.8 s/rotation}, \mbox{pitch 0.984:1}, and reconstructed in \mbox{360 mm} field-of-view (FOV).

\setlength\tabcolsep{0in}

\section{Training for GM-MRFs with various numbers of components}

Table~\ref{tab:train_diffK} lists the number of GM components used for each subgroup
in each GM-MRF with $5\times5\times3$ patch models, as described in Sec. V-A.
Different groups roughly capture different materials or tissue types in the image,
as group 1 for air, group 2 for lung tissue, group 3 for smooth soft tissue, group 4 for low-contrast soft-tissue edge,
group 5 for high-contrast edge, and group 6 for bone.

\begin{table*}[!t]
\renewcommand{\arraystretch}{1.3}
\caption{Number of GM components for each subgroup in each $5\times5\times3$ GM-MRF model.}
\label{tab:train_diffK}
\centering
\begin{tabular}{|C{1.5in}|C{0.8in}|C{0.8in}|C{0.8in}|C{0.8in}|C{0.8in}|C{0.8in}|} \hline
Group index & 1 & 2 & 3 & 4 & 5 & 6 \\ \hline
6 components &  1 & 1 & 1 & 1 & 1 & 1 \\ \hline
15 components &  1 & 3 & 2 & 3 & 3 & 3 \\ \hline
31 components &  1 & 7 & 5 & 7 & 7 & 7 \\ \hline
66 components &  1 & 15 & 5 & 15 & 15 & 15 \\ \hline
131 components &  1 & 30 & 10 & 30 & 30 & 30 \\ \hline
\end{tabular}
\end{table*}

\section{Supplemental results}
\subsection{2-D image denoising}

\begin{figure*}[!t]
\centerline{
\subfloat[original $3\times3$ GM-MRF (14.05 HU)]{\includegraphics[width=2.1in]{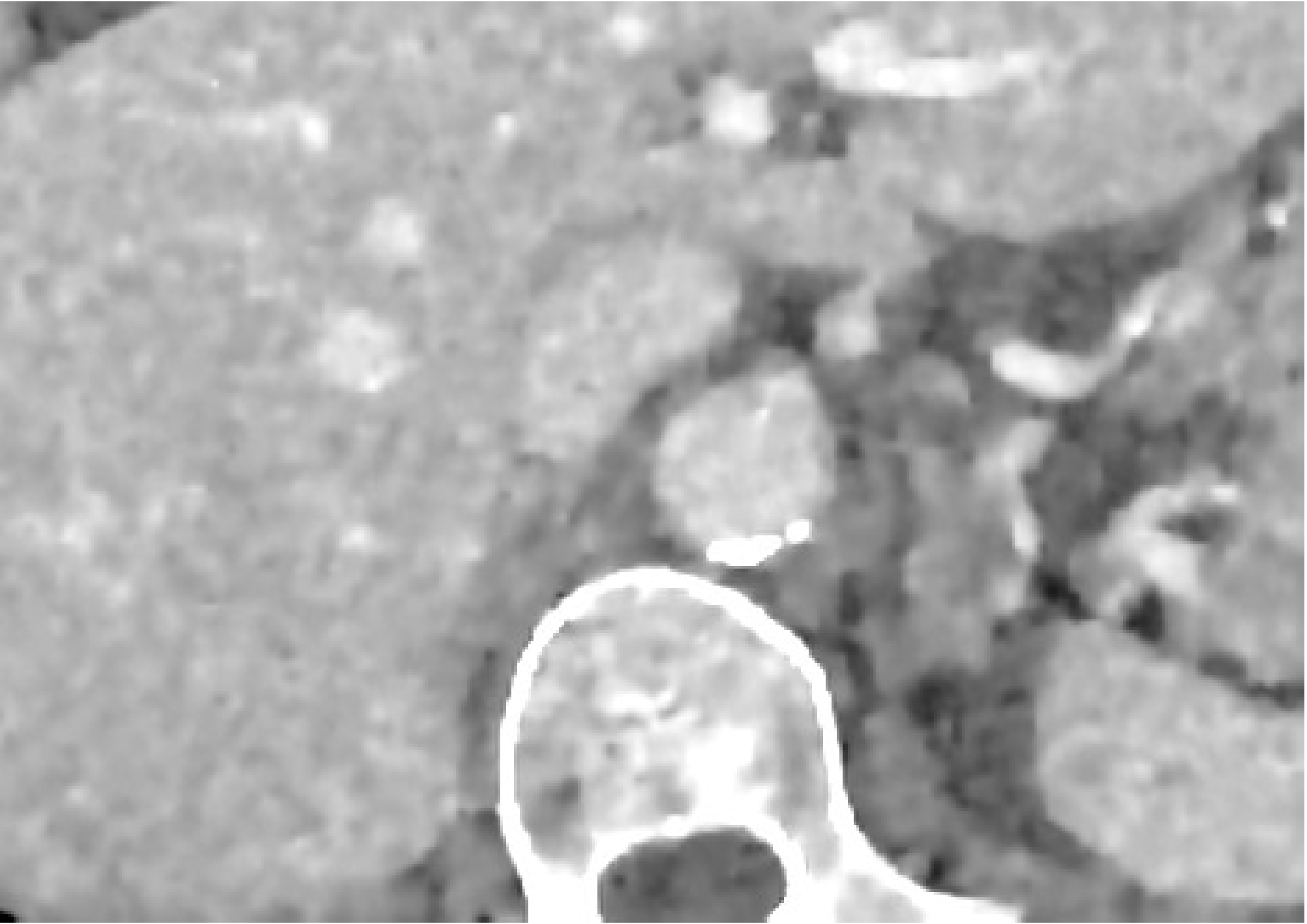}}\
\subfloat[original $5\times5$ GM-MRF (13.78 HU)]{\includegraphics[width=2.1in]{figs/zhang25.pdf}}\
\subfloat[original $7\times7$ GM-MRF (13.62 HU)]{\includegraphics[width=2.1in]{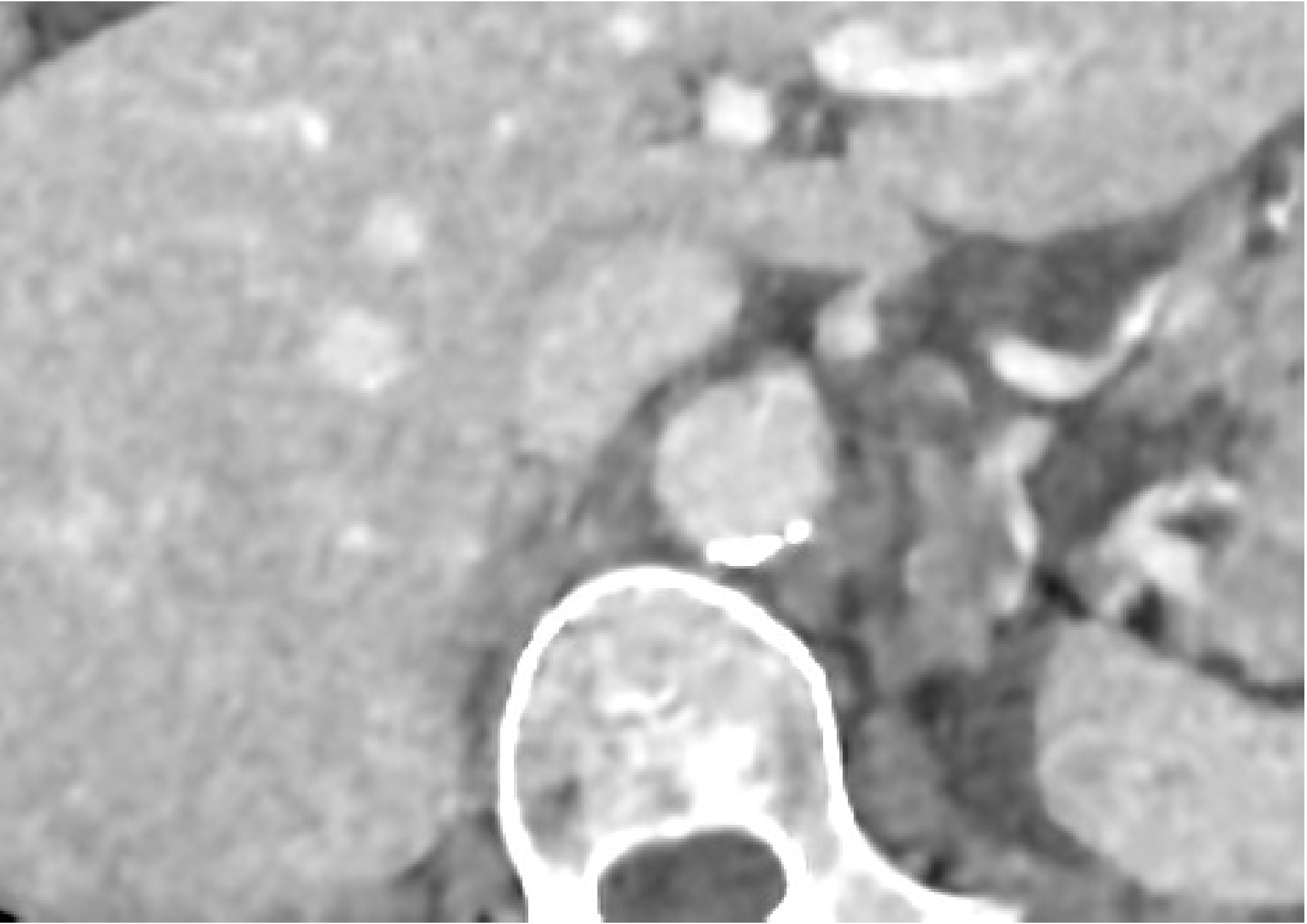}}}
\centerline{
\subfloat[adjusted $3\times3$ GM-MRF (14.84 HU)]{\includegraphics[width=2.1in]{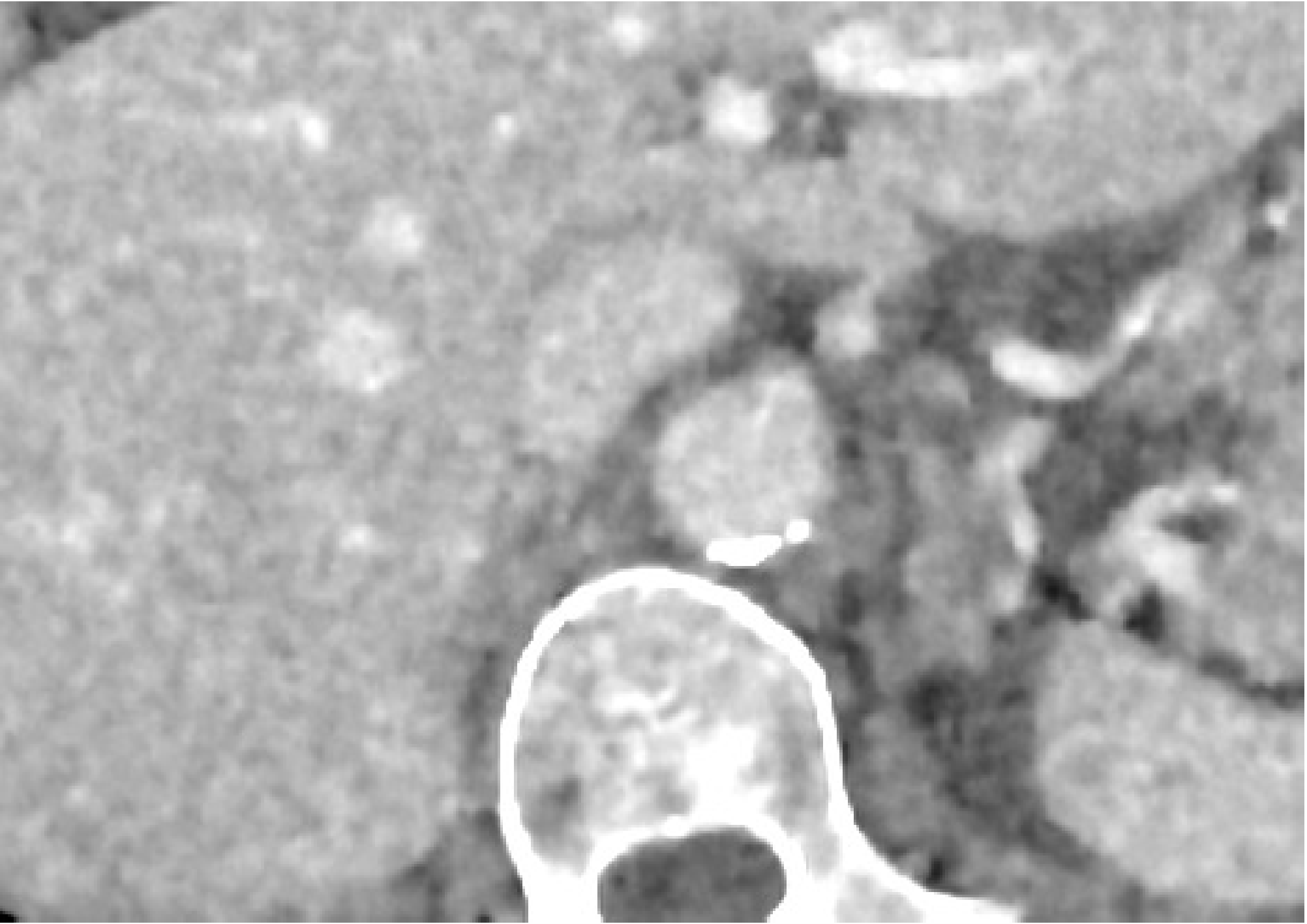}}\
\subfloat[adjusted $5\times5$ GM-MRF (14.33 HU)]{\includegraphics[width=2.1in]{figs/zhang26.pdf}}\
\subfloat[adjusted $7\times7$ GM-MRF (14.26 HU)]{\includegraphics[width=2.1in]{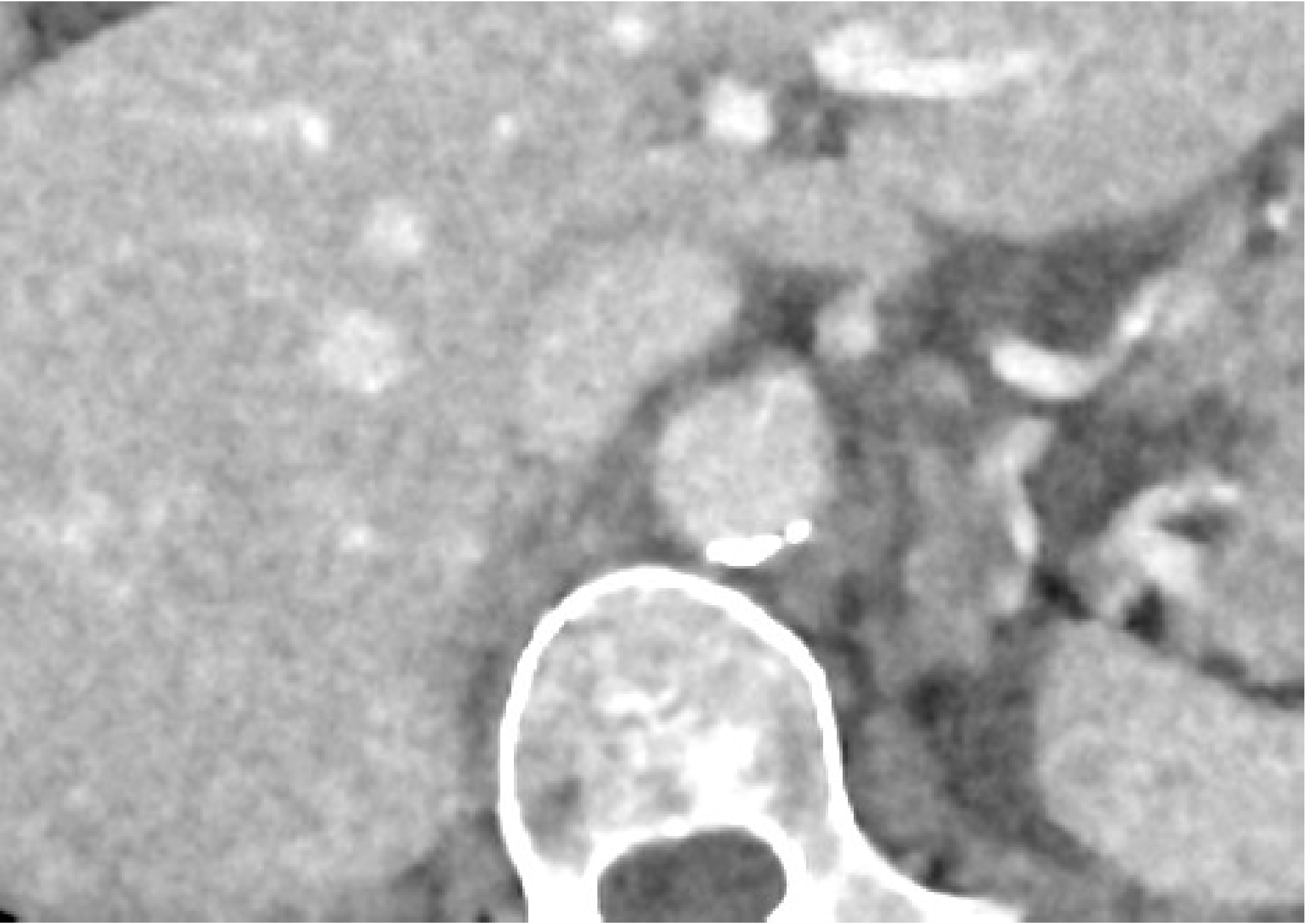}}}
\caption{GM-MRF denoising results with different sizes of patch models (with RMSE value). 
Individual image is zoomed to a region containing soft tissue, contrast, and bone for display purposes.
RMSE value between each reconstructed image and the ground truth is reported. 
Display window: [-100 200] HU.
The RMSE value decreases as the patch size increases in GM-MRF models.
Moreover, the $3\times3$ result appears slightly blocky and therefore seems less natural than the $5\times5$ and $7\times7$ results.}
\label{fig:denoising_diffsize}
\end{figure*}

Fig.~\ref{fig:denoising_diffsize} presents the denoising results by using GM-MRF methods with different sizes of patch models.
It is observed that the RMSE value decreases as the patch size increases in the GM-MRF models, 
since the model becomes more expressive as patch size increases.
The 3x3 result appears slightly artificially blocky, while the 7x7 result appears more natural.
It is also worth mentioning that the improvement gained from increasing patch size from $3\times3$ to $5\times5$ 
is more significant than that from $5\times5$ to $7\times7$.

\subsection{3-D phantom reconstruction}

\begin{figure*}[!t]
\centerline{
\subfloat[Quantitative measurement of noise and resolution]{\includegraphics[width=3in]{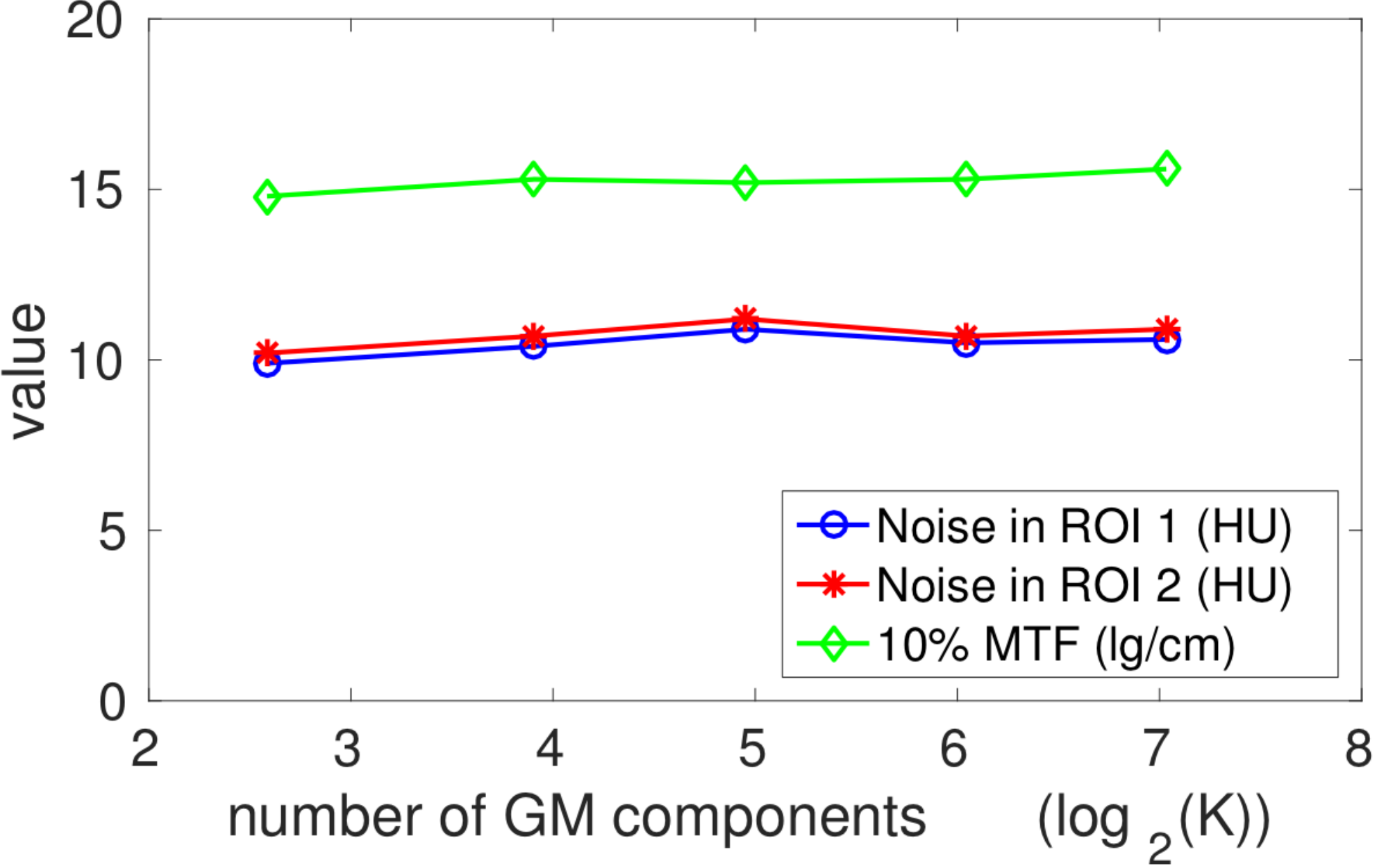}}}\
\centerline{
\subfloat[6 components]{\includegraphics[width=1.4in]{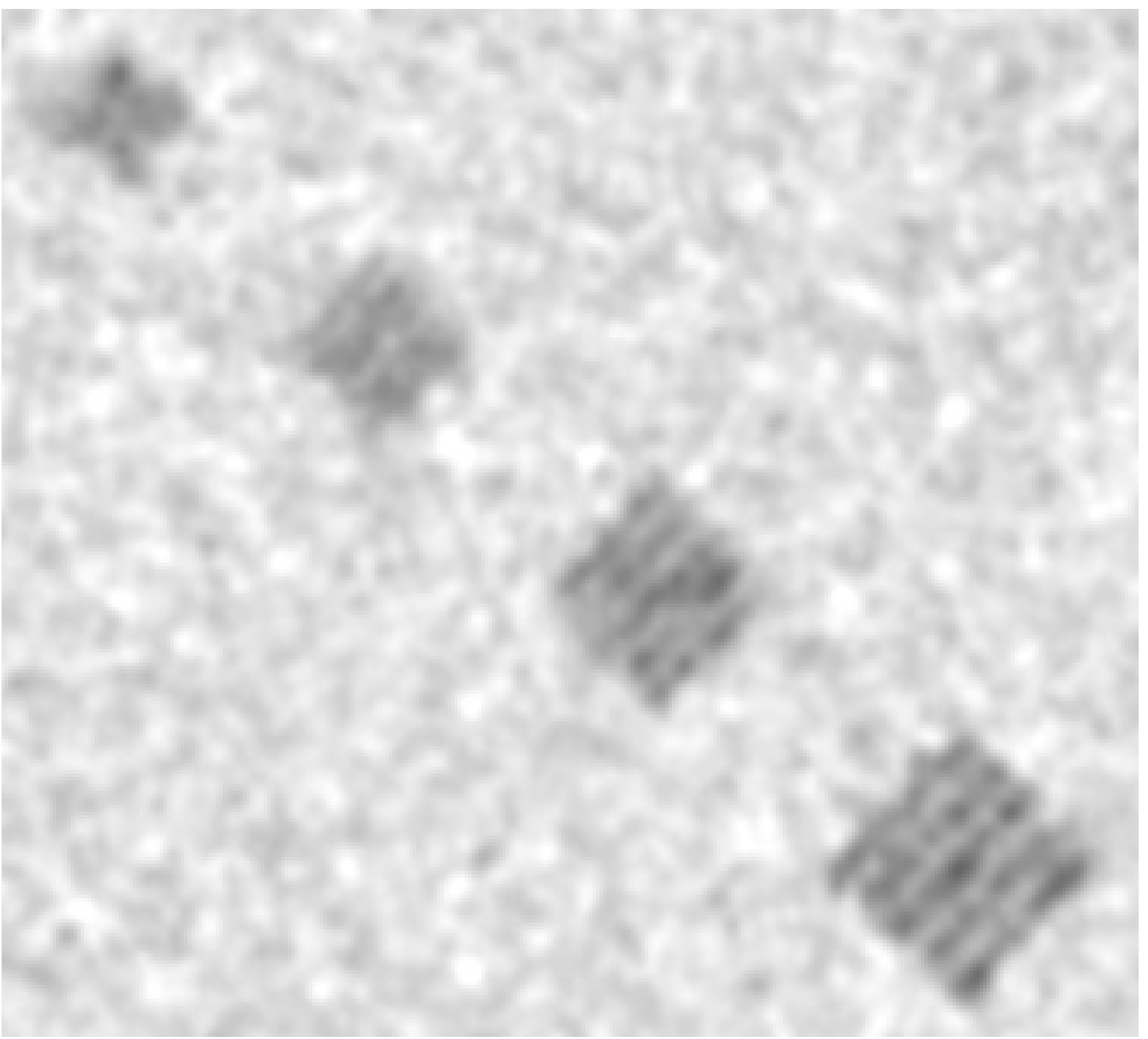}}\ 
\subfloat[15 components]{\includegraphics[width=1.4in]{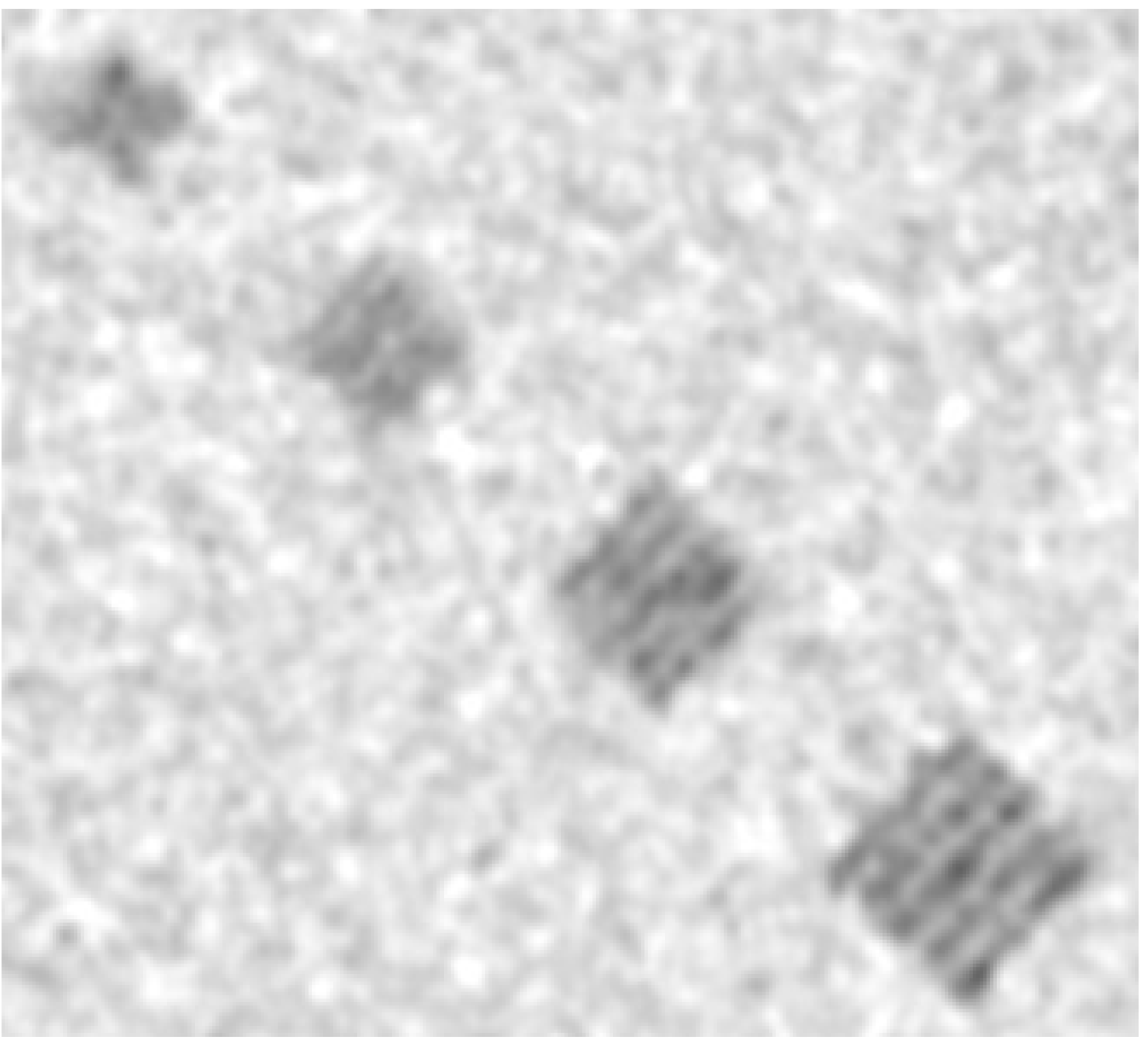}}\
\subfloat[31 components]{\includegraphics[width=1.4in]{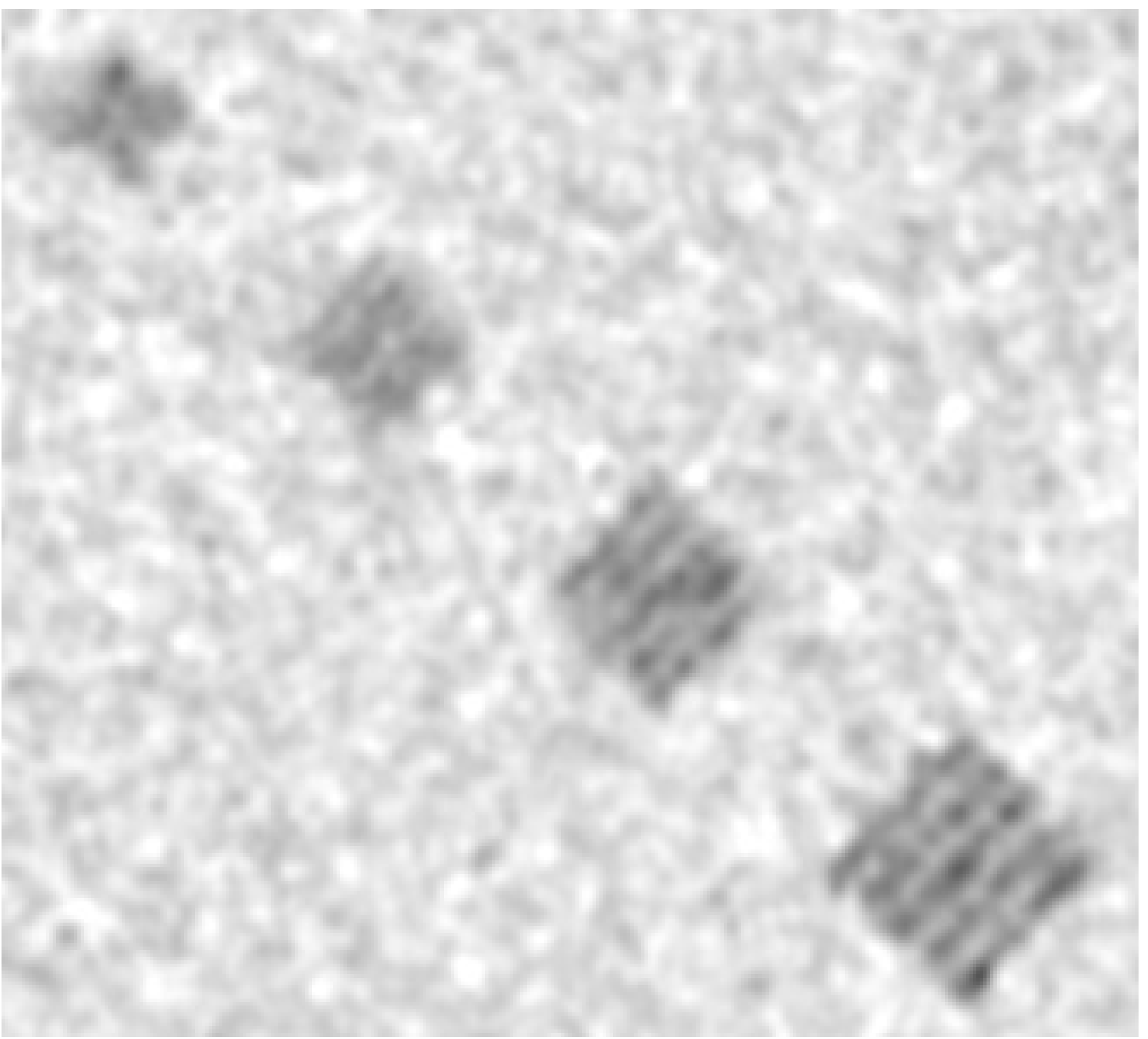}}\
\subfloat[66 components]{\includegraphics[width=1.4in]{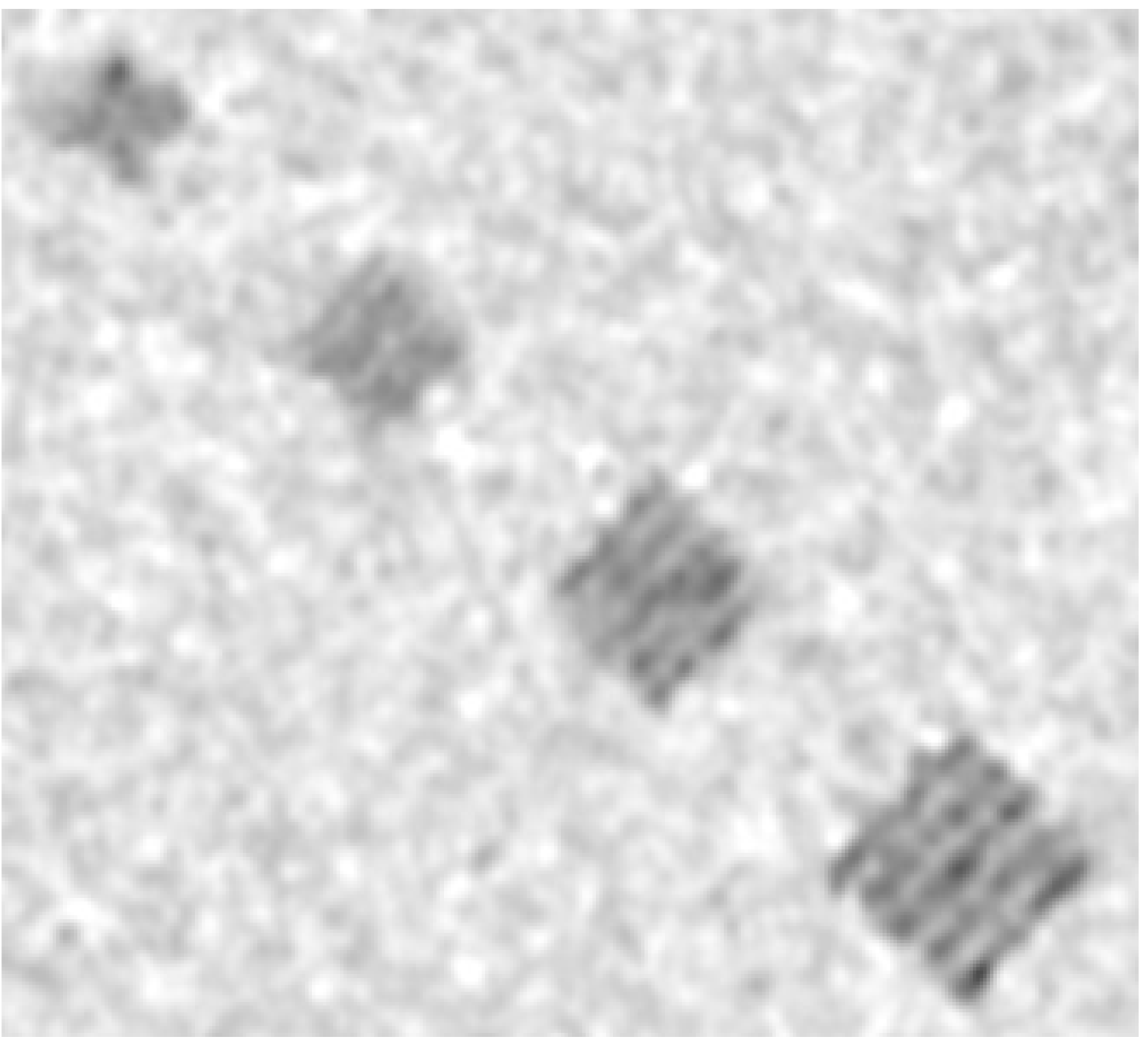}}\
\subfloat[131 components]{\includegraphics[width=1.4in]{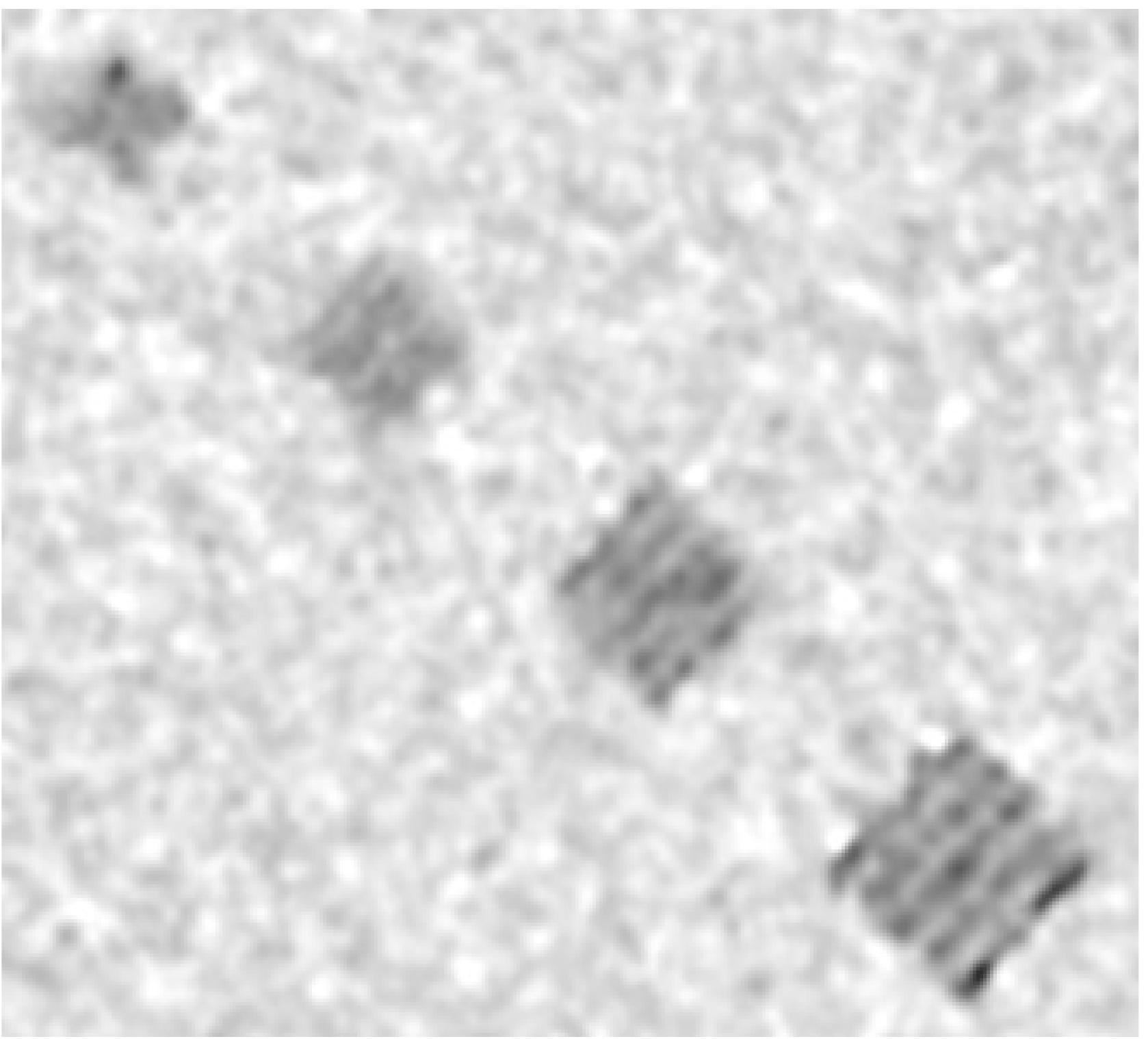}}}
\centerline{ 
\subfloat[Diff: (c) -- (b)]{\includegraphics[width=1.4in]{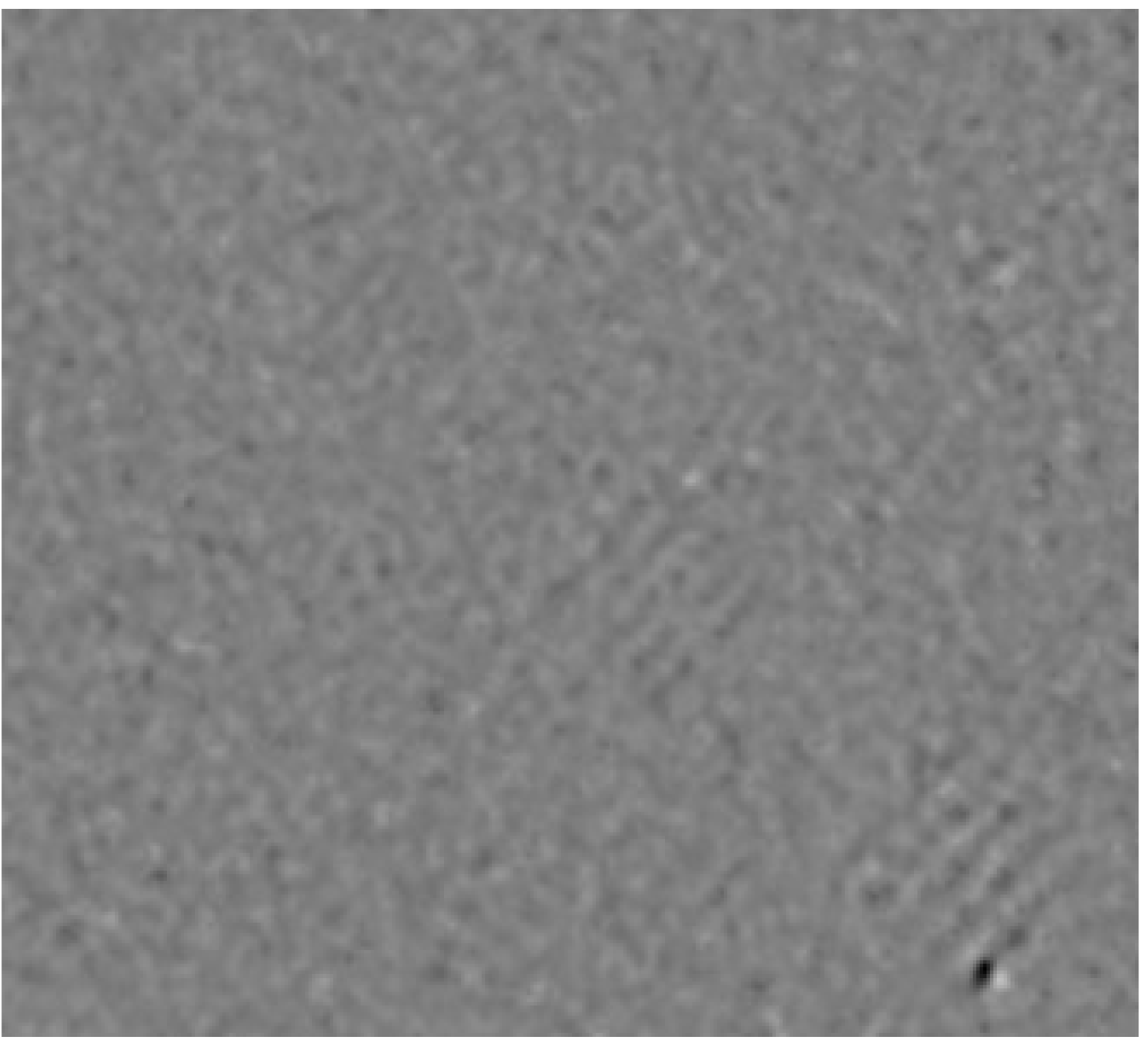}}\
\subfloat[Diff: (d) -- (c)]{\includegraphics[width=1.4in]{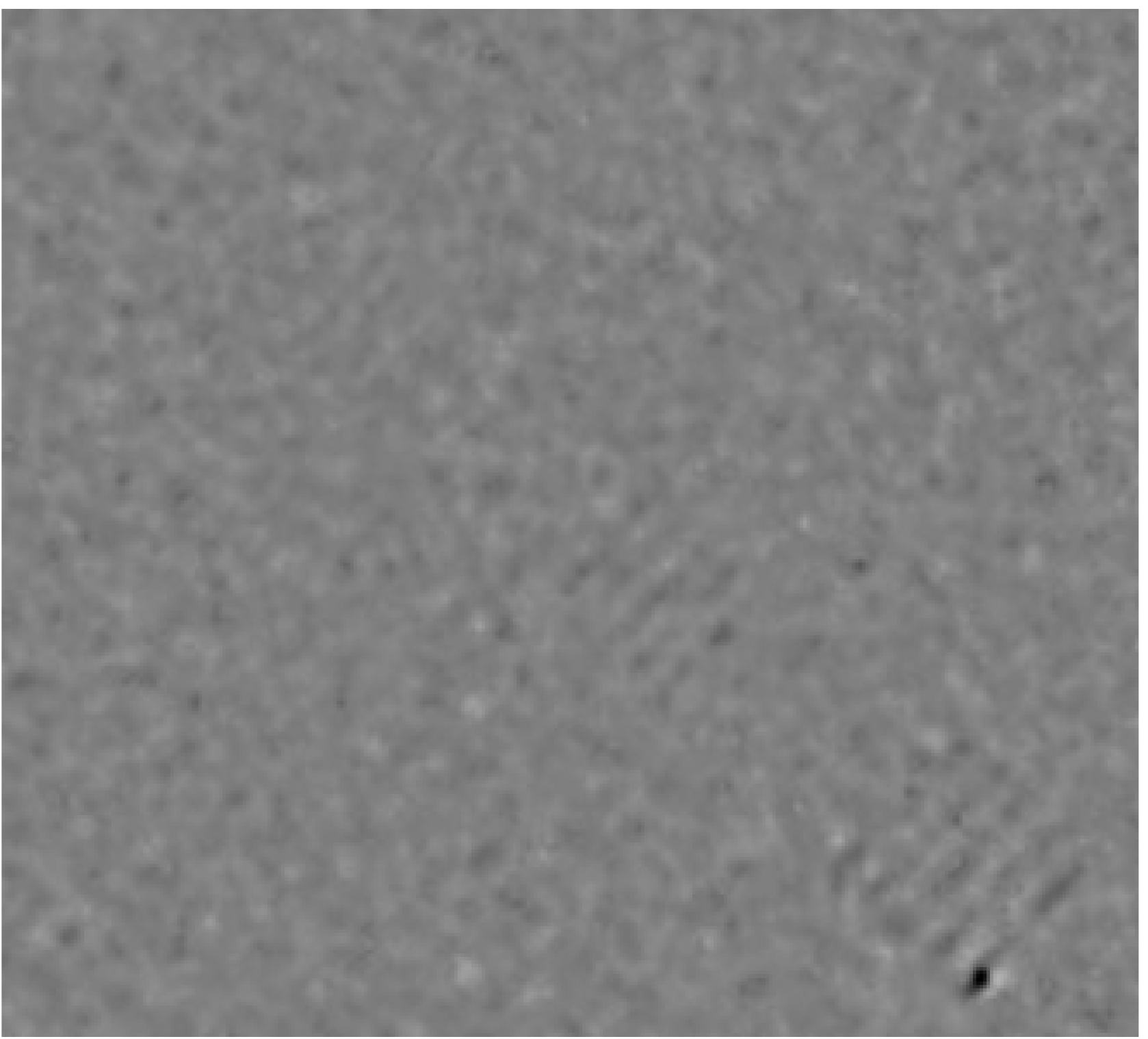}}\
\subfloat[Diff: (e) -- (d)]{\includegraphics[width=1.4in]{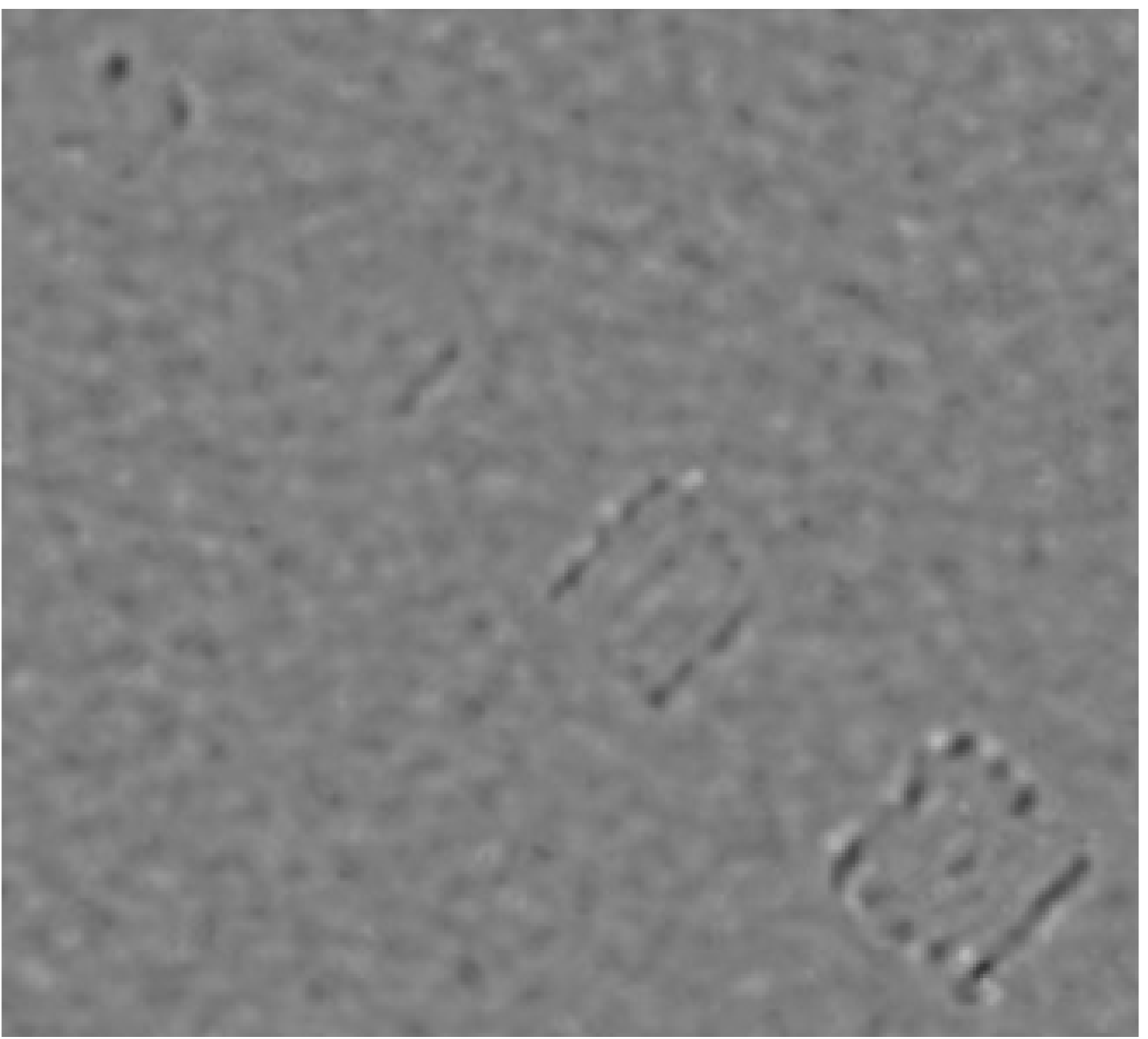}}\
\subfloat[Diff: (f) -- (e)]{\includegraphics[width=1.4in]{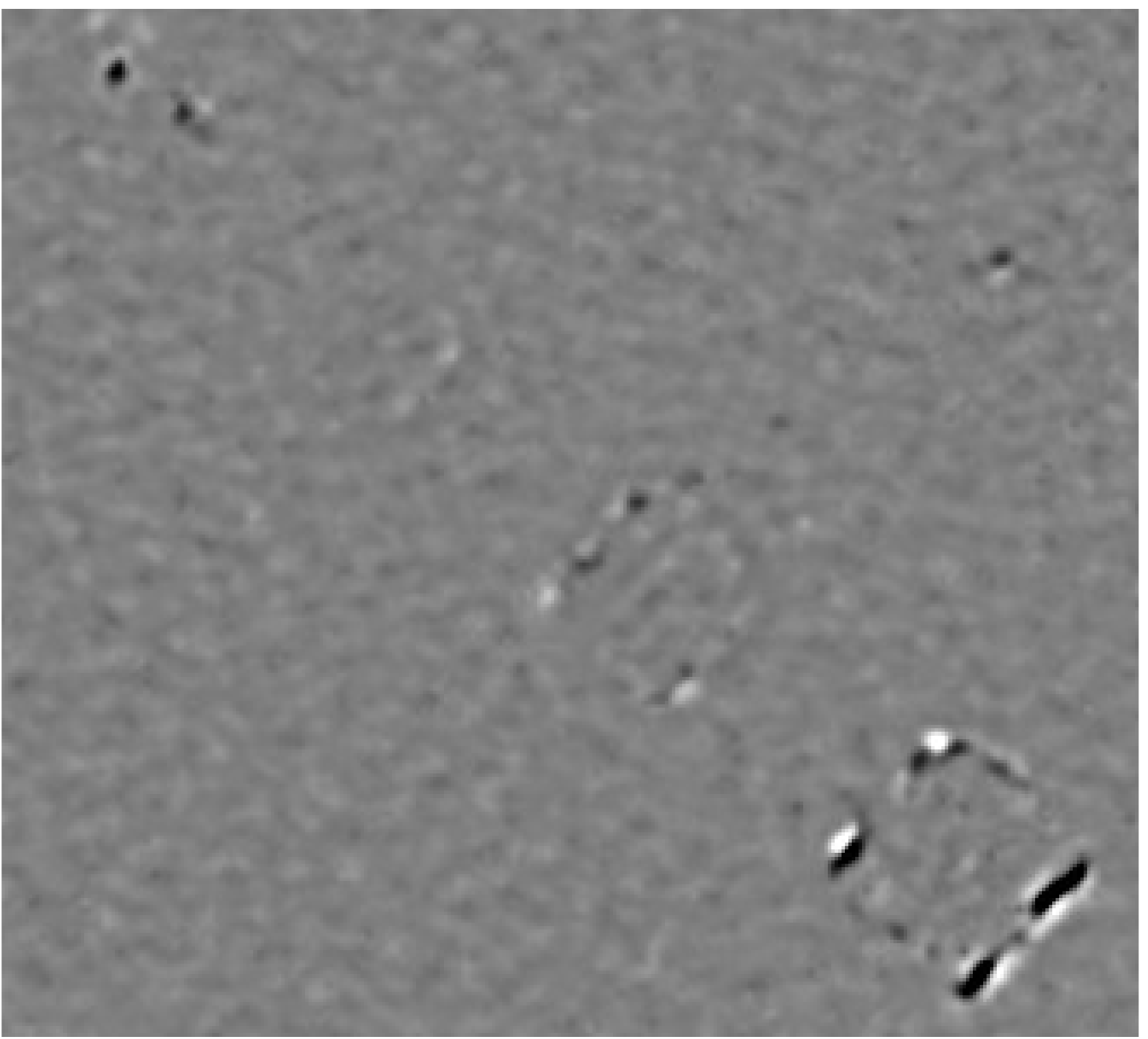}}}
\caption{GEPP reconstructions for 75 mA data, using $5\times5\times3$ GM-MRF model with various number of GM components. Individual image is zoomed to a small FOV containing cyclic bars for display purposes. 
Display window: for GEPP images: [-85 165] HU; for difference images: [-15 15] HU.
As number of GM components increases, the GM-MRF model gradually produce more enhanced cyclic bars.}
\label{fig:gepp_diffK}
\end{figure*}

Fig.~\ref{fig:gepp_diffK} presents the GEPP reconstructions for 75 mA data produced by using the GM-MRF with various number of components. 
Fig. 3(a) shows that, with matched noise level in homogeneous regions, 
trained GM-MRF models with different number of components can achieve similar high-contrast resolution.
Fig. 3(b)--(j) show that the GM-MRF model tends to produce more enhanced cyclic bars with increasing number of components. 
This implies that additional GM components will capture the behavior of edges and structures.

\begin{figure*}[!t]
\centerline{
\subfloat[average eigenvalues with fixed $\alpha$ and different values of $p$]{\includegraphics[height=1.6in]
{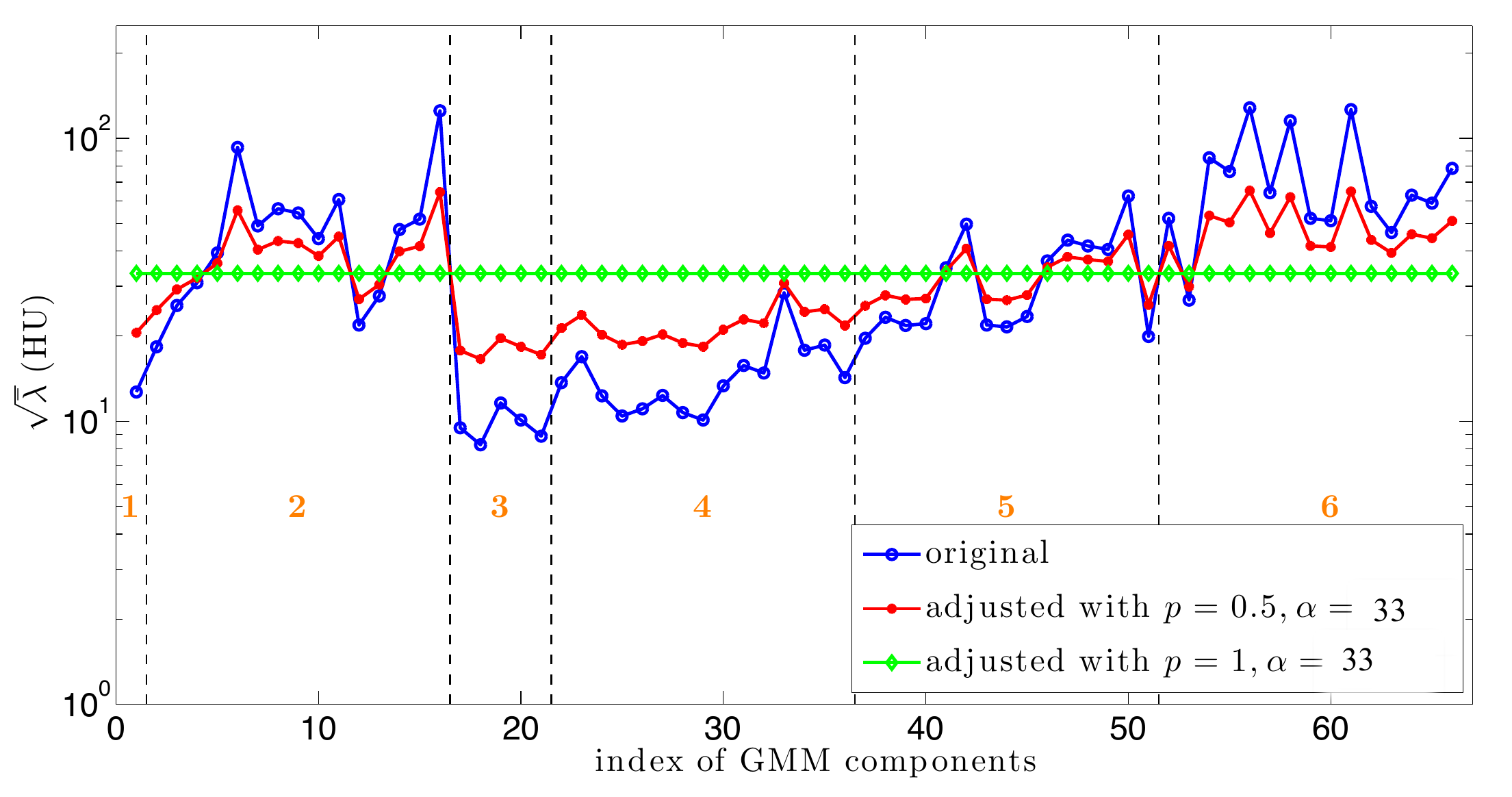}}\qquad 
\subfloat[noise and resolution measurements]{\includegraphics[height=1.6in,trim={0  0 -1in 0},clip]{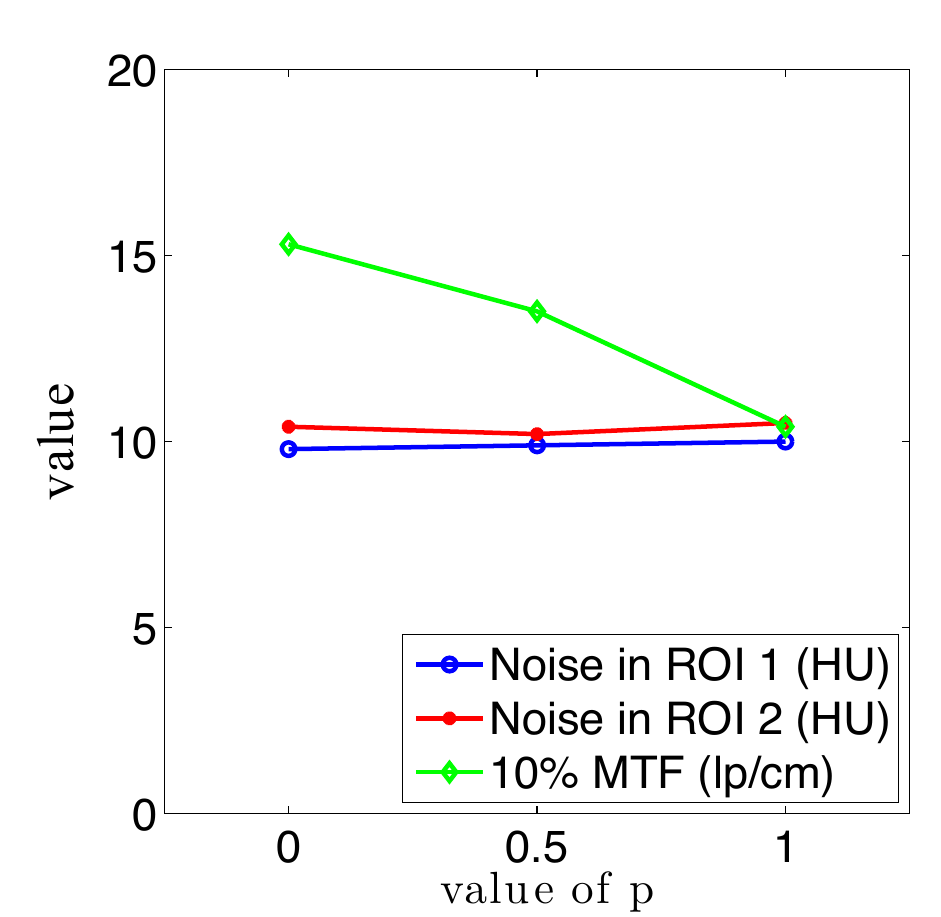}} }
\centerline{ 
\subfloat[$p=0$]{\includegraphics[width=1.4in]{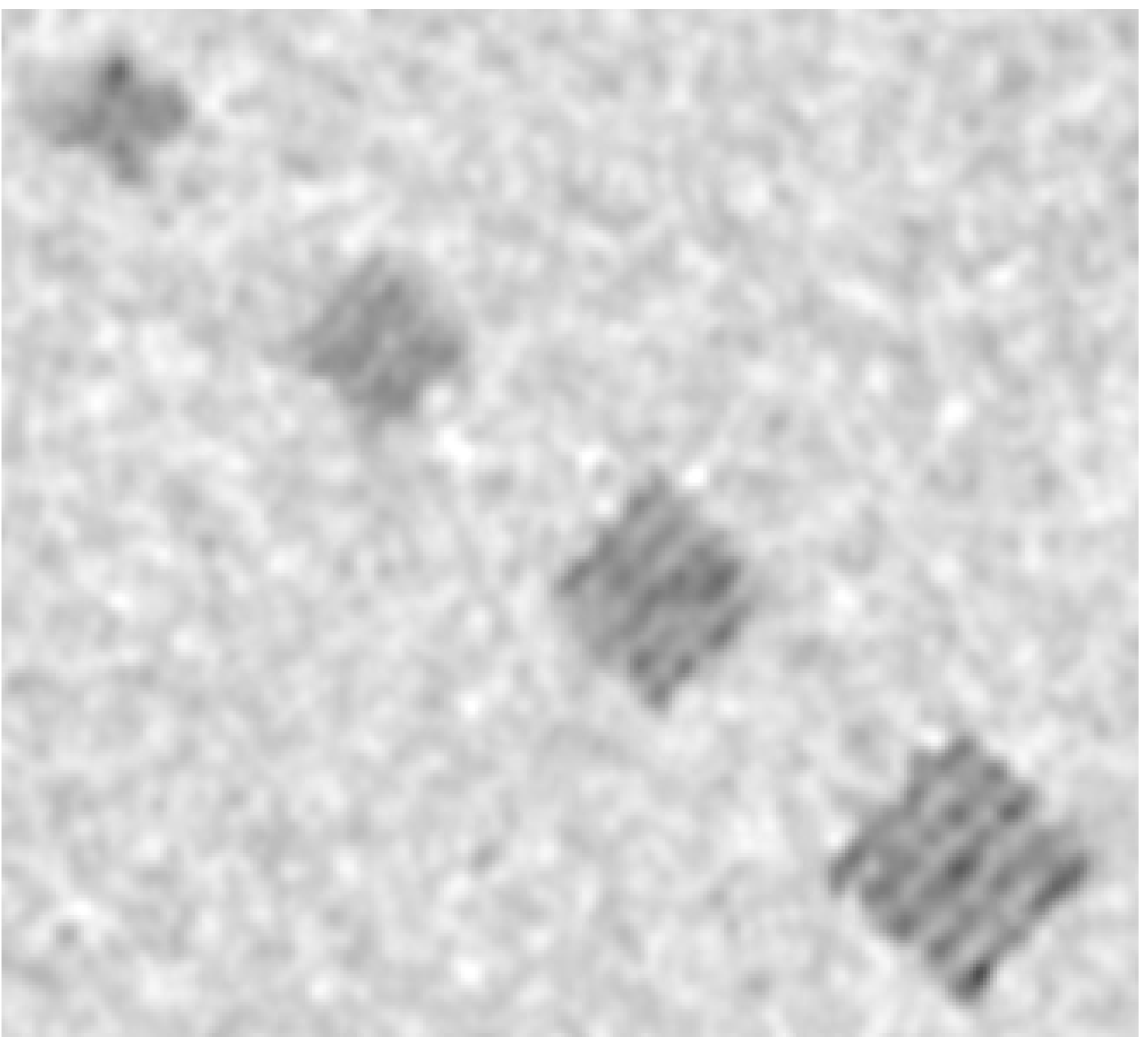}}\ 
\subfloat[$p=0.5$]{\includegraphics[width=1.4in]{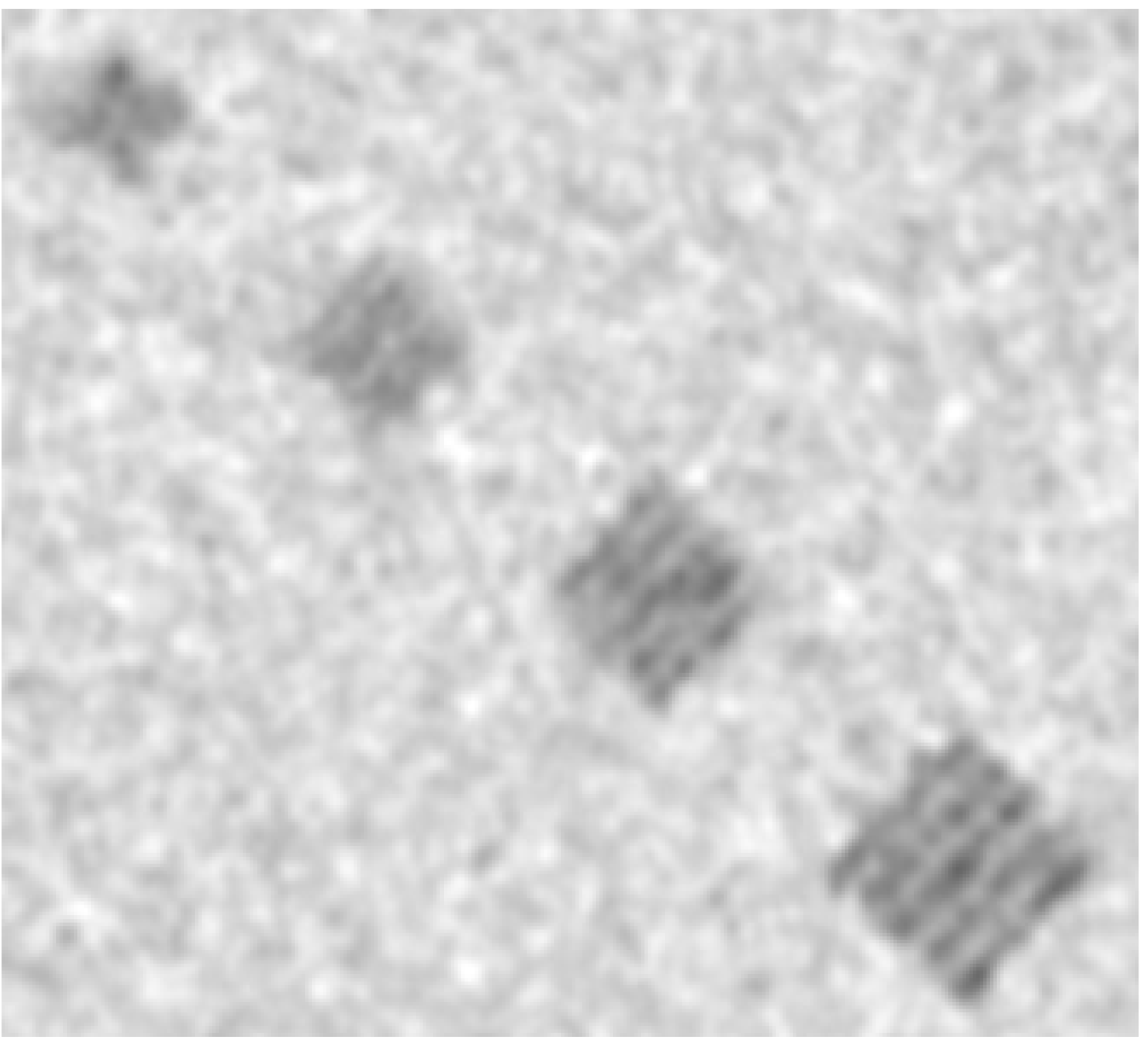}}\
\subfloat[$p=1$]{\includegraphics[width=1.4in]{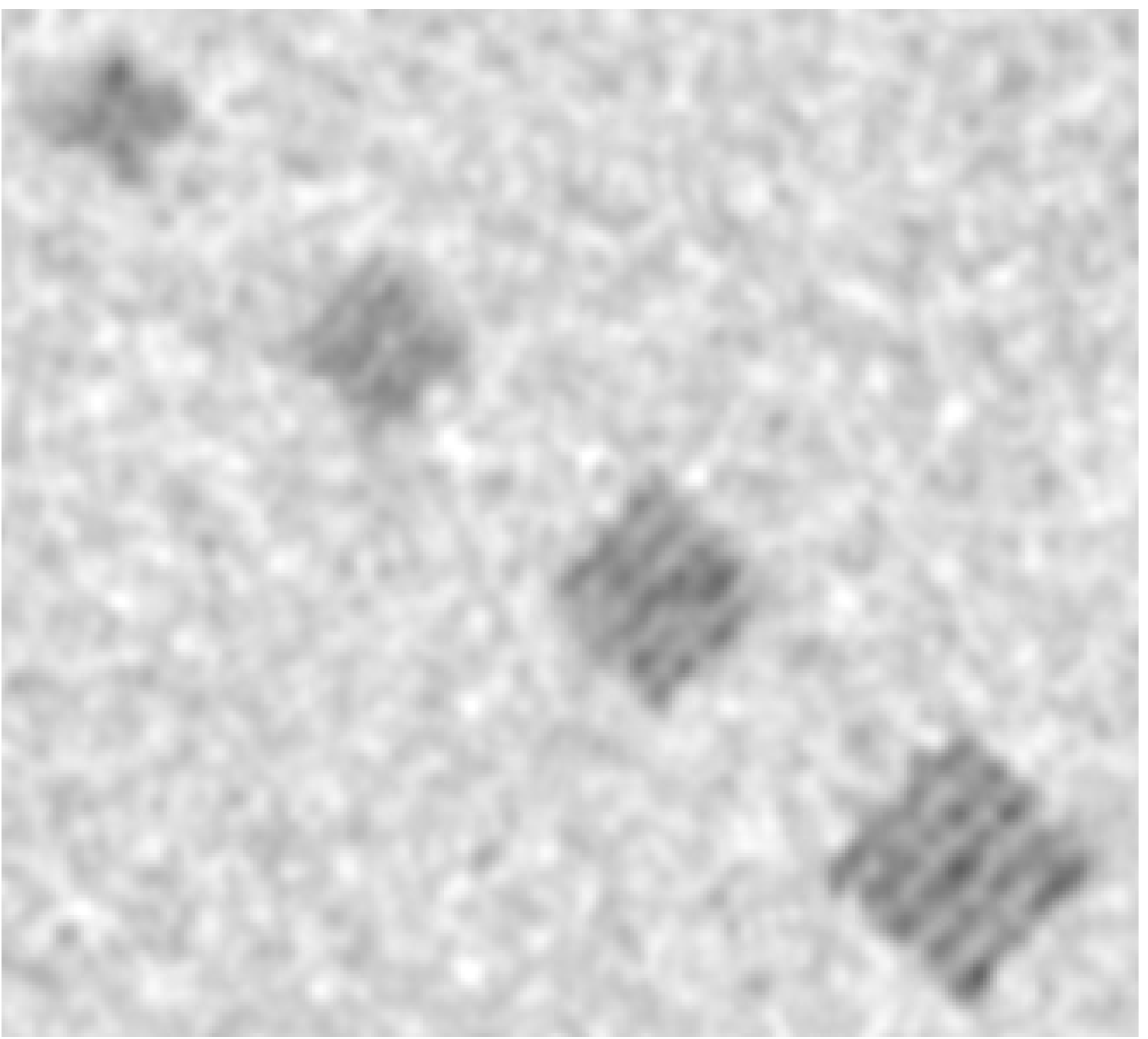}}\
\subfloat[Diff: (c) -- (d)]{\includegraphics[width=1.4in]{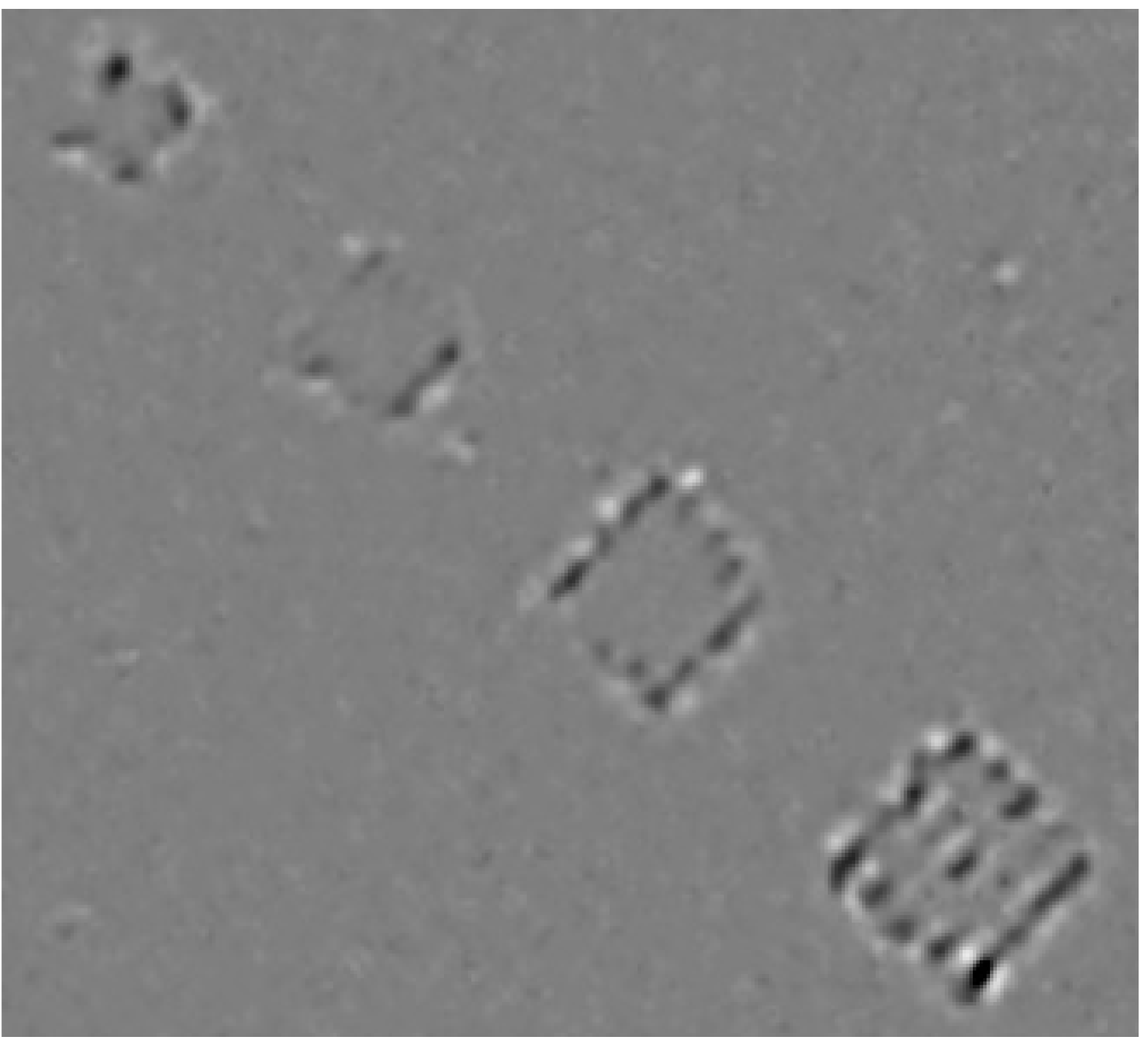}}\
\subfloat[Diff: (d) -- (e)]{\includegraphics[width=1.4in]{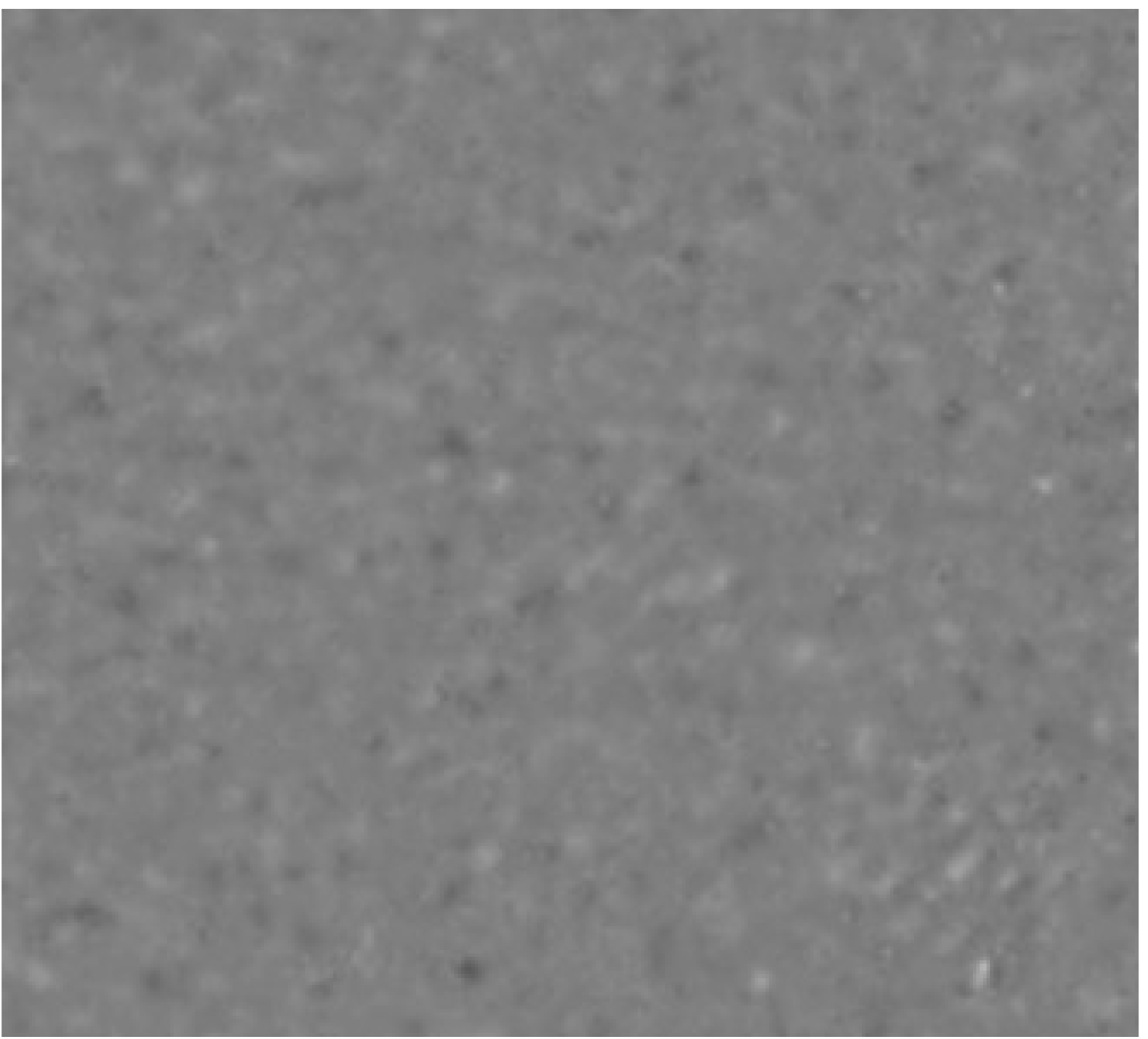}}}
\caption{GEPP reconstructions for 75 mA data, using $5\times5\times3$ GM-MRF model with varying value of parameter $p$. Individual image is zoomed to a small FOV containing cyclic bars for display purposes.
Display window: for GEPP images: [-85 165] HU; for difference images: [-15 15] HU.
With matched noise level in homogeneous regions, higher $p$ value leads to lower MTF in subplot (b)
and introduces more blurriness to cyclic bars.}
\label{fig:gepp_diffp}
\end{figure*}

\begin{figure*}[!t]
\centerline{
\subfloat[average eigenvalues with fixed $p$ and different values of $\alpha$]{\includegraphics[height=1.6in]{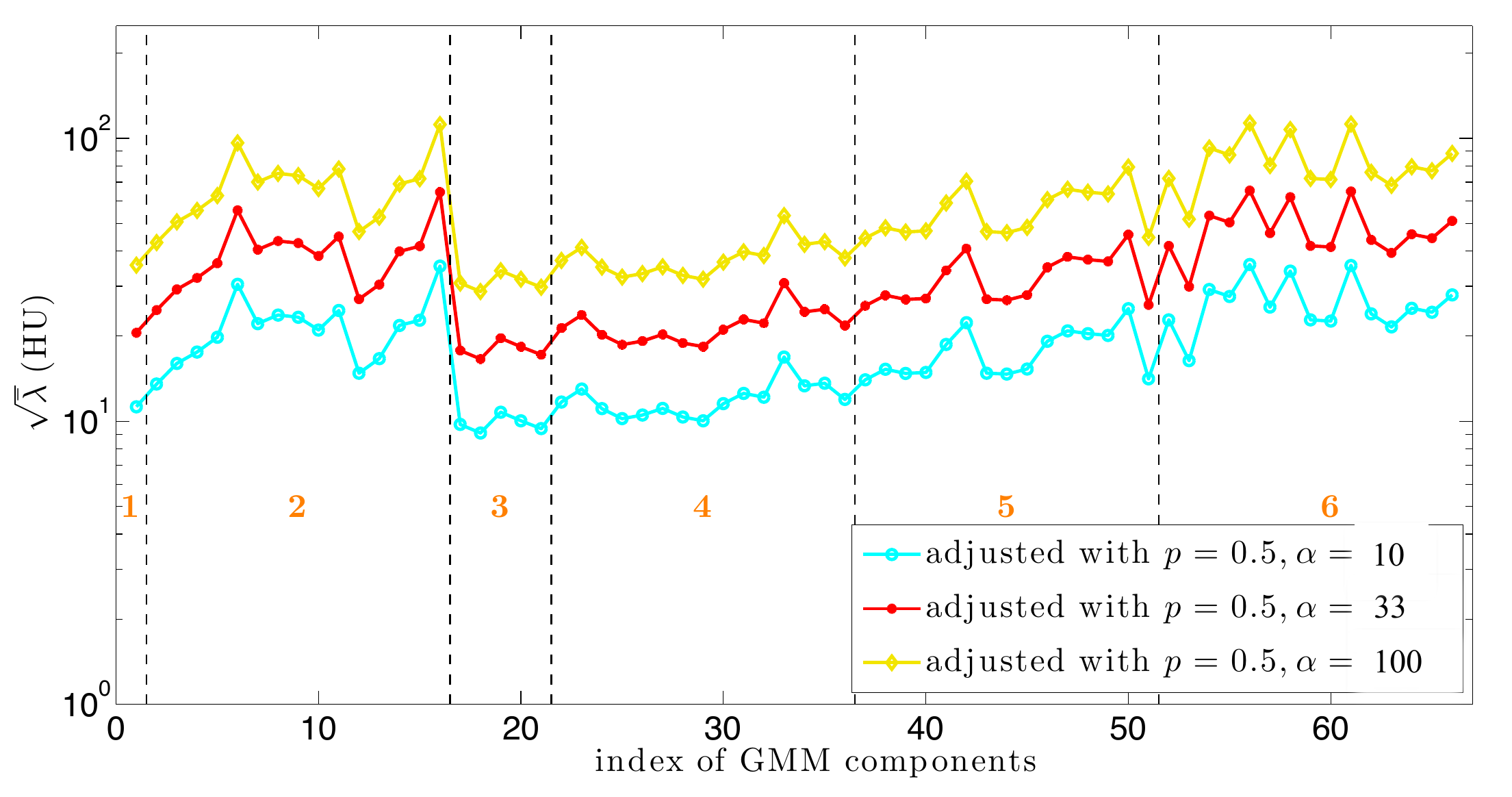}}\qquad
\subfloat[noise and resolution measurements]{\includegraphics[height=1.6in,trim={0  0.5in -1in 0},clip]{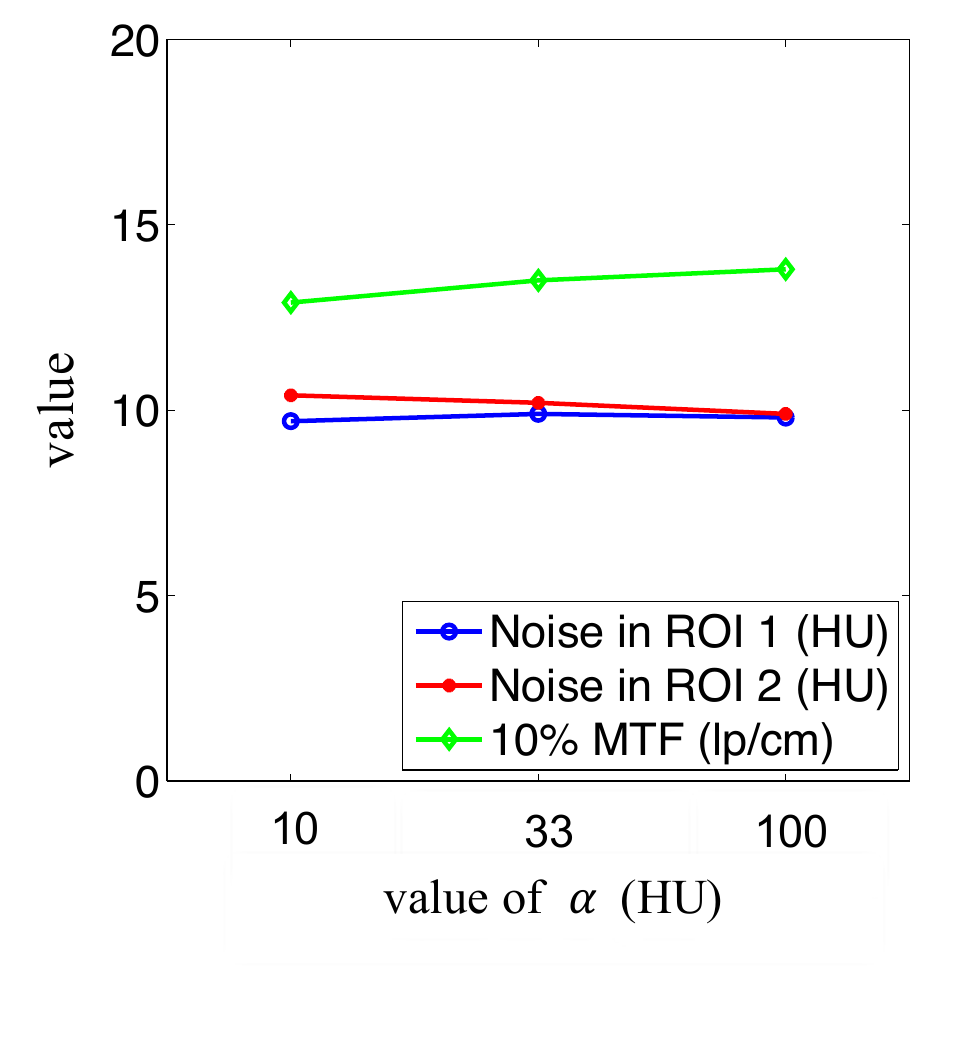}} } 
\centerline{ 
\subfloat[$\alpha=10\ {\rm HU}$]{\includegraphics[width=1.4in]{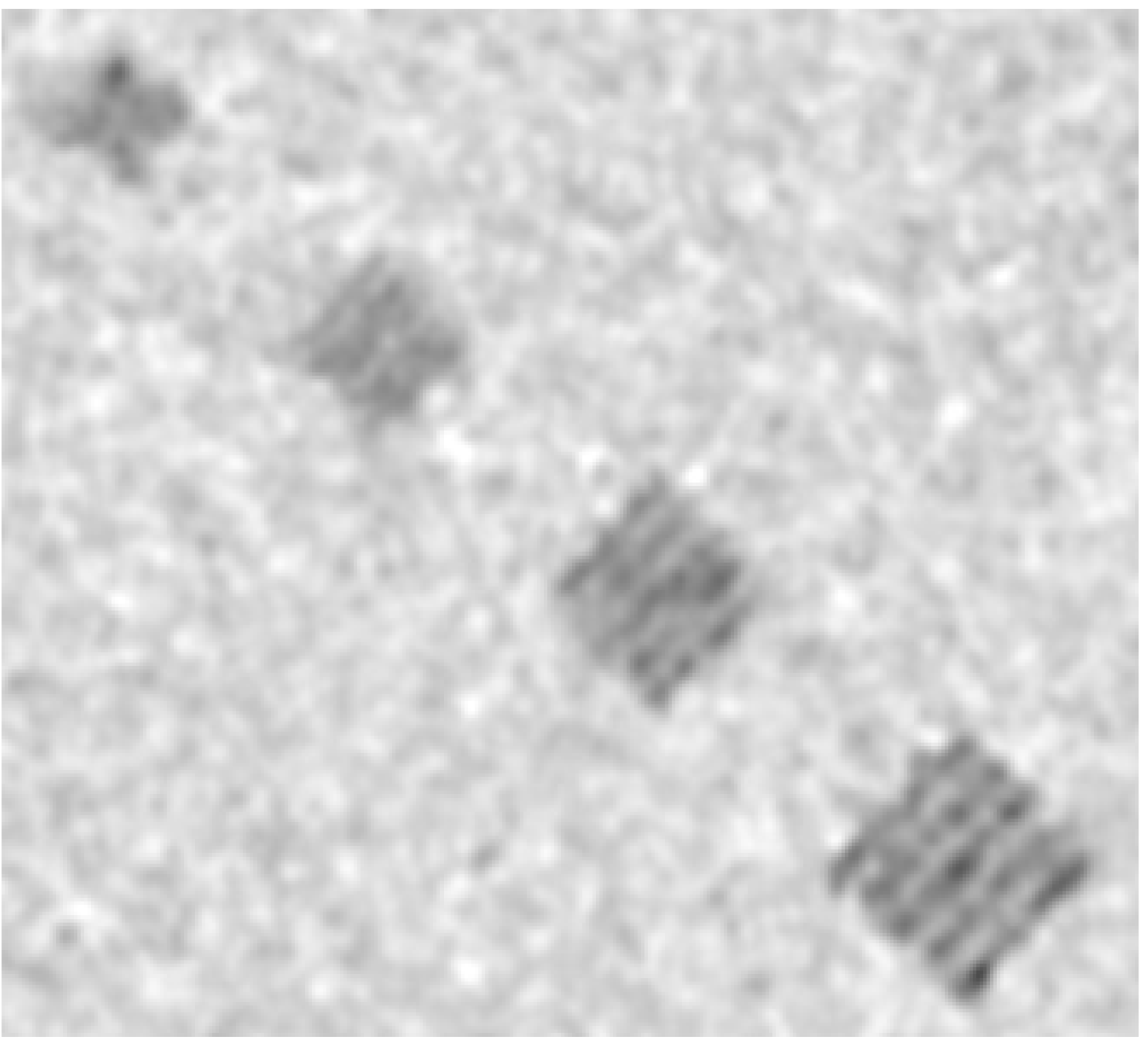}}\ 
\subfloat[$\alpha=33\ {\rm HU}$]{\includegraphics[width=1.4in]{figs/zhang107.pdf}}\
\subfloat[$\alpha=100\ {\rm HU}$]{\includegraphics[width=1.4in]{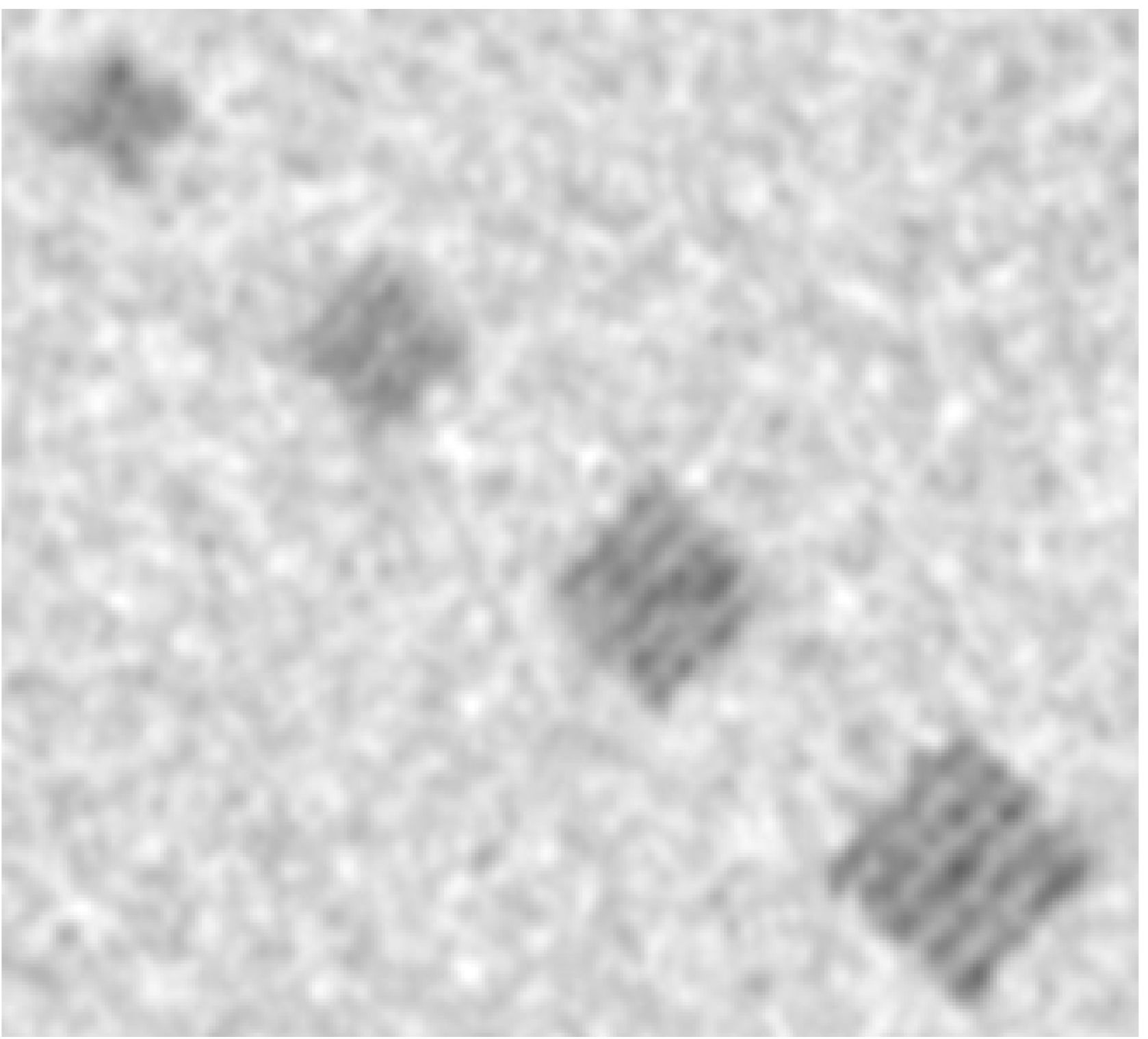}}\
\subfloat[Diff: (c) -- (d)]{\includegraphics[width=1.4in]{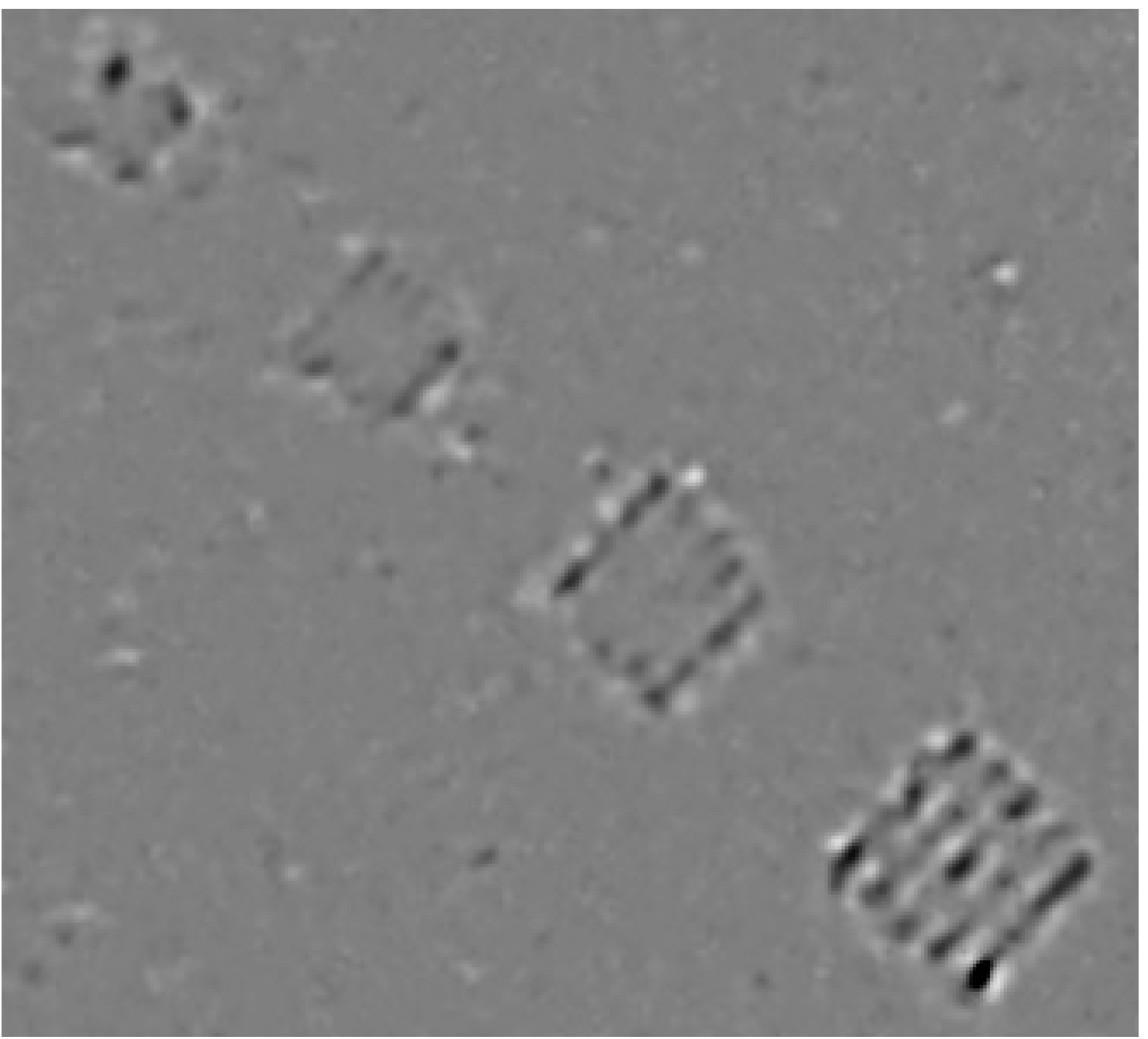}}\
\subfloat[Diff: (d) -- (e)]{\includegraphics[width=1.4in]{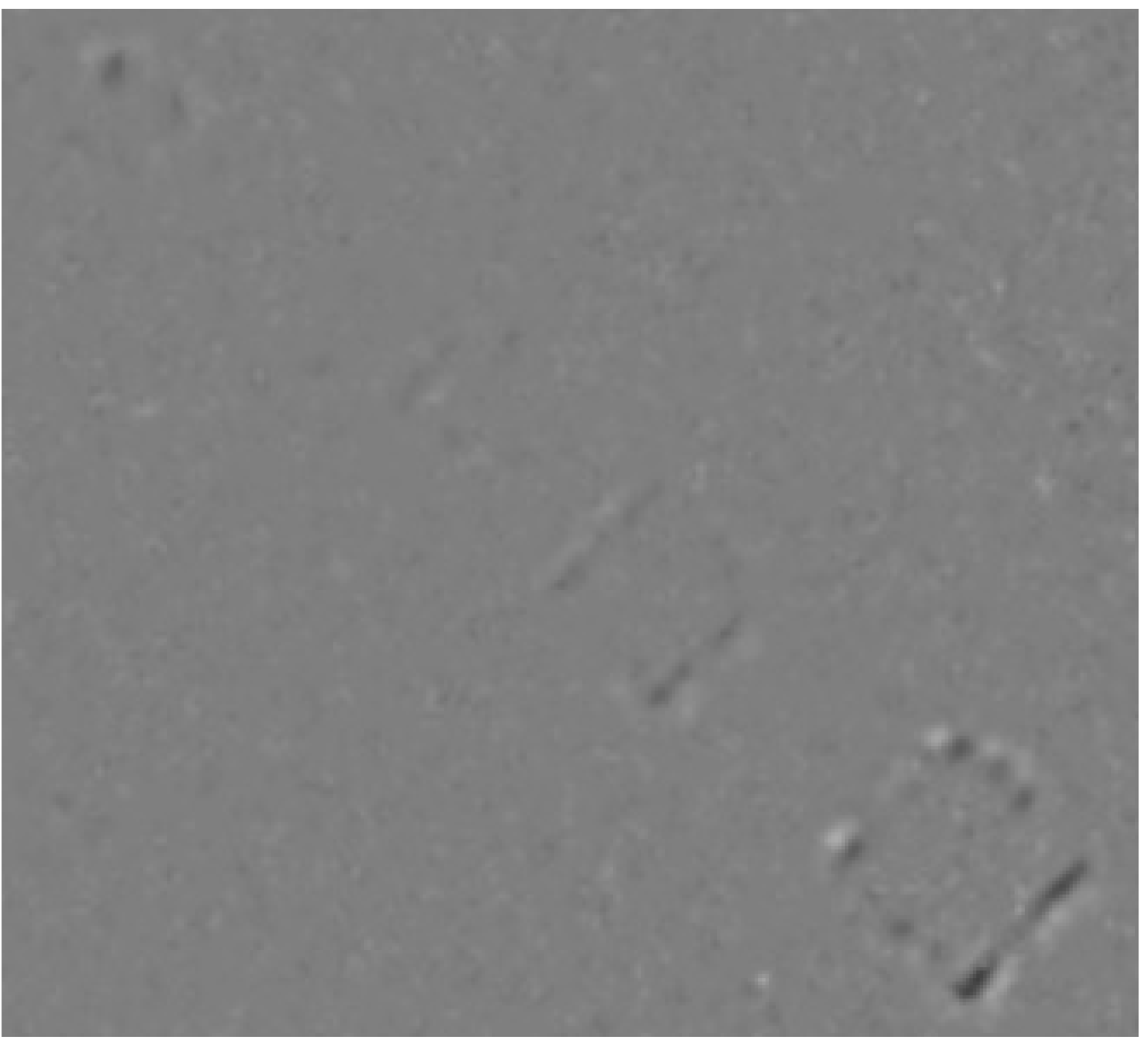}}}
\caption{GEPP reconstructions for 75 mA data, using $5\times5\times3$ GM-MRF model with varying value of parameter $\alpha$. Individual image is zoomed to a small FOV containing cyclic bars for display purposes. Display window: for GEPP images: [-85 165] HU; for difference images: [-15 15] HU.
Larger value of $\alpha$ introduces more blurriness to cyclic bars.}
\label{fig:gepp_diffa}
\end{figure*}

Fig. \ref{fig:gepp_diffp} and \ref{fig:gepp_diffa} present the study on the impact of parameters $p$ and $\alpha$ 
of GM covariance scaling on the reconstructed images, respectively.
Fig. \ref{fig:gepp_diffp} shows that, with matched noise level in homogeneous regions,
increasing the value of $p$ compresses the dynamic range of the average eigenvalues of GM covariance matrices, 
which in this case reduces the covariances of high-contrast edges (group 5) and bones (group 6) 
and therefore leads to decreasing high-contrast resolution as indicated by 10\% MTF in Fig. \ref{fig:gepp_diffp}(b).
Moreover, increased $p$ value with fixed $\alpha = 33\ {\rm HU}$ also increases the covariances of smooth soft tissues (group 3) and low-contrast edges (group 4),
which introduces more overlapping in the distribution and may consequently reduce the specification of structures and edges.
Therefore, we observe the blurring in the cyclic bars as $p$ increases. 
Fig. \ref{fig:gepp_diffa} shows that, with fixed compression rate $p=0.5$,
increasing the value of $\alpha$ increases the covariances for all GM components,
and therefore encourages distribution overlapping and may consequently reduce the specification of structures and edges, 
which leads to blurriness in cyclic bars.

\subsection{3-D clinical reconstruction}

\begin{figure*}[!t]
\centering
\begin{tabular}{C{2.2in}C{2.2in}C{2.2in}} 
\includegraphics[width=2.2in]{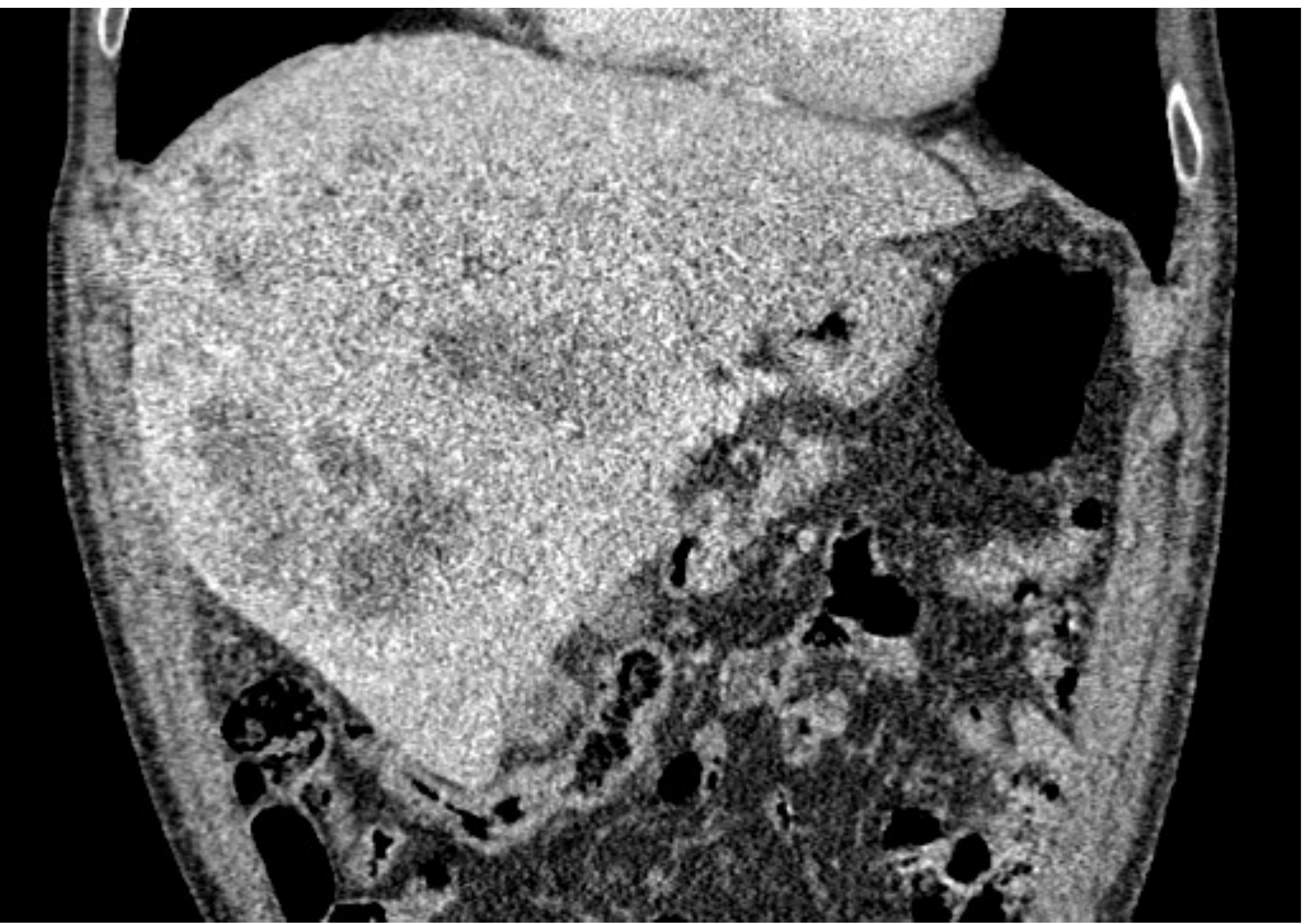} &
\includegraphics[width=2.2in]{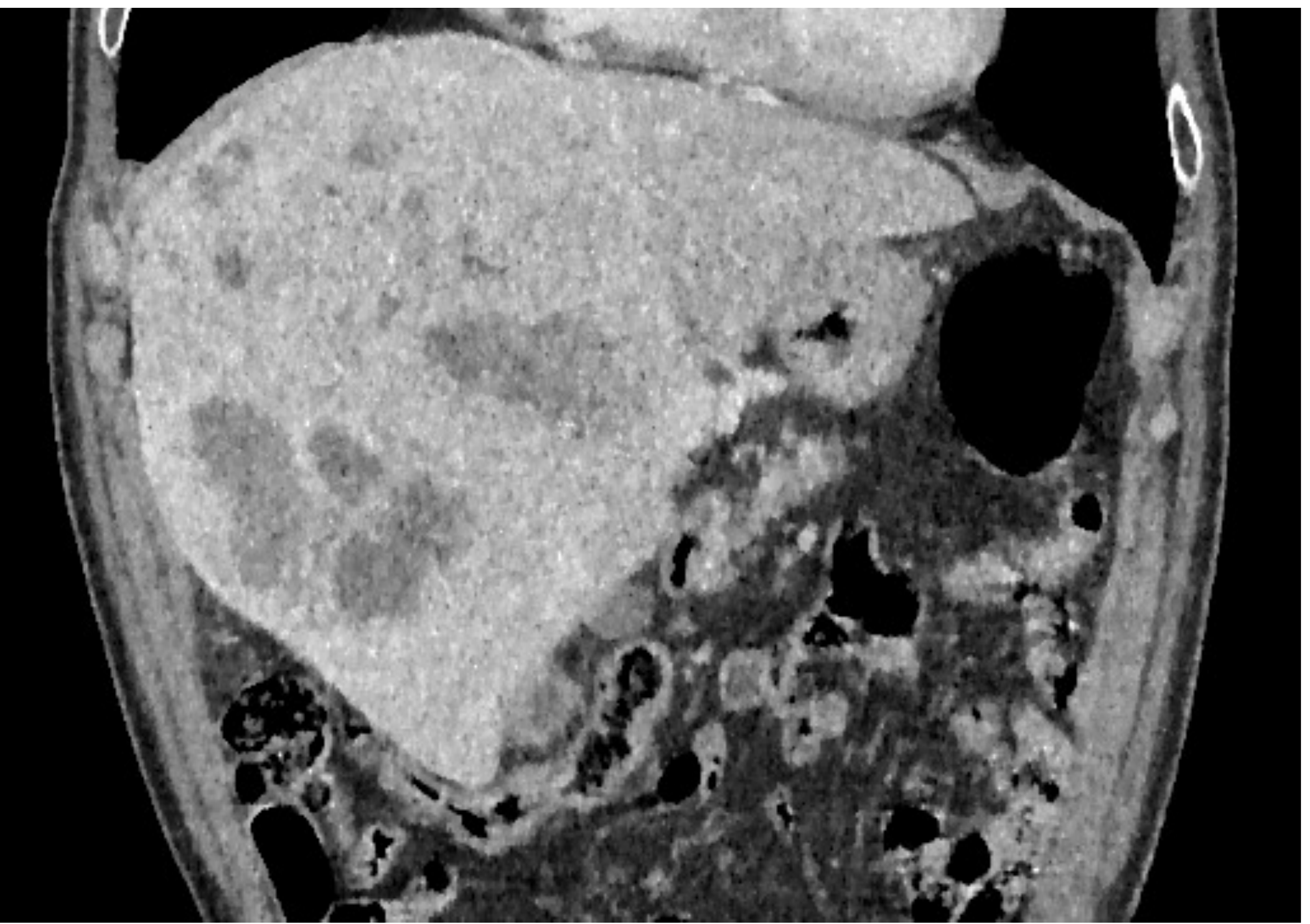} &
\includegraphics[width=2.2in]{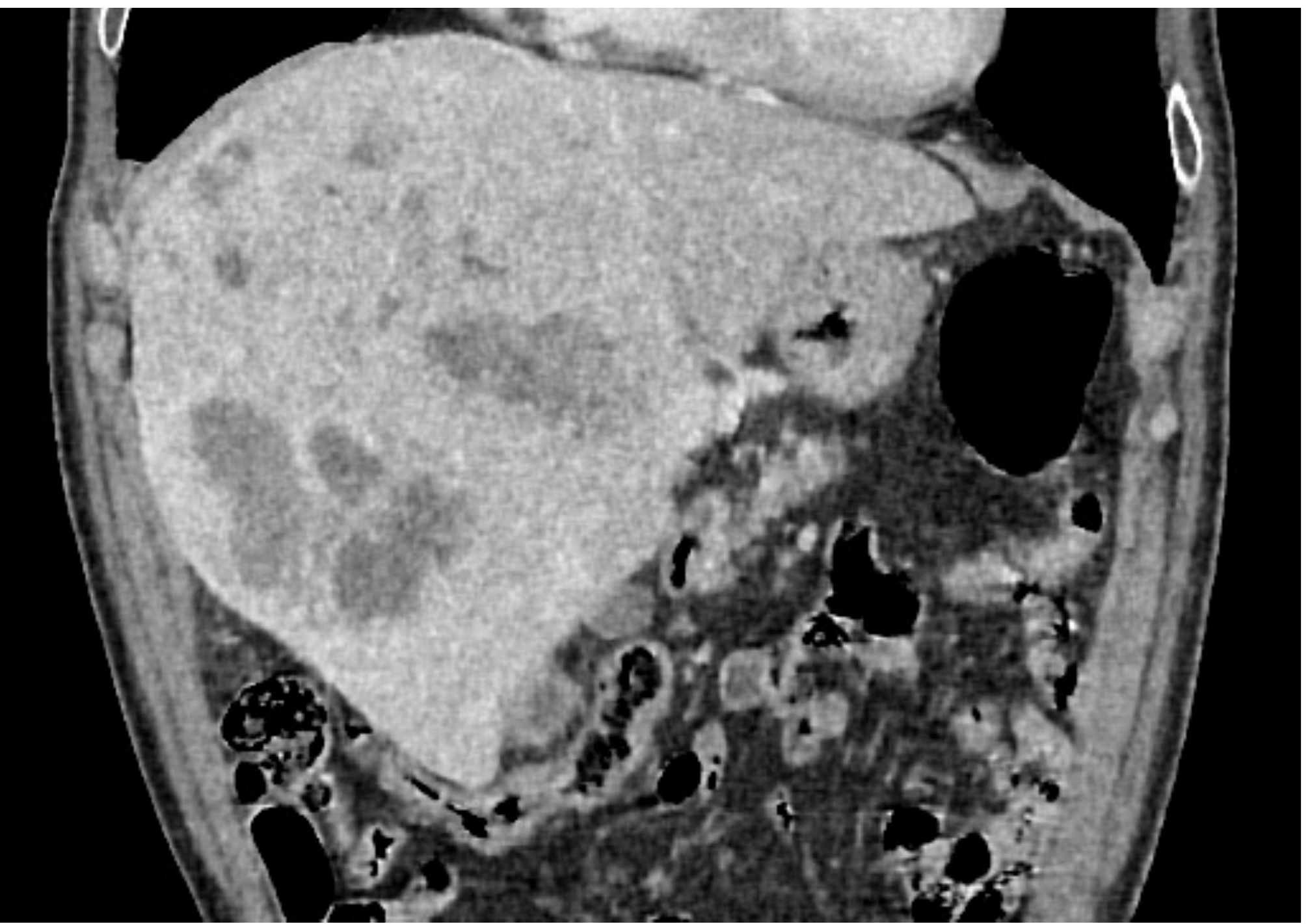} \\
\includegraphics[width=2.2in]{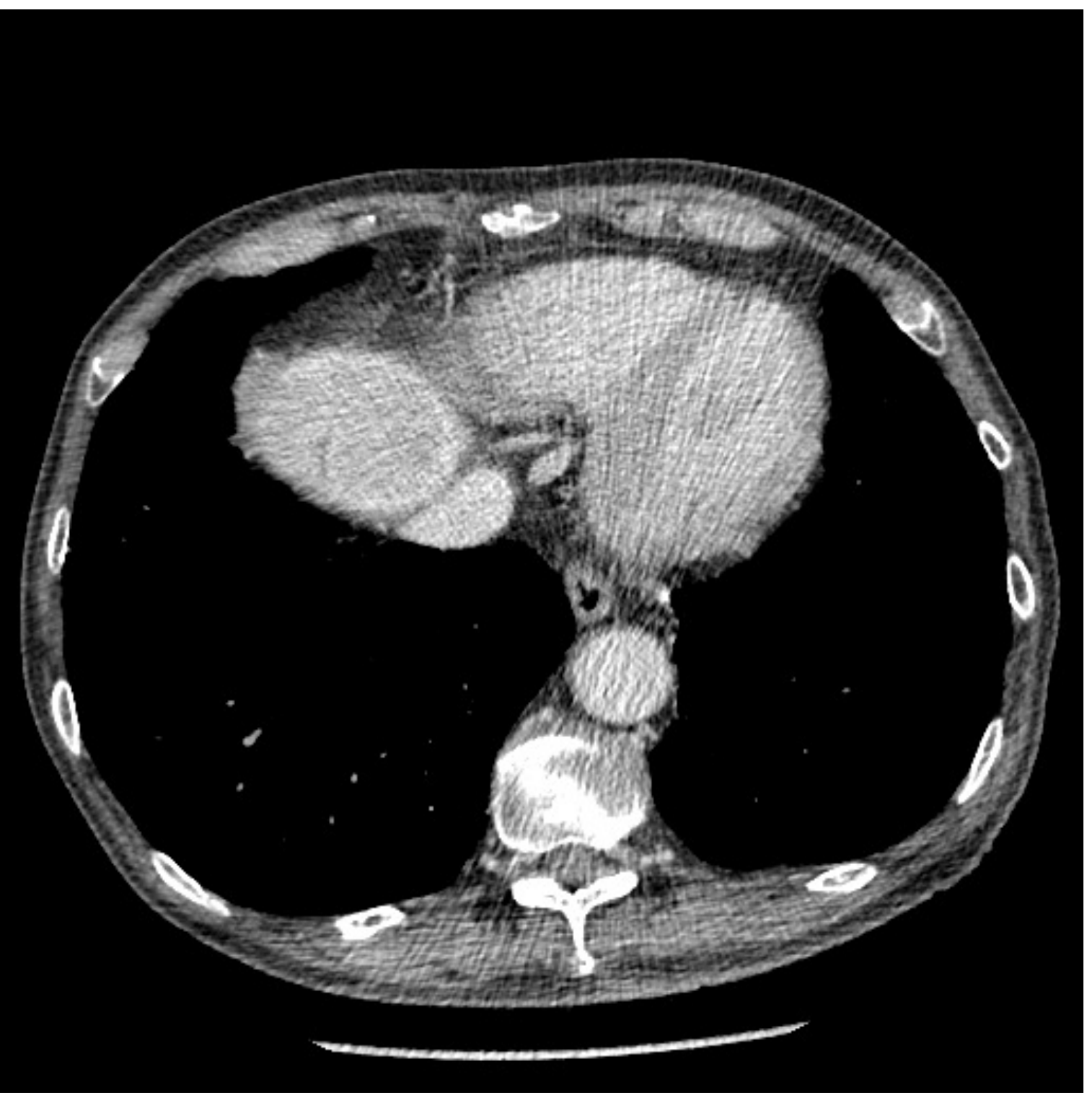} &
\includegraphics[width=2.2in]{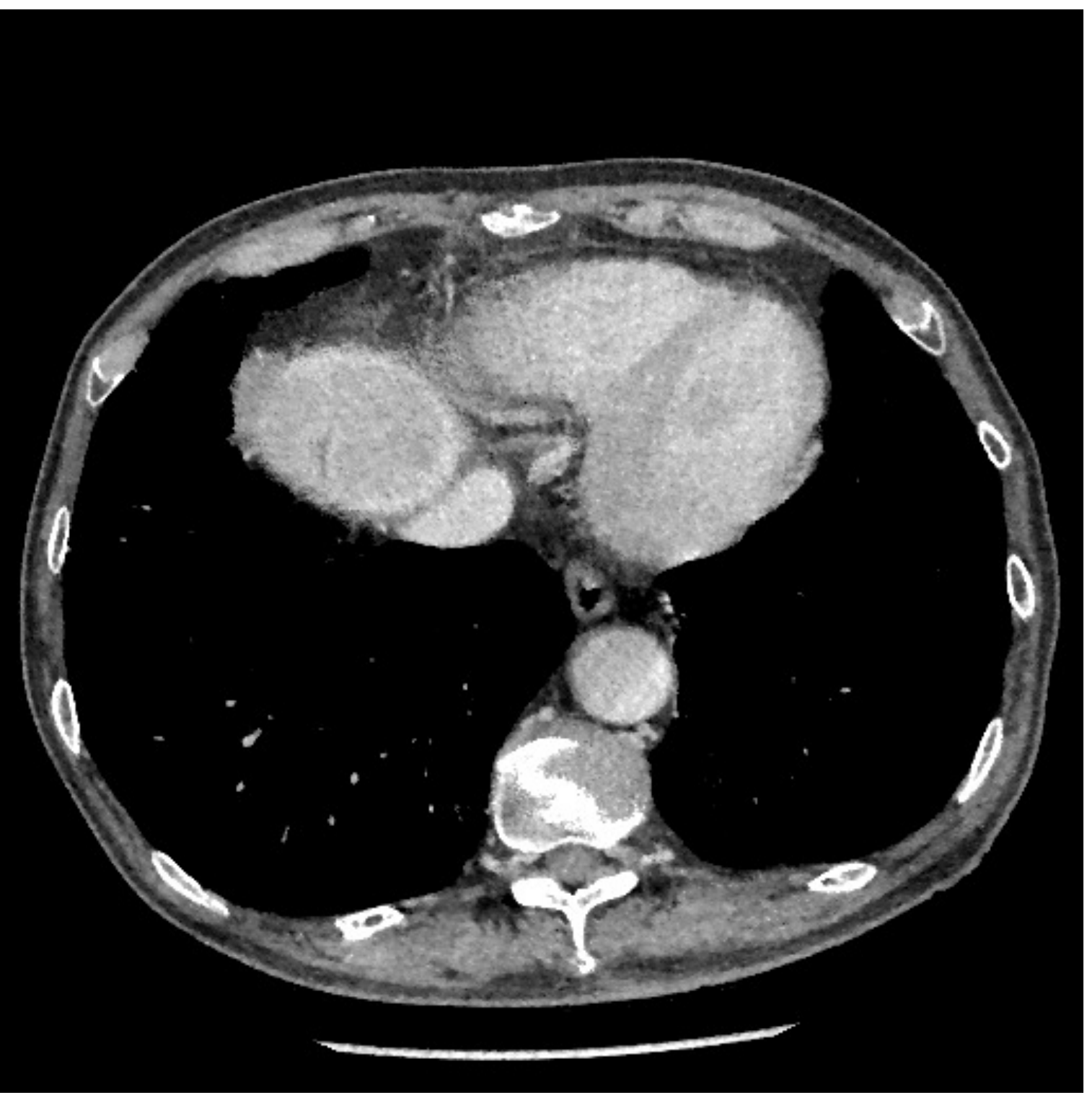} &
\includegraphics[width=2.2in]{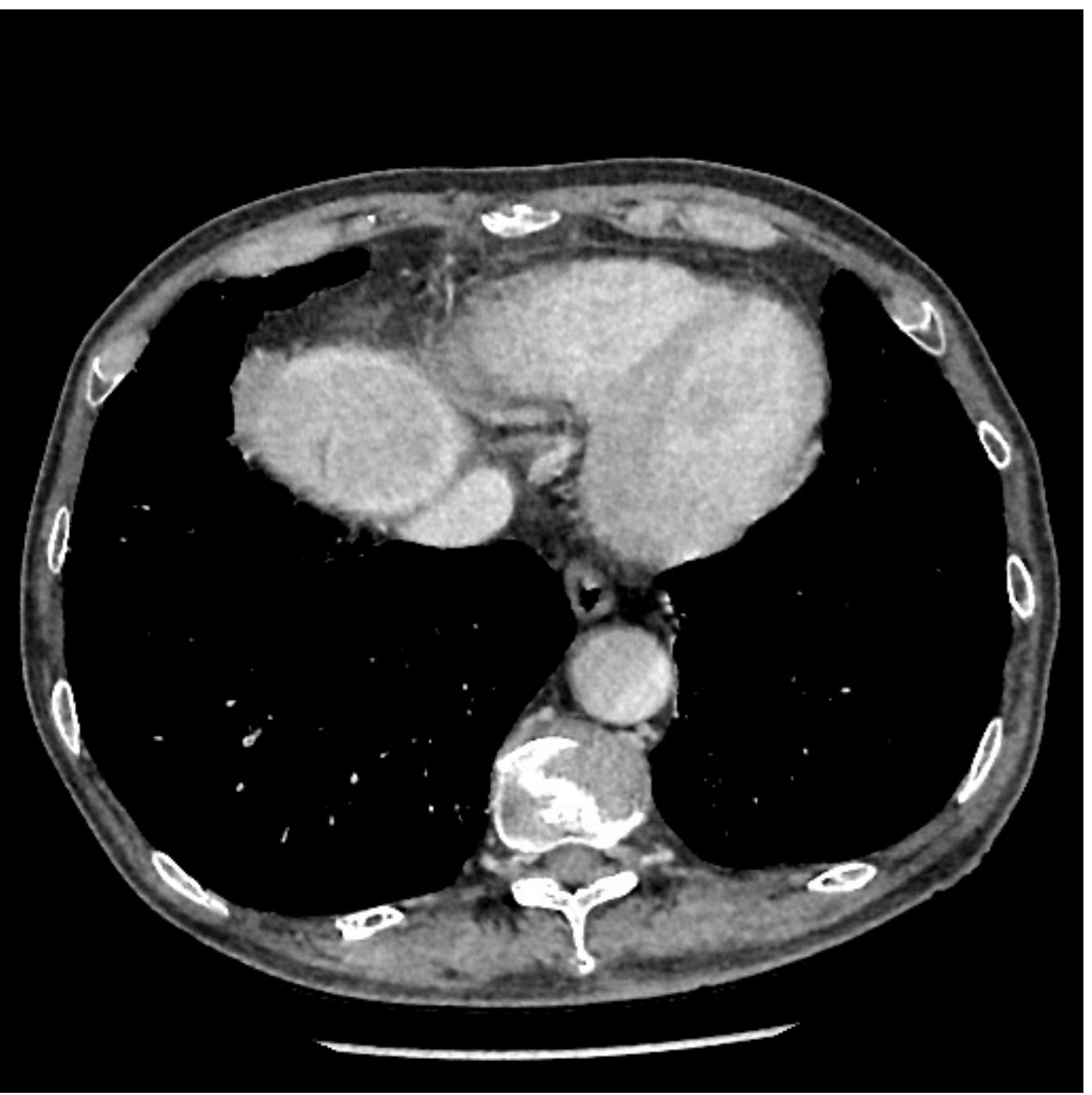} \\
\includegraphics[width=2.2in]{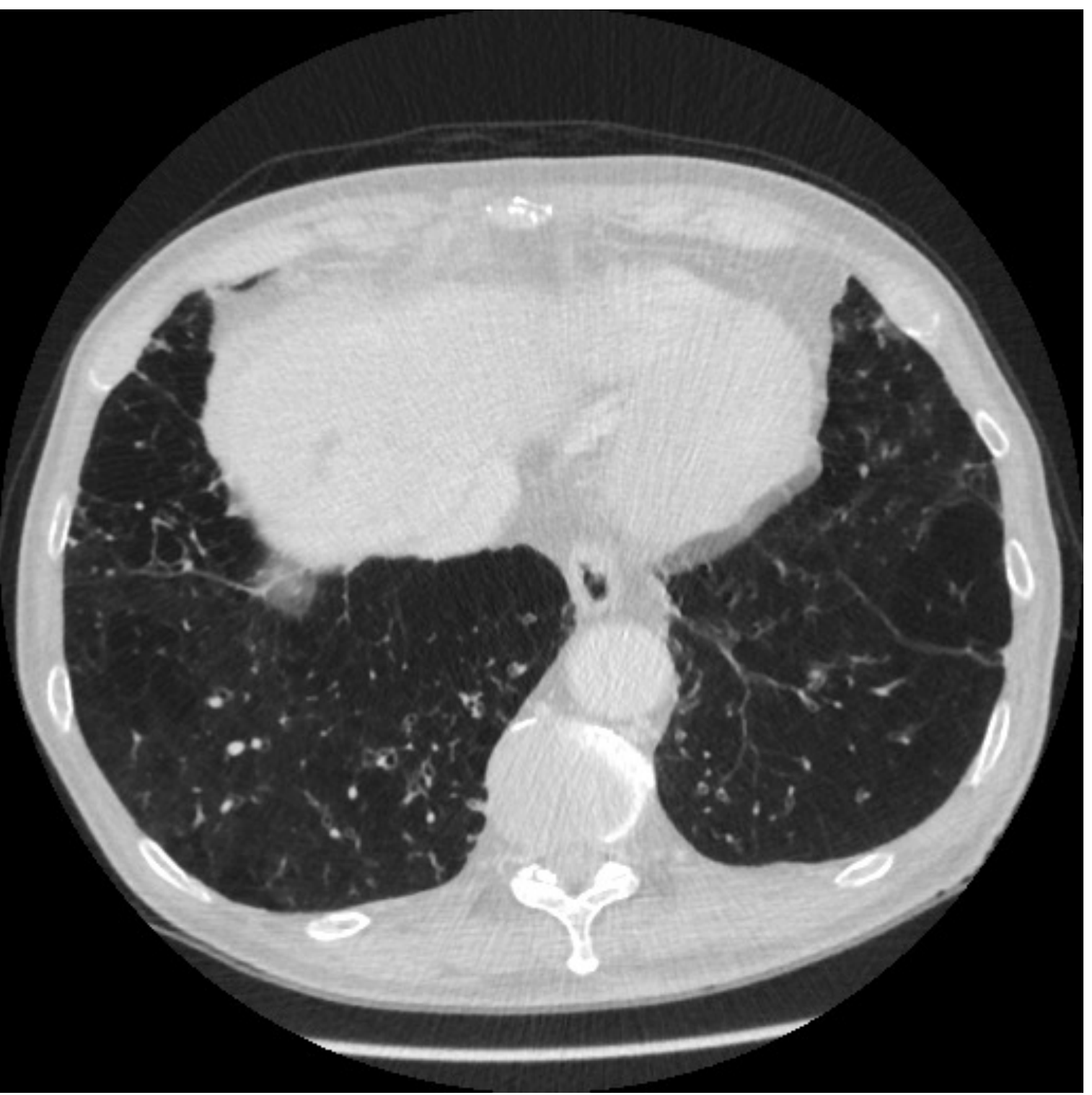} &
\includegraphics[width=2.2in]{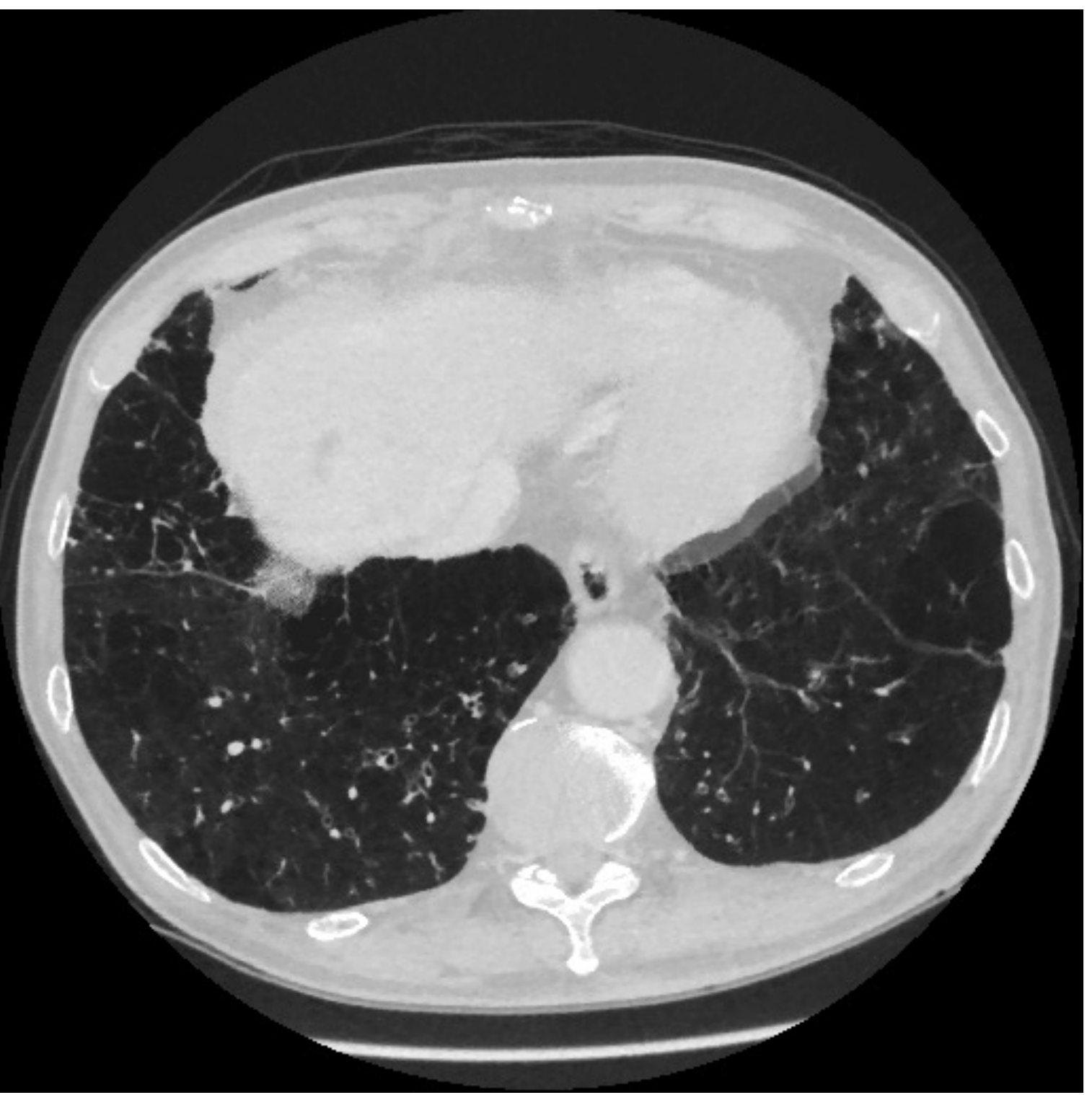} &
\includegraphics[width=2.2in]{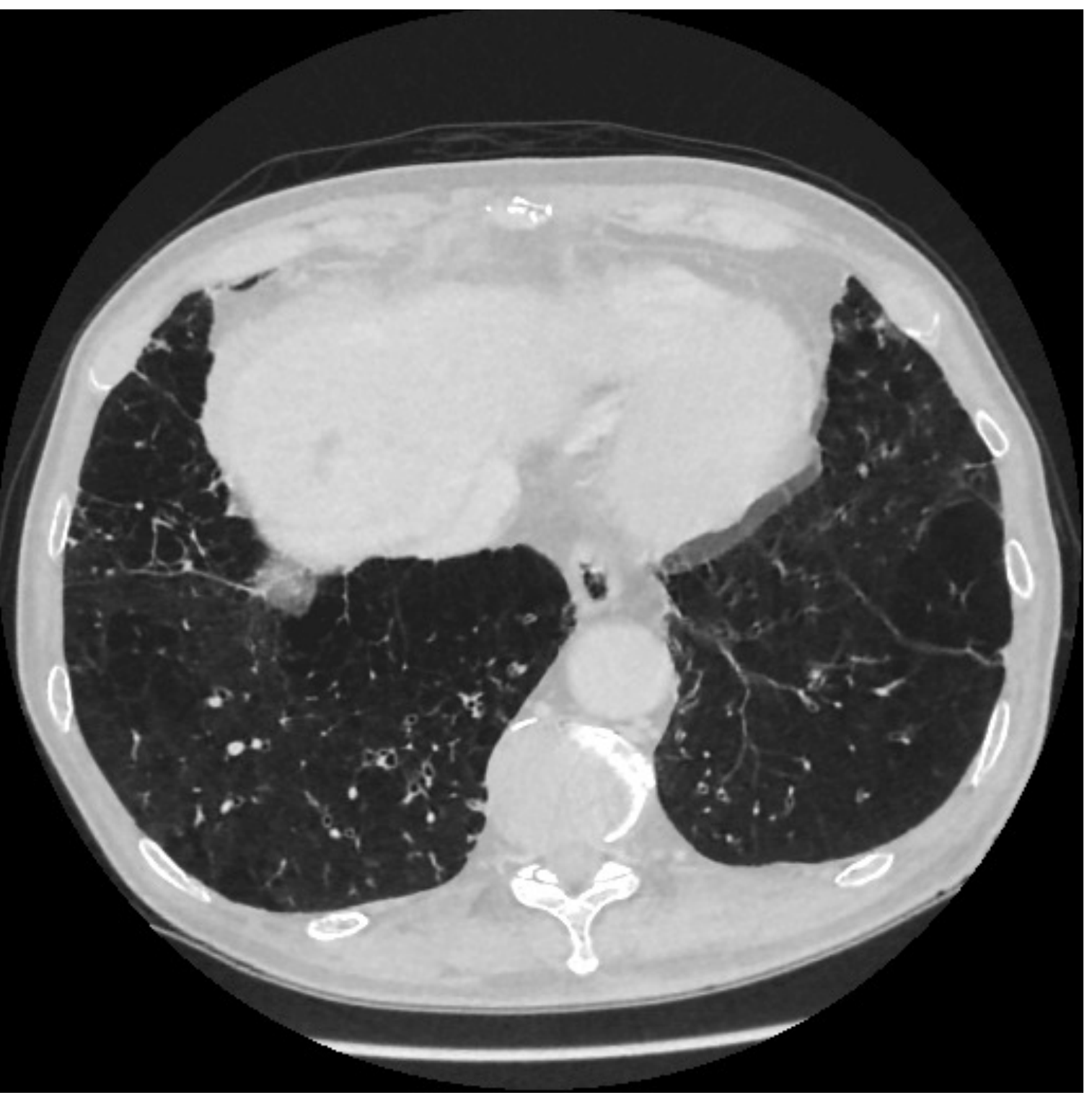} \\
\includegraphics[width=2.2in,trim={0 0.25in 0 0},clip]{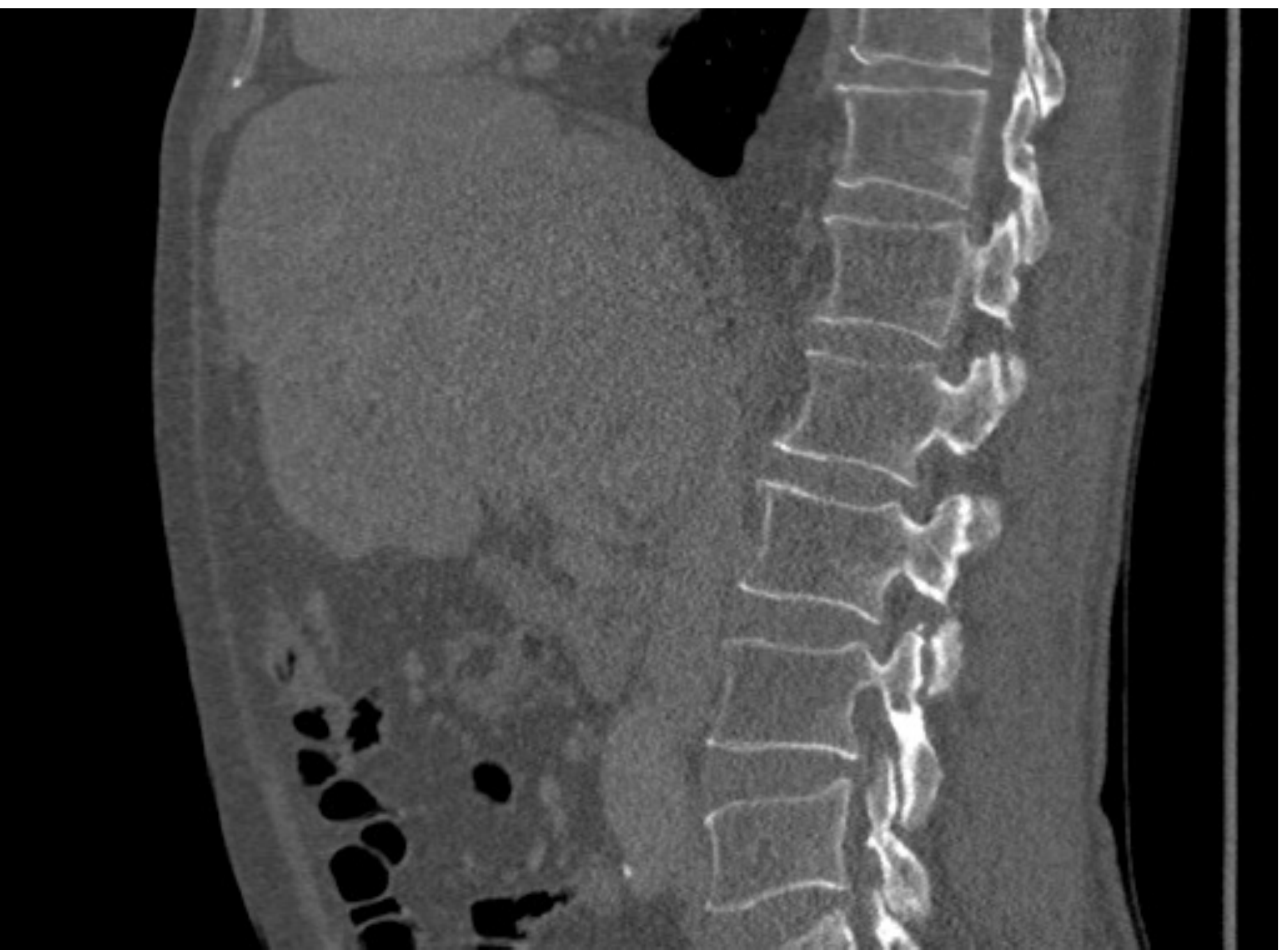} &
\includegraphics[width=2.2in]{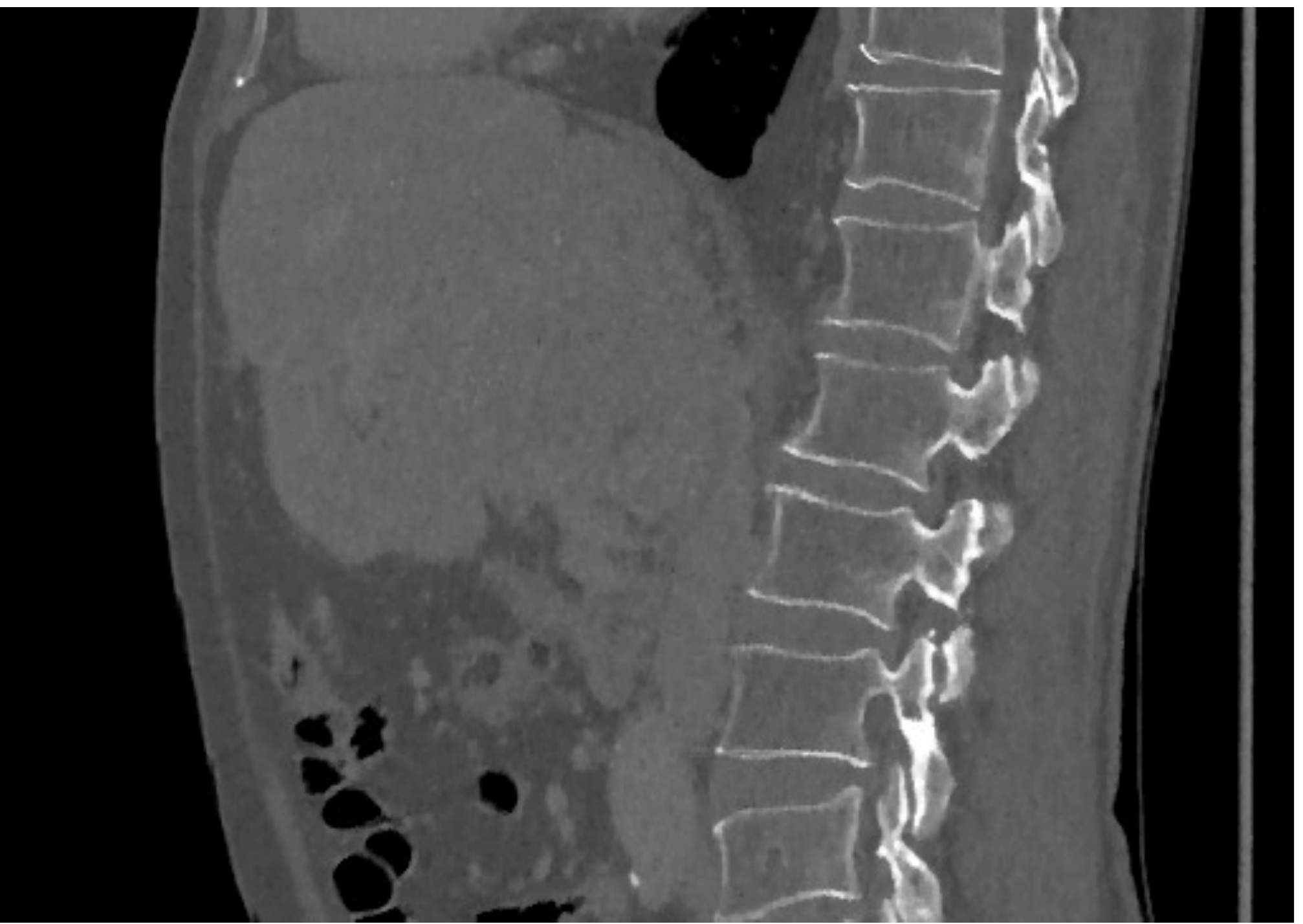} &
\includegraphics[width=2.2in]{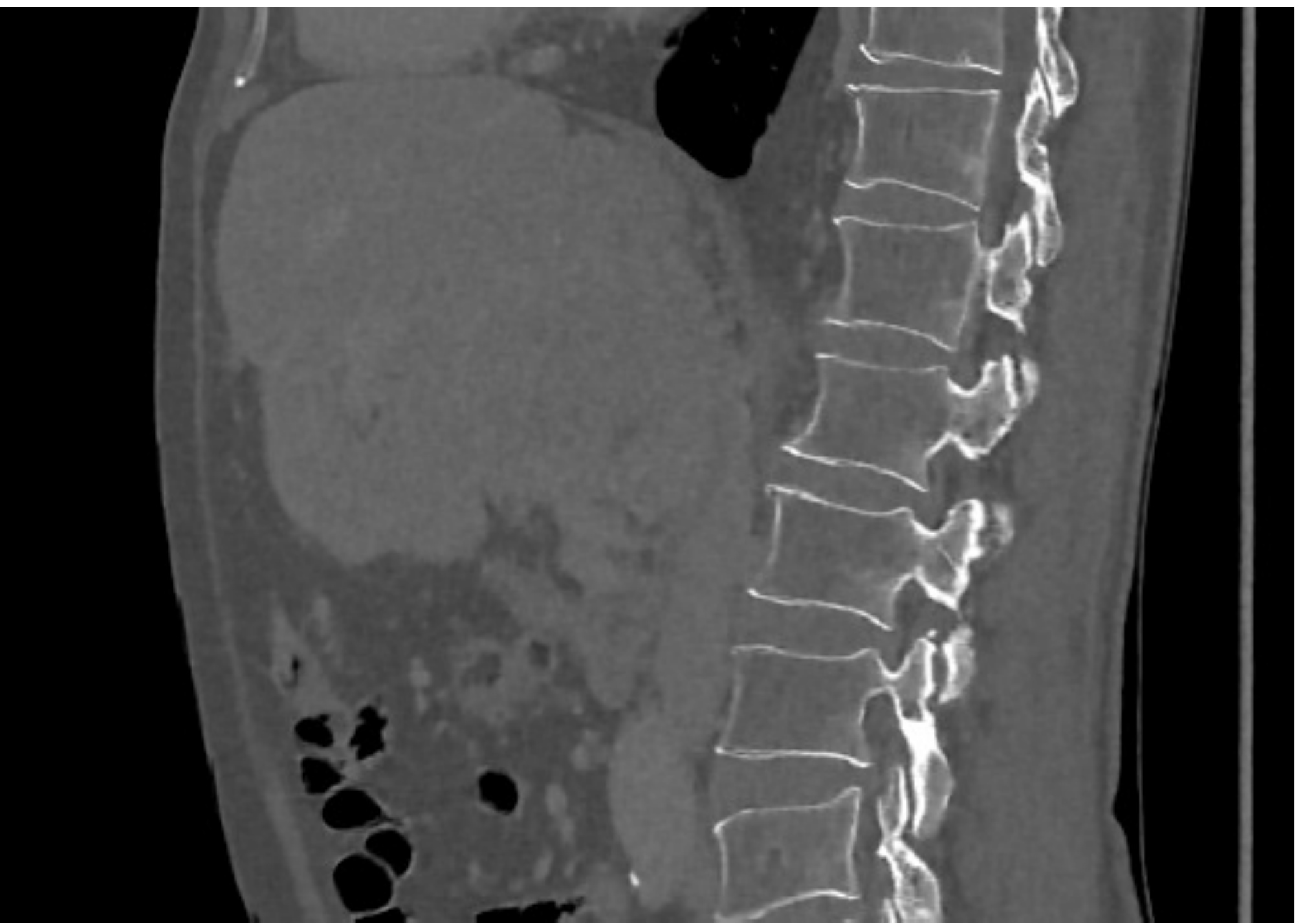} \\
(a) FBP  & (b) MBIR w/ $q$-GGMRF w/ reduced regularization & (c) MBIR w/ adjusted GM-MRF
\end{tabular}
\caption{From left to right, the columns represent (a) FBP, (b) MBIR with $q$-GGMRF with reduced regularization, and
(c) MBIR with adjusted GM-MRF with $p=0.5, \alpha=33\ {\rm HU}$.
Display window from top to bottom row: \mbox{[-160 240] HU}, \mbox{[-160 240] HU}, \mbox{[-1200 400] HU}, and \mbox{[-600 1200] HU}.
Note that MBIR with GM-MRF prior suppresses excessive noise spikes while maintaining detail such as contrast and edges in soft tissues, and also improves the sharpness in lung tissues and bones. 
}
\label{fig:clinical}
\end{figure*}

Fig.~\ref{fig:clinical} presents the 3-D reconstruction results of a low-dose scan of a new patient whose data was not used for training the GM-MRF model.
Experimental data was collected from a GE Discovery CT750 HD scanner,
which was acquired in \mbox{$64\times0.625$ mm} helical mode with \mbox{120 kVp}, \mbox{100 mA}, \mbox{0.4 s/rotation}, 
\mbox{pitch 0.984:1}, and reconstructed in \mbox{337 mm} FOV.

Fig.~\ref{fig:clinical} shows that the GM-MRF prior improves the texture in soft tissue 
without compromising the fine structures and details, as compared to the other methods.
Particularly, when compared to MBIR with traditional $q$-GGMRF prior, 
MBIR with GM-MRF prior reduces the speckle noise in liver (first row) while still maintaining the normal texture and edge definition.  
Moreover, the second row shows that MBIR with GM-MRF prior also improves the fuzzy edges 
as shown in the images associated with MBIR with traditional $q$-GGMRF prior.
The third and fourth rows show the improved resolution in lung and bone as produced by MBIR with GM-MRF prior.

\setlength\tabcolsep{0in}

\end{document}